\documentclass{article}

\usepackage{microtype}
\usepackage{graphicx}
\usepackage{booktabs} %

\usepackage{hyperref}

\usepackage{amsmath}
\usepackage{cleveref}
\usepackage{subcaption}

\usepackage{wrapfig}
\usepackage{capt-of}
\usepackage{tabularx}

\usepackage{textcomp}
\usepackage{etoolbox}

\makeatletter
\newcommand{\multiline}[1]{%
  \begin{tabularx}{\dimexpr\linewidth-\ALG@thistlm}[t]{@{}X@{}}
    #1
  \end{tabularx}
}
\makeatother

\usepackage{color}
\usepackage{amsmath}
\usepackage{amssymb}
\usepackage{multirow, varwidth}
\usepackage{dblfloatfix}
\usepackage{verbatim}
\makeatletter
\newif\if@restonecol
\makeatother

\usepackage{xr}
\usepackage[amsmath,thmmarks]{ntheorem}

\DeclareRobustCommand\onedot{\futurelet\@let@token\@onedot}
\def\onedot{. }
\def\eg{\emph{e.g}\onedot} 
\def\ie{\emph{i.e}\onedot}

\makeatletter
\patchcmd{\@algocf@start}%
  {-1.5em}%
  {0pt}%
  {}{}%
\makeatother

\definecolor{candypink}{rgb}{0.89, 0.44, 0.48}
\definecolor{mediumaquamarine}{rgb}{0.4, 0.8, 0.67}
\definecolor{azure}{rgb}{0.0, 0.5, 1.0}
\definecolor{awesome}{rgb}{1.0, 0.13, 0.32}

\newcommand{\pienv}{\pi^{\text{lin}}}
\newcommand{\samplesub}{\theta_{\text{s}}}
\newcommand{\pisample}{\pi_{\samplesub}}
\newcommand{\optionsub}{\theta_{\text{z}}}
\newcommand{\pioption}{\pi_{\optionsub}}
\newcommand{\trajencoder}{p_\phi}

\newcommand{\rewardenv}{R^{\text{lin}}}
\newcommand{\Mod}[1]{\ (\mathrm{mod}\ #1)}

\DeclareMathOperator*{\maximize}{maximize}

\usepackage[accepted]{icml2021}

\icmltitlerunning{Unsupervised Skill Discovery with Bottleneck Option Learning}

\begin{document}

\twocolumn[
\icmltitle{Unsupervised Skill Discovery with Bottleneck Option Learning}

\icmlsetsymbol{equal}{*}

\begin{icmlauthorlist}
\icmlauthor{Jaekyeom Kim}{equal,snucse}
\icmlauthor{Seohong Park}{equal,snucse}
\icmlauthor{Gunhee Kim}{snucse}
\end{icmlauthorlist}

\icmlaffiliation{snucse}{Department of Computer Science and Engineering, Seoul National University, South Korea}

\icmlcorrespondingauthor{Gunhee Kim}{gunhee@snu.ac.kr}

\icmlkeywords{Reinforcement Learning, Skill Discovery}

\vskip 0.3in
]

\printAffiliationsAndNotice{\icmlEqualContribution} %

\begin{abstract}
Having the ability to acquire inherent skills from environments without any external rewards or supervision like humans is an important problem.
We propose a novel unsupervised skill discovery method named \textit{Information Bottleneck Option Learning (IBOL)}.
On top of the linearization of environments that promotes more various and distant state transitions,
IBOL enables the discovery of diverse skills.
It provides the abstraction of the skills learned with the information bottleneck framework for the options with improved stability and encouraged disentanglement.
We empirically demonstrate that IBOL outperforms multiple state-of-the-art unsupervised skill discovery methods on the information-theoretic evaluations and downstream tasks in MuJoCo environments, including Ant, HalfCheetah, Hopper and D'Kitty. 
Our code is available at \url{https://vision.snu.ac.kr/projects/ibol}.
\end{abstract}

\section{Introduction}
Deep reinforcement learning (RL) has recently shown great advancement in solving various tasks,
from playing video games \cite{atari_mnih2013,atarinature_mnih2015,openaidota_berner2019} to controlling robot navigation \cite{rlrobot_kahn2018}.
While the standard RL is to maximize rewards from environments as a form of supervision,
there has been a surge of interest in unsupervised learning without the assumption of extrinsic rewards \cite{intrinsic_sukhbaatar2018,max_shyam2019}.
Discovering inherent skills in environments without supervision %
is important and desirable for multiple reasons.
First, since it is still challenging to define an effective reward function for practical tasks \cite{ird_hadfield2017,realworld_dulac2019}, %
unsupervised skill discovery helps reduce the burden of it by identifying effective skills for environments. 
Second, in sparse-reward environments, learned skills can encourage the exploration for encountering rewards,
not only by providing useful primitives for the exploration but also by reducing the effective horizon.
Third, those skills can be directly used to solve downstream tasks, for example, by 
employing a meta-controller on top of the discovered skills in a hierarchical manner \cite{valor_achiam2018,diayn_eysenbach2019,dads_sharma2020}.
Finally, discovered skills could help better understand environments by providing interpretable sets of behaviors.

Unsupervised skill discovery can be formalized with the \textit{options} framework \cite{hrl_sutton1999},
which generalizes primitive actions with the notion of \textit{options}.
For ease of learning, options, or synonymously \emph{skills}, are often formulated by introducing a skill latent parameter $z$ to an ordinary policy,
resulting in a skill policy with a form of $\pi(a | s, z)$ keeping the same $z$ for multiple steps or the full episode horizon \cite{vic_gregor2016,valor_achiam2018,diayn_eysenbach2019,dads_sharma2020}.
In recent research on the unsupervised skill discovery problem, 
information-theoretic approaches have been prevalent \cite{vic_gregor2016,valor_achiam2018,diayn_eysenbach2019,dads_sharma2020}.

\begin{figure*}[t!]
  \centering
  \includegraphics[width=\linewidth]{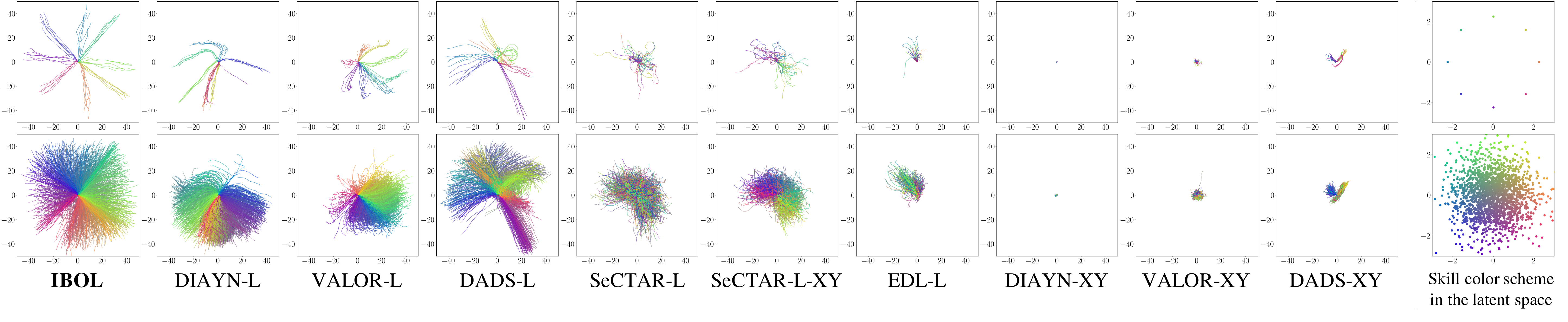}
  \caption{
    Visualization of the $x$-$y$ traces of skills discovered by each algorithm in Ant,
    where the colors represent the two-dimensional skill latents used for the sampling of the skills (see the color scheme on the right).
    (Top) Trajectories of the six roll-outs from each of the eight different skill latents.
    (Bottom) Trajectories of $2000$ skill latents sampled from the standard normal distribution.
  }
  \label{fig:ant_xy}
\end{figure*}

In this work, we propose a novel unsupervised skill discovery method named \textit{Information Bottleneck Option Learning (IBOL)},
whose two major novelties over existing approaches are (i) the \textit{linearizer} and (ii) the \textit{information bottleneck}-based skill learning. 
First, the linearizer is a kind of low-level policy to be suitable for skill discovery by converting a given environment into one with simplified dynamics. %
It reduces the skill discovery algorithm's burden to learn how to make transitions to diverse states in a given environment without any external rewards,
which is not a straightforward job with fairly complex dynamics such as Ant and Humanoid from MuJoCo \cite{mujoco_todorov2012}.
Once the linearizer is trained, it can be reused   for multiple training sessions with different skill discovery approaches.
Figure \ref{fig:ant_xy} compares the qualitative visualization of the skills learned by different methods in the locomotion (\ie $x$-$y$) plane, in Ant.
As shown, DIAYN \cite{diayn_eysenbach2019}, VALOR \cite{valor_achiam2018} and DADS \cite{dads_sharma2020} with the linearizer (with suffix `-L') learn far more diverse skills than the same methods without the linearizer.

Leveraging the environment simplified with the linearizer,
IBOL discovers and learns skills based on the information bottleneck (IB) framework \cite{ib_tishby2000,vib_alemi2016}. %
Compared to prior approaches, IBOL can introduce some desirable properties to the learned skills.
It discovers and learns skills with the skill latent variable $Z$ in a more disentangled way,
which makes the learned skills better interpretable with respect to $Z$.
Interpretable models help understand their behaviors and provide intuition about their further uses \cite{adel2018_interpretable}.
Figure \ref{fig:ant_xy} demonstrates that the skill trajectories instantiated by IBOL have a visually simpler and more predictable mapping with the skill latents,
which is one of the main requirements for increasing interpretability \cite{adel2018_interpretable}.
Moreover, the skills learned by IBOL cover the locomotion plane more uniformly and widely.
Finally, with the IB-style objective, the skill latent variable $Z$ is learned to be not only informative about the discovered skills but also parsimonious to keep unrelated information about the skills.

Our key contributions can be summarized as follows.

\begin{itemize}
  \item To the best of our knowledge, our method is the first to separate the problem of making transitions in the state space from skill discovery, simplifying the environment dynamics with independent pre-training, whose learning cost is amortized across multiple skill discovery trainings.
    It aids skill discovery methods to learn diverse skills by making the environment dynamics as linear as possible.
  \item We propose a novel skill discovery method with information bottleneck,
    which provides multiple benefits including learning skills in a more disentangled and interpretable way with respect to skill latents and being robust to nuisance information.
  \item Our method shows superior performance to various state-of-the-art unsupervised skill discovery methods including DADS \cite{dads_sharma2020}, DIAYN \cite{diayn_eysenbach2019} and VALOR \cite{valor_achiam2018} in multiple MuJoCo \cite{mujoco_todorov2012} environments.
    To verify this, we measure the information-theoretic metrics and the performance on four downstream tasks.
\end{itemize}

\section{Preliminaries and Related Work}
\label{sec:prior_works}
We review previous information-theoretic approaches to unsupervised skill discovery and discuss their limitations.

\textbf{Preliminaries}.
We consider a Markov Decision Process (MDP) $\mathcal{M} = (\mathcal{S}, \mathcal{A}, p)$ \emph{without external rewards}.
$\mathcal{S}$ and $\mathcal{A}$ respectively denote the state and action spaces, and 
$p(s_{t+1}|s_t,a_t)$ is the transition function
where $s_t, s_{t+1} \in \mathcal{S}$ and $a_t \in \mathcal{A}$.
Given a policy $\pi(a_t|s_t)$, a trajectory $\tau = (s_0, a_0, \ldots, s_T)$ follows the distribution $\tau \sim p(\tau) = p(s_0)\prod_{t=0}^{T-1} \pi(a_t|s_t) p(s_{t+1}|s_t,a_t)$.
Within the options framework \citep{hrl_sutton1999}, we formulate the unsupervised skill discovery problem as learning a latent-conditioned skill policy $\pi(a_t|s_t,z)$ where $z \in \mathcal{Z}$ represents the \emph{skill latent}.
We consider continuous skill latents $z \in \mathbb{R}^d$.
$h(\cdot)$ and $I(\cdot;\cdot)$ denote differential entropy and mutual information, respectively.

We introduce existing skill discovery methods in two groups: \emph{latent-first} and \emph{trajectory-first} methods.

\begin{figure*}[t!]
  \begin{subfigure}[t]{0.29\linewidth}
    \centering
    \includegraphics[width=1.0\columnwidth]{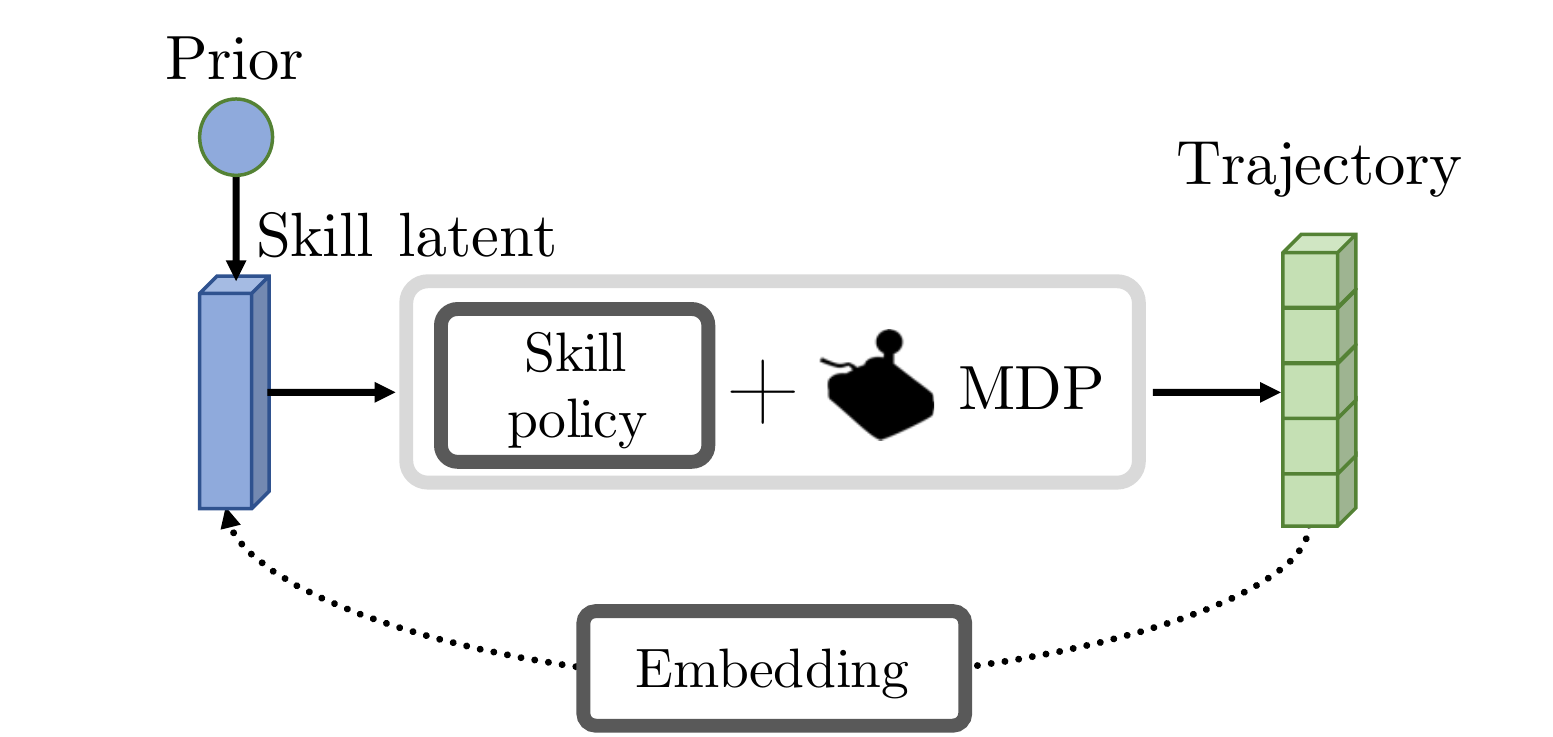}
    \caption{Latent-first methods.}
    \label{fig:model_latentfirst}
  \end{subfigure}
  \begin{subfigure}[t]{0.35\linewidth}
    \centering
    \includegraphics[width=1.0\columnwidth]{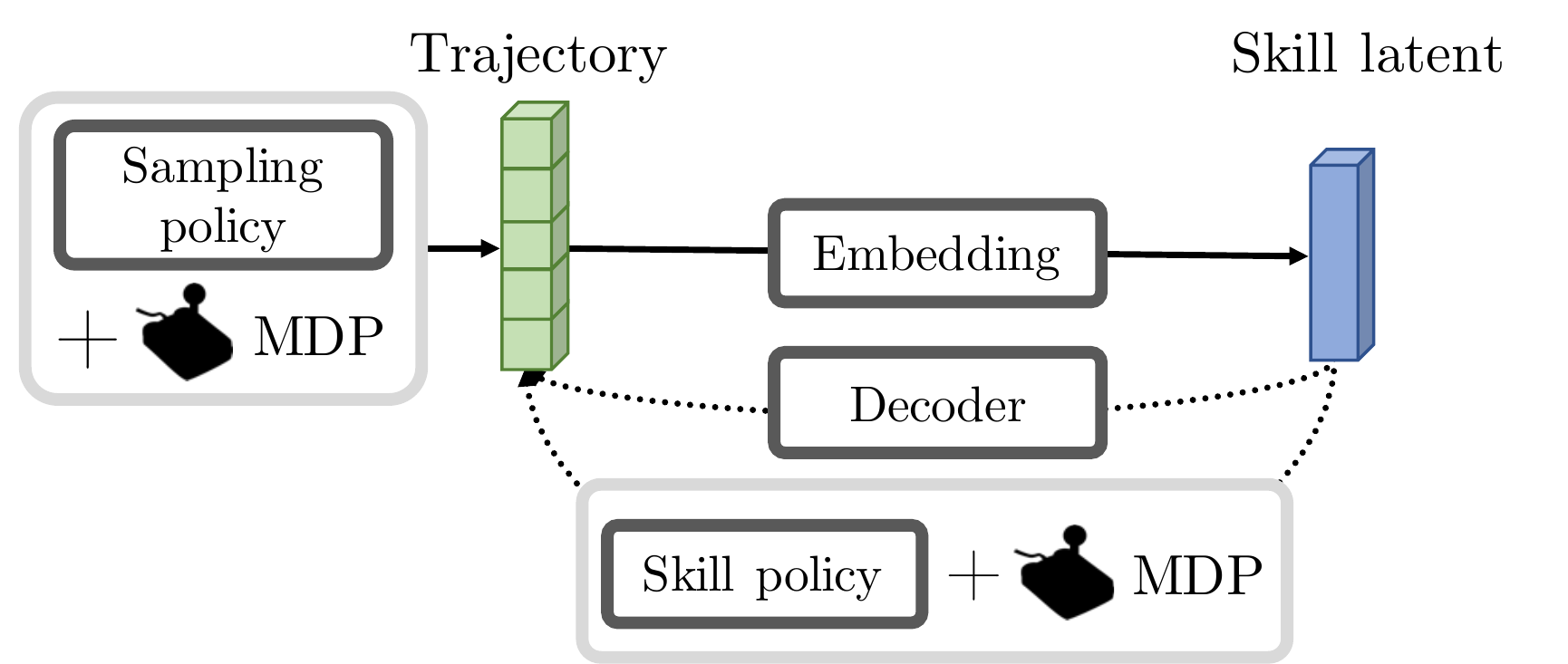}
    \caption{Trajectory-first methods.}
    \label{fig:model_trajfirst}
  \end{subfigure}
  \begin{subfigure}[t]{0.35\linewidth}
    \centering
    \includegraphics[width=1.0\columnwidth]{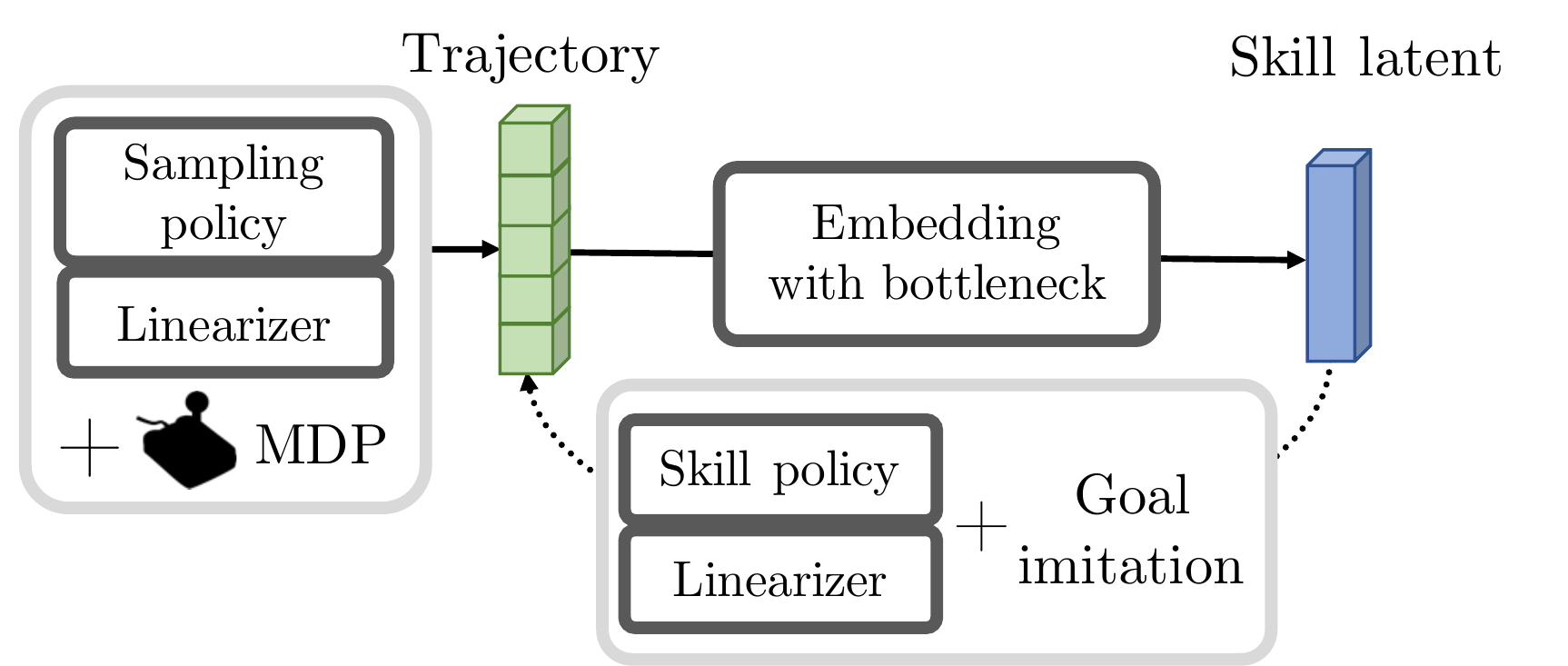}
    \caption{IBOL.}
    \label{fig:model_ibol}
  \end{subfigure}
  \caption{
    Architecture overview of (a) latent-first methods, (b) trajectory-first methods and (c) IBOL.
  }
\end{figure*}

\textbf{Latent-first methods}.
Skill discovery methods in this category, such as VIC \cite{vic_gregor2016}, DIAYN \cite{diayn_eysenbach2019}, VALOR \cite{valor_achiam2018}, DADS \cite{dads_sharma2020} and HIDIO \cite{hidio_zhang2021}, first sample a skill latent $z$ and then trajectories conditioned on $z$,
as illustrated in Figure \ref{fig:model_latentfirst}.
They aim to increase $I(Z; S)$, the mutual information between the skill latent and state variables.
VALOR \cite{valor_achiam2018}, which incorporates VIC and DIAYN as its special forms \cite{valor_achiam2018},
 optimizes a lower bound of the identity $I(Z;S) = h(Z) - h(Z|S)$. Its objective is to maximize
\begin{align}
  \mathbb{E}_{z \sim p(z)} \Bigg[ \mathbb{E}_{\tau \sim p(\tau | z)} [ \log p_D(z | s_{0:T})]
  + \beta {\cdot} \sum_{t=0}^{T-1} h(A_t) \Bigg],
  \nonumber
\end{align}
where $A_t$ is the action variable that follows $\pi(a_t|s_t,z)$, $\beta$ is the entropy coefficient, $p(z)$ is the prior distribution over $z$,
and $p_D(z|s_{0:T})$ is a trainable decoder that reconstructs the original $z$ given $s_{0:T}$.
\citet{valor_achiam2018} show that this objective has an equivalency to $\beta$-VAE \cite{betavae_higgins2017}
with the structure of $z$ (input) $\to$ $\tau$ (latent) $\to$ $z$ (reconstruction).
However, this objective does not take advantage of the benefits that the VAE formulations can provide,
such as the theoretical connection to more disentangled and interpretable $z$  \cite{infodropout_achille2018,emergenceof_achille2018,isolating_chen2018}.
DADS \cite{dads_sharma2020} optimizes the other identity $I(Z;S) = h(S) - h(S|Z)$,
using a skill dynamics model $q(s_{t+1}|s_t, z)$ that predicts the next state conditioned on $z$.
While the learned dynamics model enables model-based planning, %
it lacks an explicit mapping from states to skill latents, and thus hardly obtains disentangled skill latents $z$.

\textbf{Trajectory-first methods}.
Another group of methods first samples trajectories and then encodes them into skill latents using the variational autoencoder (VAE) \cite{vae_kingma2014},
as visualized in Figure \ref{fig:model_trajfirst}.
This category includes SeCTAR \cite{sectar_coreyes2018}, EDL \cite{edl_campos2020} and OPAL \cite{opal_ajay2020}.
SeCTAR and EDL have separate objectives for their \emph{exploratory policies} to sample diverse trajectories by maximizing $h(p(\tau))$ or $h(S)$.
OPAL assumes an offline RL setting where a fixed set of trajectories is priorly given.
While these methods employ the VAE with the usual direction of $\tau \to z \to \tau$ (EDL has $s \to z \to s$) that encourages disentangled representations,
they have a limitation that the exploratory policy \emph{only} maximizes the diversity of trajectories.
On the contrary, our IBOL method, which also falls into this category,
\emph{jointly} maximizes both the diversity and discriminability of trajectories (\Cref{sec:training}),
which leads to a significant improvement in performance (\Cref{sec:exp}).

Finally, all of the prior works learn the skill policies on top of raw environment dynamics.
Although dealing with raw dynamics is not highly demanding in simple environments, it could hinder the skill learning in environments with complex dynamics 
such as Ant and Humanoid from MuJoCo \cite{mujoco_todorov2012}. %
IBOL solves the issue by \emph{linearizing} the environment dynamics ahead of skill discovery so that it can acquire diverse skills by reaching different states more easily in the simplified environment dynamics. %
Furthermore, we find that the linearization benefits other existing skill discovery methods too (\Cref{sec:exp}).

\section{Information Bottleneck Option Learning (IBOL)}
\label{sec:method}

\begin{figure*}
\begin{minipage}[t]{0.46\textwidth}
\begin{algorithm}[H]
  \caption{(Phase 1) Training Linearizer}
  \label{alg:phase1}
  \begin{algorithmic}
    \STATE Initialize linearizer $\pienv$.
    \WHILE{\textit{not converged}}
      \FOR{$i=1$ {\bfseries to} $n$}
        \STATE Sample goals $(g_0^{(i)}, g_\ell^{(i)}, g_{2\ell}^{(i)}, \ldots)$.
        \STATE Sample trajectory using $\pienv$ and goals.
        \STATE Compute linearizer reward $R^{\text{lin}}$ using \Cref{eq:lin}.
        \STATE Add trajectory to replay buffer.
      \ENDFOR
      \STATE Update $\pienv$ using collected samples from replay buffer with SAC \cite{sac_haarnoja2018}.
    \ENDWHILE
  \end{algorithmic}
\end{algorithm}
\end{minipage}
\hfill
\begin{minipage}[t]{0.52\textwidth}
\begin{algorithm}[H]
  \caption{(Phase 2) Skill Discovery}
  \label{alg:phase2}
  \begin{algorithmic}
    \STATE Load pre-trained linearizer $\pienv$.
    \STATE Initialize sampling policy $\pisample$, trajectory encoder $\trajencoder$, skill policy $\pioption$.
    \WHILE{\textit{not converged}}
      \FOR{$i=1$ {\bfseries to} $n$}
        \STATE Sample trajectory using $\pisample$\ on top of $\pienv$.
      \ENDFOR
      \STATE Compute objective from \Cref{eq:obj_final}.
      \STATE Compute its gradient w.r.t. $\phi$, $\optionsub$.
      \STATE Compute its policy gradient w.r.t. $\samplesub$.
      \STATE Jointly update $\pisample$, $\trajencoder$, $\pioption$ with gradients. %
    \ENDWHILE
  \end{algorithmic}
\end{algorithm}
\end{minipage}
\end{figure*}

We decompose the skill discovery problem into two separate phases.
Firstly, IBOL trains the \emph{linearizer} that lifts the burden from the skill discovery algorithm to generate diverse states and trajectories under complex environment dynamics (\Cref{sec:flattening}).
Secondly, on top of the pre-trained linearizer, IBOL learns to map trajectories into the continuous skill latent space, %
with the information bottleneck principle \cite{ib_tishby2000,vib_alemi2016} (\Cref{sec:skill_discovery}). %
Figure \ref{fig:model_ibol} provides the conceptual illustration of IBOL.
\Cref{alg:phase1} overviews the training of the linearizer in the first phase and \Cref{alg:phase2} describes the skill discovery process in the second.

\subsection{Linearization of Environments}
\label{sec:flattening}

The \emph{linearizer} $\pienv$ is a pre-trained low-level policy that aims to ``linearize'' the environment dynamics.
It takes as input \emph{goals} produced by IBOL's policies for skill discovery (will be discussed in \Cref{sec:skill_discovery}),
and translates them into raw actions in the direction of a given goal while interacting with the environment. %
We define the linearizer $\pienv (a_t|s_t,g_t)$ as a goal-conditioned policy \cite{uvfa_schaul2015},
which takes both a state $s_t \in \mathcal{S}$ and a goal $g_t \in \mathcal{G}$ as input and outputs a probability distribution over actions $a_t \in \mathcal{A}$.
The goal space $\mathcal{G}$ is defined as $\mathcal{G} = [-1, 1]^{dim(\mathcal{S})}$,
which has the same dimensionality as the state space (up to $47$ in our experiments).
Each goal dimension provides a signal for the direction in the corresponding state dimension.
We assume that a goal $g_t \in \mathcal{G}$ is given at every $\ell$-th time step such that $t \equiv 0 \Mod{\ell}$ (called a \emph{macro} time step),
and otherwise kept fixed, \ie $g_t = g_{t-1}$ for $t \not\equiv 0 \Mod{\ell}$.

We sample goals $(g_0, g_\ell, g_{2\ell}, \ldots)$ at the beginning of each roll-out and train the linearizer with a reward function of
\begin{align}
    \rewardenv(s_t, g_t, a_t, s_{t+1}) &= \frac{1}{\ell} (s_{(c + 1) \cdot \ell} - s_{c \cdot \ell})^\top g_t, \label{eq:lin}
\end{align}
where $c = \left \lfloor \frac {t} {\ell} \right \rfloor$.
It corresponds to the inner product of the goal $g_t$ and
the state difference between macro time steps: $(s_{(c + 1) \cdot \ell} - s_{c \cdot \ell})$.
Intuitively, each goal dimension value (ranging from $-1$ to $+1$) indicates the desired direction and the degree of change in the corresponding state dimension.

The inner product in the reward function has several advantages for skill discovery compared to the Euclidean distance in prior approaches \citep{hiro_nachum2018,nearoptimal_nachum2019}.
First, unlike the Euclidean distance that needs to specify the valid range of each state dimension, the inner product only takes care of directions in the state space.
Thus, training of the linearizer requires no additional supervision on specifying valid goal spaces or state ranges.
Second, by setting some dimensions of a goal to be (near-)zero values,
we can ignore changes in the corresponding state dimensions,
which is not achievable with the Euclidean distance.
This enables IBOL's policies for skill discovery to ignore nuisance dimensions without manually specifying them (\Cref{sec:skill_discovery}).

We find that the linearizer benefits not only IBOL but also other skill discovery methods since it can promote reaching diverse and distant states easier, as shown in Figure \ref{fig:ant_xy}.

\subsection{Skill Discovery with Bottleneck Learning}
\label{sec:skill_discovery}

On top of the pre-trained and fixed linearizer $\pienv$,
we learn policies that produce \emph{goals} and acquire a continuous set of skills that is not only distinguishable and diverse but also disentangled and interpretable.
The linearizer alone is highly limited to discovery abstractive and informative skills,
since  it is trained with the inner product reward function and thus optimized for transitioning to distant states rather than the mapping with the latent space.
Additionally, IBOL can fix possibly imperfect linearization with the linearizer by combining appropriate high-level goals. %
In \Cref{sec:exp}, we will demonstrate that how such limitations of the linearizer can be resolved by the following skill discovery process.

In contrast to previous skill discovery methods that maximize $I(S;Z)$ \citep{vic_gregor2016,diayn_eysenbach2019,valor_achiam2018,dads_sharma2020},
IBOL consists of the following three learnable components based on the information bottleneck:
\begin{enumerate}
    \item The sampling policy $\pisample(g_t|s_t)$ produces diverse and easily mappable trajectories.
    \item The trajectory encoder $\trajencoder(z|s_{0:T})$ encodes the state trajectories into the skill latent space.
    \item The skill policy $\pioption(g_t|s_t,z)$ learns to imitate the skills given their latents.
\end{enumerate}
Note that the sampling and skill policies produce goals $g_t$ instead of raw actions $a$, as they operate on top of the linearizer.
We will first start with the sampling policy $\pisample$ and introduce our IB objective for the trajectory encoder $\trajencoder$.
We then show that it naturally leads to the emergence of the skill policy $\pioption$ as a variational approximation to the sampling policy $\pisample$.

\textbf{IBOL's objective}. 
Assuming trajectories generated by the sampling policy, $\{\tau^{(1)}, \tau^{(2)}, \ldots, \tau^{(n)}\}$,
our objective is to embed the state trajectories $\{s_{0:T}^{(1)}, \ldots, s_{0:T}^{(n)}\}$ into the skill latent space $\mathcal{Z}$.
We encode the \emph{state} trajectory $s_{0:T}$, not the \emph{whole} trajectory $\tau$, because an outside observer can only see the agent's state, not its underlying actions or goals.
However, the encoded skill latent $z$ should contain sufficient information about the underlying goals so that the whole trajectory is reproducible from $z$.
Furthermore, since raw states often contain nuisance information not pertaining to skill discovery,
$z$ is encouraged to minimally contain unnecessary or noisy information in the states irrelevant to the goals.
This leads to the Information Bottleneck objective \citep{ib_tishby2000,vib_alemi2016} over the structure of $S_{0:T}$ (input) $\to$ $Z$ (latent) $\to$ $G_{0:T-1}$ (target).

Formally, let us first define the \emph{sampling policy} parameterized by $\samplesub$ as $\pisample(g_t|s_t) \colon \mathcal{S} \to \mathcal{P}(\mathcal{G})$,
which maps a state to a probability distribution over goals. %
A trajectory $\tau = (s_0, g_0, s_1, \ldots, g_{T-1}, s_T)$ obtained by the sampling policy
is acquired from the distribution
$\tau \sim p_{\samplesub}(\tau) = p(s_0)\prod_{t=0}^{T-1}\pisample(g_t|s_t)p(s_{t+1}|s_t,g_t)$.
Under the distribution $p_{\samplesub}(\tau)$, let $S_t$ be a random variable corresponding to $s_t$ and $G_t$ be a random variable for $g_t$.
We define the \emph{trajectory encoder} parameterized by $\phi$ as $\trajencoder(z|s_{0:T}) \colon \mathcal{S}^{T + 1} \to \mathcal{P}(\mathcal{Z})$
that maps a state trajectory to a probability distribution over skill latents $z$ in the skill space $\mathcal{Z}$.
Let $Z$ be a random variable for $z$.

We formulate our IB objective as follows.
First, given $S_t$, the skill latent $Z$ should be informative about the goal $G_t$ that the sampling policy has produced, which leads to the \emph{prediction term} $I(Z; G_t | S_t)$.
Second, $Z$ should be penalized for preserving information about the state trajectory but unrelated to the goals,
which corresponds to the \emph{compression term} $I(Z; S_{0:T})$.
Summing these up, we obtain the following objective:
\begin{align}
  &\maximize~ \mathbb{E}_{t} [ I(Z; G_t | S_t) - \beta \cdot I(Z; S_{0:T}) ],
\end{align}
where $\mathbb{E}_t$ is the expectation over $\{0, 1, \ldots, T-1\}$, and $\beta$ is a constant that controls the weight of the compression term.

\textbf{Lower bound optimization}.  Since the objective is practically intractable, 
we derive its lower bound \citep{vib_alemi2016} as follows (see \Cref{sec:lowerbound_derivation} for the full derivation):
\begin{align}
  & \mathbb{E}_{t} [ I(Z; G_t | S_t) - \beta \cdot I(Z; S_{0:T}) ] \label{eq:obj_ib} \\
  &= \mathbb{E}_{\substack{\tau \sim p_{\samplesub}(\tau), t, \\ z \sim \trajencoder(z|s_{0:T})} } \bigg[ \log \frac{p_{\samplesub}(g_t | s_t, z)}{\pisample(g_t | s_t)}
  - \beta \log \frac{\trajencoder(z | s_{0:T})}{\trajencoder(z)} \bigg] \nonumber \\
  & \geq \mathbb{E}_{\tau \sim p_{\samplesub}(\tau)}\bigg[\mathbb{E}_{z \sim \trajencoder(z|s_{0:T}), t} \Big[ \log \pioption(g_t | s_t, z) \label{eq:obj_lowerbound2} \\
  & \hspace{25pt} - \log \pisample(g_t | s_t) \Big]
  - \beta \hspace{-1pt}\cdot \hspace{-1pt} D_{\text{KL}} (\trajencoder(Z|s_{0:T}) \| r(Z)) \bigg], \nonumber
\end{align}
where $D_{\text{KL}}$ denotes the Kullback-Leibler (KL) divergence.
Here we use two variational approximations:
the \emph{skill policy}'s output distribution $\pioption(g_t|s_t,z)$ is a variational approximation of $p_{\samplesub}(g_t|s_t,z)$ and
$r(z)$ is that of the marginal distribution $\trajencoder(z)$.
In \Cref{eq:obj_lowerbound2}, the first term $\log \pioption (g_t|s_t,z)$ makes the skill policy $\pioption(g_t|s_t,z)$ imitate the sampling policy's output given the skill latent $z$; thus we call this the \emph{imitation term}.
The third term $-\beta D_{\text{KL}}(\trajencoder(Z|s_{0:T}) \| r(Z))$ is the \emph{compression term} that forces the output distributions of the trajectory encoder to be close to $r(z)$.
We will revisit the second term $-\log \pisample(g_t|s_t)$ later.

We fix $r(z)$ to $\mathcal{N}(0, I)$ as in \citet{vib_alemi2016} for the following reasons.
First, it enables us to analytically compute the KL divergence.
Second, more importantly, it induces disentanglement between the dimensions of $z$ \cite{infodropout_achille2018,emergenceof_achille2018,isolating_chen2018}.
Disentangled representations lead to more interpretable skills with respect to their skill latents $z$. %
In \Cref{sec:disentanglement}, we provide further details on how the compression term encourages the disentanglement of skill latent dimensions.

It is worth noting that the first and third terms in \Cref{eq:obj_lowerbound2} are related to the $\beta$-VAE objective \citep{vae_kingma2014,betavae_higgins2017,vib_alemi2016} and previous skill discovery methods that use trajectory VAEs \citep{sectar_coreyes2018,opal_ajay2020}.
The first and the third term correspond to the reconstruction loss and the KL divergence loss in $\beta$-VAEs, respectively.
One important difference is that we reconstruct not the original state trajectories but their underlying goals.
It eliminates the need for state decoders or sampling with the skill policy during training.

\subsection{Training}
\label{sec:training}

The trajectory encoder and the skill policy can be trained using the reparameterization trick as in VAEs \citep{vae_kingma2014}; we optimize those two terms in \Cref{eq:obj_lowerbound2} with respect to their parameters, $\optionsub$ and $\phi$. 
Note that the skill policy does not interact with the environment during training and 
the second term $-\log \pisample(g_t|s_t)$ is independent of these parameters.

The sampling policy $\pisample(g_t|s_t)$ can be trained with the same objective of \Cref{eq:obj_lowerbound2}.
This is the key difference with prior trajectory-first methods that employ similar VAE architectures \citep{edl_campos2020,sectar_coreyes2018,opal_ajay2020} (\Cref{sec:prior_works}).
They either have a separate objective for training their sampling policies \citep{edl_campos2020,sectar_coreyes2018}
or assume the offline RL setting \citep{opal_ajay2020}. In contrast, we jointly train all the components with the same objective.

There are several merits of using the same objective.
First, the second term $-\log \pisample(g_t|s_t)$, referred to as the \emph{entropy term}, encourages the sampling policy to produce diverse trajectories.
In deterministic environments, maximizing this term is equivalent to maximizing the entropy of whole trajectories, as $h(p_{\samplesub}(\tau)) = T \cdot \mathbb{E}_{\tau \sim p_{\samplesub}(\tau),t} [-\log \pisample(g_t|s_t)] + (\text{const})$.
Note that this entropy term often remains constant in IB literature \citep{vib_alemi2016}, assuming that the training data (\eg images) are given,
whereas we can diversify the ``training data'' too.
Second, optimizing the whole \Cref{eq:obj_lowerbound2} makes the sampling policy generate trajectories that are not only diverse but also \emph{easily encoded into the skill space for the trajectory encoder and skill policy} thanks to the first and third terms,
which helps the learning of the two components as well.
This is not achievable when the sampling policy is trained with a diversity maximizing objective only.
In \Cref{sec:exp}, we will demonstrate how taking into account both diversity and encodability leads to a huge difference in performance, comparing with baselines without such consideration.

\textbf{Practical training}.
Since the expectation in \Cref{eq:obj_lowerbound2} involves the sampling policy's roll-outs in the environment,
we optimize the sampling policy via the policy gradient method.
However, there exists one practical difficulty when training IBOL.
Since the sampling policy $\pisample(g_t|s_t)$ lacks a variable about the context (\eg $z$) compared to the skill policy $\pioption(g_t|s_t,z)$,
$\pisample$ is less expressive than $\pioption$, which could end up with a suboptimal convergence.
To solve this issue, we introduce a new context parameter $u \in \mathcal{U}$ with its prior $p(u)$ to the sampling policy,
redefining it as $\pisample(g_t|s_t, u): \mathcal{S} \times \mathcal{U} \to \mathcal{P}(\mathcal{G})$.
The new parameter $u$ for $\pisample$ plays a similar role to the skill latent $z$ for $\pioption$.
We also fix $p(u)=\mathcal{N}(0, I)$ as in $r(z)$.
To obtain roll-outs from the sampling policy, we first sample $u$ from its prior, and then keep sampling goals with the fixed $u$.

Given that $r(z)$ and $p(u)$ are identical,
we additionally include an auxiliary term
$\mathbb{E}_{u \sim p(u), \tau \sim p_{\samplesub}(\tau | u)} [ \lambda \cdot \trajencoder(u | s_{0:T}) ]$ 
to further stabilize the training.
This term guides the output of the trajectory encoder $\trajencoder$ to $u$, which is from $p(u)=r(z)$,
operating compatibly with the compression term.

Finally, with the revised sampling policy, 
we approximate the second term in \Cref{eq:obj_lowerbound2} as done in DADS \cite{dads_sharma2020}:
$\pisample(g_t | s_t) = \int_{u} \pisample(g_t | s_t, u) p(u | s_t) du \approx \int_{u} \pisample(g_t | s_t, u) p(u) du \approx \frac{1}{L} \sum_{i=1}^L \pisample(g_t | s_t, u_i)$ for $u_i \overset{\text{i.i.d.}}{\sim} p(u)$,
where $L$ is the number of samples from the prior.
Therefore, the final objective of our method is
\begin{align}
  \mathbb{E}_{\substack{u \sim p(u), \\ \tau \sim p_{\samplesub}(\tau | u)}} &\bigg[
      \mathbb{E}_{\substack{z \sim \trajencoder(z|s_{0:T}), t \\ u_i \overset{\text{i.i.d.}}{\sim} p(u)}} \big[ J^{\text{P}} \big] - \beta {\cdot} J^{\text{C}} + \lambda {\cdot} \trajencoder(u | s_{0:T})
  \bigg] \nonumber \\
  \text{where~} J^{\text{P}} = {}& \log \pioption(g_t | s_t, z) - \log \bigg(\frac{1}{L} \sum_{i=1}^L \pisample(g_t | s_t, u_i)\bigg) \nonumber \\
  J^{\text{C}} = {}& D_{\text{KL}}(\trajencoder(Z|s_{0:T}) \| r(Z)) \label{eq:obj_final}.
\end{align}

\begin{figure*}[t!]
  \centering

  \hspace{0.7cm}
  \begin{subfigure}[t]{0.18\linewidth}
    \includegraphics[width=1.0\linewidth]{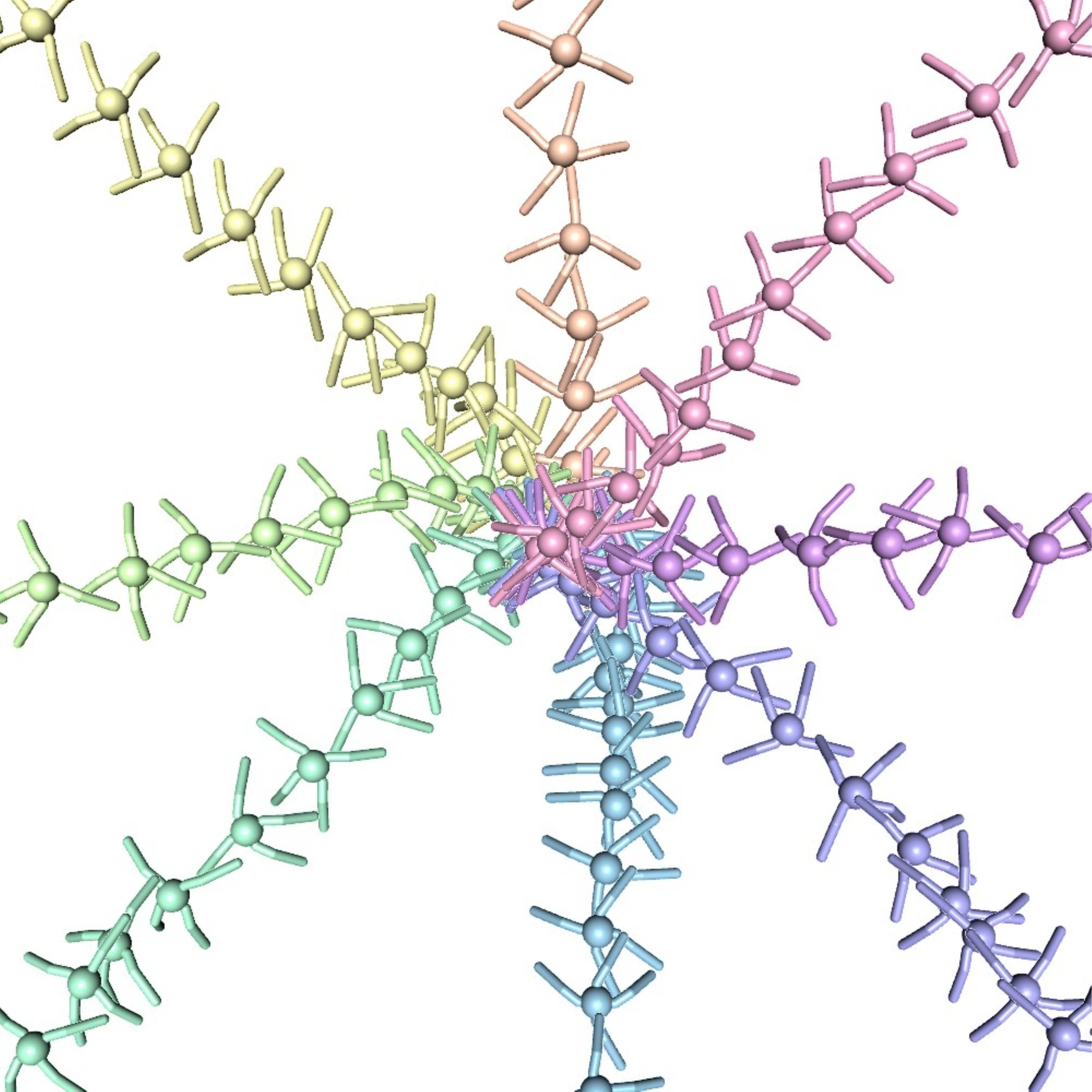}
    \caption{Ant}
  \end{subfigure}
  \hfill
  \begin{subfigure}[t]{0.20\linewidth}
    \includegraphics[width=1.0\linewidth]{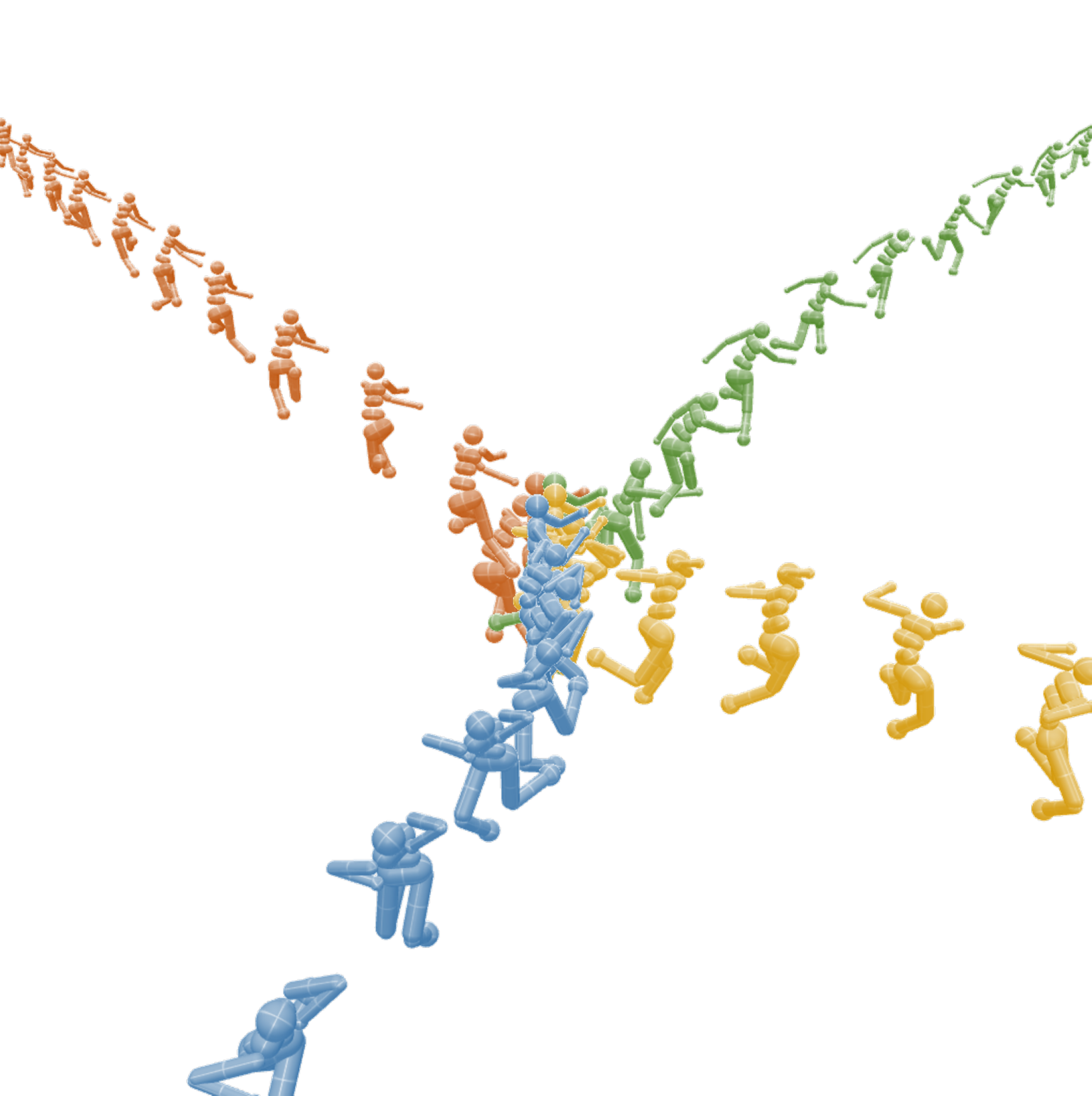}
    \caption{Humanoid}
  \end{subfigure}
  \hfill
  \begin{subfigure}[t]{0.27\linewidth}
    \includegraphics[width=1.0\linewidth]{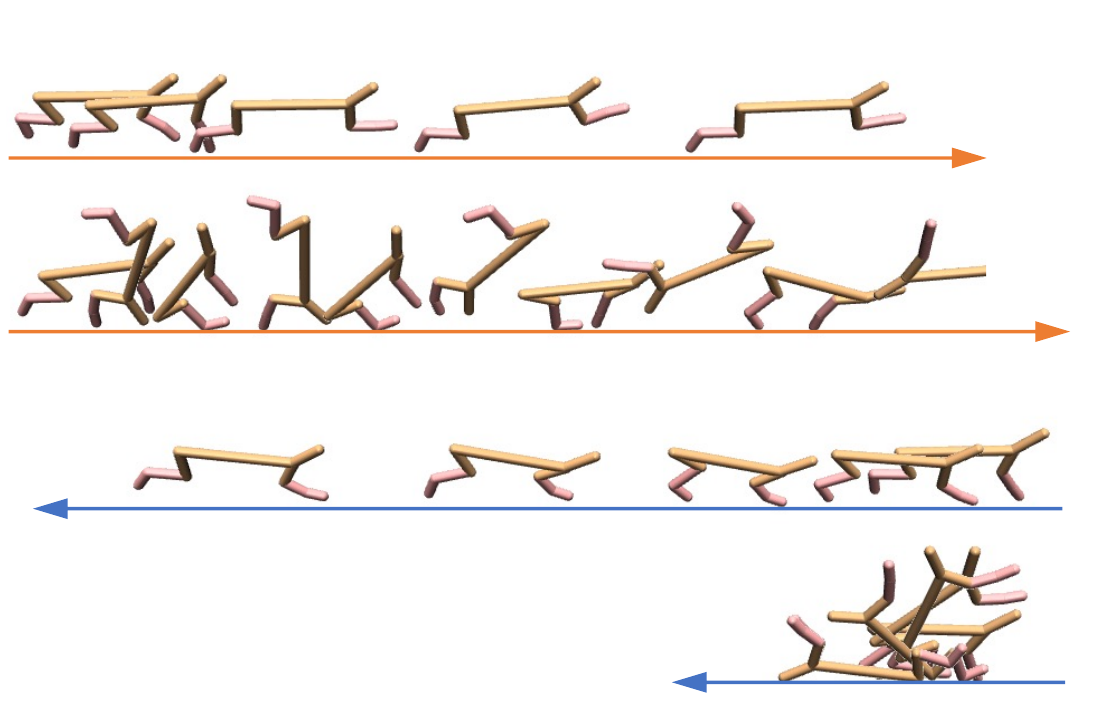}
    \caption{HalfCheetah}
  \end{subfigure}
  \hfill
  \begin{subfigure}[t]{0.27\linewidth}
    \includegraphics[width=1.0\linewidth]{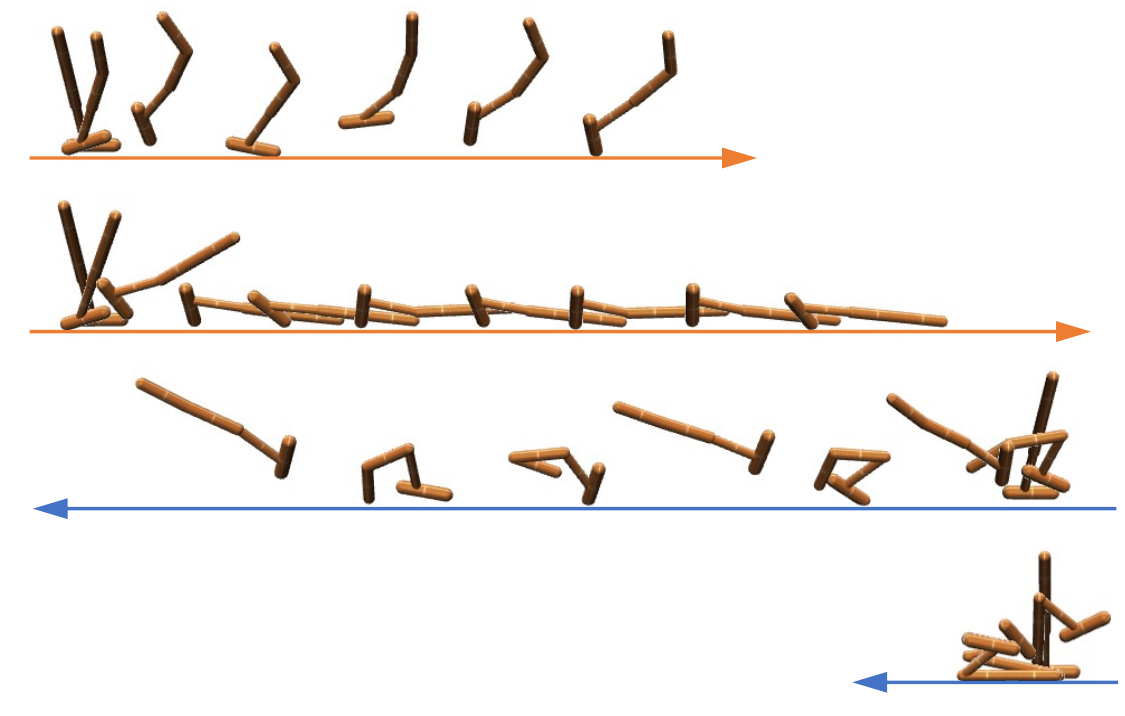}
    \caption{Hopper}
  \end{subfigure}
  \hspace{0.7cm}

  \caption{
    Examples of rendered scenes illustrating the skills that IBOL discovers with no rewards in MuJoCo environments.
    (a) Ant moving in various directions.
    (b) Humanoid running in different directions.
    (c) (Top to Bottom) HalfCheetah running forward, rolling forward, running backward and flipping backward.
    (d) (Top to Bottom) Hopper hopping forward, crawling forward, jumping backward and flipping backward.
  }
  \label{fig:mujoco_traj}
\end{figure*}

\section{Experiments}
\label{sec:exp}

We compare our IBOL with other state-of-the-art methods in multiple aspects.
First, we visualize the learned skills with the trajectory plots and the rendered scenes from environments (\Cref{sec:qual_results}).
Second, we quantitatively evaluate the skill discovery methods in terms of multiple information-theoretic metrics (\Cref{sec:info_eval}).
Third, we evaluate the trained policies on the downstream tasks with different configurations (\Cref{sec:downstream}).
Finally, we present additional behaviors of IBOL in the absence of the locomotion signals and with the distorted goal space (\Cref{sec:additional_obs}).
Please refer to Appendix for additional results.

\textbf{Experiment setup and baselines}.
We experiment with MuJoCo environments \cite{mujoco_todorov2012} %
for multiple tasks: 
Ant, HalfCheetah, Hopper and Humanoid from OpenAI Gym \cite{openaigym_brockman2016} 
with the setups by \citet{dads_sharma2020} 
and D'Kitty from ROBEL \cite{robel_ahn2020} adopting the configurations by \citet{sharma2020_emergent}.
We use D'Kitty with the random dynamics setting; 
in each episode, multiple properties of the environment, such as its joint dynamics, friction and height field, are randomized, 
which provides an additional challenge to agents.
We mainly compare our method with recent information-theoretic unsupervised skill discovery methods, VALOR \cite{valor_achiam2018}, DIAYN \cite{diayn_eysenbach2019} and DADS \cite{dads_sharma2020}.
Since IBOL operates on top of the linearized environments, we also consider a variant of each algorithm that uses the same linearizer,
denoted with the suffix `-L' (\eg VALOR-L). %
In Ant experiments, we use the suffix `-XY' to refer to the methods with the \textit{$x$-$y$ prior} \cite{dads_sharma2020}, which 
forces them to focus exclusively on the locomotion skills by restricting the observation space of the trajectory encoder (or the skill dynamics model in DADS) to the $x$-$y$ coordinates.

\textbf{Implementation}.
For experiments, we use pre-trained linearizers with two different random seeds on each environment.
When training the linearizers, we sample a goal $g$ at the beginning of each roll-out and fix it within that episode to learn consistent behaviors, as in SNN4HRL \cite{snn4hrl_florensa2016}.
We consider continuous priors for skill discovery methods. Especially, we use the standard normal distribution for $p(u)$ and $r(z)$ in IBOL and for $p(z)$ in other methods.
Further details are described in \Cref{sec:exp_details}. %

\subsection{Visualization of Learned Skills}
\label{sec:qual_results}

Figure \ref{fig:mujoco_traj} shows that IBOL, with no extrinsic rewards, discovers diverse locomotion skills for Ant and Humanoid and multiple skills with various speeds and poses in both directions for HalfCheetah and Hopper.
We present the discovery of orientation primitives for Ant in \Cref{sec:additional_obs} and
additional results including the videos of the discovered skills at \url{https://vision.snu.ac.kr/projects/ibol}.

Figure \ref{fig:ant_xy} demonstrates that while all the algorithms mainly discover locomotion skills, IBOL discovers visually less entangled primitives with the most diverse directions compared to the latent-first and trajectory-first baselines.
We train IBOL, DIAYN-L, VALOR-L, DADS-L, SeCTAR-L, SeCTAR-L-XY and EDL-L on Ant with the skill latent variables of $d = 2$,
where SeCTAR-L-XY is equipped with the \textit{$x$-$y$ prior} \cite{dads_sharma2020}.
We qualitatively examine their trajectories in the $x$-$y$ plane; since the $x$-$y$ dimensions are interpretable and have a large range of values, they can illustrate the characteristic differences between skill discovery algorithms well.
We also train DIAYN-XY, VALOR-XY and DADS-XY to enforce them to discover skills on the $x$-$y$ plane without the linearizer.
We observe that the linearizer significantly improves not only the diversity of trajectories but also the correspondence between skill latents and trajectories by reducing the burden of making transitions in the $x$-$y$ dimensions.

\subsection{Information-Theoretic Evaluations}
\label{sec:info_eval}

We present the metrics that evaluate the unsupervised skill discovery methods without the need for external tasks.
While the quantities between skill latents $Z$ and state sequences $S_{0:T}$ generated with $\pioption$ are attractive,
the high dimensionality of $S_{0:T}$ makes it a less viable choice.
One workaround  is to examine only the last states $S_T$ instead of the whole sequences,
as $S_T$ still  characterizes skills in environments to some degree.
That is, we can simply estimate $I(Z; S_T)$ instead of $I(Z; S_{0:T})$ to measure how informative $Z$ is.
This can also be viewed as follows:
in $I(Z; S_{0:T}) = I(Z; S_T) + \sum_{i=0}^{T-1} I(Z; S_i | S_{i+1:T})$,
only the first term $I(Z; S_T)$ is taken into account,
as $I(Z; S_i | S_{i+1:T}) = h(Z | S_{i+1:T}) - h(Z | S_{i:T})$
and adding $S_i$ to $S_{i+1:T}$ to the condition would change only little entropy of $Z$. %

We also consider metrics for measuring the disentanglement of $Z$.
We find \citet{disentangletheory_do2019} provide a helpful viewpoint to our evaluation.
They suggest that the concept of disentanglement has three considerations: \textit{informativeness}, \textit{separability} and \textit{interpretability}.
Informativeness denotes how much information each latent dimension contains about the data,
and separability is a concept about \textit{no} information sharing between two latent dimensions on the data.
Interpretability considers the alignment between the ground-truth and learned factors.
Among them, we do not employ the interpretability measure because the lack of supervision in unsupervised skill discovery prevents achieving a high value \cite{challengedisentangle_locatello2019}.
For example, if data points are uniformly distributed in a two-dimensional circle, there can be infinite equally good ways to disentangle the data into two axes.
To measure informativeness and separability, we use the SEPIN$@k$ and WSEPIN metrics \citep{disentangletheory_do2019} evaluated for skill latents and the last states (detailed in \Cref{sec:info_eval_metrics}).

We compare the skill policies trained by IBOL, DIAYN-L, VALOR-L and DADS-L with $d = 2$.
We use the three evaluation metrics, $I(Z; S_T^{\text{(loc)}})$, SEPIN$@1$ and WSEPIN on Ant, HalfCheetah, Hopper and D'Kitty, 
keeping only the state dimensions for the agent's locomotion (\ie $x$-$y$ coordinates for Ant and D'Kitty and $x$ for the rest) denoted as $\text{(loc)}$.
One rationale behind it is that the algorithms on the linearized environments successfully discover the locomotion skills (\eg Figure \ref{fig:ant_xy}). 
The locomotion coordinates are also suitable for assessing skill \textit{discovery}, since these values can vary in large ranges.

\begin{figure*}[t!]
\begin{minipage}{0.58\textwidth}
  \hfill
  \begin{subfigure}[t]{0.2617328072\linewidth}
    \includegraphics[width=1.0\columnwidth]{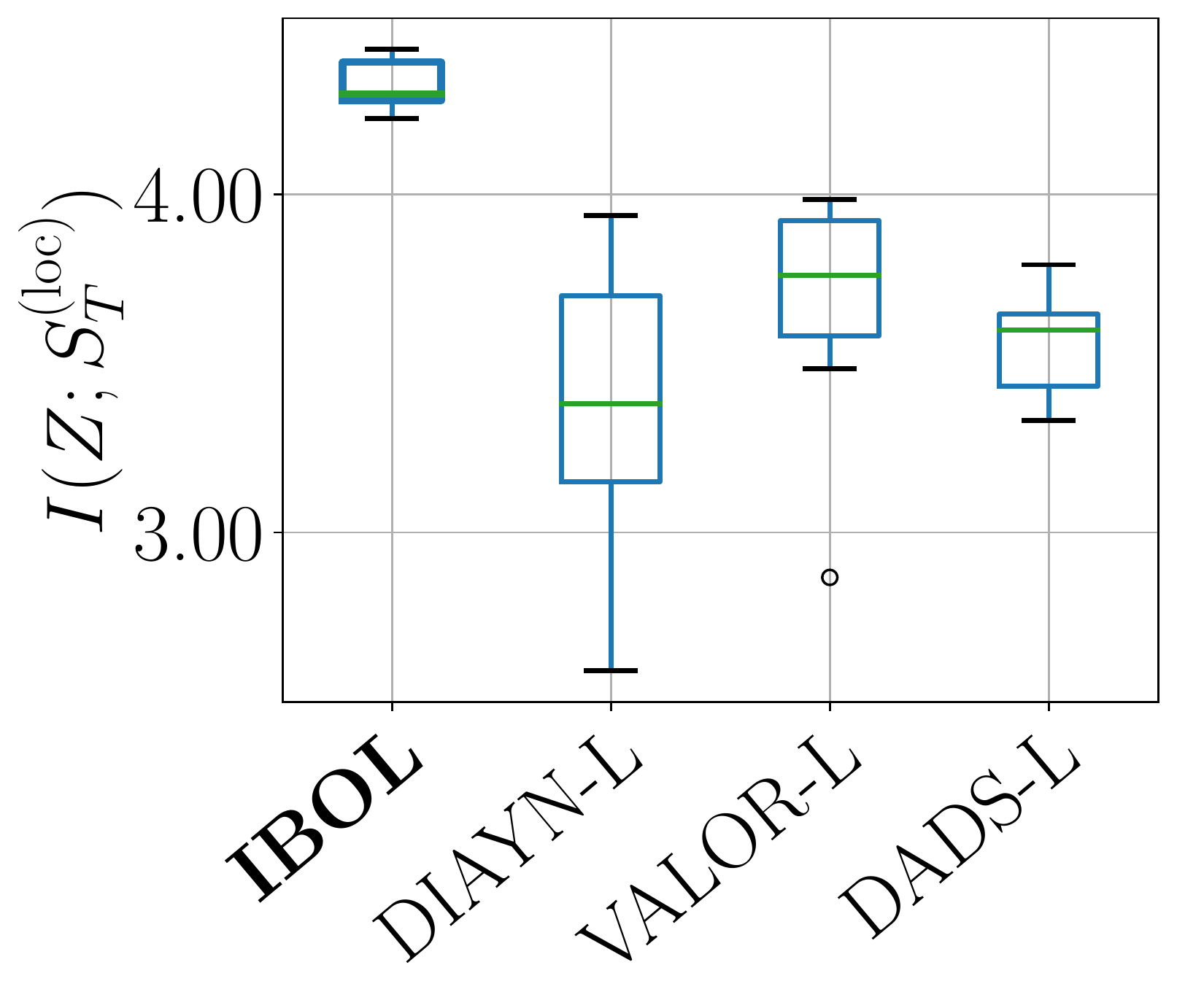}
  \end{subfigure}
  \begin{subfigure}[t]{0.2360890643\linewidth}
    \includegraphics[width=1.0\columnwidth]{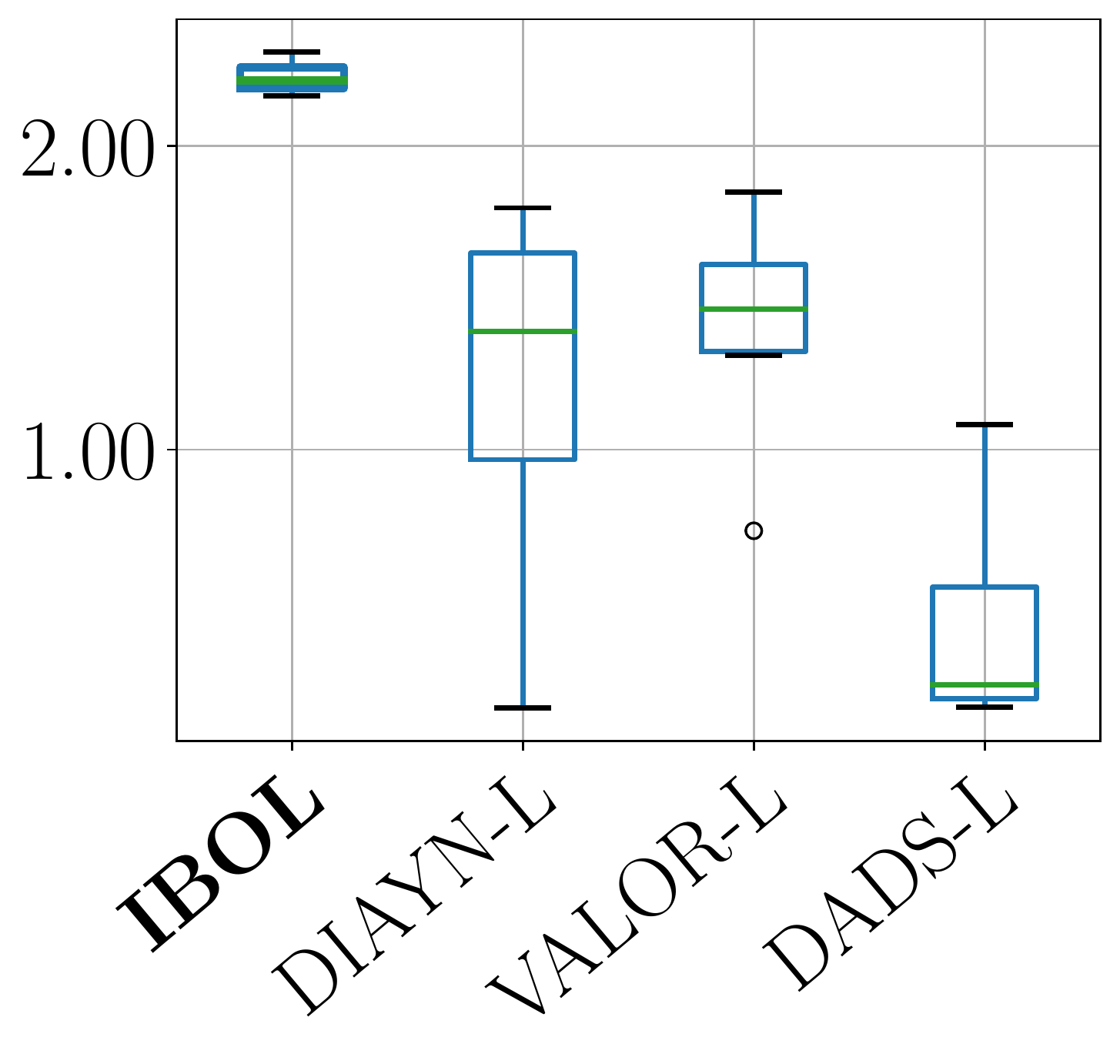}
  \end{subfigure}
  \begin{subfigure}[t]{0.2360890643\linewidth}
    \includegraphics[width=1.0\columnwidth]{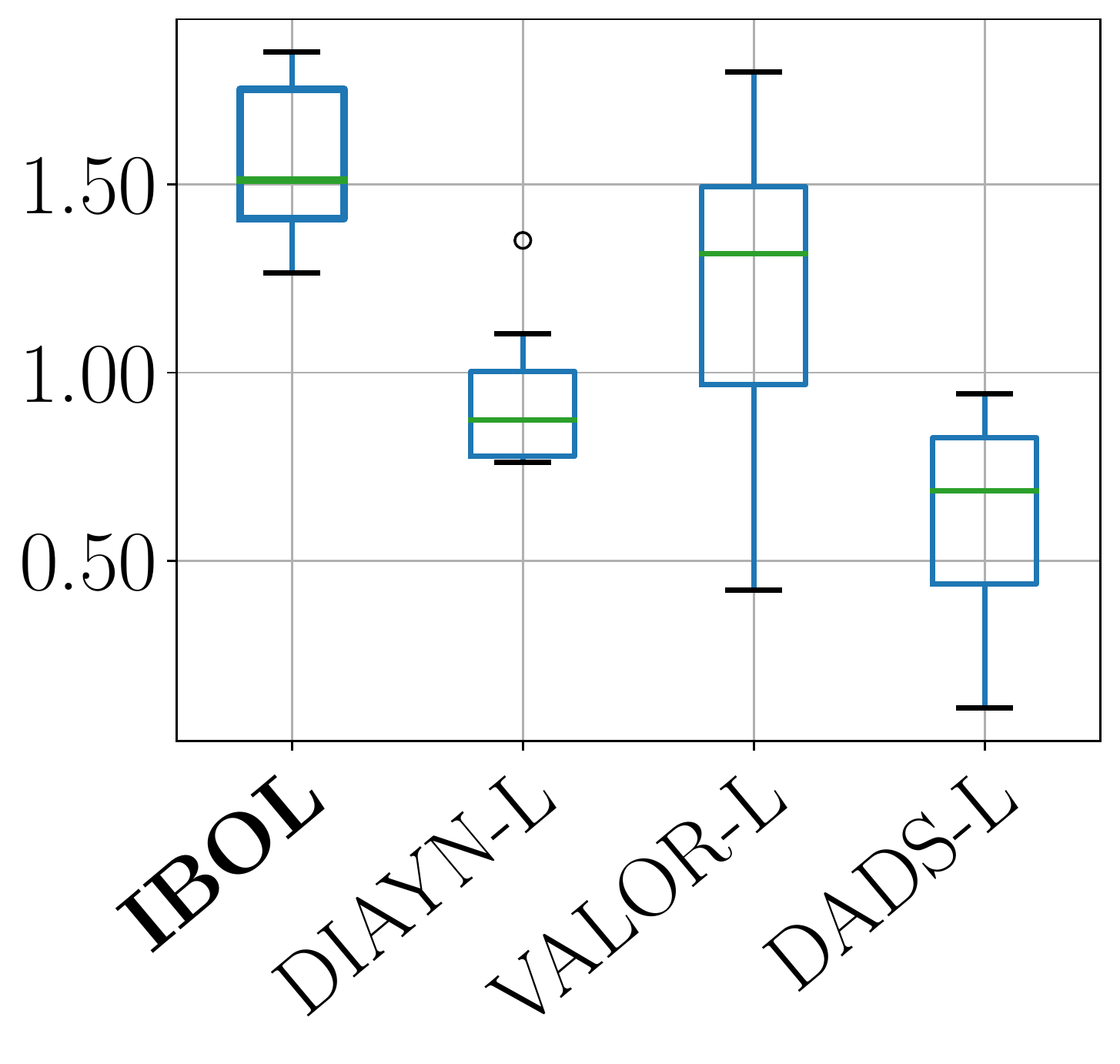}
  \end{subfigure}
  \begin{subfigure}[t]{0.2360890643\linewidth}
    \includegraphics[width=1.0\columnwidth]{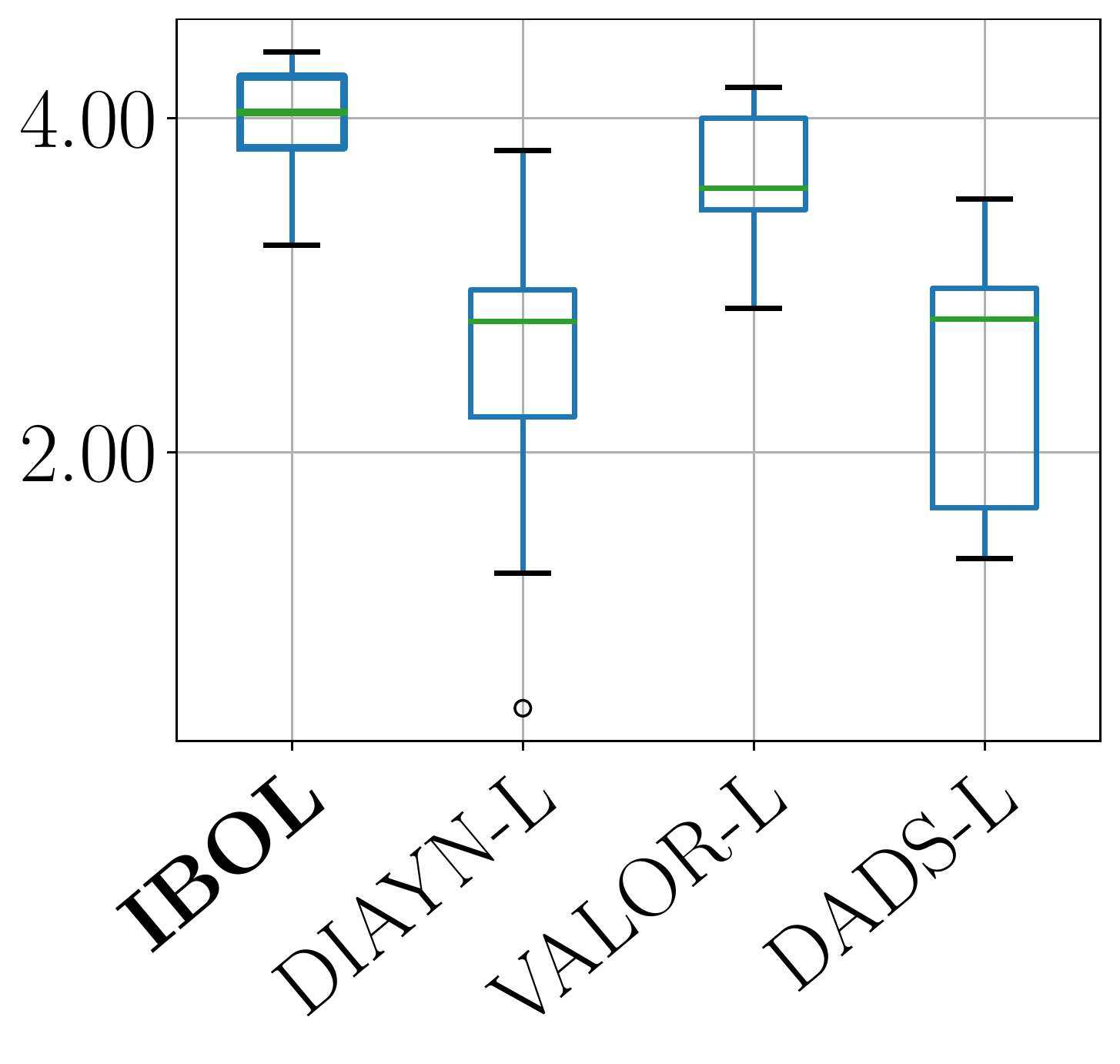}
  \end{subfigure}
  \\

  \vspace{-0.2cm}
  \hfill
  \begin{subfigure}[t]{0.254319053\linewidth}
    \includegraphics[width=1.0\columnwidth]{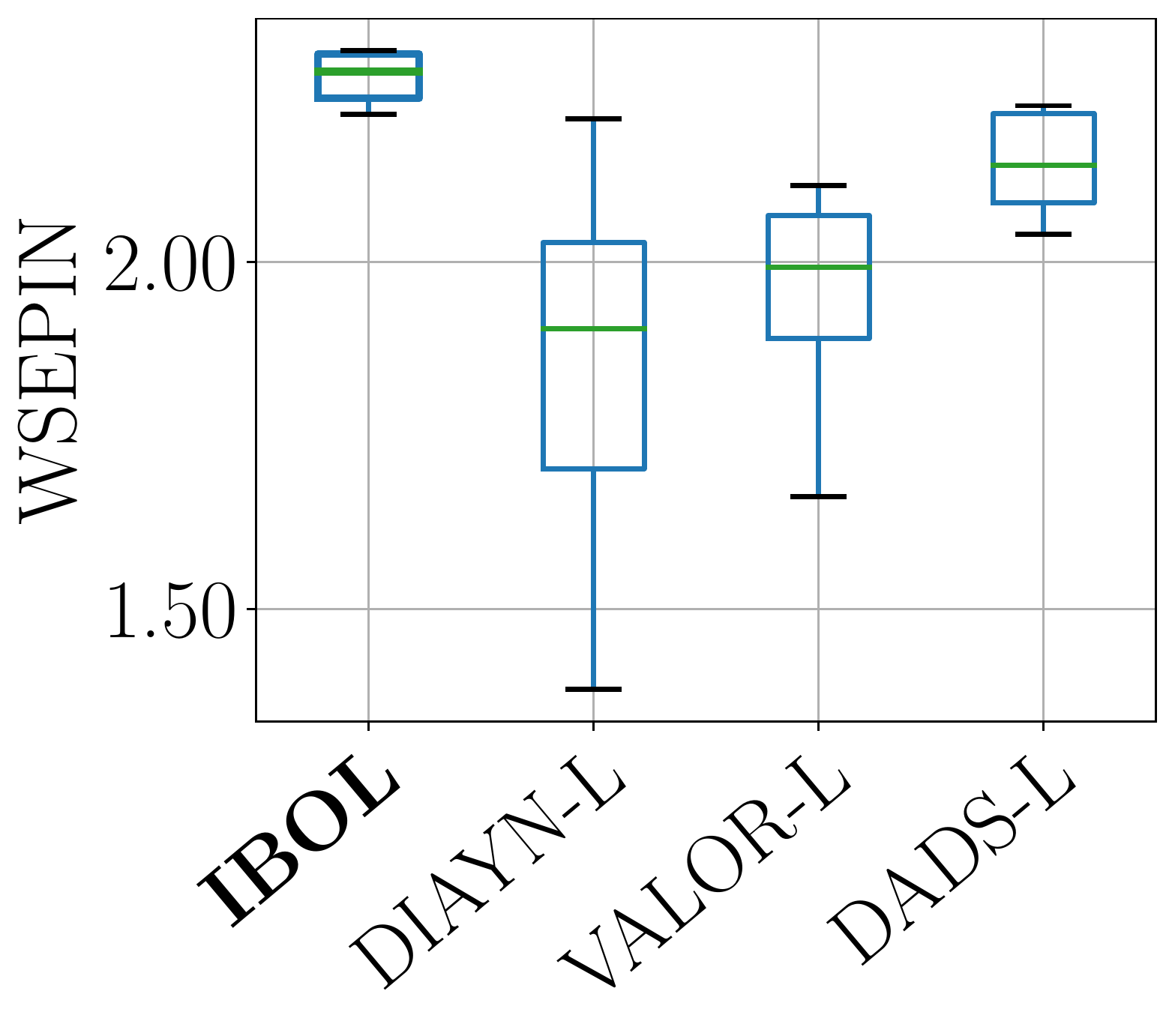}
  \end{subfigure}
  \begin{subfigure}[t]{0.2360890643\linewidth}
    \includegraphics[width=1.0\columnwidth]{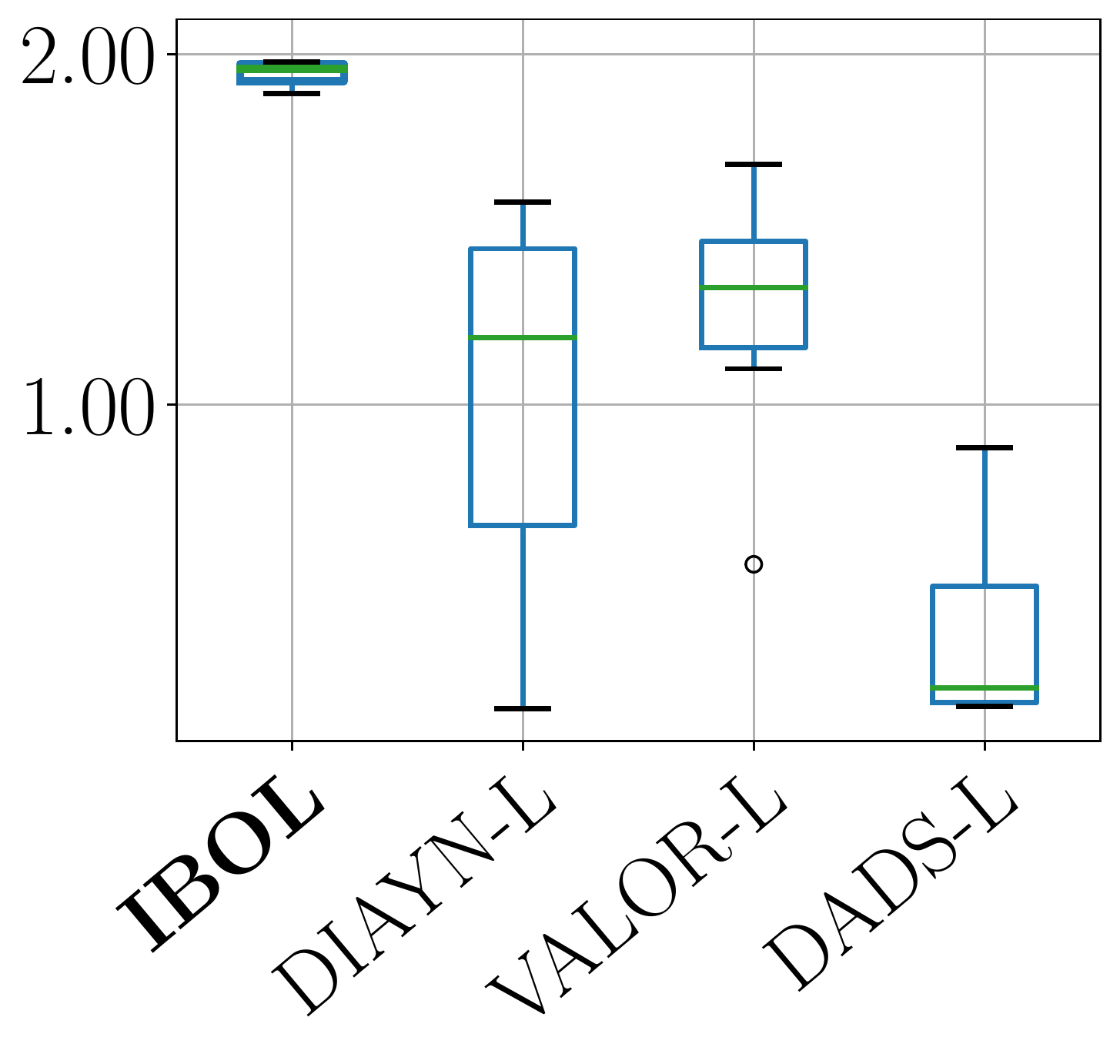}
  \end{subfigure}
  \begin{subfigure}[t]{0.2360890643\linewidth}
    \includegraphics[width=1.0\columnwidth]{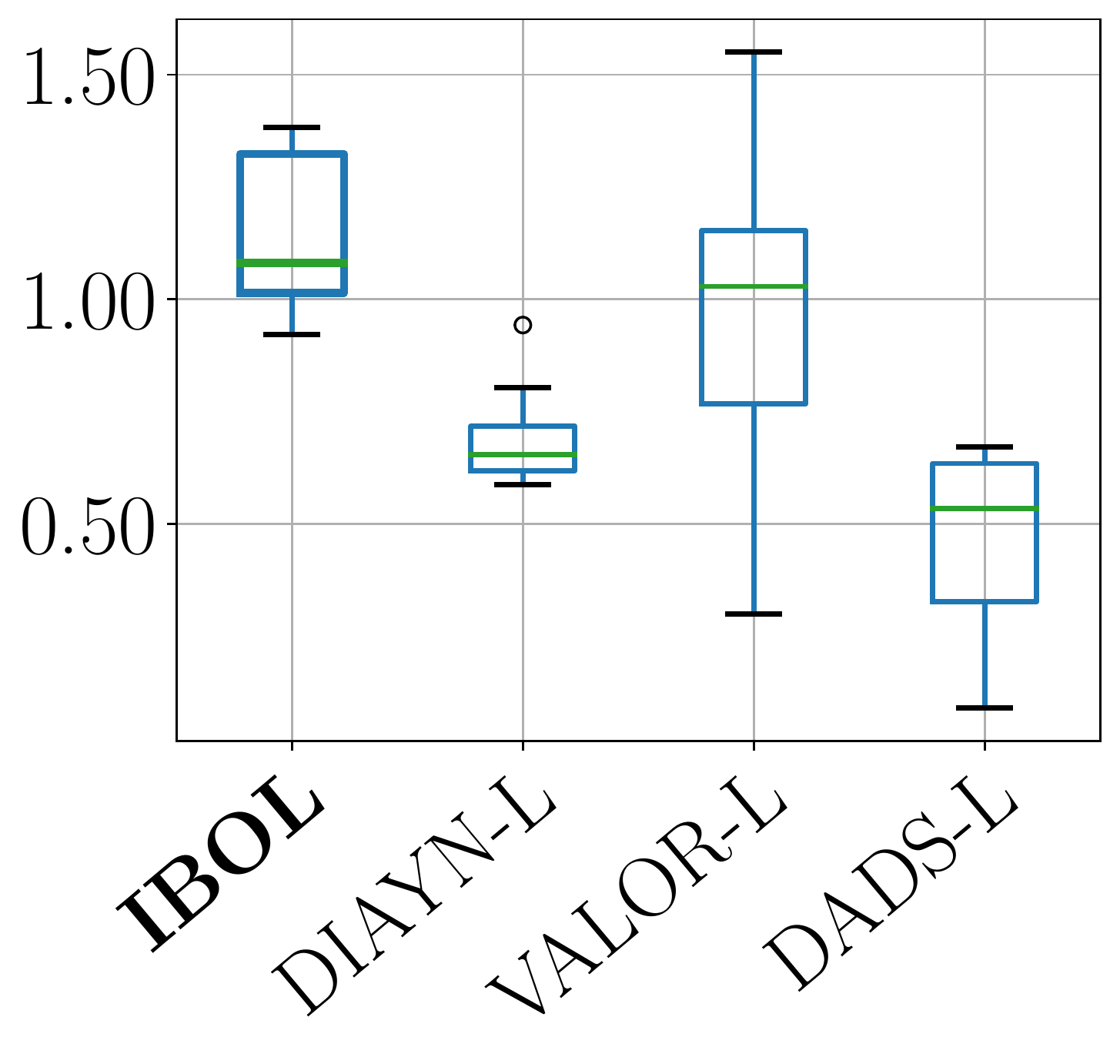}
  \end{subfigure}
  \begin{subfigure}[t]{0.2360890643\linewidth}
    \includegraphics[width=1.0\columnwidth]{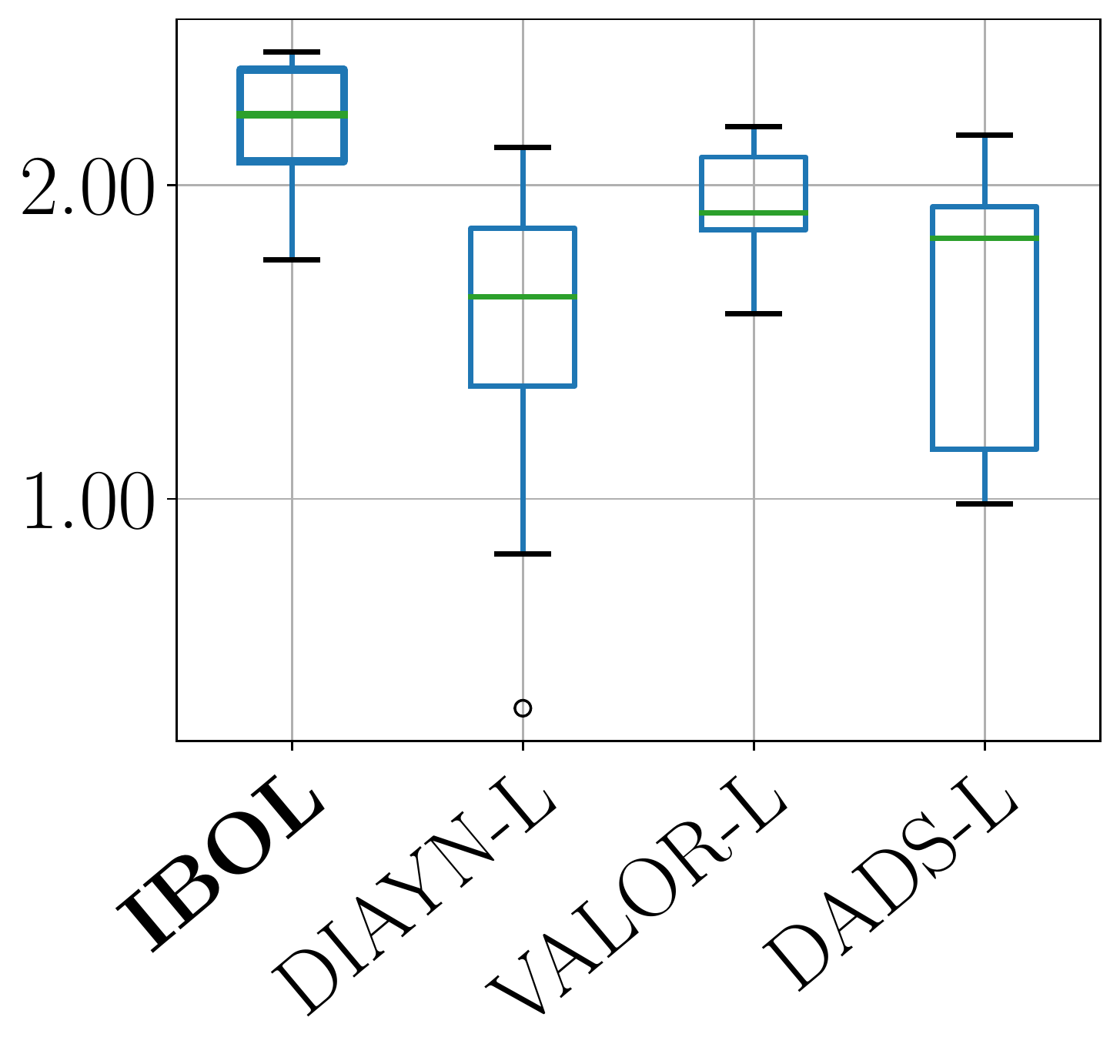}
  \end{subfigure}
  \\

  \vspace{-0.2cm}
  \hfill
  \begin{subfigure}[t]{0.254319053\linewidth}
    \includegraphics[width=1.0\columnwidth]{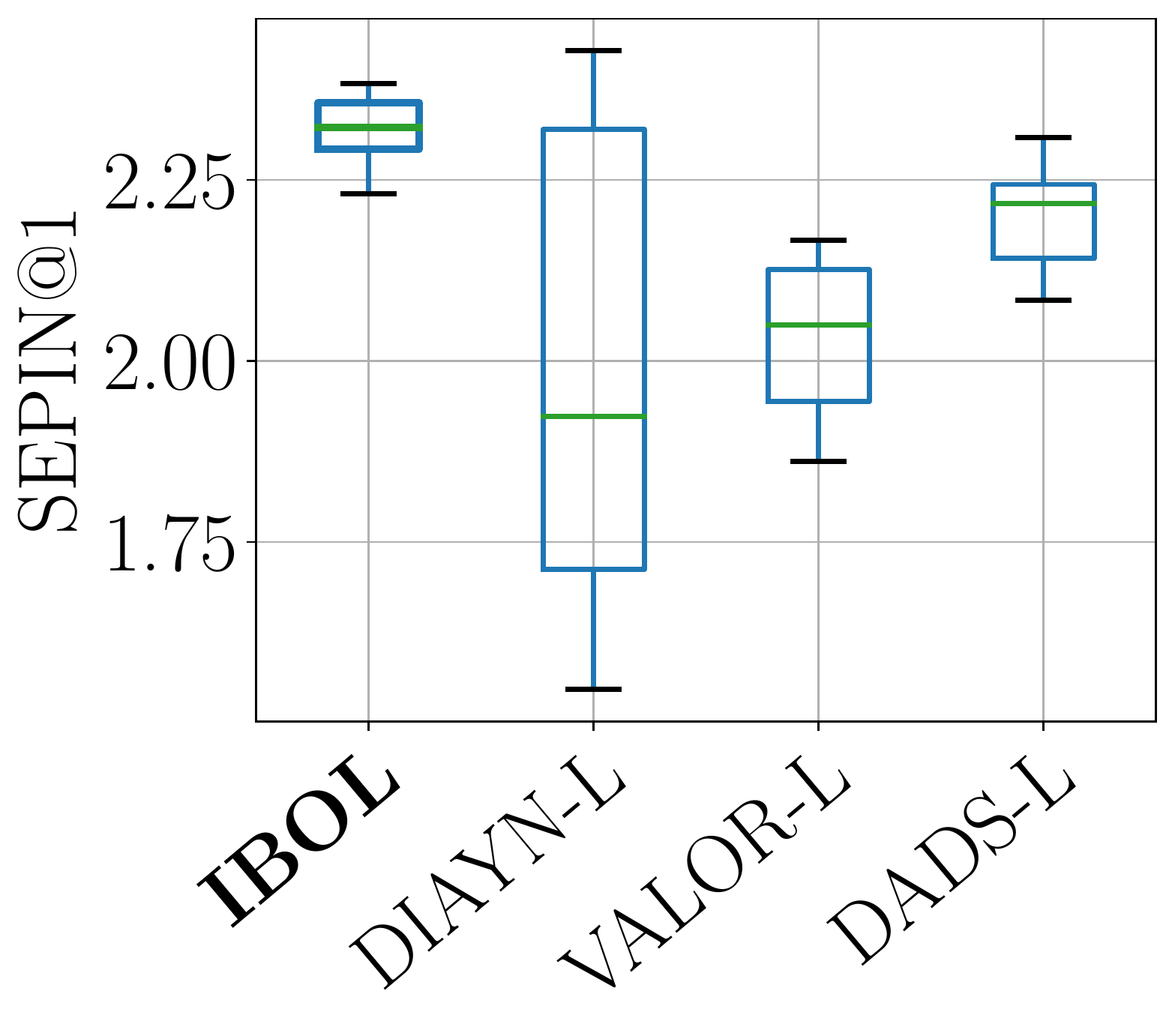}
    \caption{Ant}
  \end{subfigure}
  \begin{subfigure}[t]{0.2360890643\linewidth}
    \includegraphics[width=1.0\columnwidth]{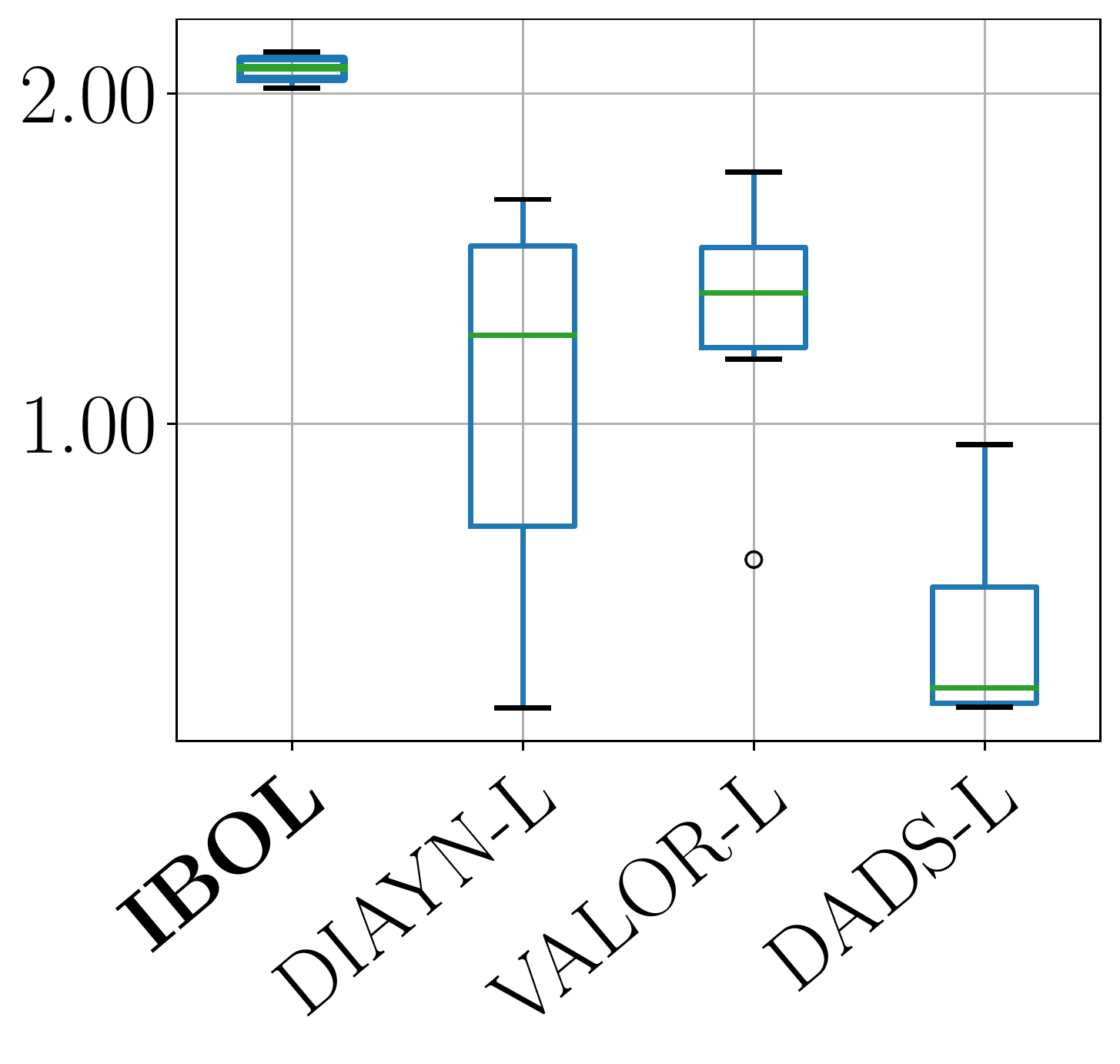}
    \caption{HalfCheetah}
  \end{subfigure}
  \begin{subfigure}[t]{0.2360890643\linewidth}
    \includegraphics[width=1.0\columnwidth]{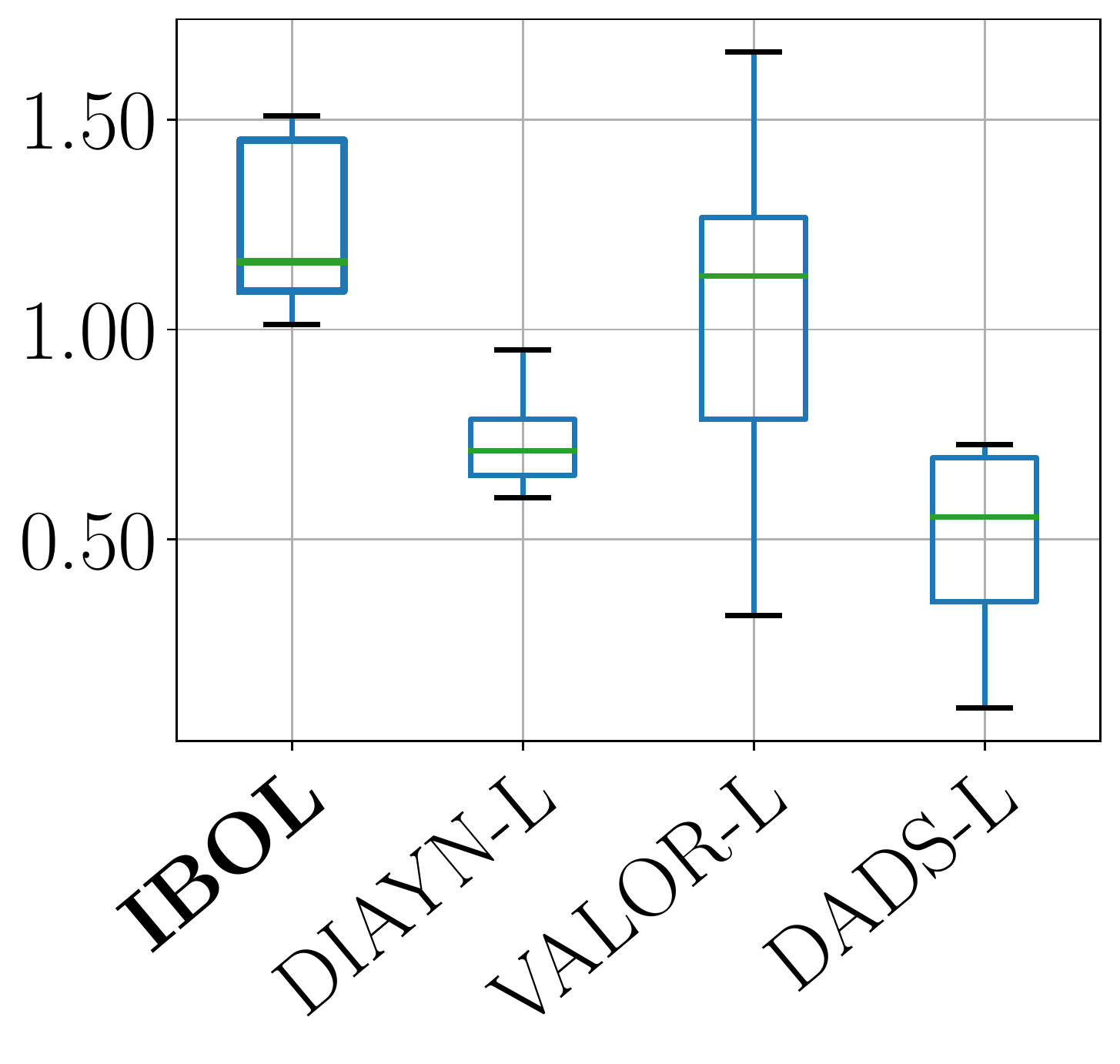}
    \caption{Hopper}
  \end{subfigure}
  \begin{subfigure}[t]{0.2360890643\linewidth}
    \includegraphics[width=1.0\columnwidth]{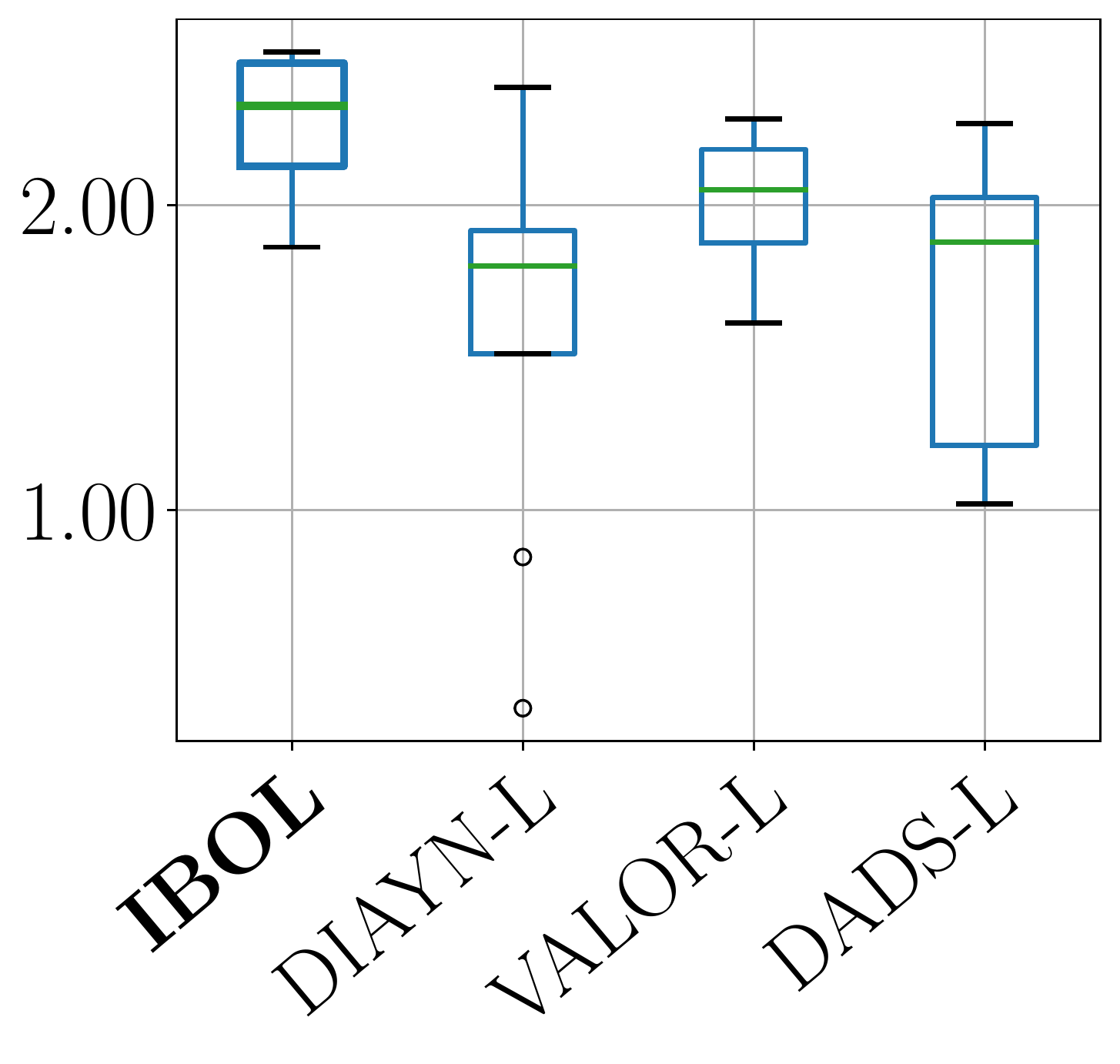}
    \caption{D'Kitty}
  \end{subfigure}

  \caption{
    Comparison of IBOL (ours) with the baseline methods, DIAYN-L, VALOR-L and DADS-L,
    in the evaluation metrics of $I(Z; S_T^{\text{(loc)}})$, WSEPIN and SEPIN$@1$, on Ant, HalfCheetah, Hopper and D'Kitty. 
    For each method, we use the eight trained skill policies.
  }
  \label{fig:eval_metrics}
\end{minipage}
\hfill
\begin{minipage}{0.39\textwidth}
  \vspace{0.80em}
  \begin{subfigure}[t]{1.0\linewidth}
    \centering
    \includegraphics[width=0.95\columnwidth]{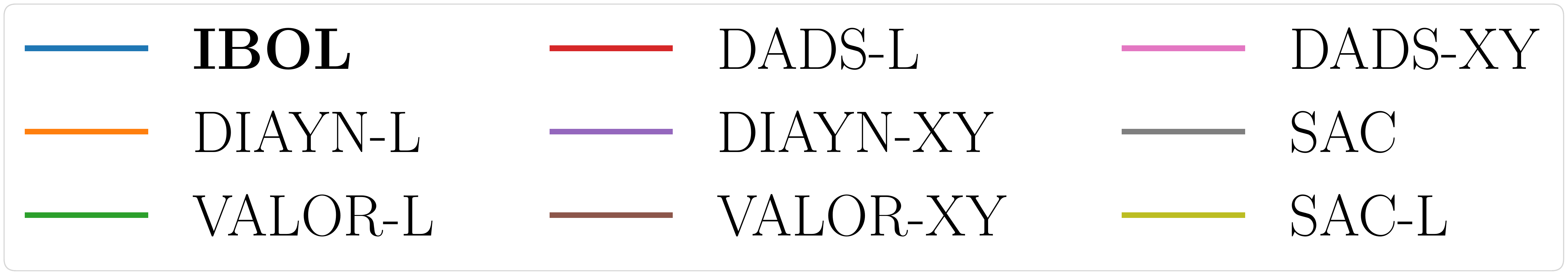}
  \end{subfigure}
  \vspace{-0.4em}

  \begin{subfigure}[t]{0.5\linewidth}
    \centering
    \includegraphics[width=0.95\columnwidth]{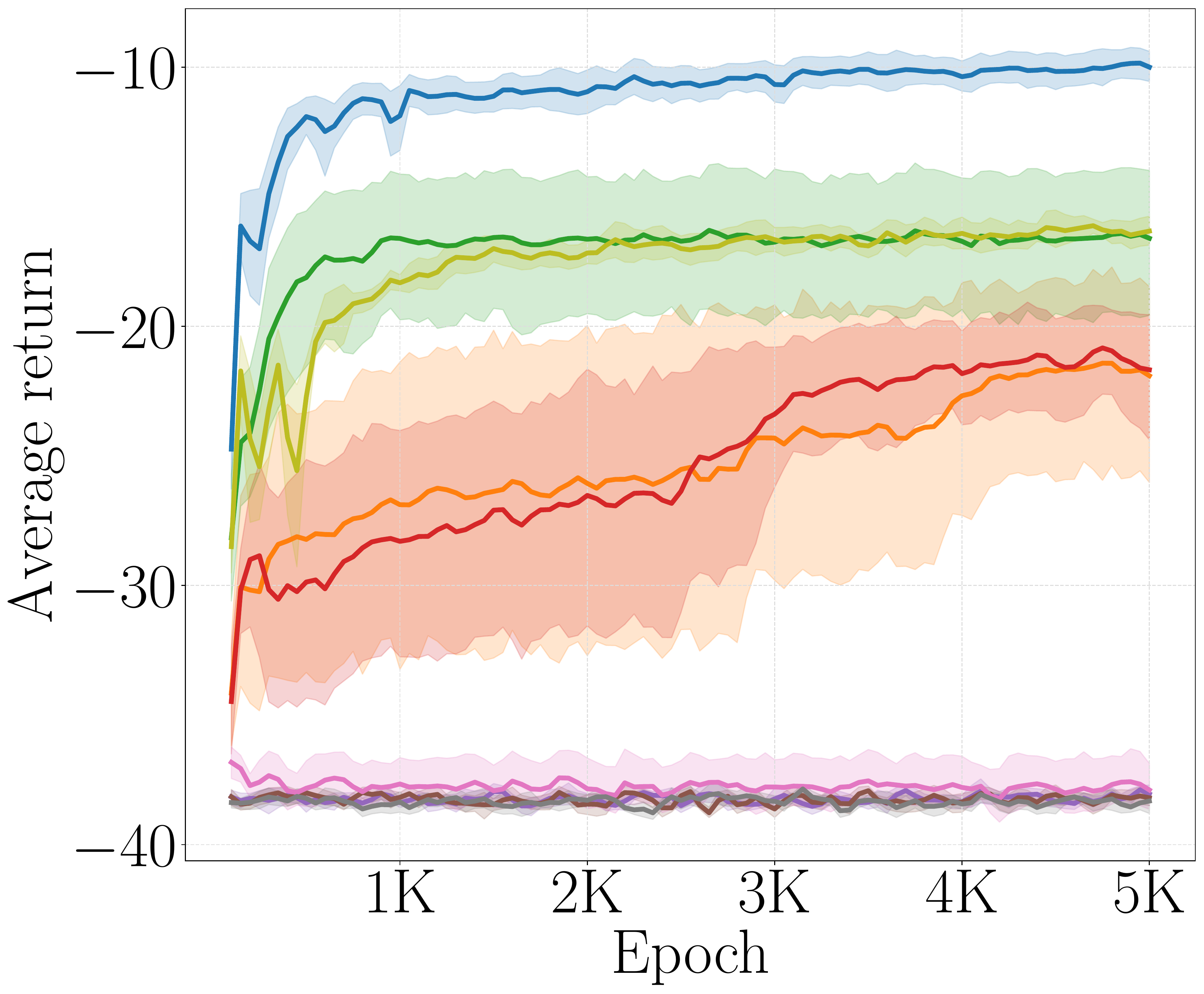}
    \vspace{-0.5em}
    \caption{\textit{AntGoal}}
    \vspace{0.8em}
    \label{fig:downstream_ag}
  \end{subfigure}
  \hspace{-0.5em}
  \begin{subfigure}[t]{0.5\linewidth}
    \centering
    \includegraphics[width=0.95\columnwidth]{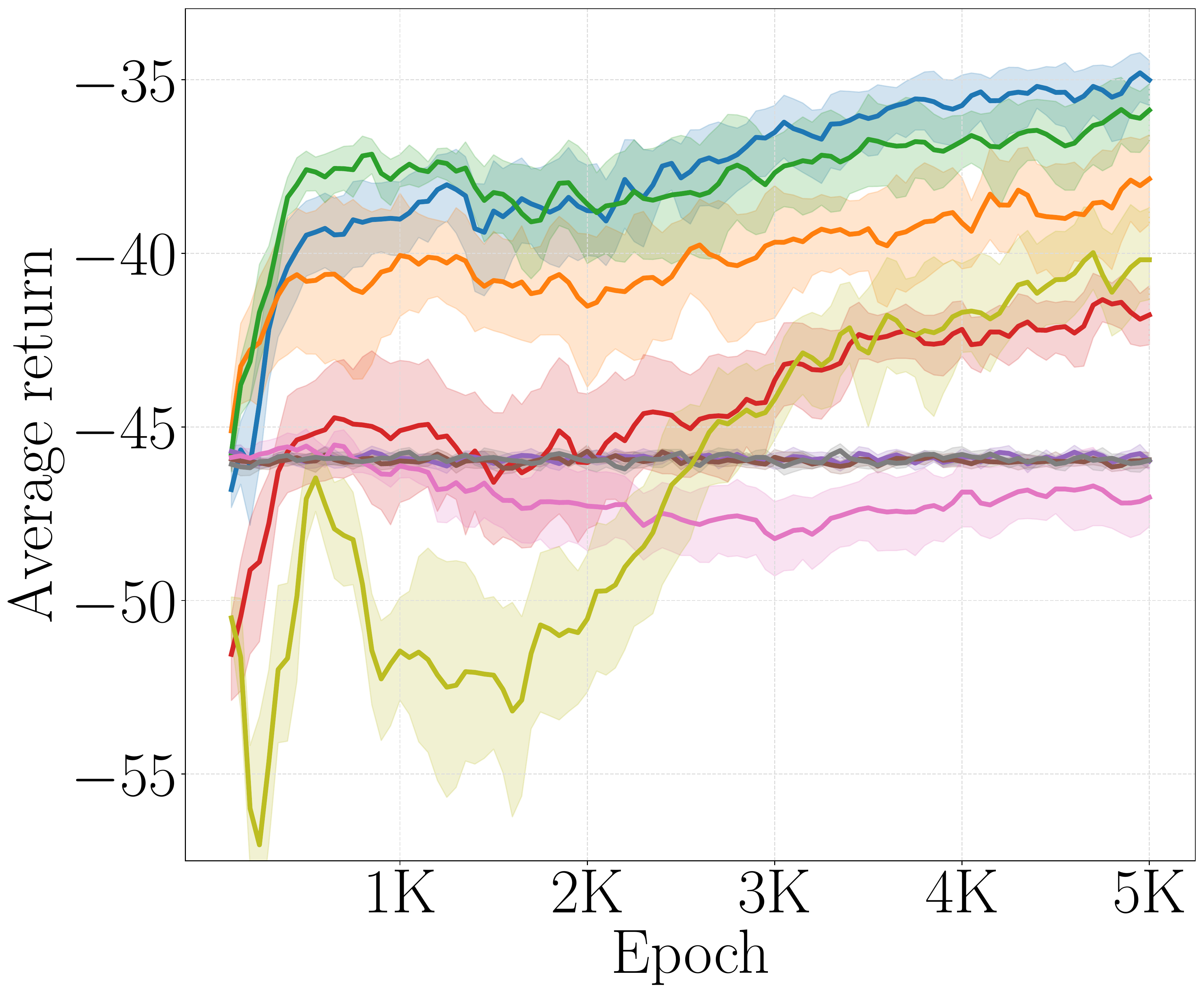}
    \vspace{-0.5em}
    \caption{\textit{AntMultiGoals}}
    \vspace{0.8em}
    \label{fig:downstream_anp}
  \end{subfigure}
  \hspace{-0.5em}

  \begin{subfigure}[t]{0.5\linewidth}
    \centering
    \includegraphics[width=0.95\columnwidth]{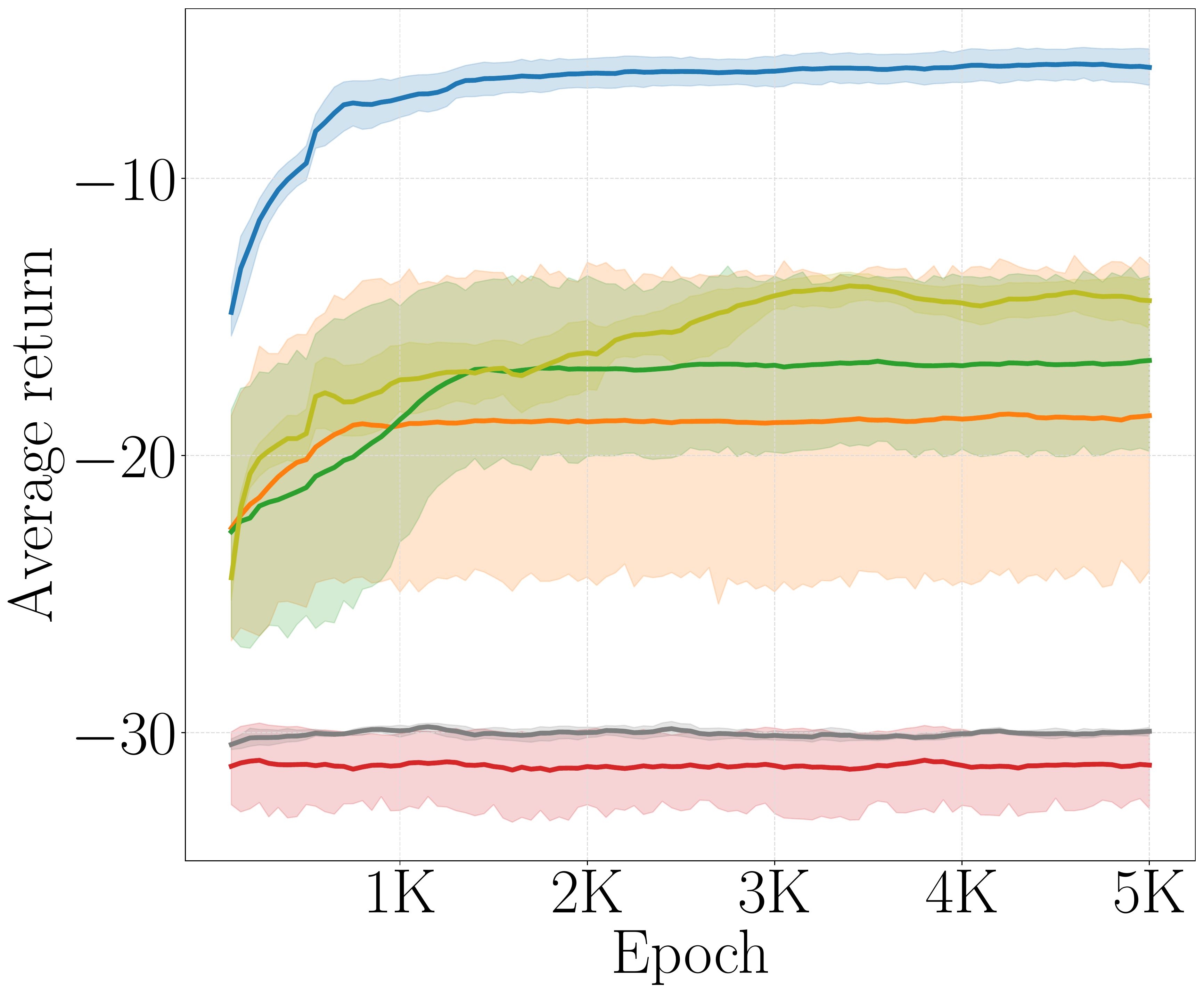}
    \vspace{-0.5em}
    \caption{\textit{CheetahGoal}}
    \label{fig:downstream_chg}
  \end{subfigure}
  \hspace{-0.5em}
  \begin{subfigure}[t]{0.5\linewidth}
    \centering
    \includegraphics[width=0.95\columnwidth]{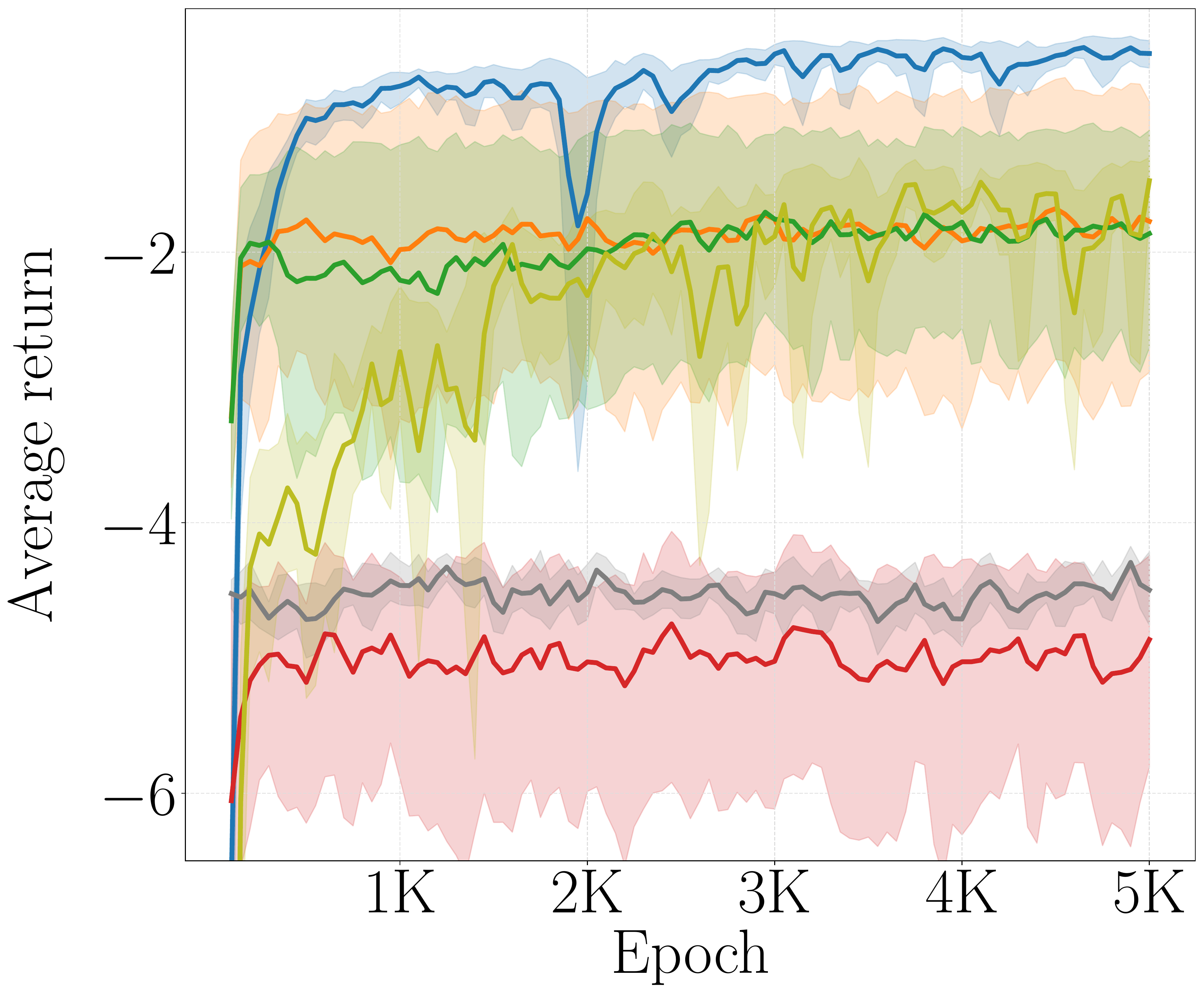}
    \vspace{-0.5em}
    \caption{\textit{{CheetahImitation}}}
    \label{fig:downstream_chi}
  \end{subfigure}

  \caption{
    Comparison of IBOL (ours) with the baseline methods on the four downstream tasks.
    Each line is the mean return over the last $100$ epochs at each time step, averaged over eight runs.
    The shaded areas denote the $95\%$ confidence interval.
  }
  \label{fig:downstream}
\end{minipage}
\end{figure*}

Figure \ref{fig:eval_metrics} shows the box plots of the results.
With the same linearizers, IBOL outperforms the three baselines, DIAYN-L, VALOR-L and DADS-L,
in all three information-theoretic evaluation metrics on Ant, HalfCheetah, Hopper and D'Kitty.
The plots for $I(Z; S_T^{\text{(loc)}})$ show that IBOL can stably discover diverse skills from the environments conditioned on the skill latent parameter $Z$.
Also, the results with WSEPIN and SEPIN$@1$ suggest that IBOL outperforms the baselines,
with regard to both informativeness and separability of $Z$'s individual dimensions.
Overall, IBOL shows the lower average deviation compared to the other methods, which demonstrates its stability in learning.
For additional analysis and details, please refer to Appendix.

\subsection{Evaluation on Downstream Tasks}
\label{sec:downstream}

We demonstrate the effectiveness of the abstraction learned by IBOL on downstream tasks.
In Ant, we modify the environment to obtain two tasks, \textit{AntGoal} and \textit{AntMultiGoals},
inspired by \citet{diayn_eysenbach2019,dads_sharma2020}.
In HalfCheetah, we test the methods on two tasks, \textit{CheetahGoal} and \textit{CheetahImitation}.

\textit{AntGoal} is a task for evaluating how capable the agent is in reaching diverse goals.
For every new episode, a goal $w = [w^{(x)}, w^{(y)}]$ is randomly sampled in the $x$-$y$ plane. 
The agent can observe the goal $w$ at every step, and receives a reward of $\big(- \|w - [s_T^{(x)}, s_T^{(y)}]\|_2\big)$ where $[s_T^{(x)}, s_T^{(y)}]$ is the agent's final position, when each episode ends.

\textit{AntMultiGoals} is a repeated version of \textit{AntGoal}.
At time step $t \equiv 0 \Mod{\eta}$ in each episode, a new goal $w = [w^{(x)}, w^{(y)}]$ is sampled based on the agent's current position, $[s_t^{(x)}, s_t^{(y)}]$,
and is held for the next $\eta$ steps.
Similarly to \textit{AntGoal}, at the end of each $\eta$-sized chunk (before sampling of a new goal),
the agent gains a reward of $\big(- \|w - [s_t^{(x)}, s_t^{(y)}]\|_2\big)$.
We set $\eta = 50$.

\textit{CheetahGoal} is a task similar to \textit{AntGoal} but in HalfCheetah.
For each episode, a goal $w^{(x)}$ in the $x$ axis is sampled and observed by the agent at every step.
At the end of the episode, the agent receives a reward of $\big(- |w^{(x)} - s_T^{(x)}|\big)$ where $s_T^{(x)}$ is the final position of the agent.

We also experiment with a different type of task, \textit{CheetahImitation}.
Each of the skill policies learned by the four skill discovery methods is used to sample $1000$ random skill trajectories, 
all of whose $x$ traces are gathered to form a set of imitation targets.
For a new episode of \textit{CheetahImitation}, 
one imitation target $w = [w_1^{(x)}, \ldots, w_T^{(x)}]$, a sequence of $T$ positions in the $x$ axis, is randomly sampled from the set.
The goal of this task is to imitate the target $w$ in the $x$ axis;
at the $t$-th step, a reward of $\big(- (w_t^{(x)} - s_t^{(x)})^2 \big)$ is given, 
where the agent perceives the target $w$ as part of its observation.
\textit{CheetahImitation} can evaluate the diversity and coverage of skill policies.

For comparison,
we employ a meta-controller on top of each skill policy learned by skill discovery methods.
The meta-controller iterates observing a state from the environment and
picking a \textit{skill} with its own meta-policy, which invokes the pre-trained skill policy with the same skill latent value $z$ for $\ell_m$ time steps.
We employ Soft Actor-Critic (SAC) \cite{sac_haarnoja2018} to train the meta-controller,
and also compare a pure SAC agent as an additional baseline method.

Figure \ref{fig:downstream} compares the performance of IBOL with the baseline methods on the four tasks:
\textit{AntGoal}, \textit{AntMultiGoals}, \textit{CheetahGoal} and \textit{CheetahImitation}.
We set $\ell_m = 5$ for \textit{AntMultiGoals} and $\ell_m = 20$ for the others.
Figures \ref{fig:downstream_ag} and \ref{fig:downstream_anp} suggest that the abstraction by IBOL is more effective for the meta-controller to learn to reach a goal from the initial state, in comparison to the baselines.
They confirm that the linearizer greatly helps different skill policies' learning of locomotion in Ant.
Figure \ref{fig:downstream_chg} shows that IBOL provides better abstraction to the meta-controller for reaching goals in HalfCheetah.
Also, Figure \ref{fig:downstream_chi} demonstrates that IBOL's skills can be used to imitate skills not only from itself but also from the other baselines.
It supports the improved diversity of skills learned by IBOL.
Overall, IBOL presents significantly smaller variances than the other baselines.

\subsection{Additional Observations}
\label{sec:additional_obs}
We present more experiments on Ant to confirm that IBOL can pick appropriate goals at different states for the linearizer in order to learn skills with high distinguishability.

\begin{figure}[t!]
  \centering

  \resizebox{\linewidth}{!}{
    \subcaptionbox*{}{
      \includegraphics[height=2.7cm]{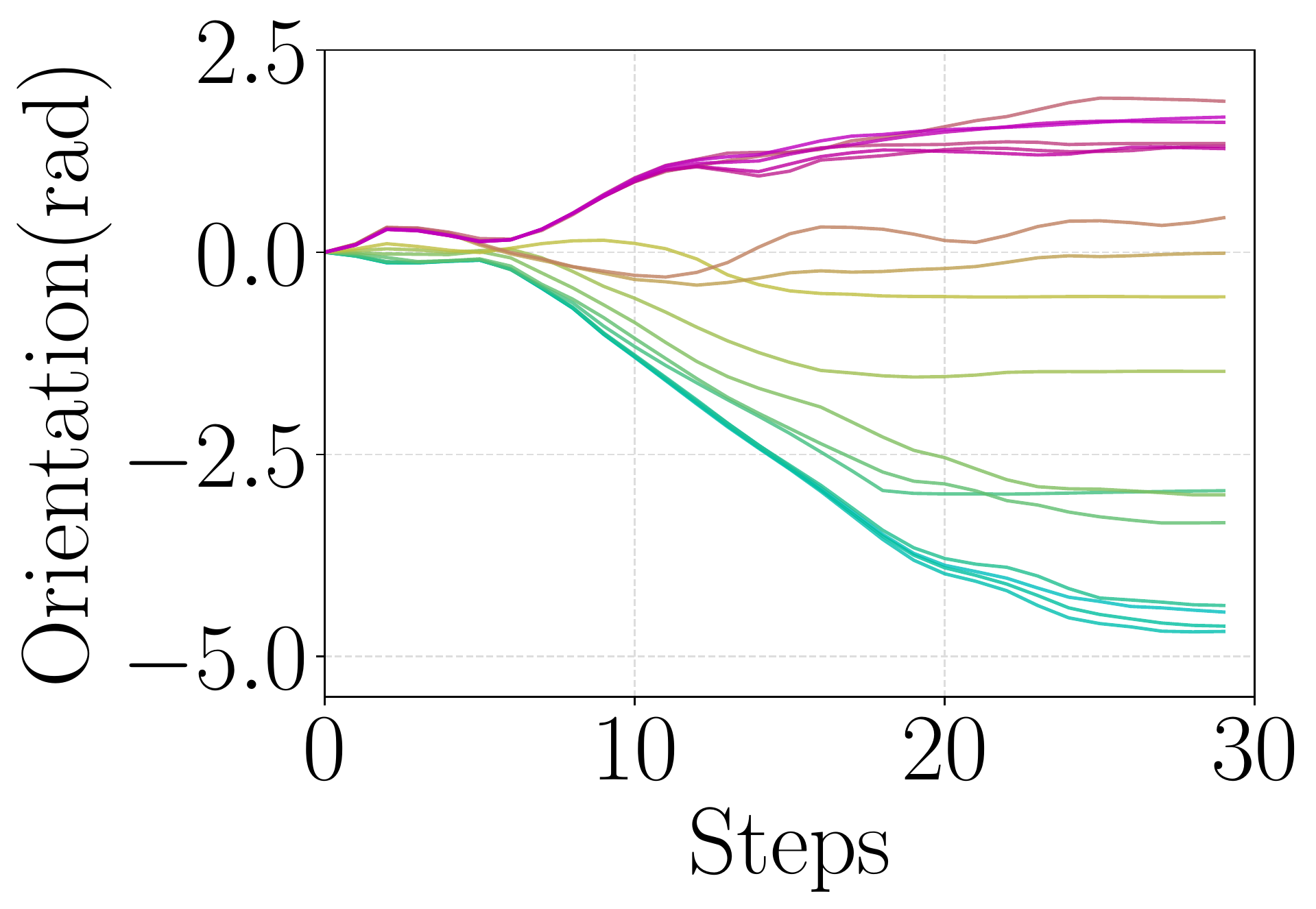}
    }
    \subcaptionbox*{}{
      \includegraphics[height=2.7cm]{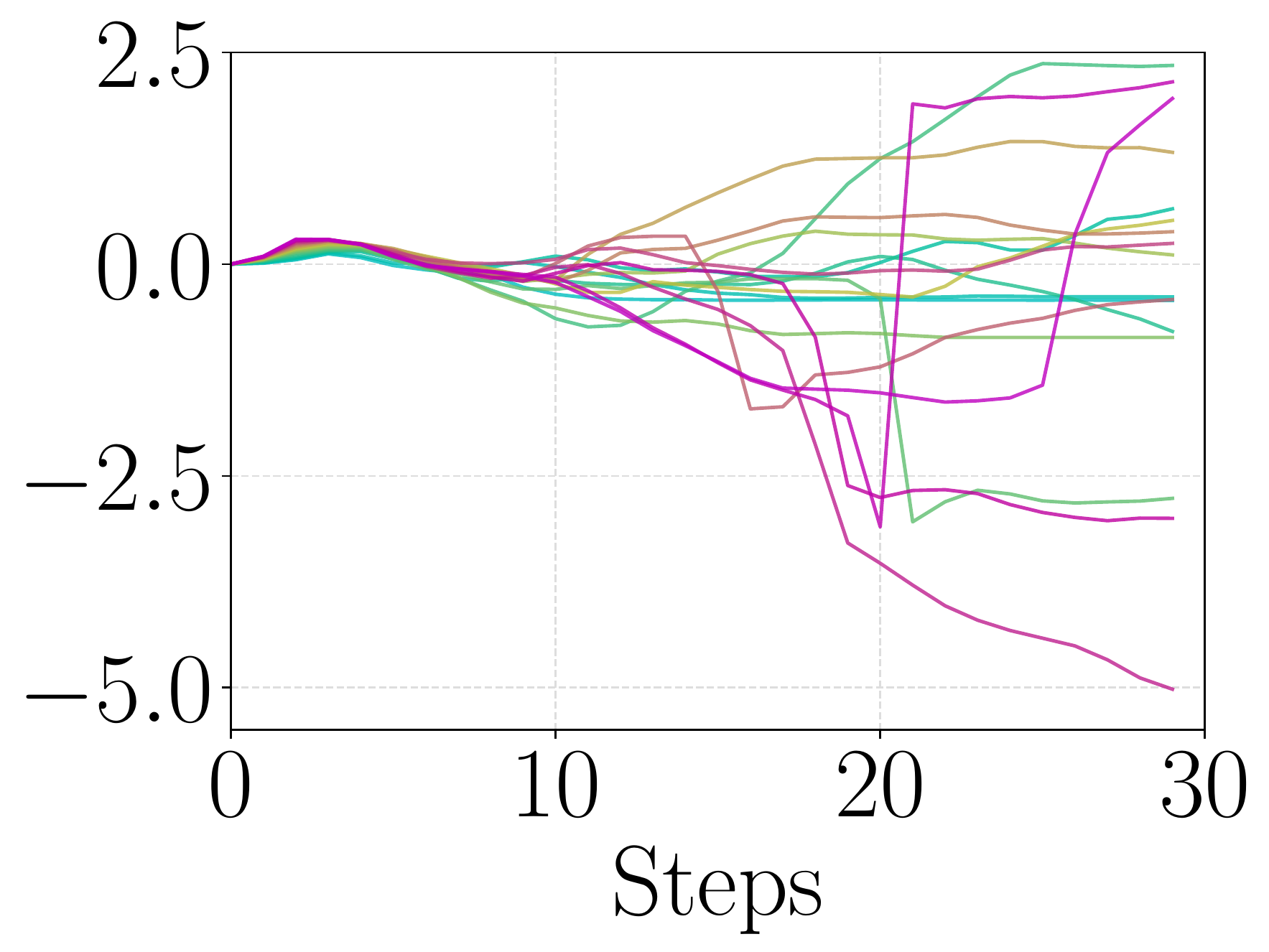}
    }
  }

  \vspace{-0.4cm}

  \begin{subfigure}[t]{0.51\linewidth}
    \caption{IBOL}
    \label{fig:ori_ibol}
  \end{subfigure}
  \begin{subfigure}[t]{0.46\linewidth}
    \caption{Linearizer only}
    \label{fig:ori_cp}
  \end{subfigure}

  \begin{subfigure}[h]{0.98\linewidth}
    \centering
    \includegraphics[width=1.00\columnwidth]{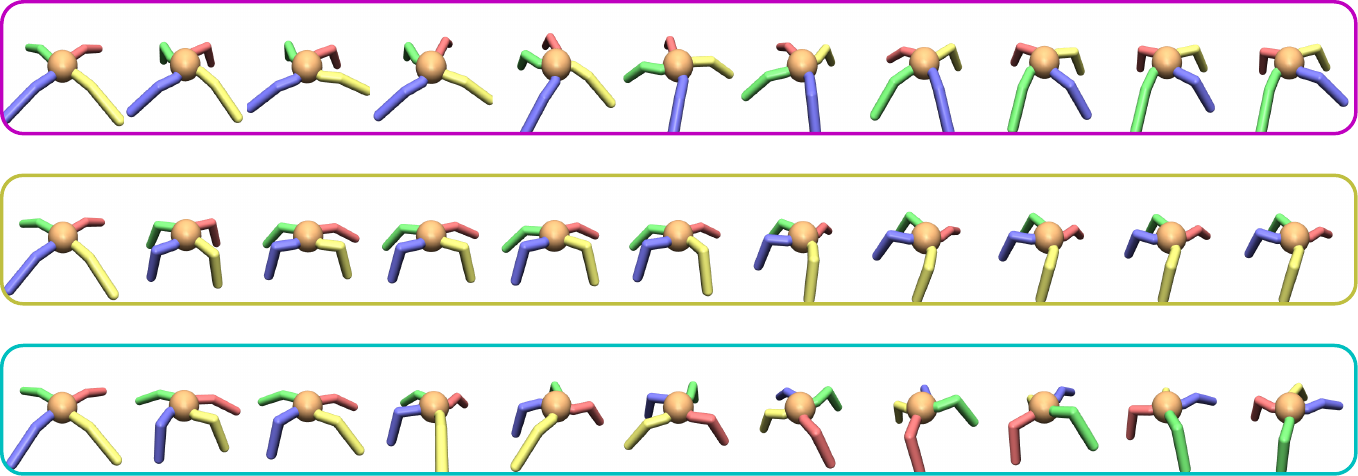}
    \caption{Rendered scenes}
    \label{fig:ori_render}
  \end{subfigure}
  \caption{
    Orientation trajectories from (a) the \textit{skill policy} of IBOL and (b) the \textit{linearizer}.
    The skill latent value is interpolated from $-4$ (cyan) to $4$ (magenta) for IBOL, while the orientation dimension value of the goal is interpolated from $-1$ (cyan) to $1$ (magenta) for the linearizer (since it is trained with the goal range of $[-1, 1]$).
    (c) Rendered scenes of IBOL's trajectories from (a).
  }
  \label{fig:ant_ori}
\end{figure}

\textbf{Learning non-locomotion skills}.
In the absence of locomotion signals, IBOL can learn orientation primitives, which is not easy unless the skill discovery algorithm produces diverse goals for the linearizer.
Figure \ref{fig:ori_ibol} shows examples of orientation skills by IBOL on Ant with $d = 1$.
Figure \ref{fig:ori_cp} depicts that using the linearizer alone fails to produce comparable results, while
IBOL utilizes various goal dimensions of the linearizer to obtain an interpolable skill set.

\begin{figure}[t!]
  \centering
  \resizebox{1.00\linewidth}{!}{
    \hspace{-0.35cm}
    \subcaptionbox*{}{
      \includegraphics[height=5.8cm]{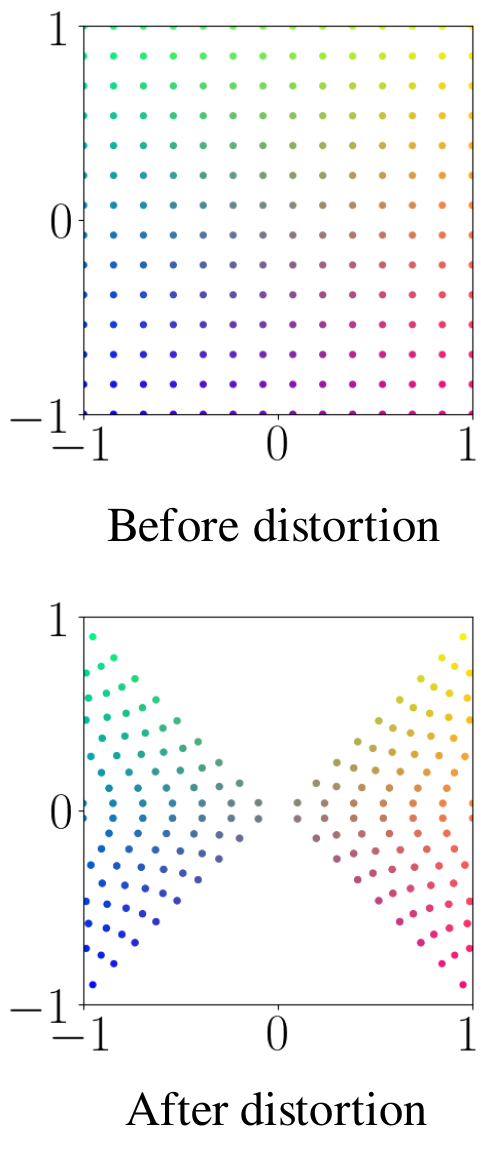}
    }
    \hspace{0.1cm}
    \rule{0.4pt}{5.8cm}
    \subcaptionbox*{}{
      \includegraphics[height=5.8cm]{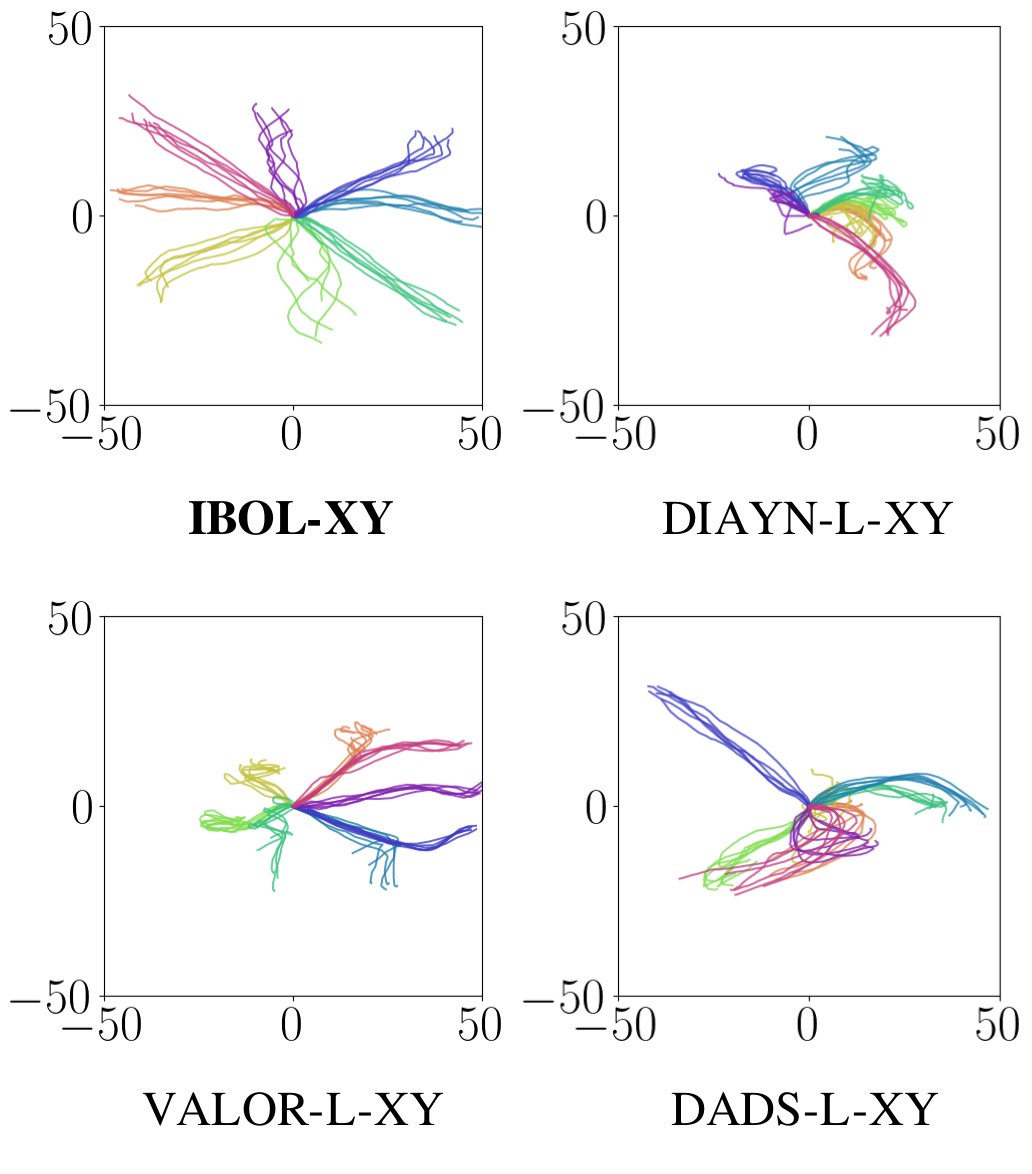}
    }
    \hspace{-0.27cm}
  }
  \vspace{-0.5em}

  \hfill
  \begin{subfigure}[h]{0.32\linewidth}
    \caption{Distortion scheme}
    \label{fig:dis_scheme}
  \end{subfigure}
  \begin{subfigure}[h]{0.63\linewidth}
    \caption{Visualization of $x$-$y$ traces}
    \label{fig:ant_xy_dis}
  \end{subfigure}
  \hfill

  \vspace{-0.3cm}
  \caption{
    (a) Distortion scheme of the linearizer. It distorts the $x$ and $y$ dimensions of goals, and produces the corresponding actions for the modified goals.
    (b) Visualization of the $x$-$y$ traces of the skills discovered by each algorithm using the distorted linearizer. %
    The same skill latents are used with the top row of Figure \ref{fig:ant_xy}.
  }
\end{figure}

\textbf{Overcoming goal space distortion}.
We conduct additional experiments to validate IBOL's capability of discovering more discriminable trajectories even under harsh conditions.
We distort the linearizer's goal space as Figure \ref{fig:dis_scheme}, so that reaching vertically distant states becomes more demanding.
We train IBOL-XY, DIAYN-L-XY, VALOR-L-XY and DADS-L-XY with $d = 2$ on top of the modified linearizer.
Figure \ref{fig:ant_xy_dis} suggests that IBOL discovers locomotion skills in various angles including vertical directions in the most visually disentangled manner.

\section{Conclusion}

We presented Information Bottleneck Option Learning (IBOL) as a novel unsupervised skill discovery method.
It first deals with the environment dynamics using the linearizer trained to transition in various directions in the state space.
It then discovers skills taking advantage of the information bottleneck framework, which learns the skill latent parameter (or the parameter of the skill policy) as the learned representations of the skills.
Our quantitative evaluation showed that the skill latent learned by IBOL provides improved abstraction measured as the disentanglement.
We also confirmed that IBOL outperforms other skill discovery methods with notably lower variances and the linearizer benefits both IBOL and other baselines on downstream tasks.

One future challenge may be to deal with environments whose state space is very high dimensional such as vision environments,
since goal directions of the linearizer in such domains might not operate well as feasible signals.
A possible solution could be adopting state representation learning techniques for RL such as \citet{nearoptimal_nachum2019}.

\section*{Acknowledgements}
We thank the anonymous reviewers for the helpful comments.
This work was supported by 
Samsung Advanced Institute of Technology,     %
the ICT R\&D program of MSIT/IITP (No. 2019-0-01309, Development of AI technology for guidance of a mobile robot to its goal with uncertain maps in indoor/outdoor environments) and     %
Institute of Information \& communications Technology Planning \& Evaluation (IITP) grant funded by the Korea government (MSIT) (No.2019-0-01082, SW StarLab).     %
Jaekyeom Kim was supported by Hyundai Motor Chung Mong-Koo Foundation.
Gunhee Kim is the corresponding author.

\bibliography{icml2021_option_cr}
\bibliographystyle{icml2021}

\newpage

\appendix

\section{Additional Qualitative Results}
As a complement to \Cref{sec:qual_results} and Figure \ref{fig:mujoco_traj} from the main paper,
we present videos that demonstrate various skills learned by IBOL at \url{https://vision.snu.ac.kr/projects/ibol}.

\section{Derivation of the Lower Bound}
\label{sec:lowerbound_derivation}
We describe the derivation of \Cref{eq:obj_lowerbound2} from the main paper.
Starting with \Cref{eq:obj_ib} from the main paper,
\begin{align}
  & \mathbb{E}_{t} [ I(Z; G_t | S_t) - \beta \cdot I(Z; S_{0:T}) ] \nonumber \\
  &= \mathbb{E}_{\substack{\tau \sim p_{\samplesub}(\tau), t, \\ z \sim \trajencoder(z|s_{0:T})} } \bigg[ \log \frac{p_{\samplesub}(g_t | s_t, z)}{\pisample(g_t | s_t)}
  - \beta \log \frac{\trajencoder(z | s_{0:T})}{\trajencoder(z)} \bigg]. \nonumber
\end{align}
For the first term, as described in the main paper, 
we use the skill policy's output distribution $\pioption(g_t|s_t,z)$ as a variational approximation of $p_{\samplesub}(g_t|s_t,z)$, 
which derives 
\begin{align}
  & \mathbb{E}_{\substack{\tau \sim p_{\samplesub}(\tau), t, \\ z \sim \trajencoder(z|s_{0:T})} } \bigg[ \log \frac{p_{\samplesub}(g_t | s_t, z)}{\pisample(g_t | s_t)} \bigg]
  \nonumber \\
  &= \mathbb{E}_{\substack{\tau \sim p_{\samplesub}(\tau), t, \\ z \sim \trajencoder(z|s_{0:T})} } \bigg[ \log \frac{\pioption(g_t|s_t,z)}{\pisample(g_t | s_t)} \bigg]
  \nonumber \\
  &\hspace{15pt} + \mathbb{E}_{\substack{s_{0:T} \sim p_{\samplesub}(\cdot), t, \\ z \sim \trajencoder(z|s_{0:T})} } \bigg[ D_{\text{KL}} (p_{\samplesub}(G_t | s_t, z) \| \pioption(G_t|s_t,z)) \bigg]
  \nonumber \\
  &\geq \mathbb{E}_{\substack{\tau \sim p_{\samplesub}(\tau), t, \\ z \sim \trajencoder(z|s_{0:T})} } \bigg[ \log \frac{\pioption(g_t|s_t,z)}{\pisample(g_t | s_t)} \bigg]. \label{eq:lowerbound_prediction}
\end{align}
Also, for the second term, we use $r(z)$ as the variational approximation of the marginal distribution $\trajencoder(z)$,
and it derives
\begin{align}
  &\mathbb{E}_{\substack{\tau \sim p_{\samplesub}(\tau), \\ z \sim \trajencoder(z|s_{0:T})} } \bigg[ \log \frac{\trajencoder(z | s_{0:T})}{\trajencoder(z)} \bigg]
  \nonumber \\
  &= \mathbb{E}_{\substack{\tau \sim p_{\samplesub}(\tau), \\ z \sim \trajencoder(z|s_{0:T})} } \bigg[ \log \frac{\trajencoder(z | s_{0:T})}{r(z)} \bigg]
  - D_{\text{KL}} (\trajencoder(Z) \| r(Z))
  \nonumber \\
  &\leq \mathbb{E}_{\substack{\tau \sim p_{\samplesub}(\tau), \\ z \sim \trajencoder(z|s_{0:T})} } \bigg[ \log \frac{\trajencoder(z | s_{0:T})}{r(z)} \bigg]
  \nonumber \\
  &= \mathbb{E}_{\substack{\tau \sim p_{\samplesub}(\tau)} } \bigg[ D_{\text{KL}} (\trajencoder(Z|s_{0:T}) \| r(Z)) \bigg]. \label{eq:lowerbound_compression}
\end{align}
Combining the derivation of \Cref{eq:lowerbound_prediction,eq:lowerbound_compression} obtains \Cref{eq:obj_lowerbound2} from the main paper.

\section{Encouraging Disentanglement}
\label{sec:disentanglement}
Disentanglement learning methods often model the generation process as $X \sim p(\cdot | Z)$,
assuming that data $X$ is generated with its underlying latent factors $Z$ \cite{disentangleby_kim2018,disentangletheory_do2019}. %
While the aggregated posterior of $Z$ is defined as $q(Z) = \int_x q(Z | x) p(x) dx$ for an encoder $q(Z | X)$ \cite{adversarialae_makhzani2015},
one of common disentanglement approaches is to penalize the total correlation of $q(Z)$, expressed as $TC(q(Z)) = D_{\text{KL}}(q(Z) \| \prod_i^d q(Z_i) )$.

The IB framework has theoretical connections to the disentanglement of representation \cite{infodropout_achille2018,emergenceof_achille2018,isolating_chen2018}.
In \Cref{sec:skill_discovery}, we derived our objective in the form of IB.
Especially, if we model the prior of $Z$ as $r(Z) = \prod_i^d r(Z_i)$, 
the term $D_{\text{KL}}(\trajencoder(Z|s_{0:T}) \| r(Z))$ in \Cref{eq:obj_lowerbound2} from the main paper is decomposed into three terms revealing the total correlation term $TC(\trajencoder(Z)) = D_{\text{KL}}(\trajencoder(Z) \| \prod_i^d \trajencoder(Z_i) )$ \cite{isolating_chen2018} (refer to \Cref{sec:kl_term_decomposition} for the proof),
where the aggregated posterior of the trajectory encoder is $\trajencoder(Z) = \int_{s_{0:T}} \trajencoder(Z | s_{0:T}) p(s_{0:T}) ds_{0:T}$.
By penalizing $TC(\trajencoder(Z))$, we encourage each dimension of the skill latent space $\mathcal{Z}$ disentangled from the others with respect to $S_{0:T}$.
As a result, the learned skill latent $Z$ can provide improved abstraction, where each dimension focuses more on only its corresponding behavior of the discovered skills.

\section{Decomposition of the KL Divergence Term}
\label{sec:kl_term_decomposition}

When the prior of $Z$ is modeled as a factorized distribution,
\ie $r(Z) = \prod_i^d r(Z_i)$, the KL divergence term in \Cref{eq:obj_lowerbound2} from the main paper can be decomposed as follows \cite{isolating_chen2018}:
\begin{align}
    &D_{\text{KL}}(\trajencoder(Z|s_{0:T}) \| r(Z)) \\
    &\qquad=\mathbb{E}_{z \sim \trajencoder(Z|s_{0:T})} \left[ \log \frac{\trajencoder(z | s_{0:T})}{r(z)} \right] \nonumber \\
    &\qquad=\mathbb{E}_{z \sim \trajencoder(Z|s_{0:T})} \left[ \log \frac{\trajencoder(z | s_{0:T})}{\trajencoder(z)} \right] \nonumber \\
    &\qquad\qquad+ \mathbb{E}_{z \sim \trajencoder(Z|s_{0:T})} \left[ \log \frac{\trajencoder(z)}{\prod_{i=1}^d \trajencoder(z_i)} \right] \nonumber \\
    &\qquad\qquad+ \mathbb{E}_{z \sim \trajencoder(Z|s_{0:T})} \left[ \log \frac{\prod_{i=1}^d \trajencoder(z_i)}{\prod_{i=1}^d r(z_i)} \right] \nonumber \\
    &\qquad= D_{\text{KL}}(\trajencoder(Z|s_{0:T}) \| \trajencoder(Z)) \nonumber \\
    &\qquad\qquad+ D_{\text{KL}}(\trajencoder(Z) \| \prod_{i=1}^d \trajencoder(Z_i)) \nonumber \\
    &\qquad\qquad+ \sum_{i=1}^d D_{\text{KL}}(\trajencoder(Z_i) \| r(Z_i)),
\end{align}
where $\trajencoder(Z) = \int_{s_{0:T}} \trajencoder(Z | s_{0:T}) p(s_{0:T}) ds_{0:T}$ denotes the aggregated posterior.

The second term corresponds to the total correlation of $Z$, as $TC(\trajencoder(Z)) = D_{\text{KL}}(\trajencoder(Z) \| \prod_i^d \trajencoder(Z_i))$,
encouraging $Z$ to have a more disentangled representation.
The third term operates as a regularizer, which pushes each dimension of the aggregated posterior $\trajencoder(Z)$ to be located in the vicinity of the prior.
The expectation of the first term can be represented in the form of mutual information, as $\mathbb{E}_{s_{0:T}} \left[ D_{\text{KL}}(\trajencoder(Z|s_{0:T}) \| \trajencoder(Z)) \right] = I(S_{0:T};Z)$.
This corresponds to the original compression term before applying the variational approximation.

\section{Information-Theoretic Evaluation Metrics}
\label{sec:info_eval_metrics}

In this section, we describe the information-theoretic metrics used for the evaluation in \Cref{sec:info_eval}.
In addition to $I(Z; S_T^{\text{(loc)}})$, we use the SEPIN$@k$ and WSEPIN metrics for measuring informativeness and separability \citep{disentangletheory_do2019}.
For SEPIN$@k$, $I(S_T^{\text{(loc)}}; Z_i | Z_{\neq i})$ quantifies the amount of information about $S_T^{\text{(loc)}}$ contained by $Z_i$ but not $Z_{\neq i}$,
and the metric is defined as 
\begin{align}
  \text{SEPIN}@k = \frac{1}{k} \sum_{j=1}^k I(S_T^{\text{(loc)}}; Z_{r_j} | Z_{\neq r_j}),
\end{align}
where $r_j$ is the dimension index with the $j$-th largest value of $I(S_T^{\text{(loc)}}; Z_i | Z_{\neq i})$ for $i = 1, \ldots, d$.
That is, SEPIN$@k$ is the average of the top $k$ values of $I(S_T^{\text{(loc)}}; Z_i | Z_{\neq i})$.
They also define WSEPIN as
\begin{align}
  \text{WSEPIN} = \sum_{i=1}^d \rho_i \cdot I(S_T^{\text{(loc)}}; Z_i | Z_{\neq i})
\end{align}
for $\rho_i = \frac{I(S_T^{\text{(loc)}}; Z_i)}{\sum_{j=1}^d I(S_T^{\text{(loc)}}; Z_j)}$.
It is the sum of $I(S_T^{\text{(loc)}}; Z_i | Z_{\neq i})$ weighted based on their informativeness, $I(S_T^{\text{(loc)}}; Z_i)$.

\section{Varying Number of Bins for MI Estimation}

\begin{figure*}[t!]
  \centering

  \hfill
  \begin{subfigure}[t]{0.2047248057\linewidth}
    \includegraphics[width=1.0\columnwidth]{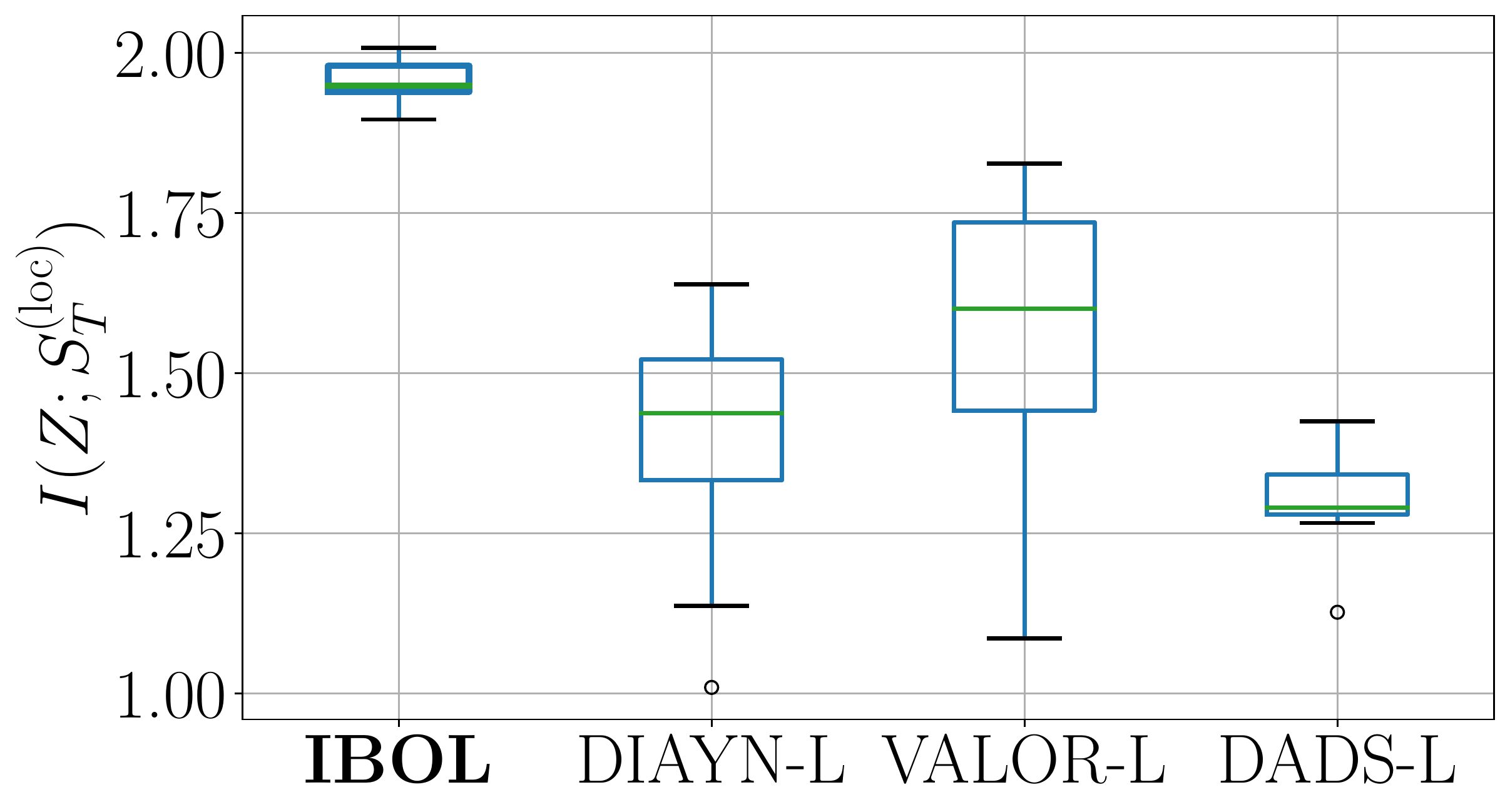}
  \end{subfigure}
  \begin{subfigure}[t]{0.1913187986\linewidth}
    \includegraphics[width=1.0\columnwidth]{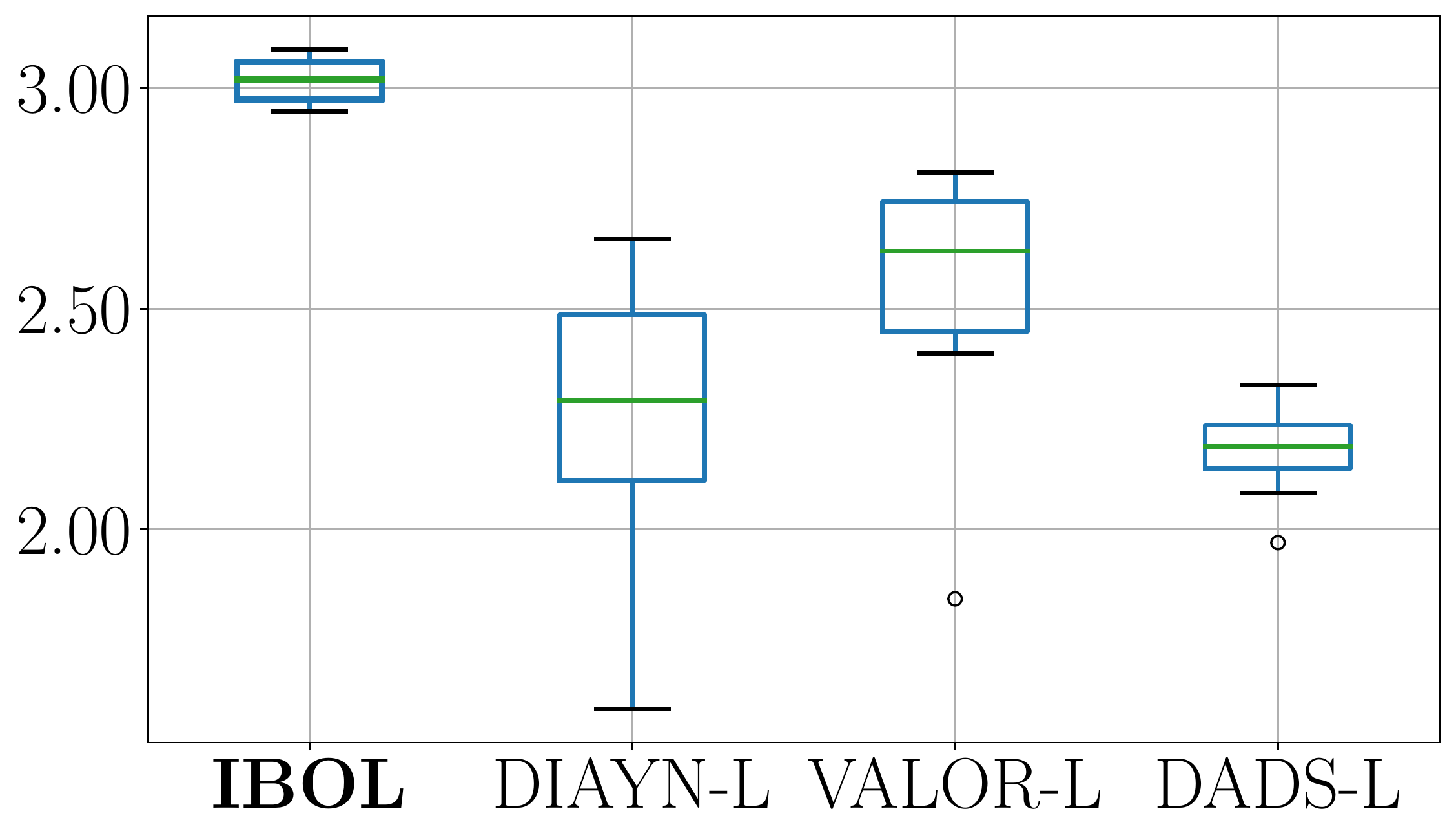}
  \end{subfigure}
  \begin{subfigure}[t]{0.1913187986\linewidth}
    \includegraphics[width=1.0\columnwidth]{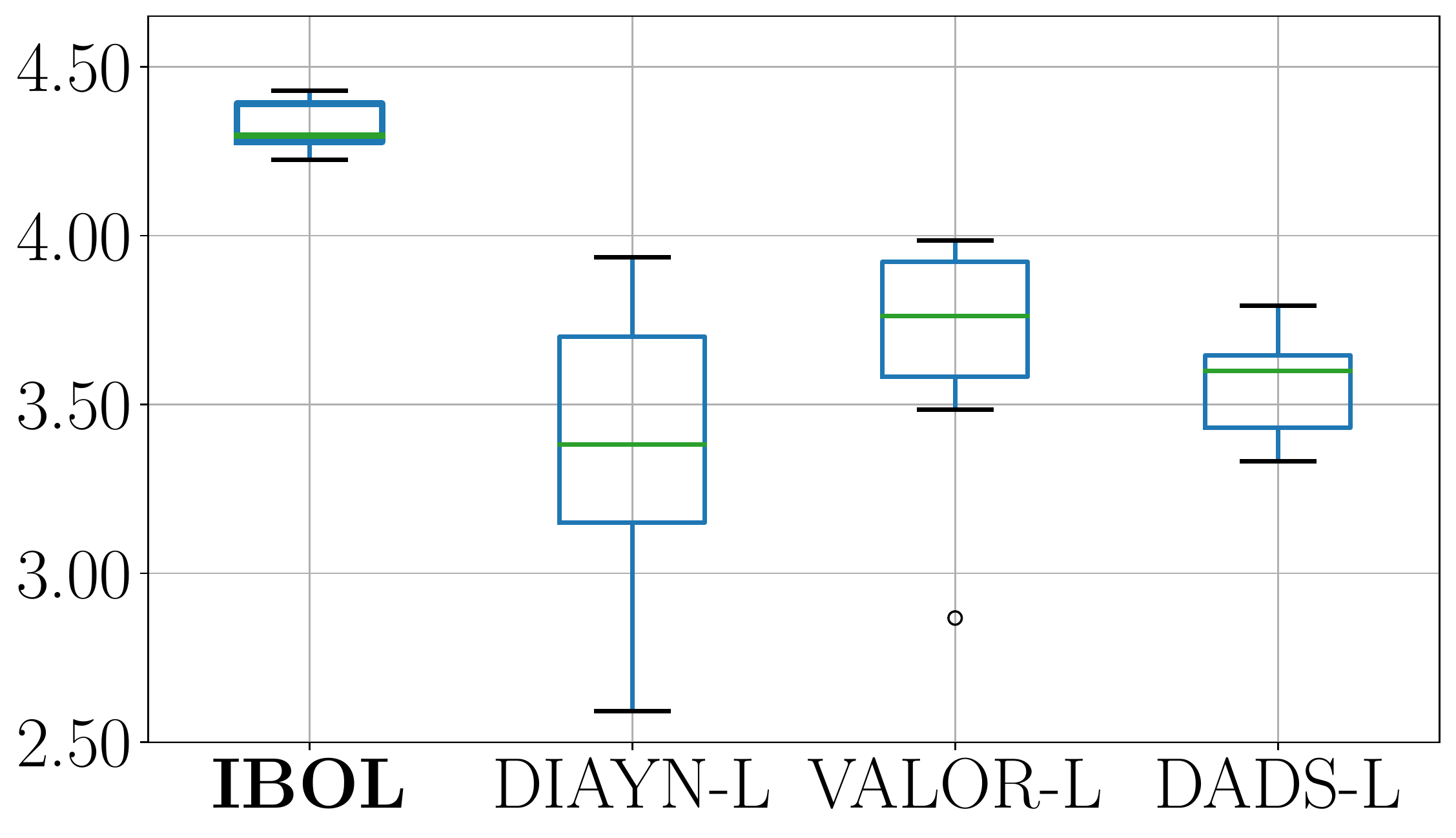}
  \end{subfigure}
  \begin{subfigure}[t]{0.1913187986\linewidth}
    \includegraphics[width=1.0\columnwidth]{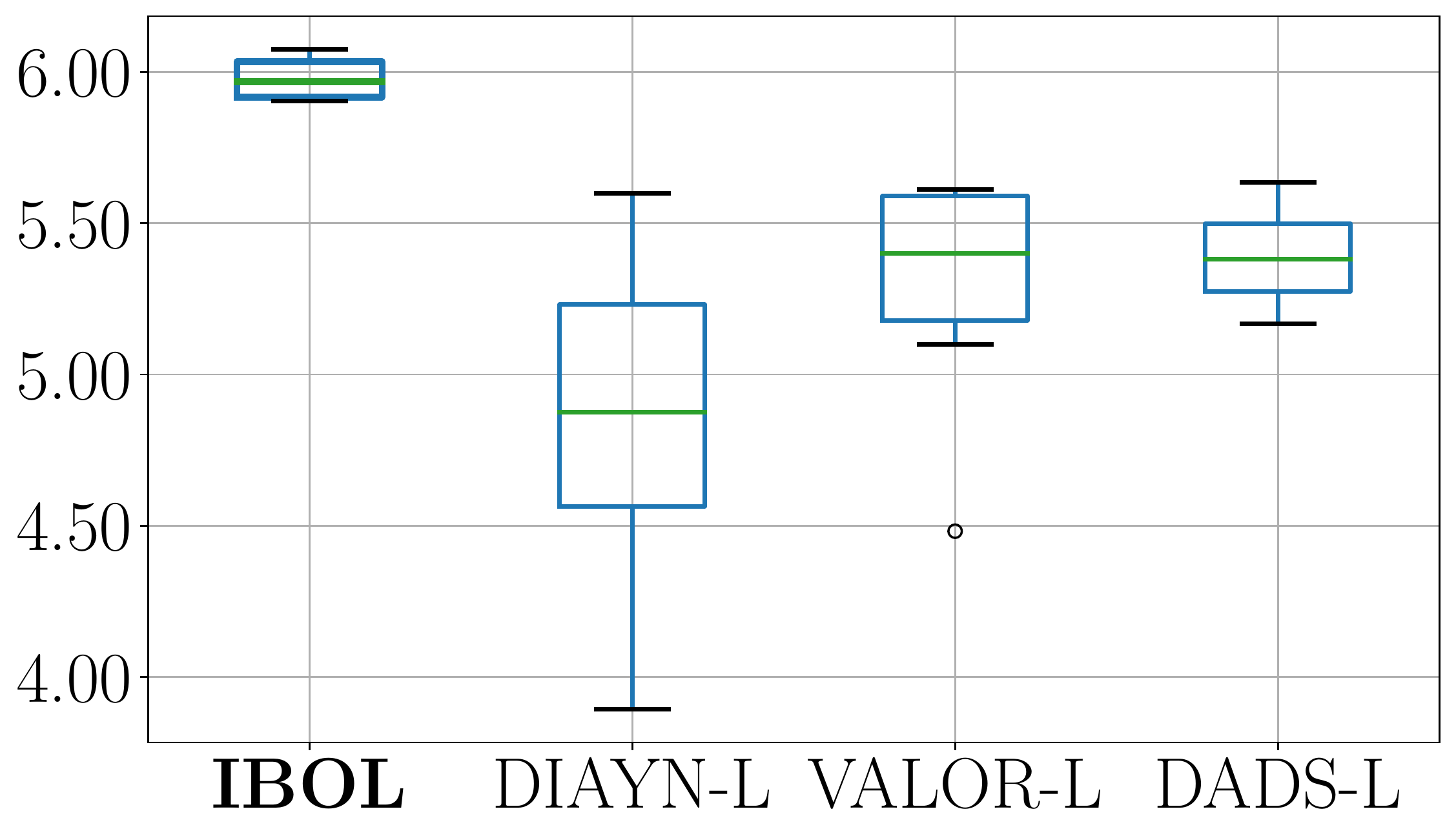}
  \end{subfigure}
  \begin{subfigure}[t]{0.1913187986\linewidth}
    \includegraphics[width=1.0\columnwidth]{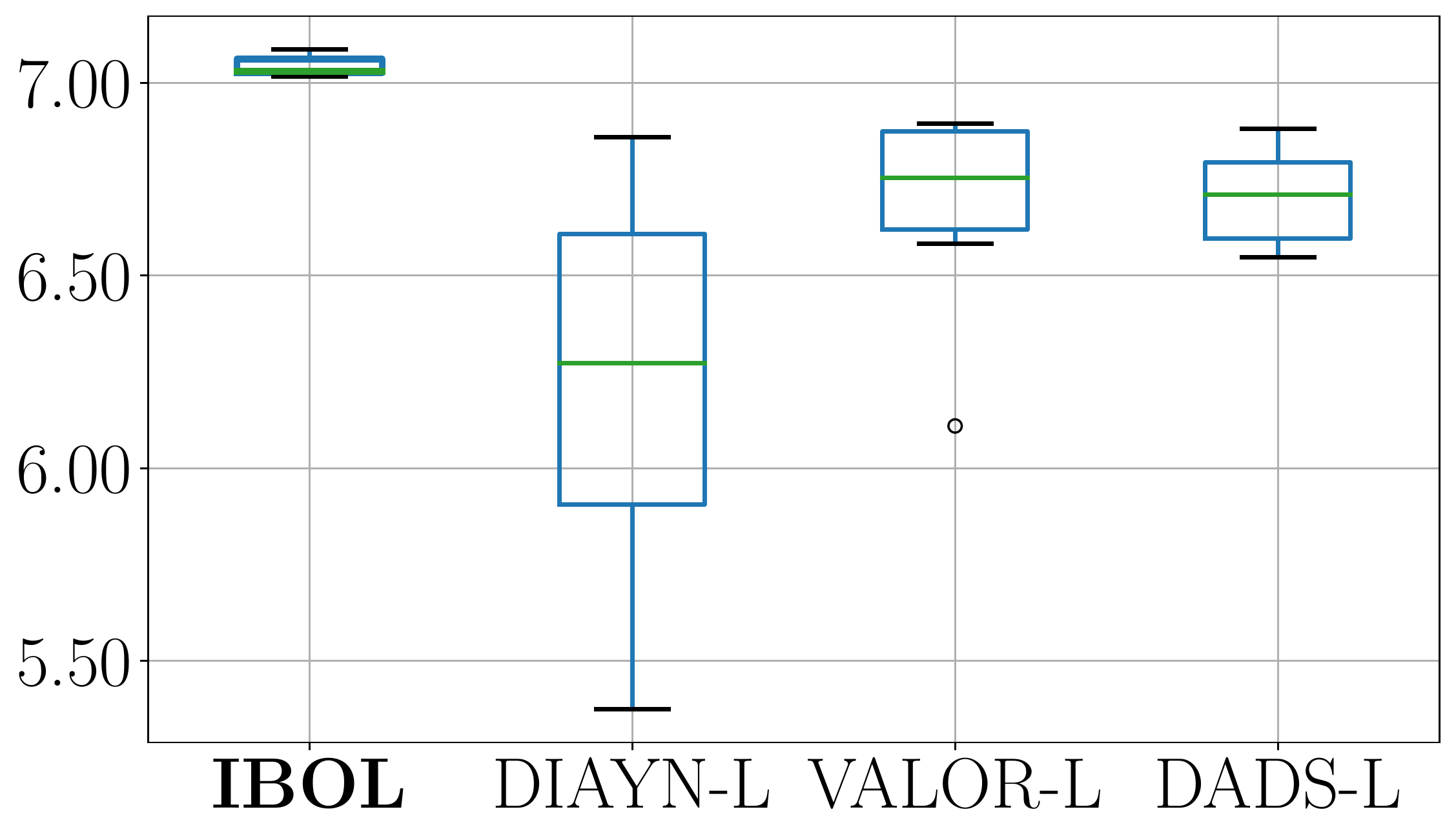}
  \end{subfigure}

  \hfill
  \begin{subfigure}[t]{0.2008490516\linewidth}
    \includegraphics[width=1.0\columnwidth]{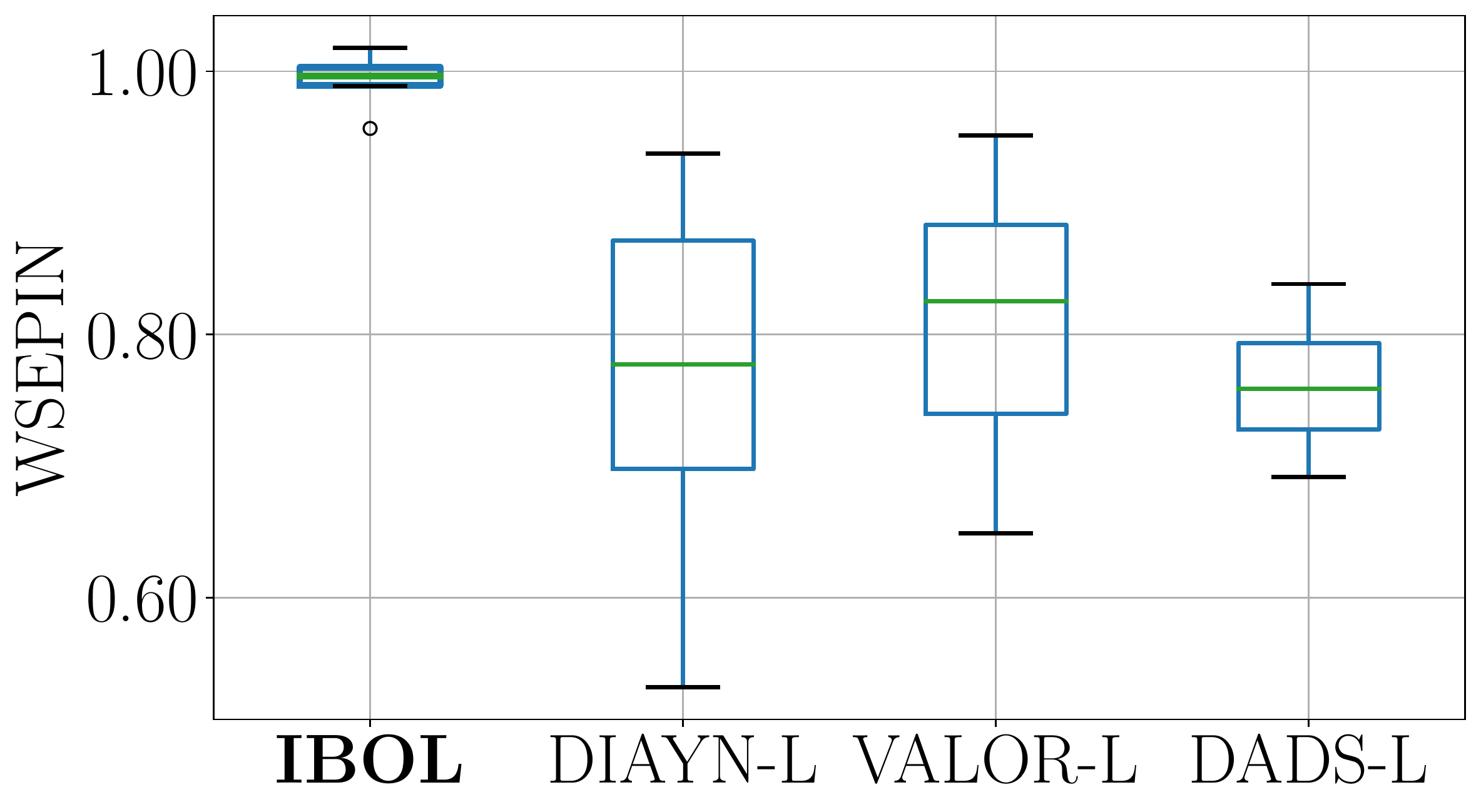}
  \end{subfigure}
  \begin{subfigure}[t]{0.1913187986\linewidth}
    \includegraphics[width=1.0\columnwidth]{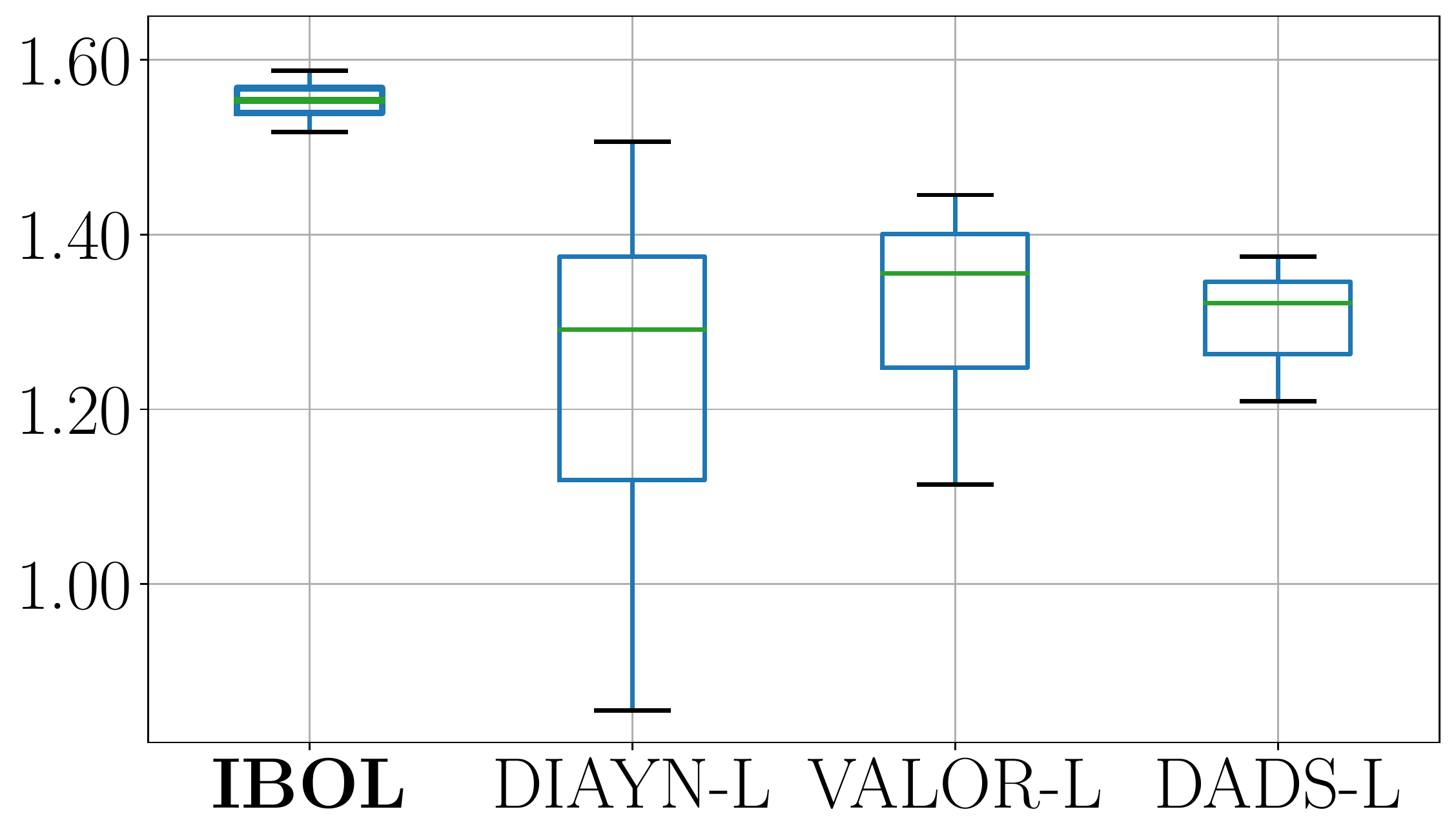}
  \end{subfigure}
  \begin{subfigure}[t]{0.1913187986\linewidth}
    \includegraphics[width=1.0\columnwidth]{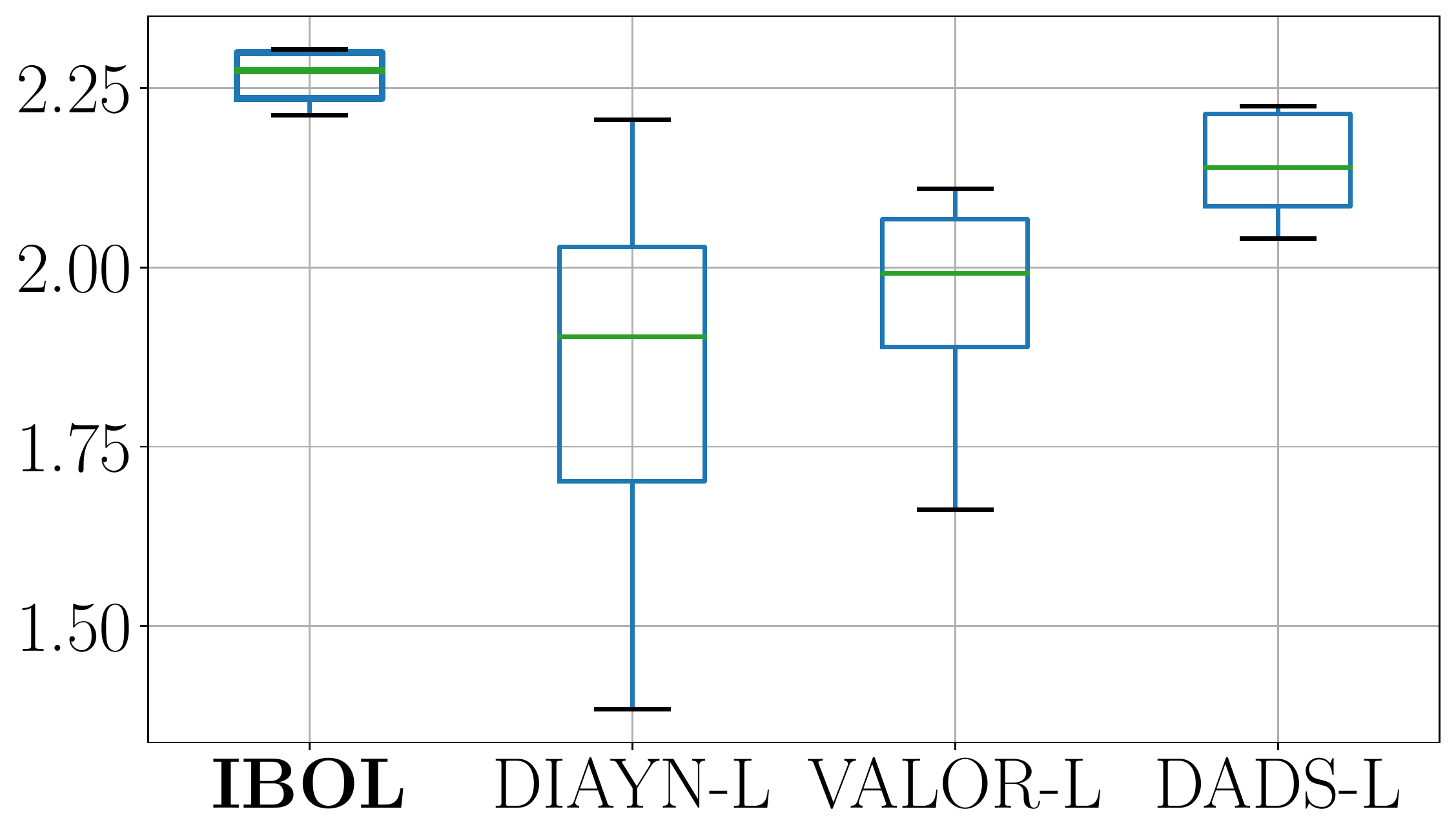}
  \end{subfigure}
  \begin{subfigure}[t]{0.1913187986\linewidth}
    \includegraphics[width=1.0\columnwidth]{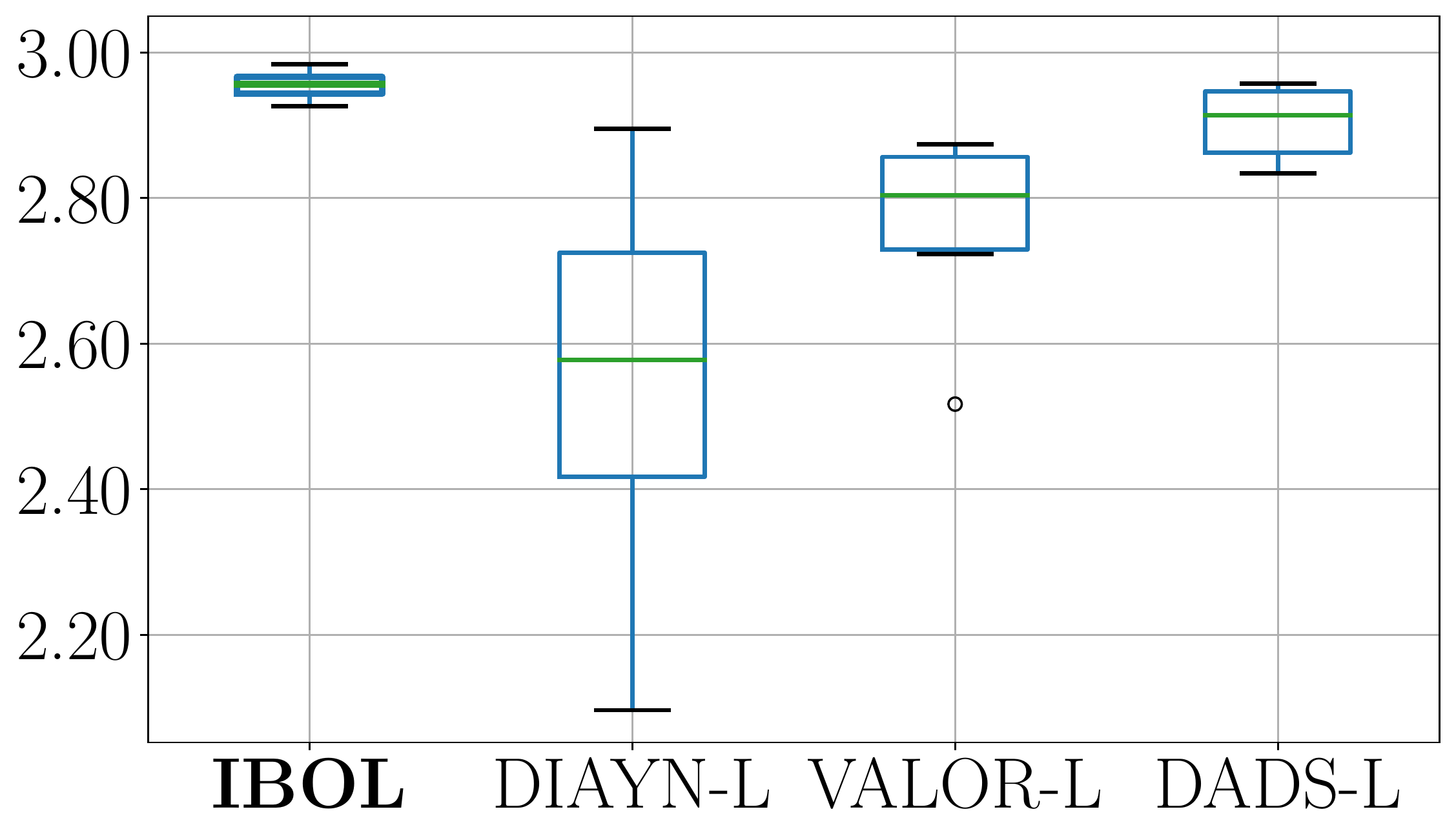}
  \end{subfigure}
  \begin{subfigure}[t]{0.1913187986\linewidth}
    \includegraphics[width=1.0\columnwidth]{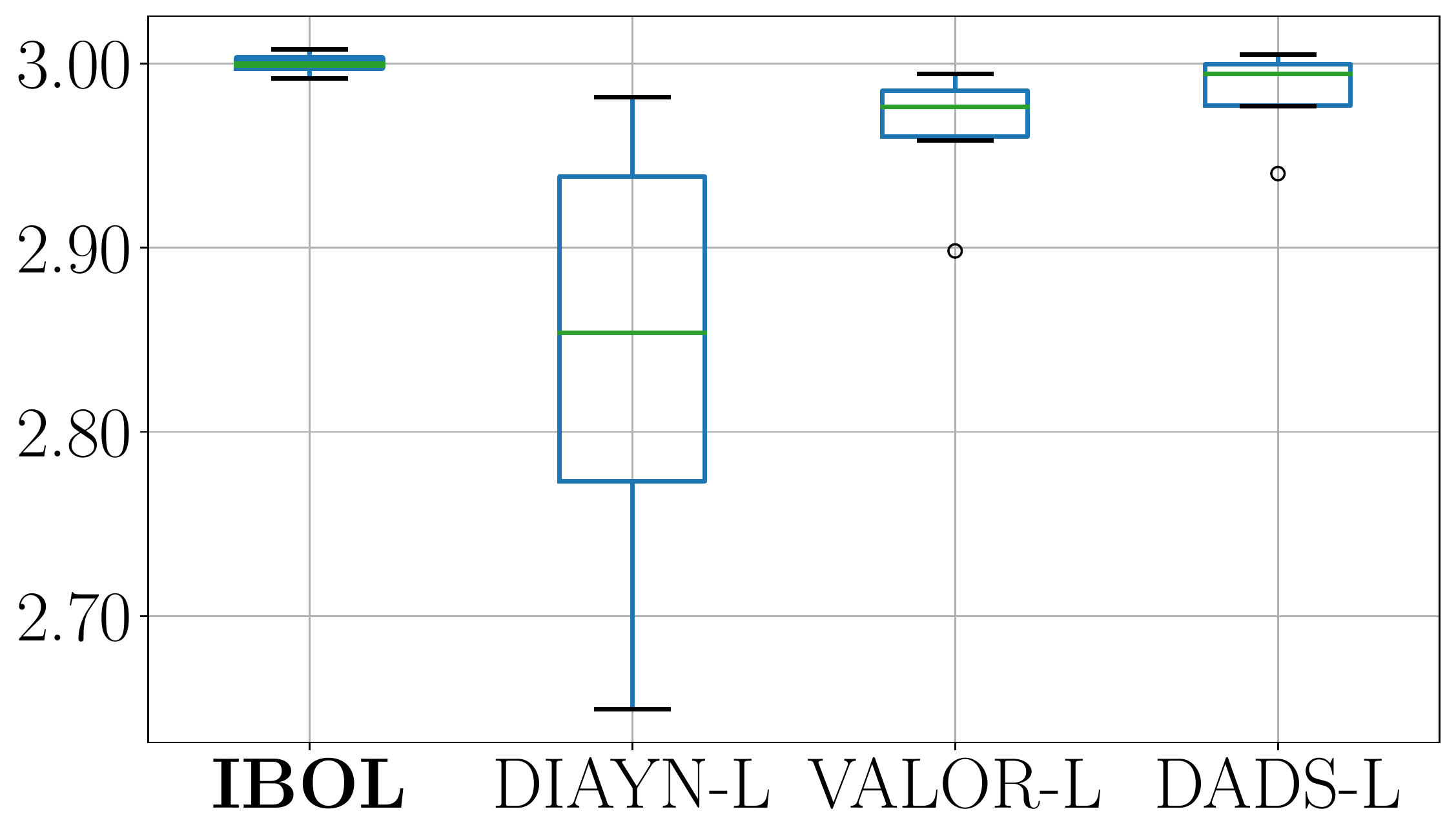}
  \end{subfigure}

  \hfill
  \begin{subfigure}[t]{0.2008490516\linewidth}
    \includegraphics[width=1.0\columnwidth]{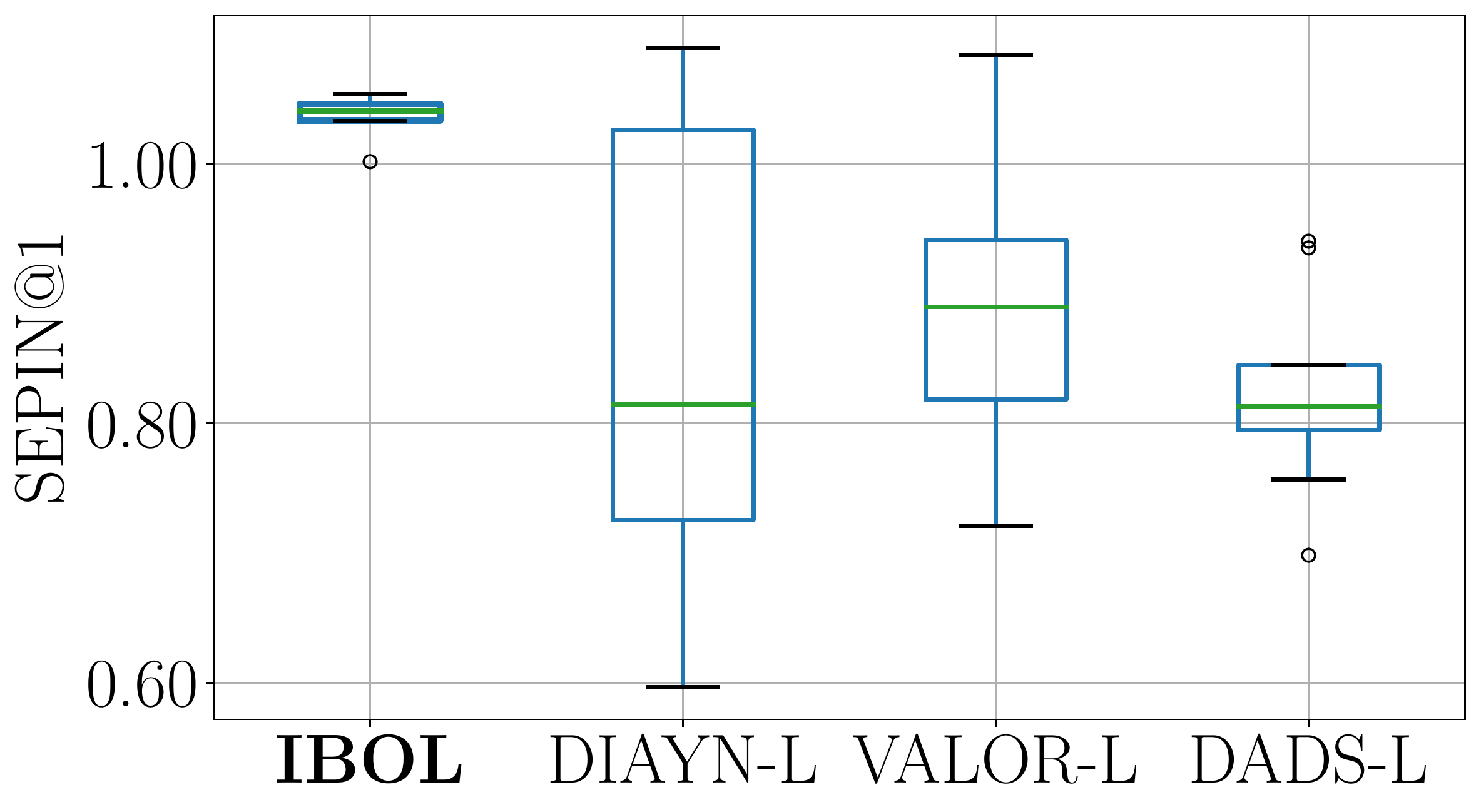}
    \caption{\# bins = 8}
  \end{subfigure}
  \begin{subfigure}[t]{0.1913187986\linewidth}
    \includegraphics[width=1.0\columnwidth]{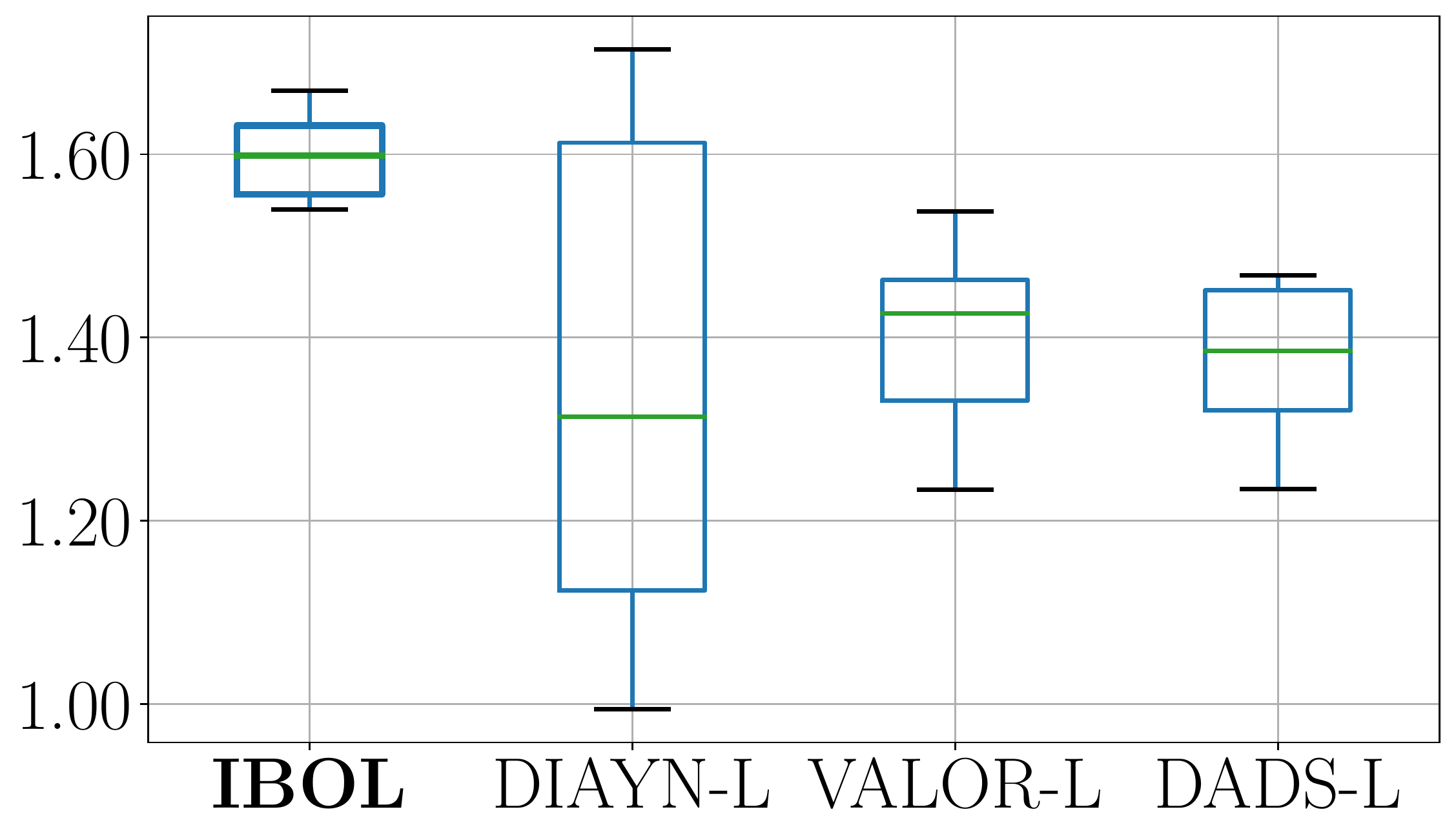}
    \caption{\# bins = 16}
  \end{subfigure}
  \begin{subfigure}[t]{0.1913187986\linewidth}
    \includegraphics[width=1.0\columnwidth]{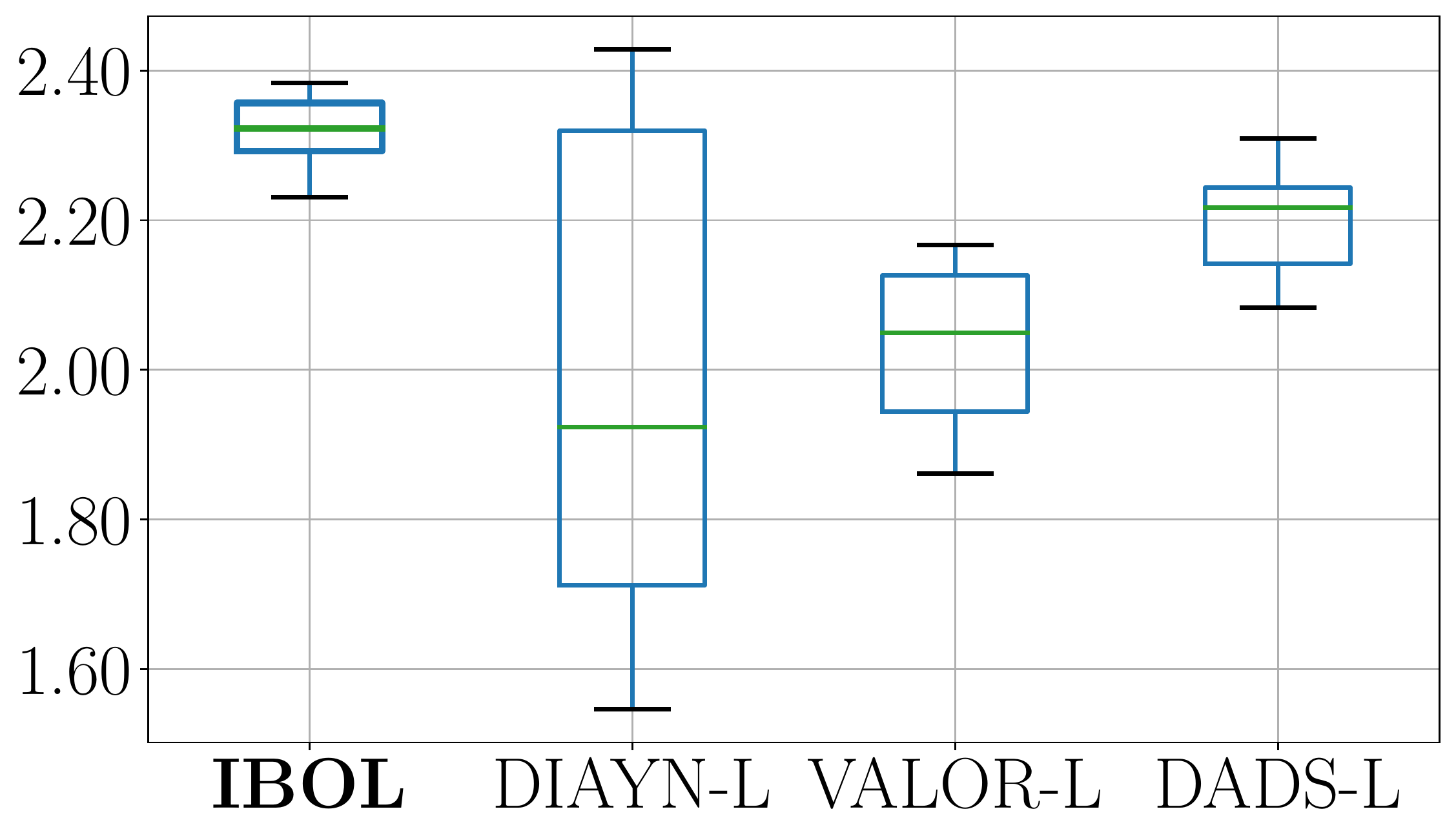}
    \caption{\# bins = 32}
  \end{subfigure}
  \begin{subfigure}[t]{0.1913187986\linewidth}
    \includegraphics[width=1.0\columnwidth]{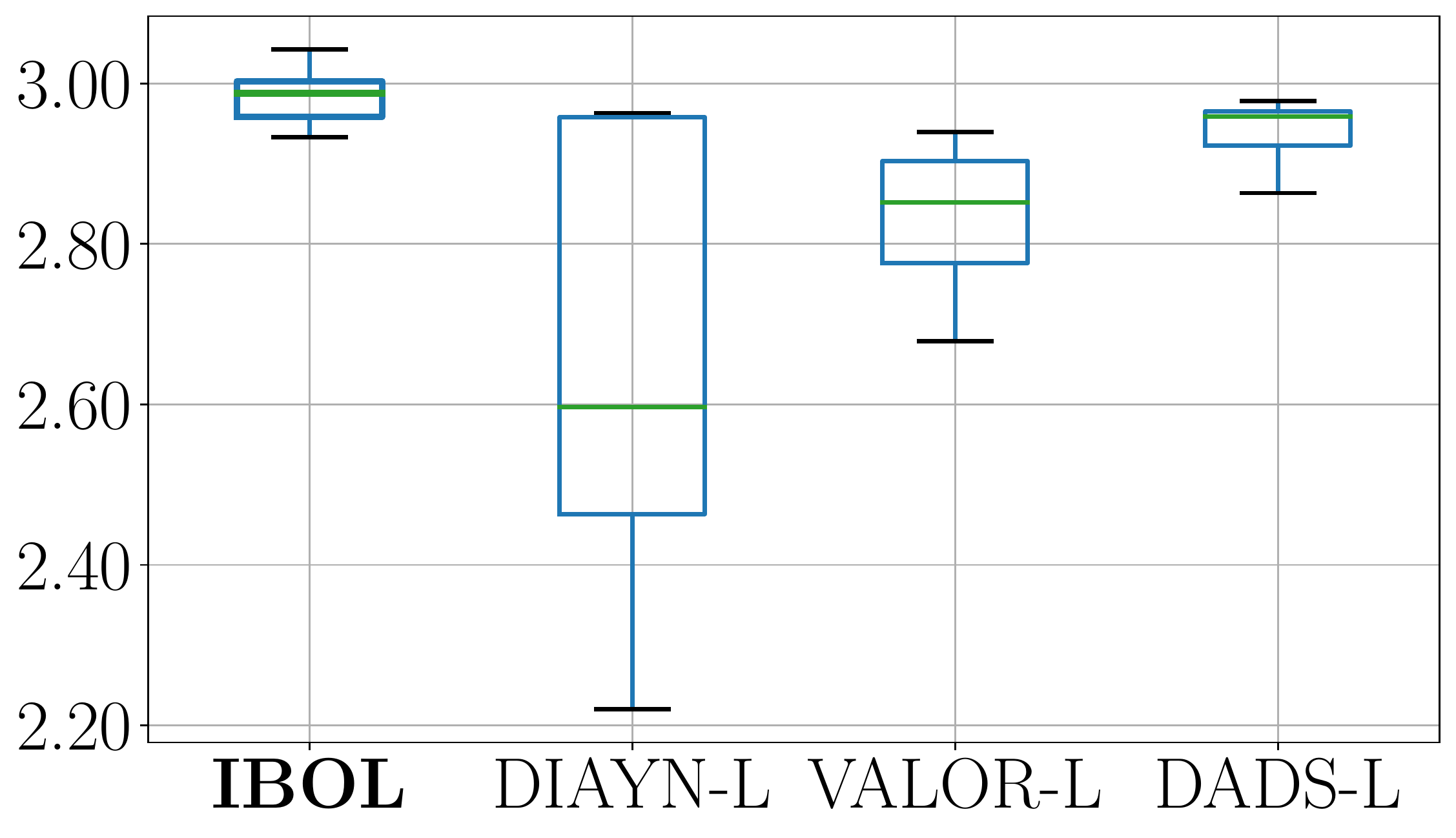}
    \caption{\# bins = 64}
  \end{subfigure}
  \begin{subfigure}[t]{0.1913187986\linewidth}
    \includegraphics[width=1.0\columnwidth]{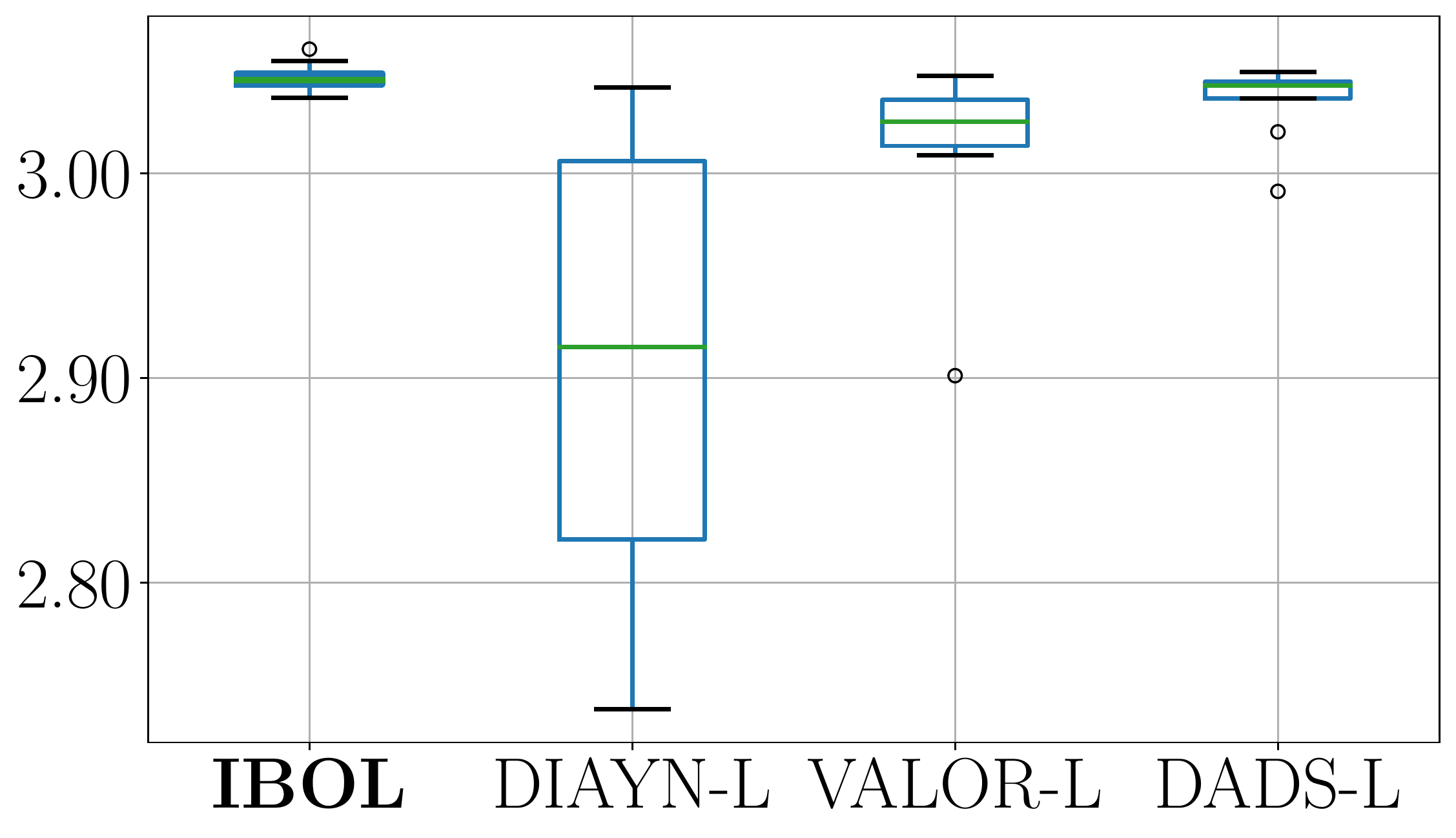}
    \caption{\# bins = 128}
  \end{subfigure}

  \caption{
    Comparison of IBOL (ours) with the baseline methods, DIAYN-L, VALOR-L and DADS-L,
    in the evaluation metrics of $I(Z; S_T^{\text{(loc)}})$, WSEPIN and SEPIN$@1$, on Ant,
    with different bin counts for the range of each variable estimating mutual information.
    For each method, we use the eight trained skill policies.
  }
  \label{fig:eval_metrics_num_bins_ant}
\end{figure*}

\begin{figure*}[t!]
  \centering

  \hfill
  \begin{subfigure}[t]{0.2047248057\linewidth}
    \includegraphics[width=1.0\columnwidth]{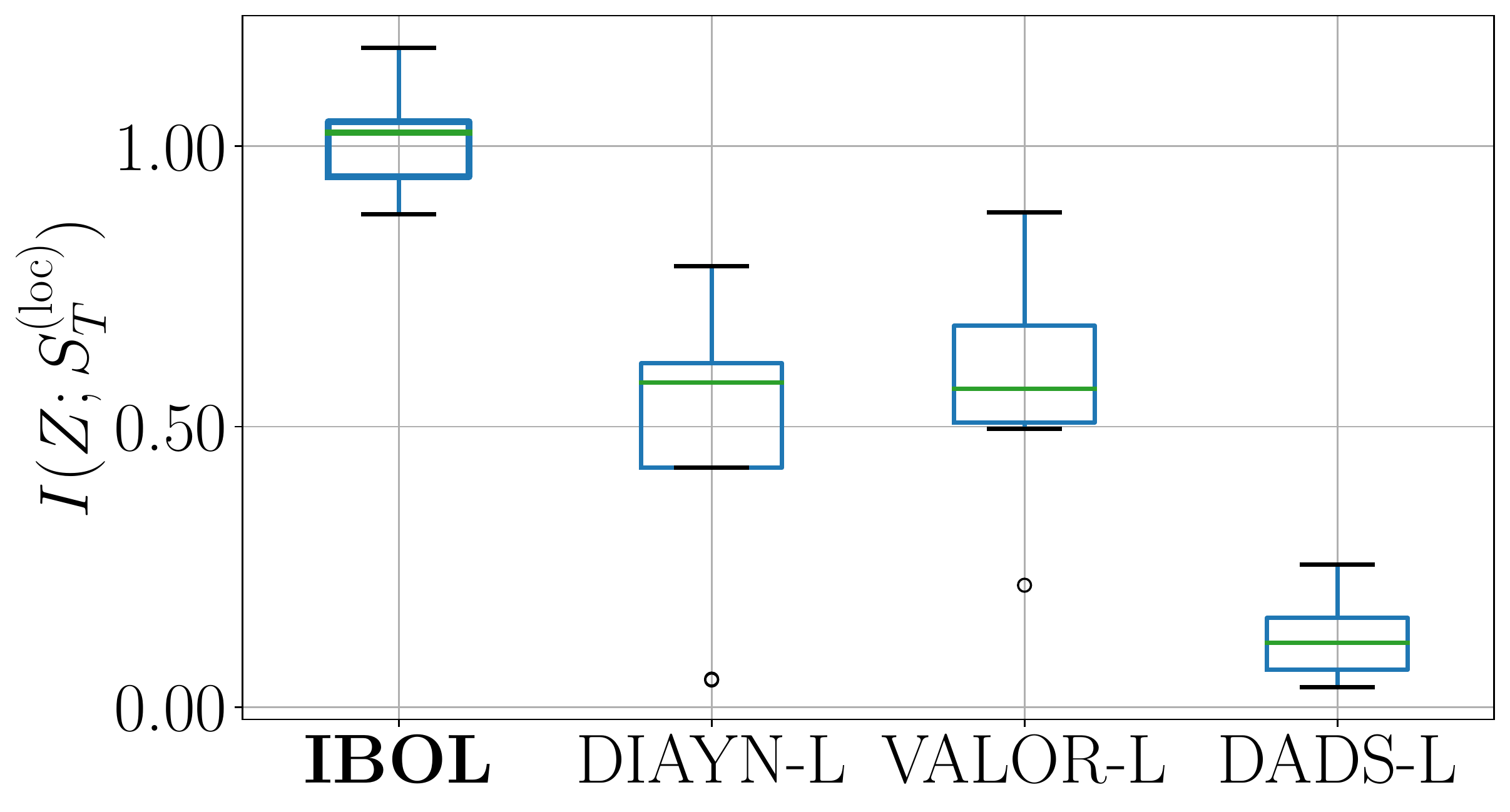}
  \end{subfigure}
  \begin{subfigure}[t]{0.1913187986\linewidth}
    \includegraphics[width=1.0\columnwidth]{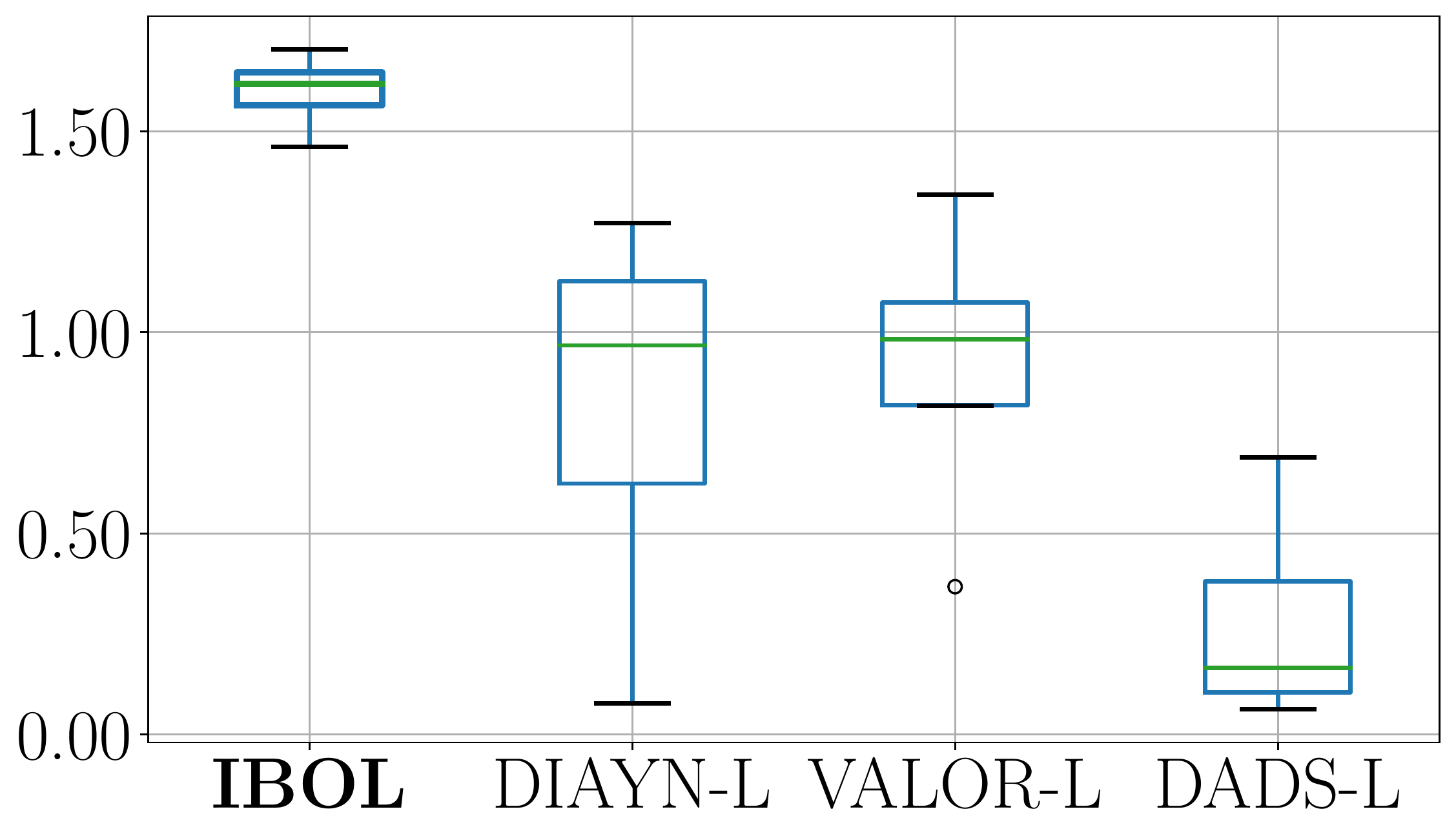}
  \end{subfigure}
  \begin{subfigure}[t]{0.1913187986\linewidth}
    \includegraphics[width=1.0\columnwidth]{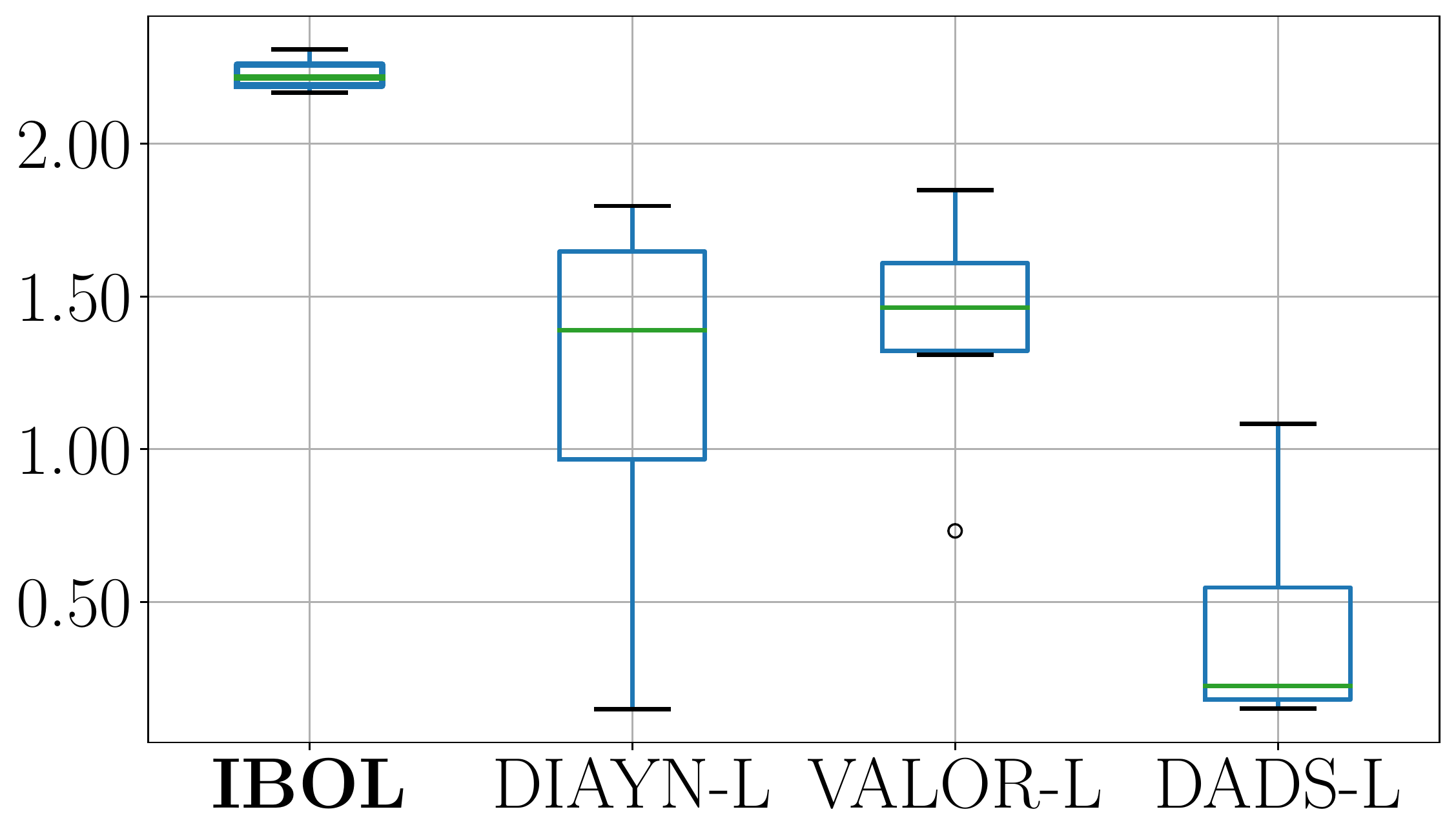}
  \end{subfigure}
  \begin{subfigure}[t]{0.1913187986\linewidth}
    \includegraphics[width=1.0\columnwidth]{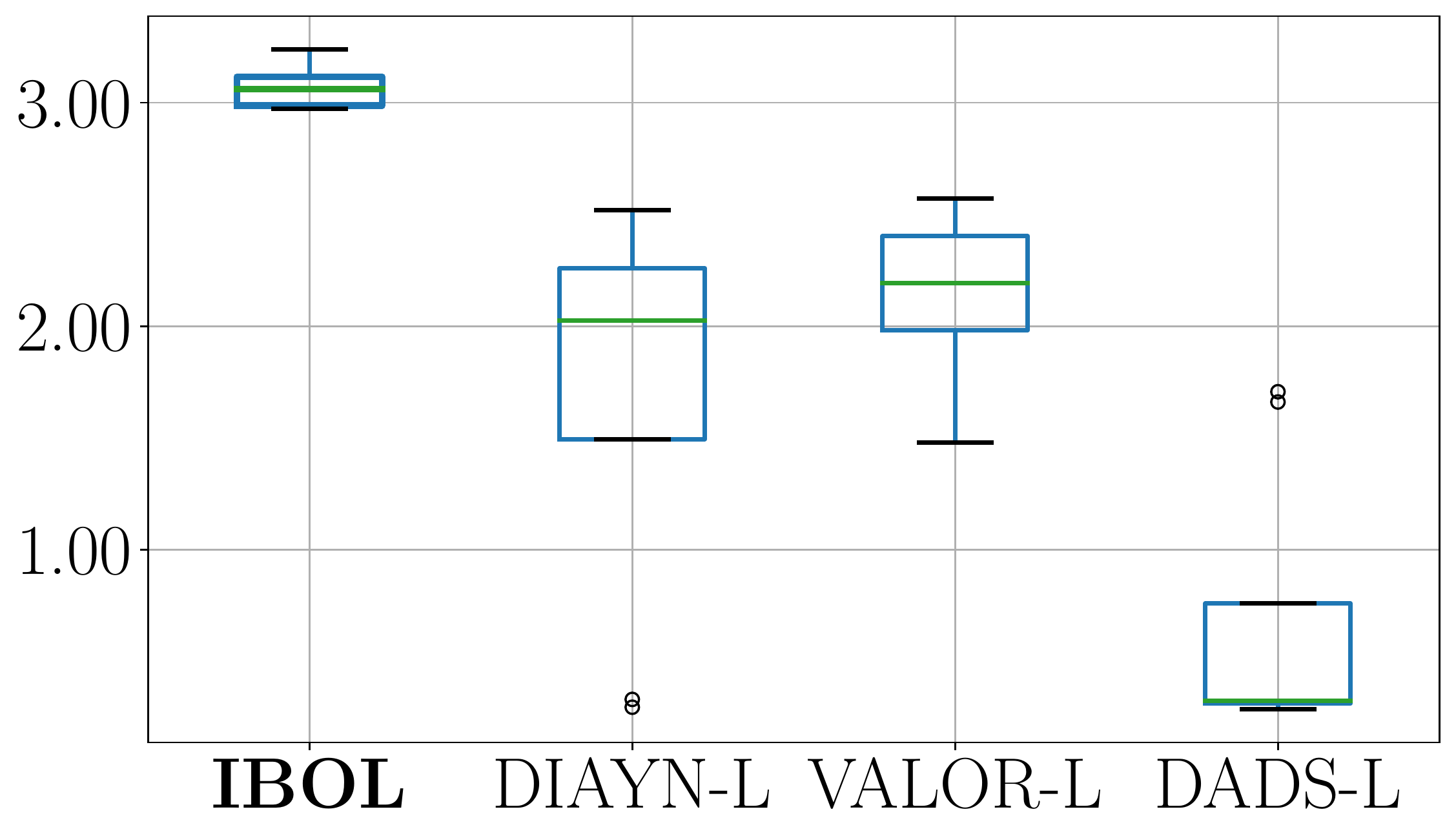}
  \end{subfigure}
  \begin{subfigure}[t]{0.1913187986\linewidth}
    \includegraphics[width=1.0\columnwidth]{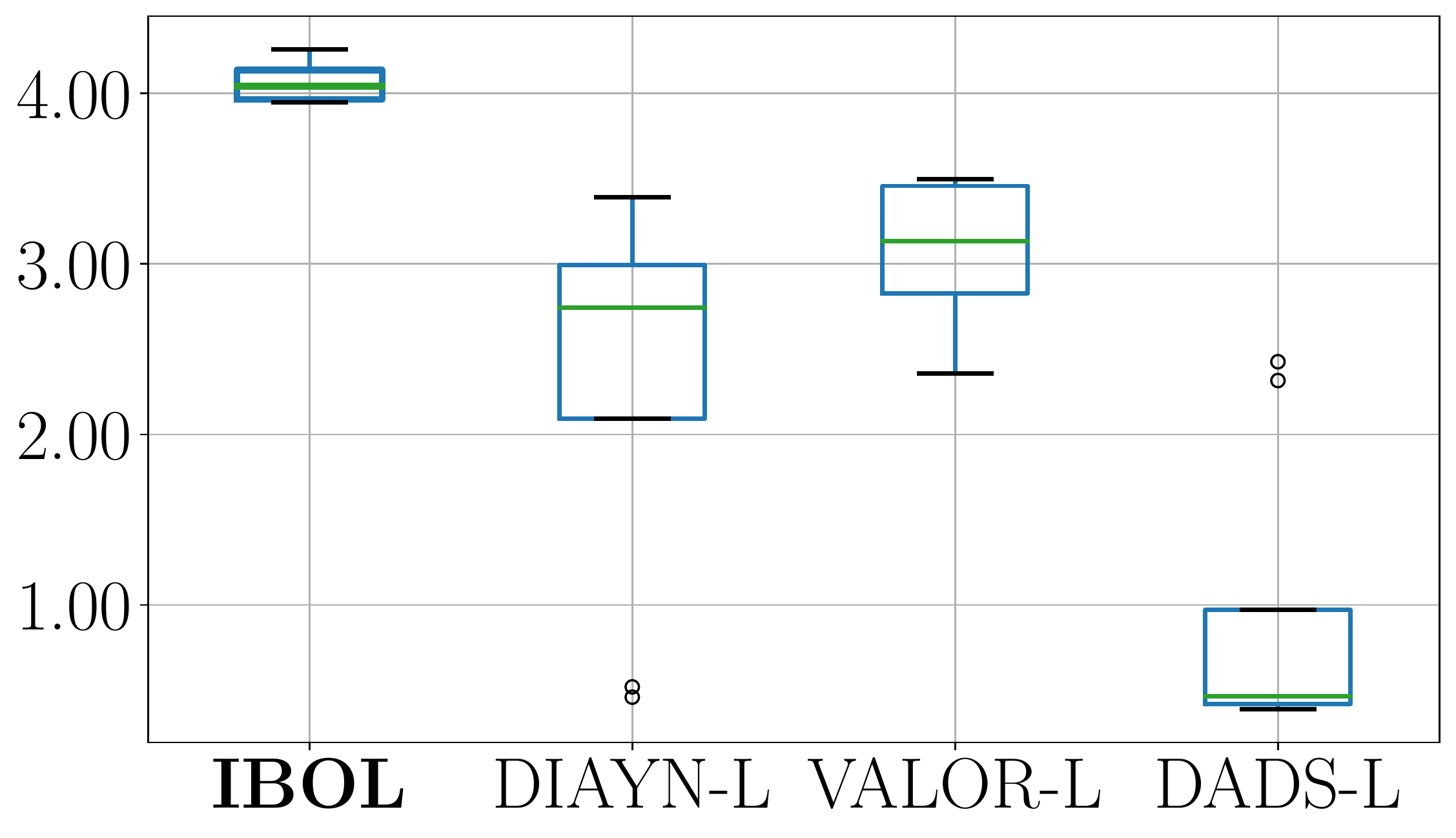}
  \end{subfigure}

  \hfill
  \begin{subfigure}[t]{0.2008490516\linewidth}
    \includegraphics[width=1.0\columnwidth]{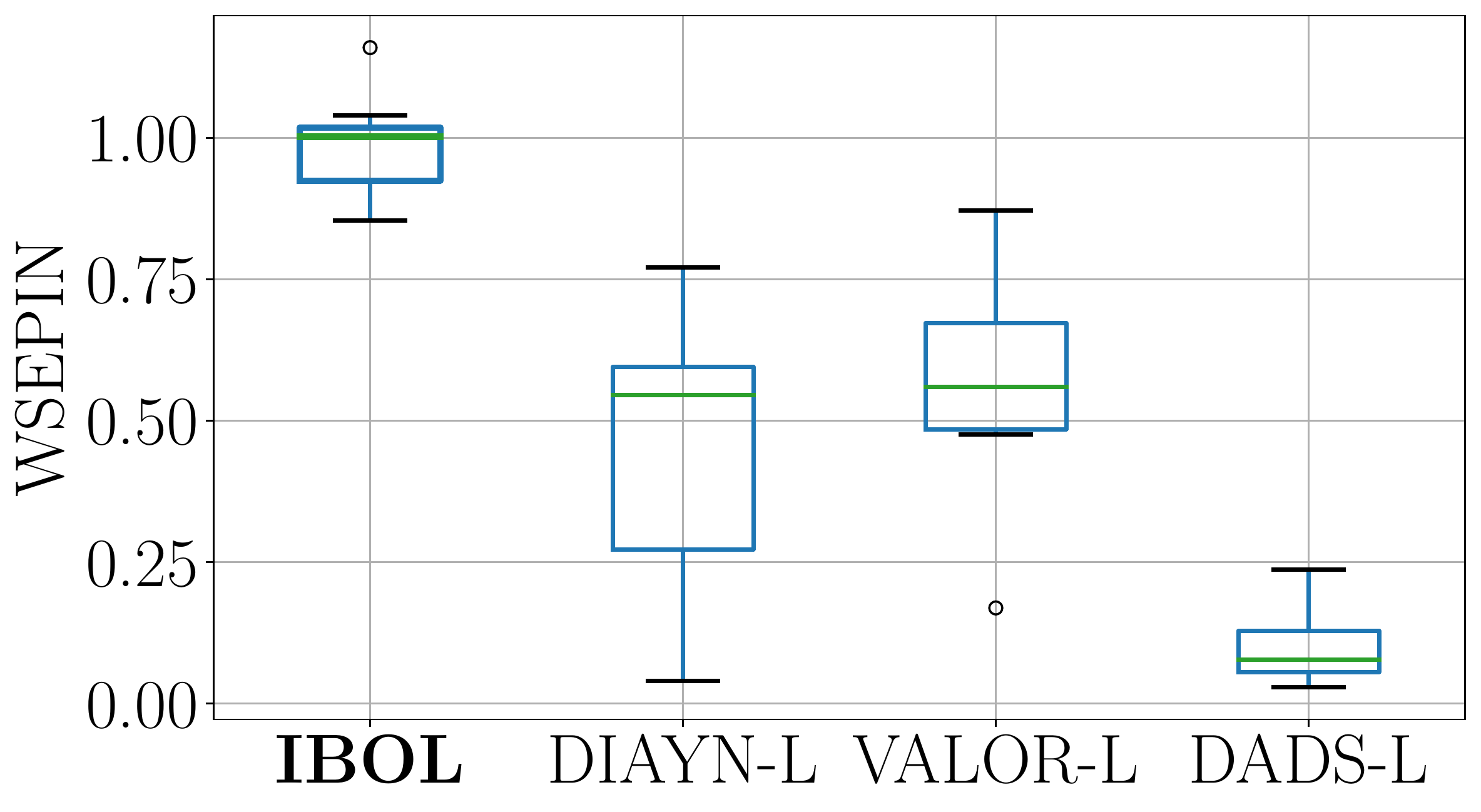}
  \end{subfigure}
  \begin{subfigure}[t]{0.1913187986\linewidth}
    \includegraphics[width=1.0\columnwidth]{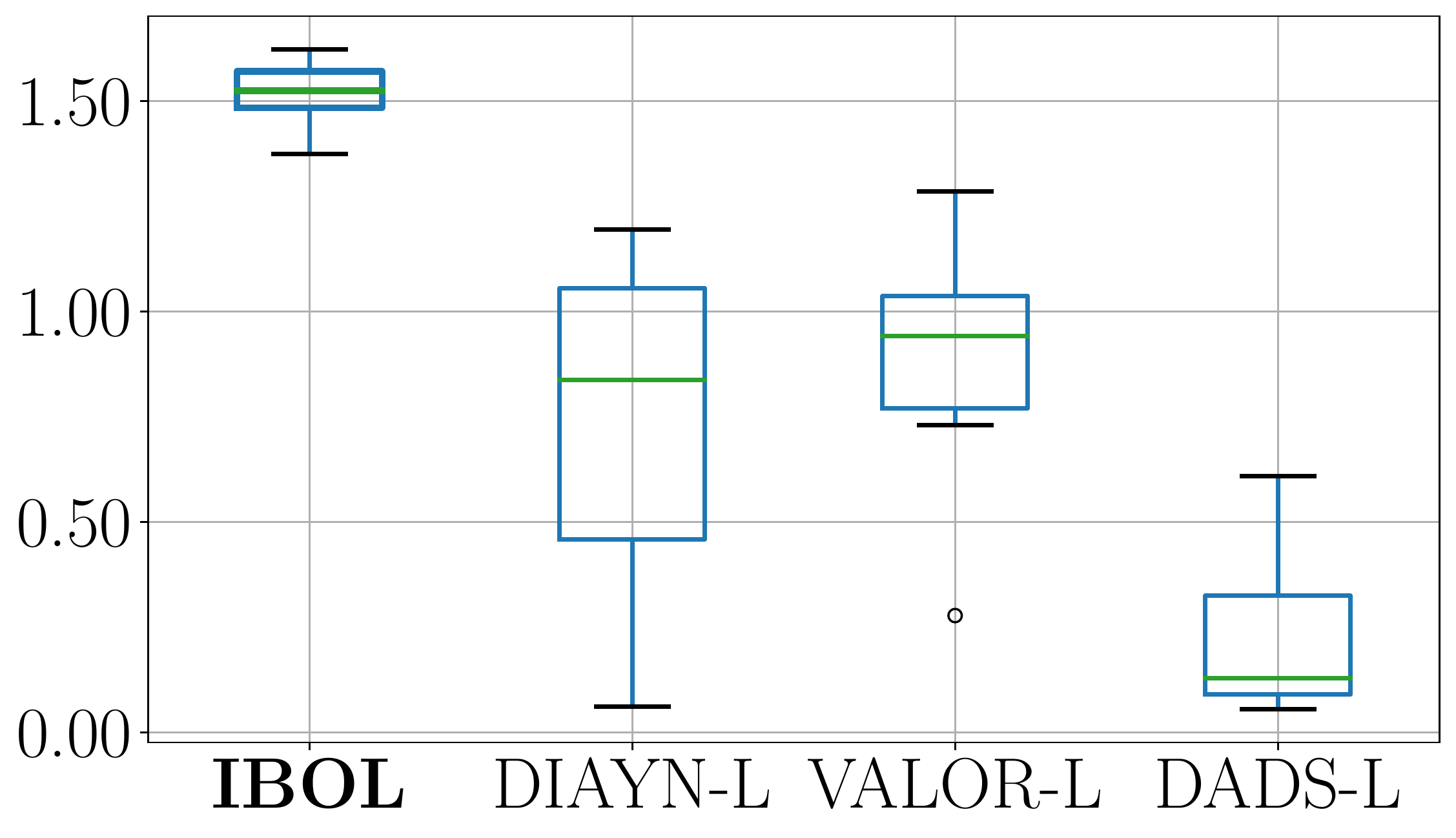}
  \end{subfigure}
  \begin{subfigure}[t]{0.1913187986\linewidth}
    \includegraphics[width=1.0\columnwidth]{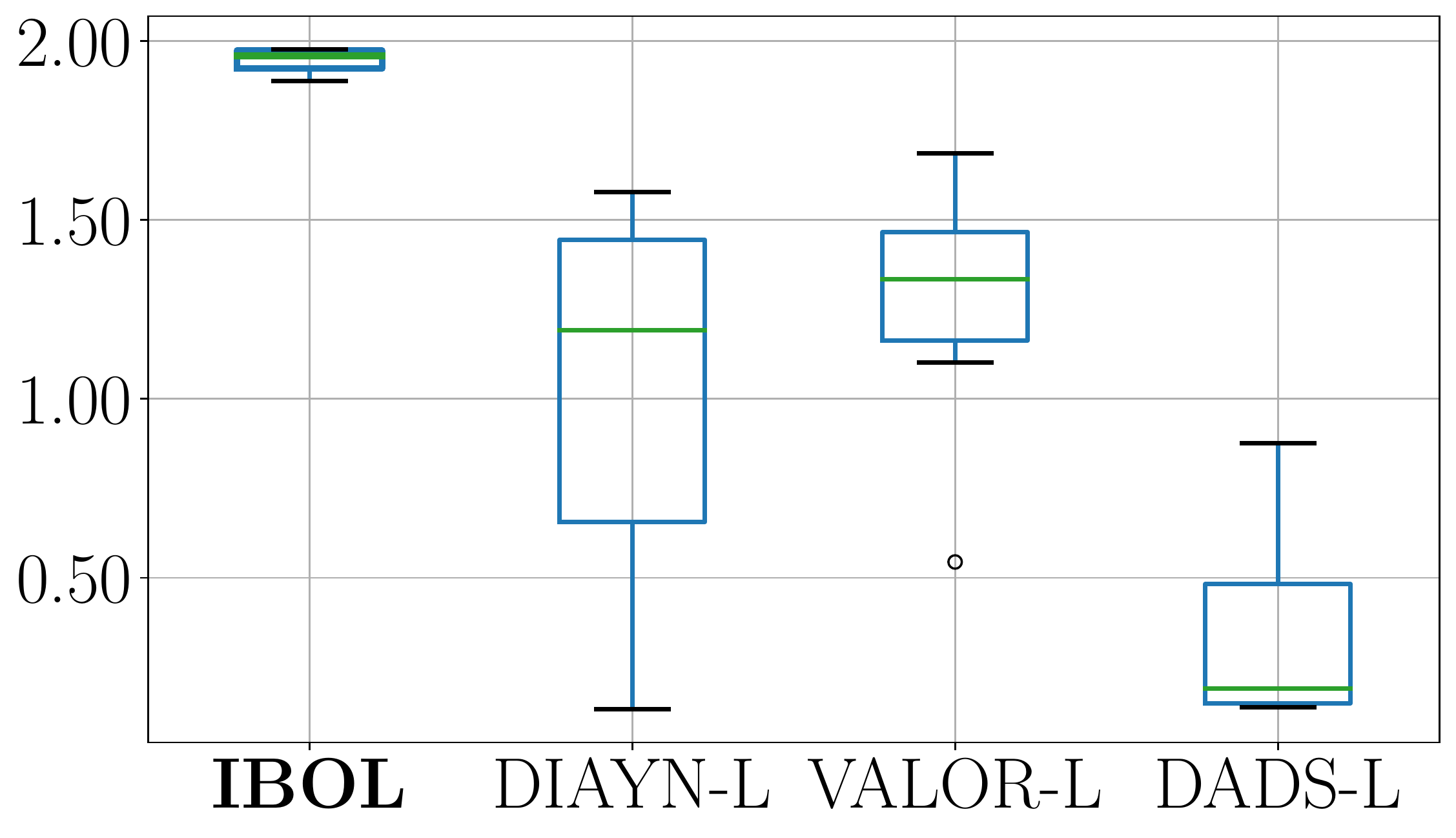}
  \end{subfigure}
  \begin{subfigure}[t]{0.1913187986\linewidth}
    \includegraphics[width=1.0\columnwidth]{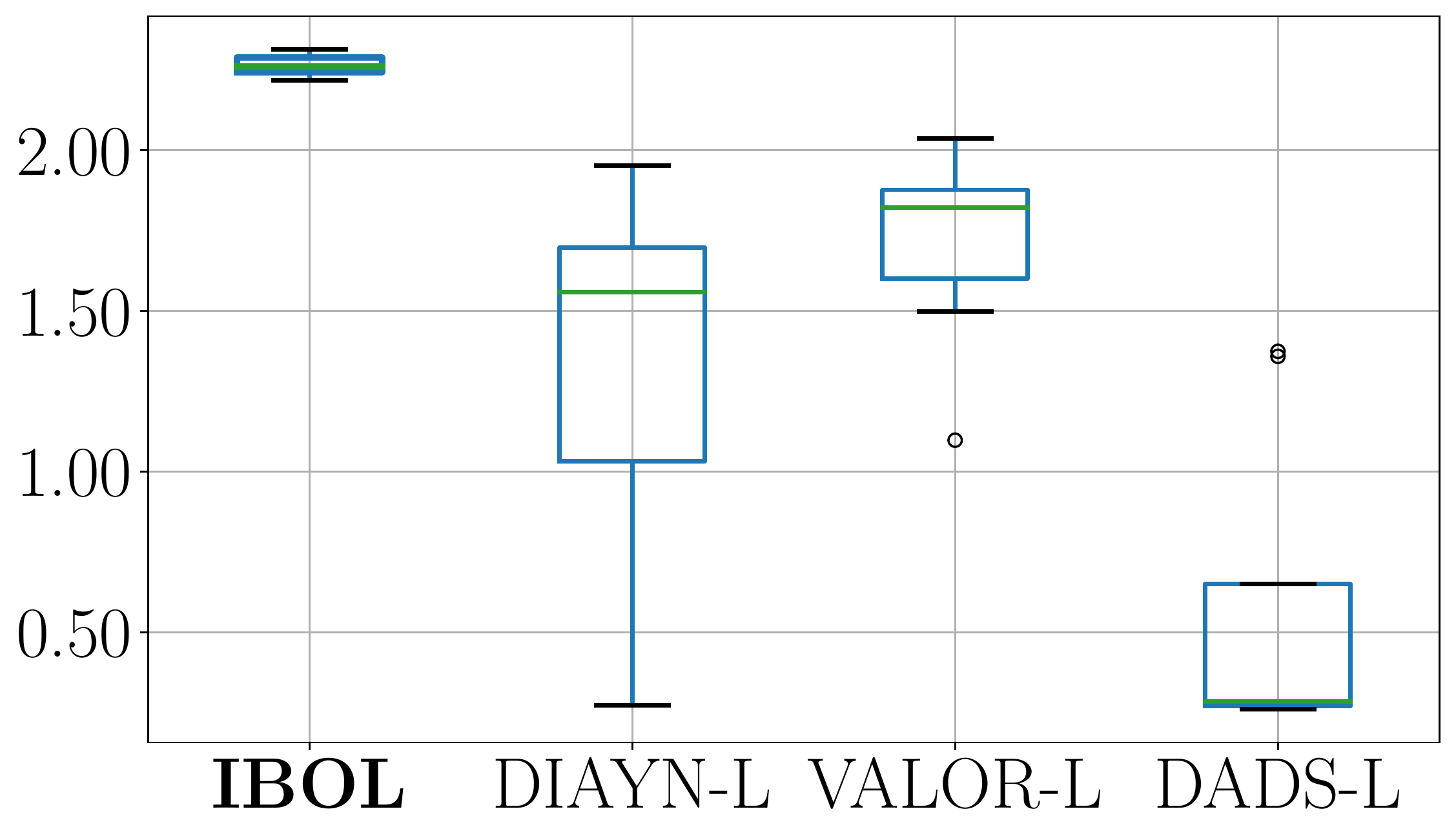}
  \end{subfigure}
  \begin{subfigure}[t]{0.1913187986\linewidth}
    \includegraphics[width=1.0\columnwidth]{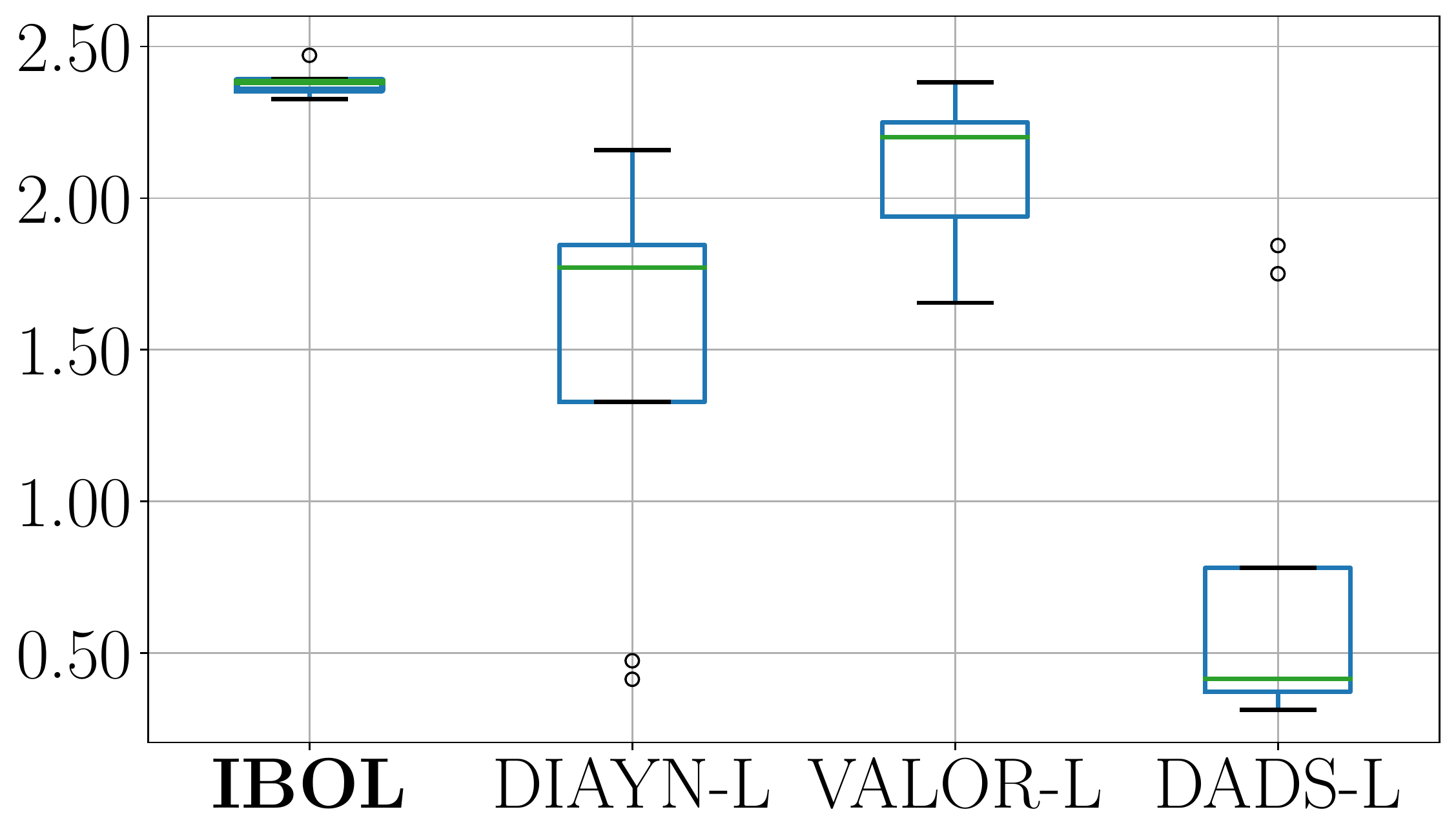}
  \end{subfigure}

  \hfill
  \begin{subfigure}[t]{0.2008490516\linewidth}
    \includegraphics[width=1.0\columnwidth]{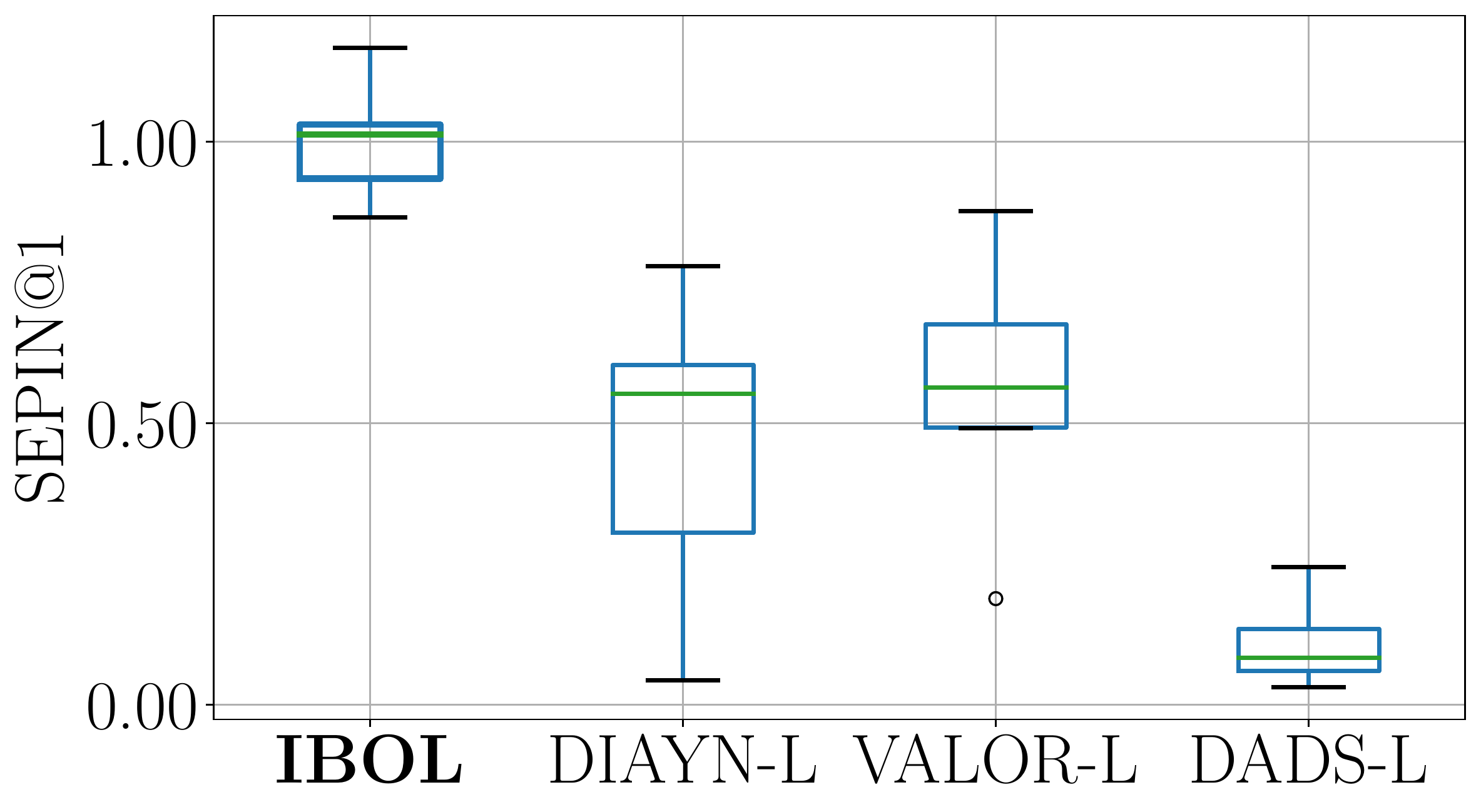}
    \caption{\# bins = 8}
  \end{subfigure}
  \begin{subfigure}[t]{0.1913187986\linewidth}
    \includegraphics[width=1.0\columnwidth]{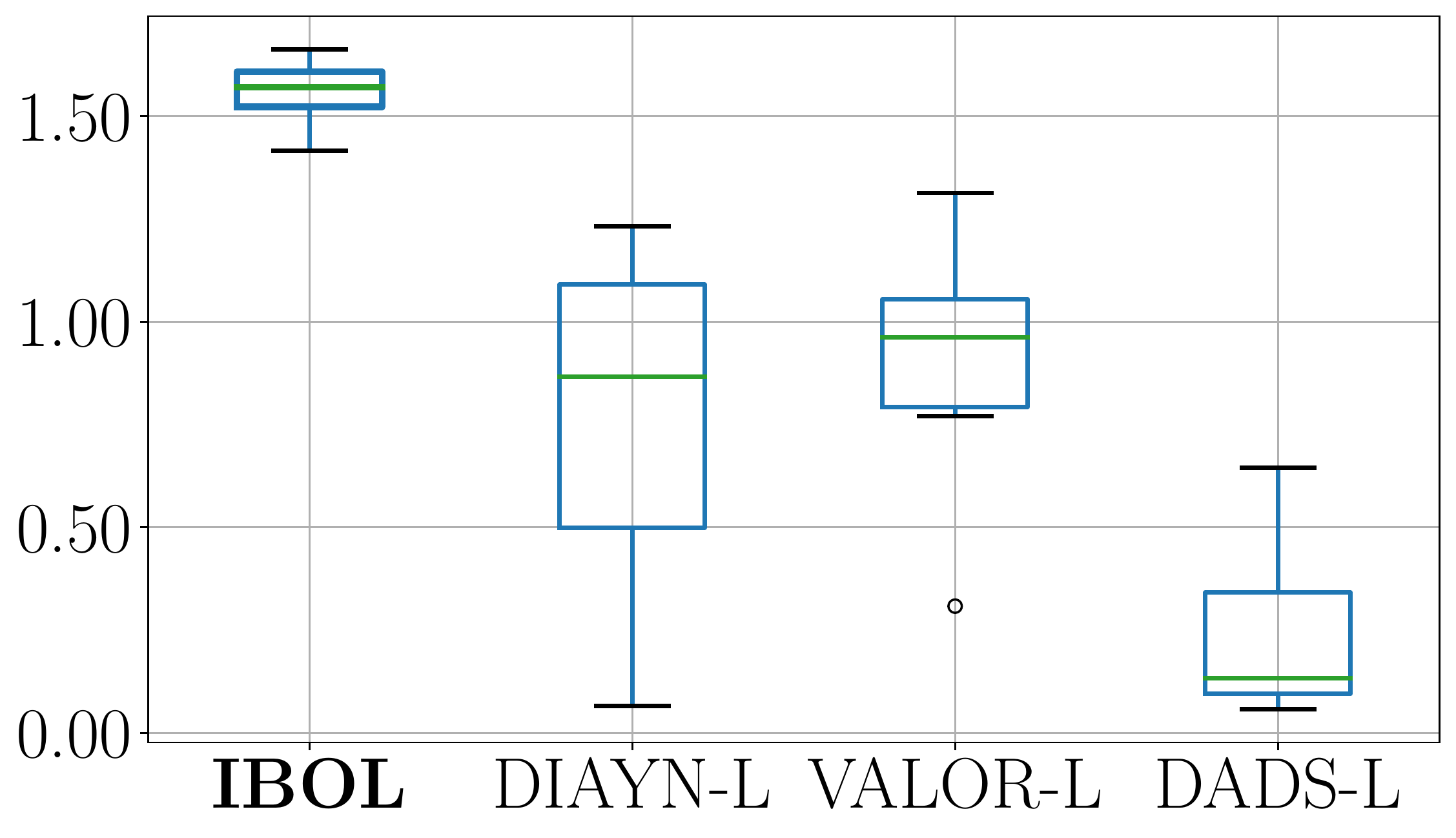}
    \caption{\# bins = 16}
  \end{subfigure}
  \begin{subfigure}[t]{0.1913187986\linewidth}
    \includegraphics[width=1.0\columnwidth]{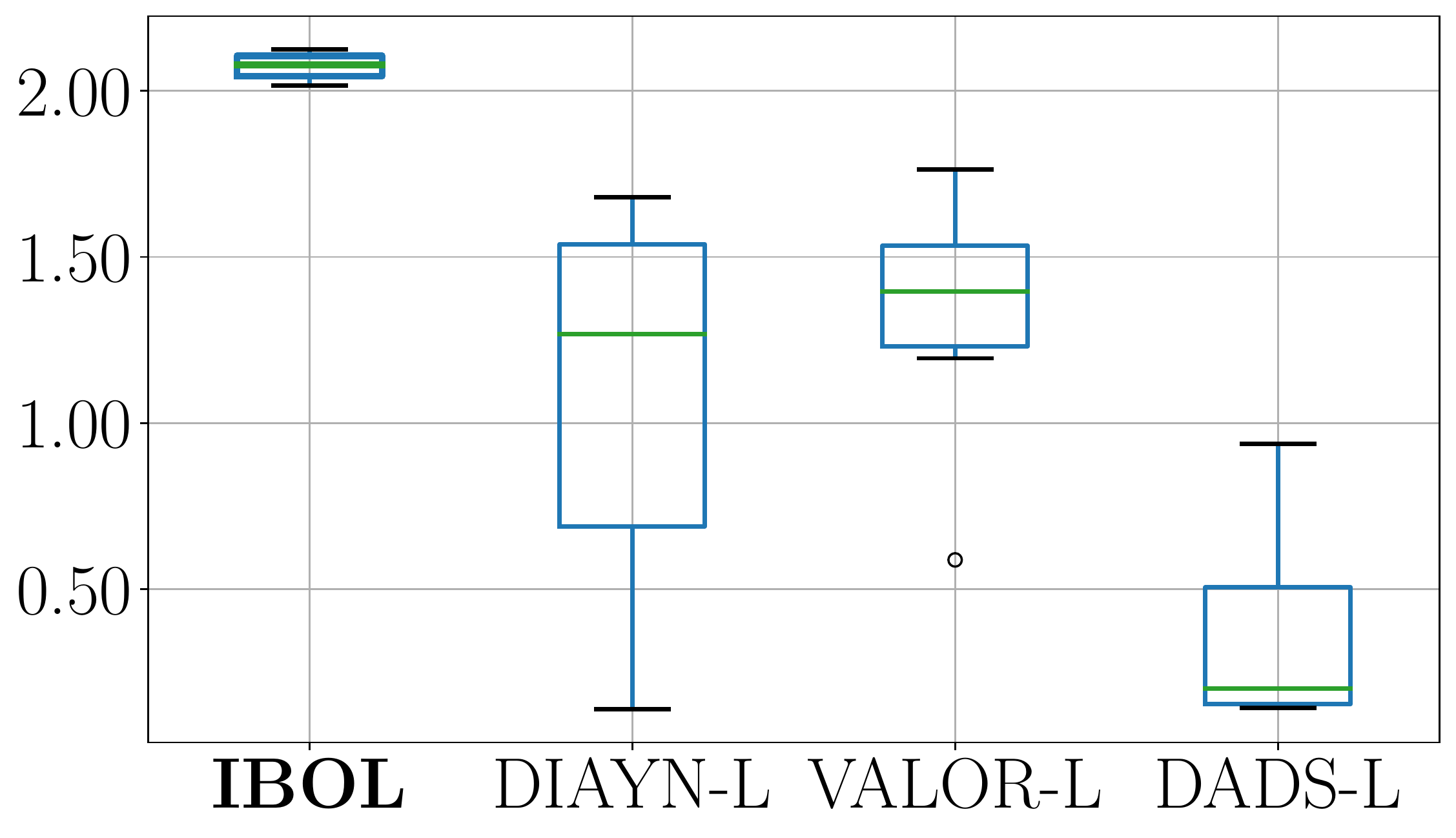}
    \caption{\# bins = 32}
  \end{subfigure}
  \begin{subfigure}[t]{0.1913187986\linewidth}
    \includegraphics[width=1.0\columnwidth]{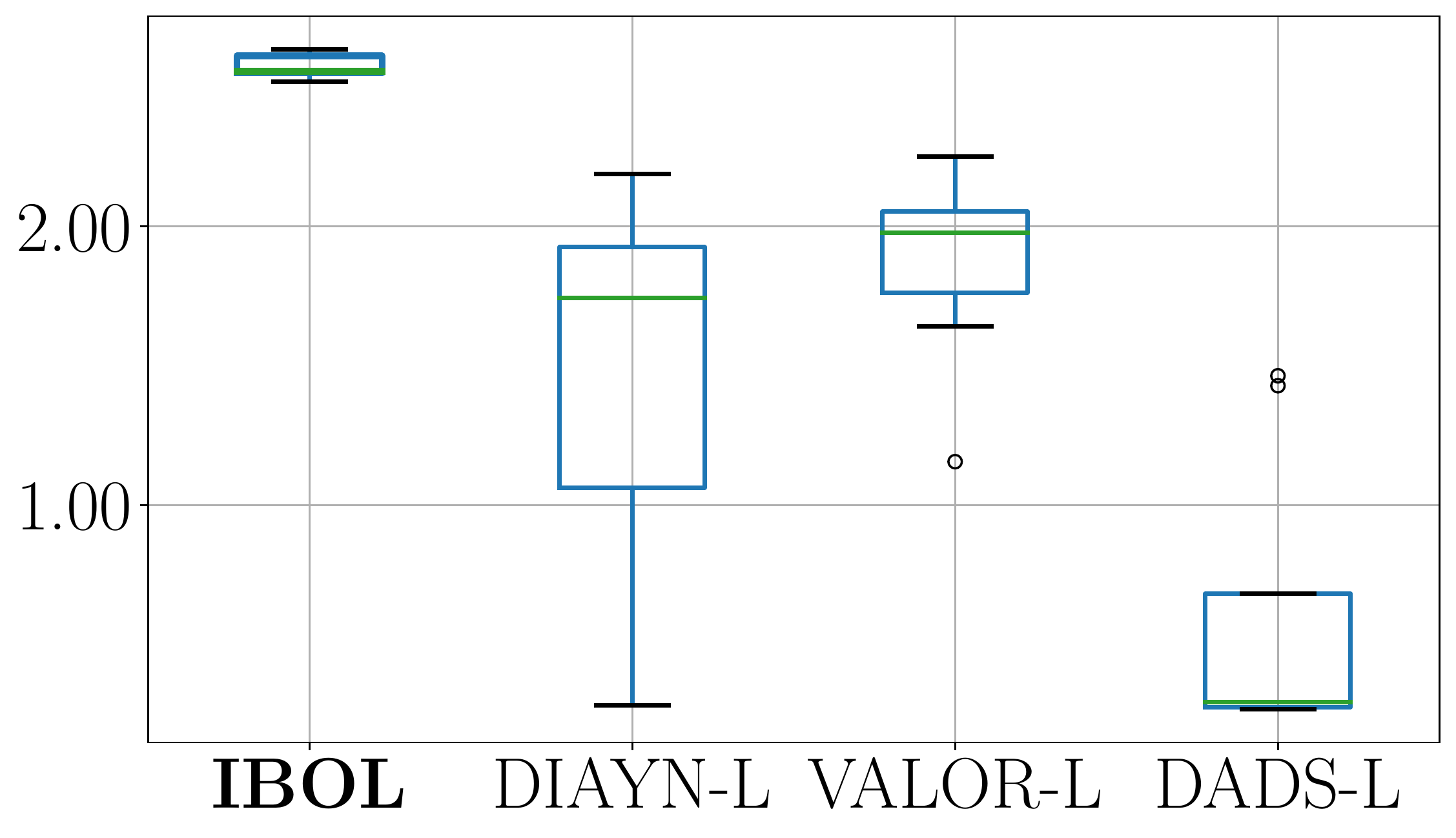}
    \caption{\# bins = 64}
  \end{subfigure}
  \begin{subfigure}[t]{0.1913187986\linewidth}
    \includegraphics[width=1.0\columnwidth]{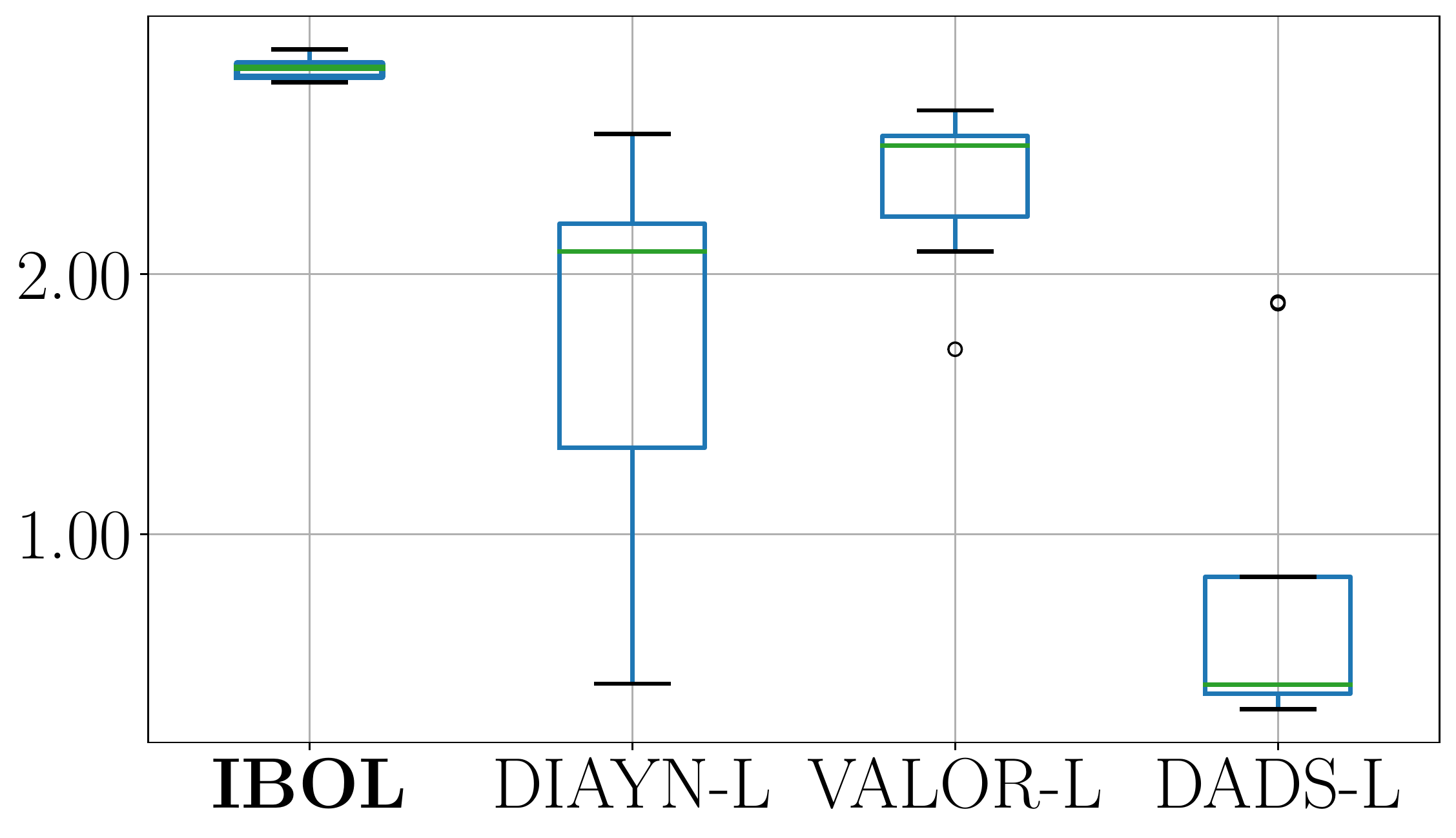}
    \caption{\# bins = 128}
  \end{subfigure}

  \caption{
    Comparison of IBOL (ours) with the baseline methods, DIAYN-L, VALOR-L and DADS-L,
    in the evaluation metrics of $I(Z; S_T^{\text{(loc)}})$, WSEPIN and SEPIN$@1$, on HalfCheetah,
    with different bin counts for the range of each variable estimating mutual information.
    For each method, we use the eight trained skill policies.
  }
  \label{fig:eval_metrics_num_bins_hc}
\end{figure*}

\begin{figure*}[t!]
  \centering

  \hfill
  \begin{subfigure}[t]{0.2047248057\linewidth}
    \includegraphics[width=1.0\columnwidth]{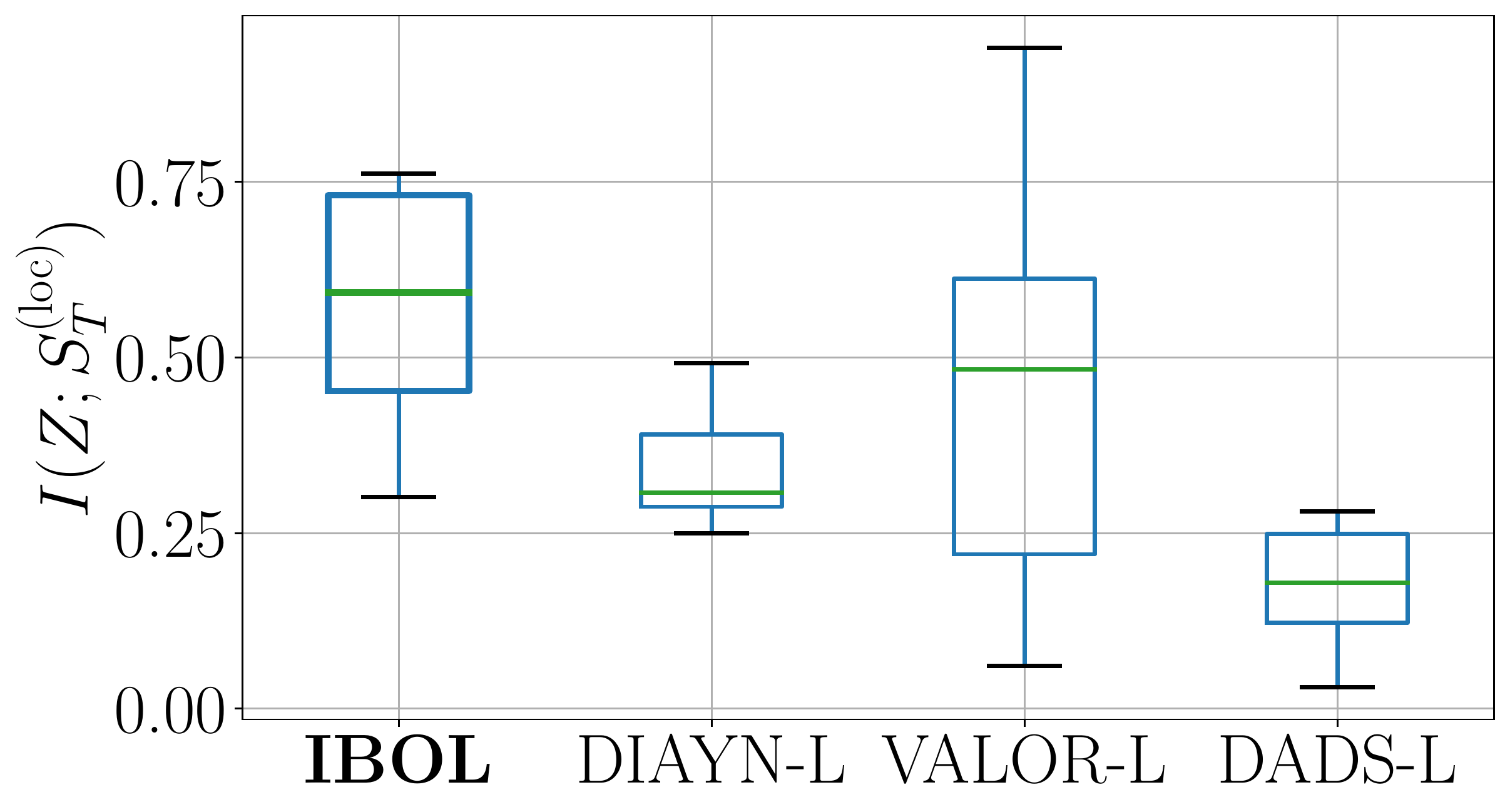}
  \end{subfigure}
  \begin{subfigure}[t]{0.1913187986\linewidth}
    \includegraphics[width=1.0\columnwidth]{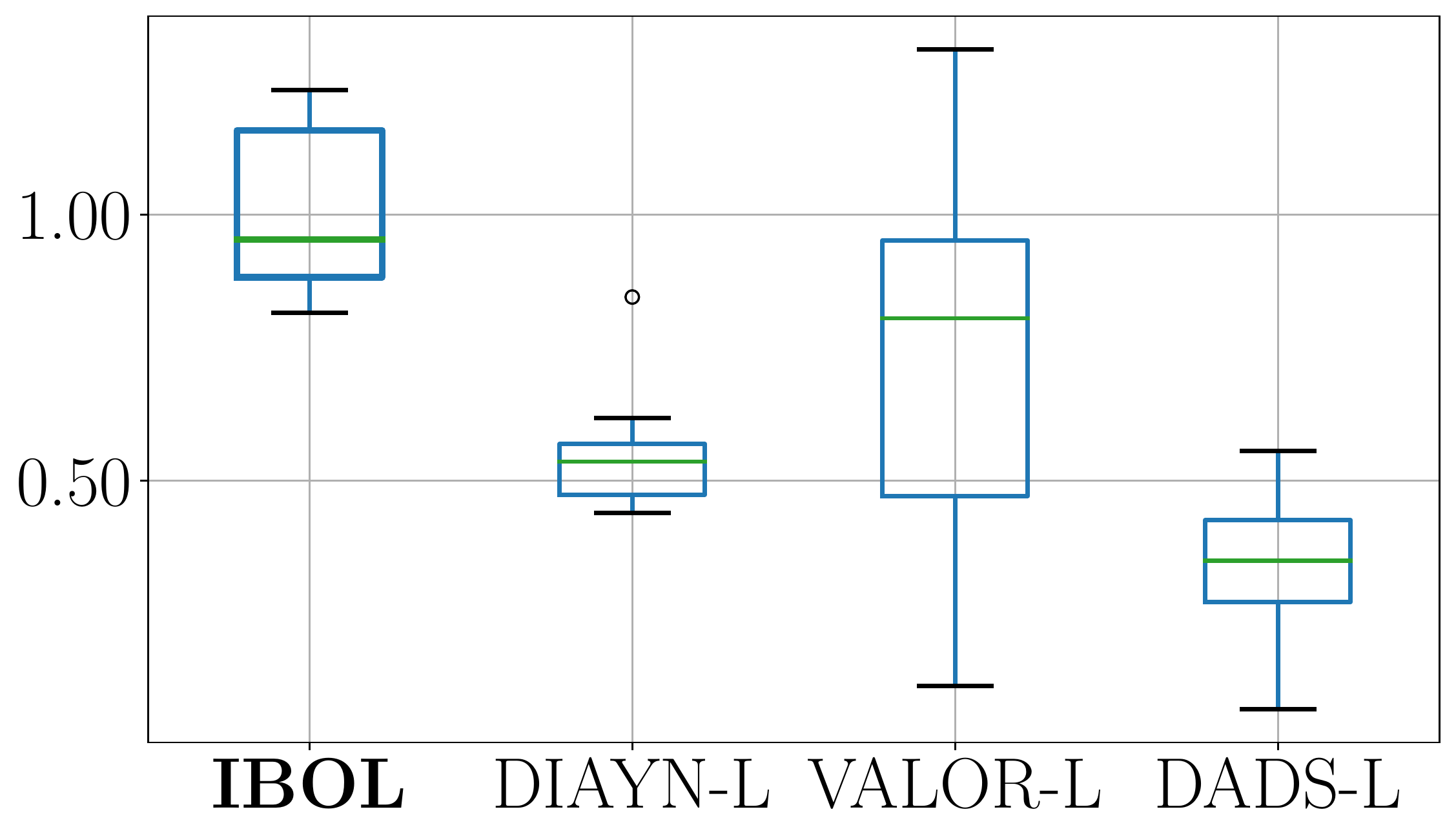}
  \end{subfigure}
  \begin{subfigure}[t]{0.1913187986\linewidth}
    \includegraphics[width=1.0\columnwidth]{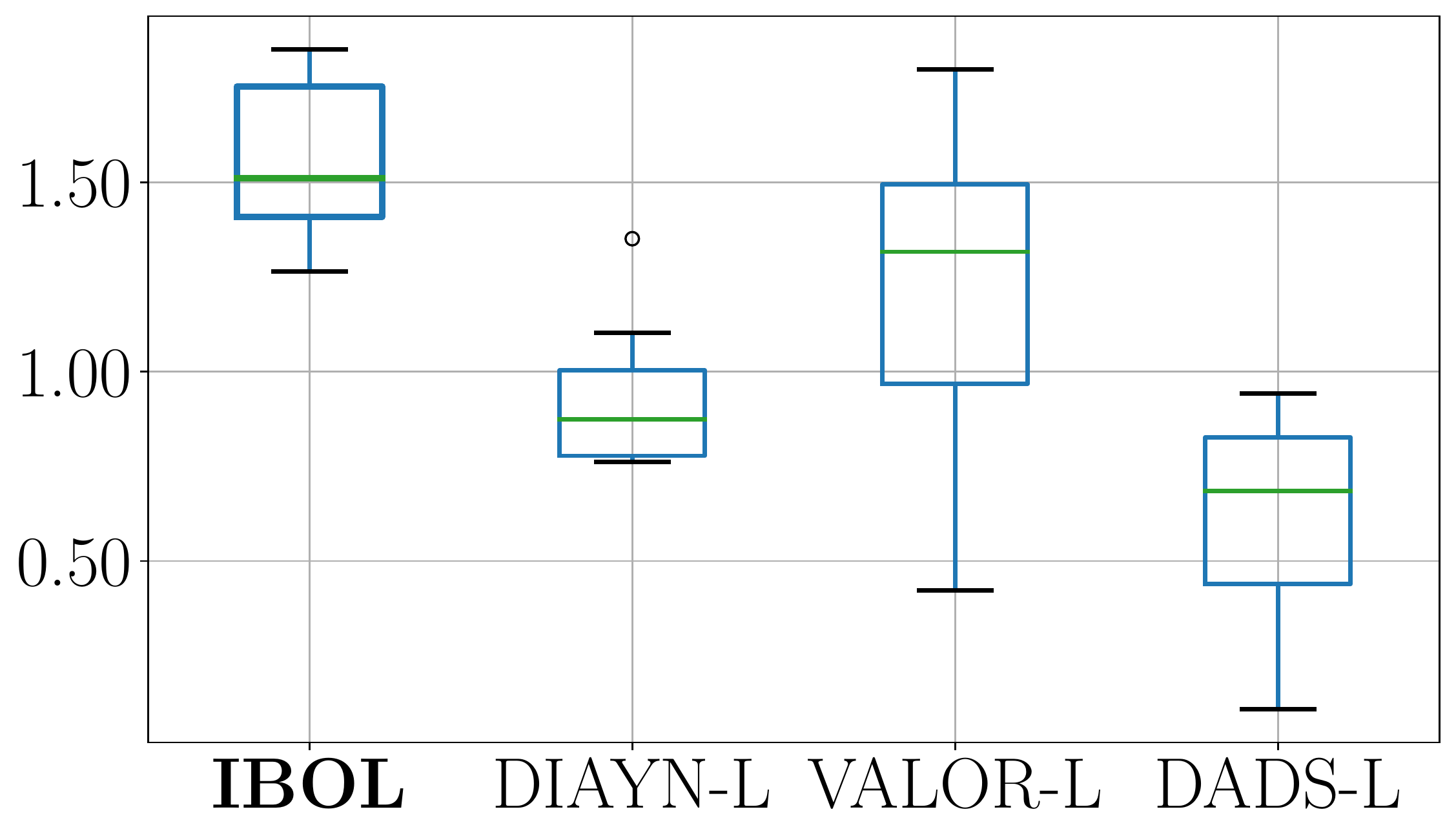}
  \end{subfigure}
  \begin{subfigure}[t]{0.1913187986\linewidth}
    \includegraphics[width=1.0\columnwidth]{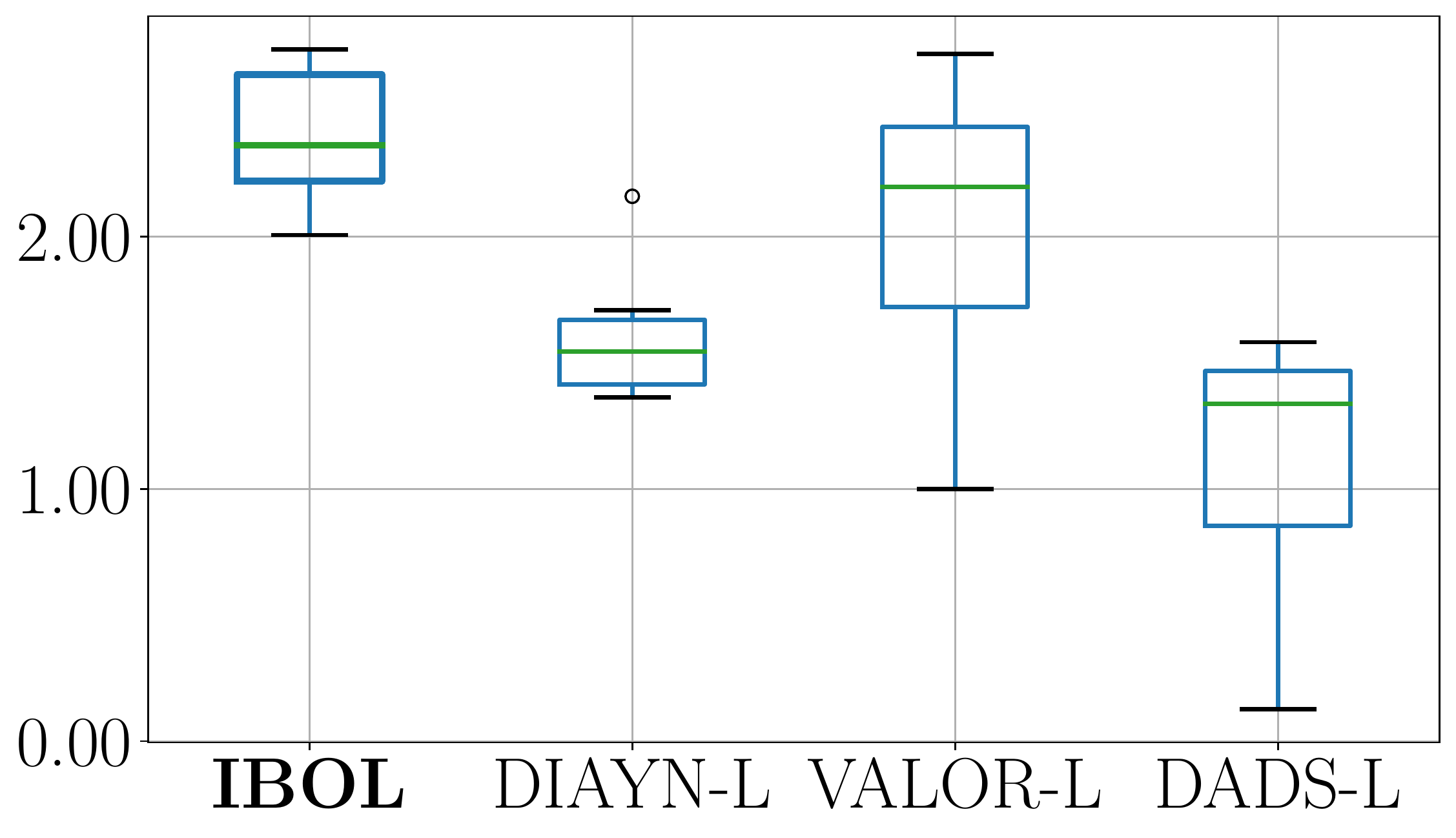}
  \end{subfigure}
  \begin{subfigure}[t]{0.1913187986\linewidth}
    \includegraphics[width=1.0\columnwidth]{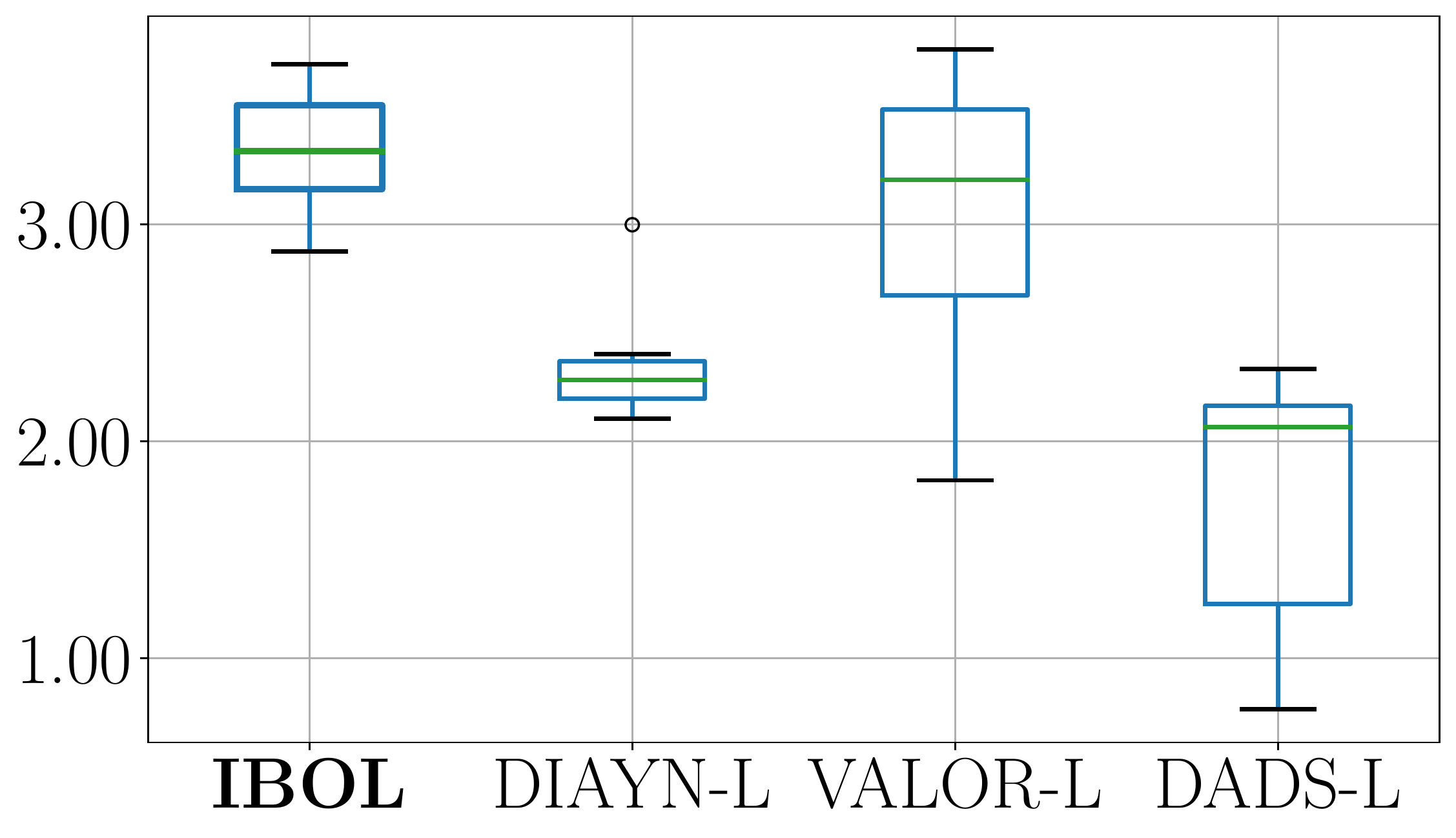}
  \end{subfigure}

  \hfill
  \begin{subfigure}[t]{0.2008490516\linewidth}
    \includegraphics[width=1.0\columnwidth]{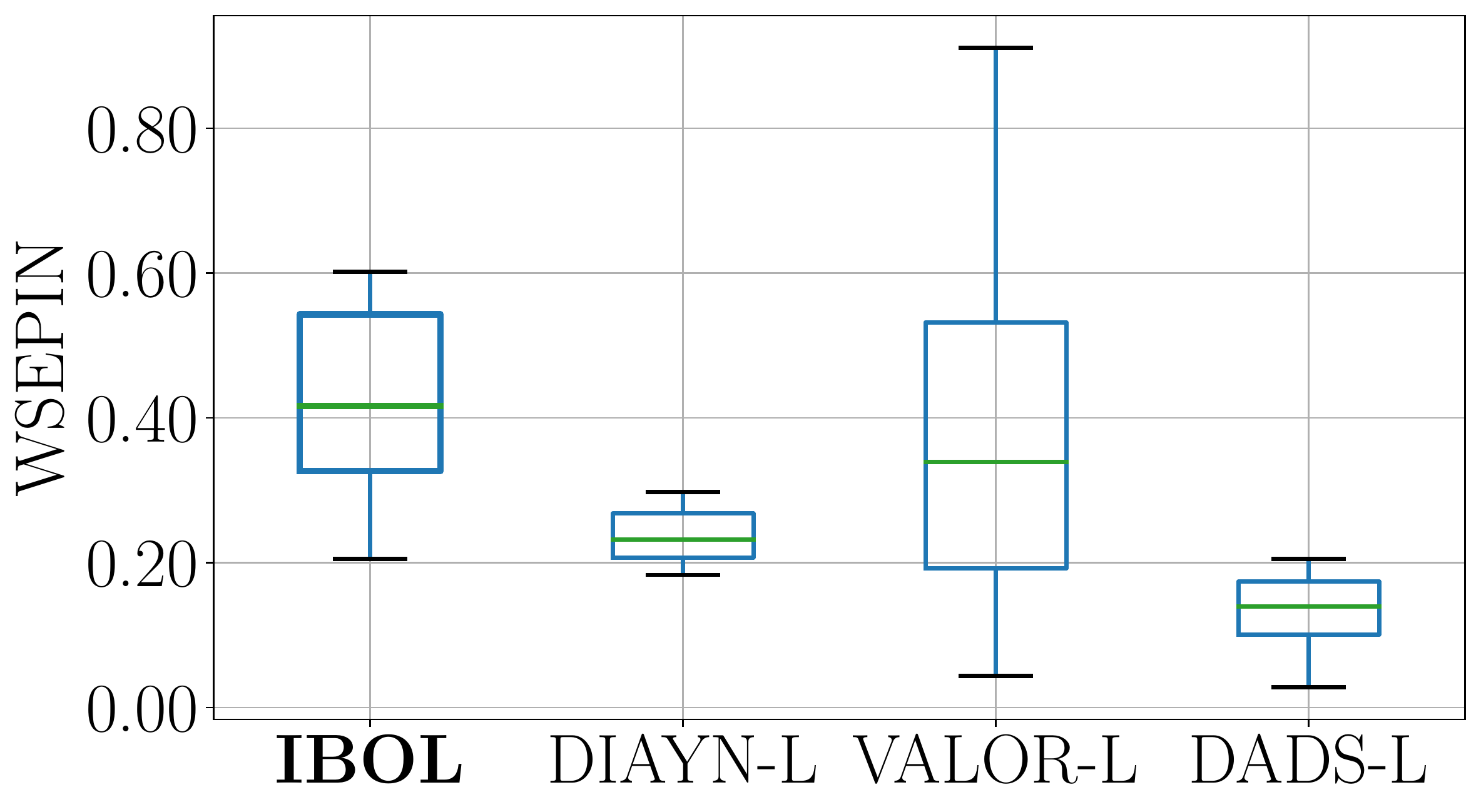}
  \end{subfigure}
  \begin{subfigure}[t]{0.1913187986\linewidth}
    \includegraphics[width=1.0\columnwidth]{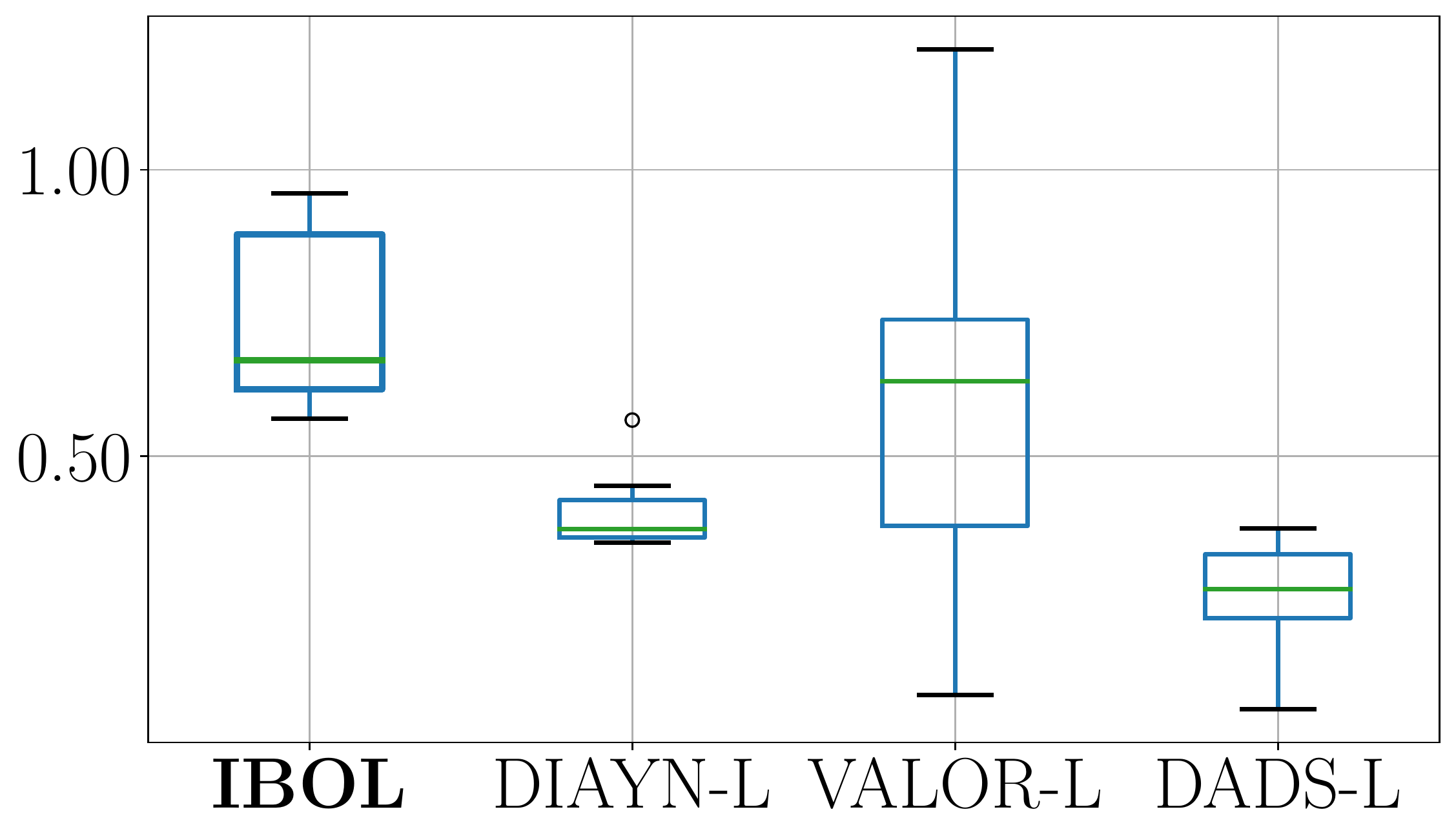}
  \end{subfigure}
  \begin{subfigure}[t]{0.1913187986\linewidth}
    \includegraphics[width=1.0\columnwidth]{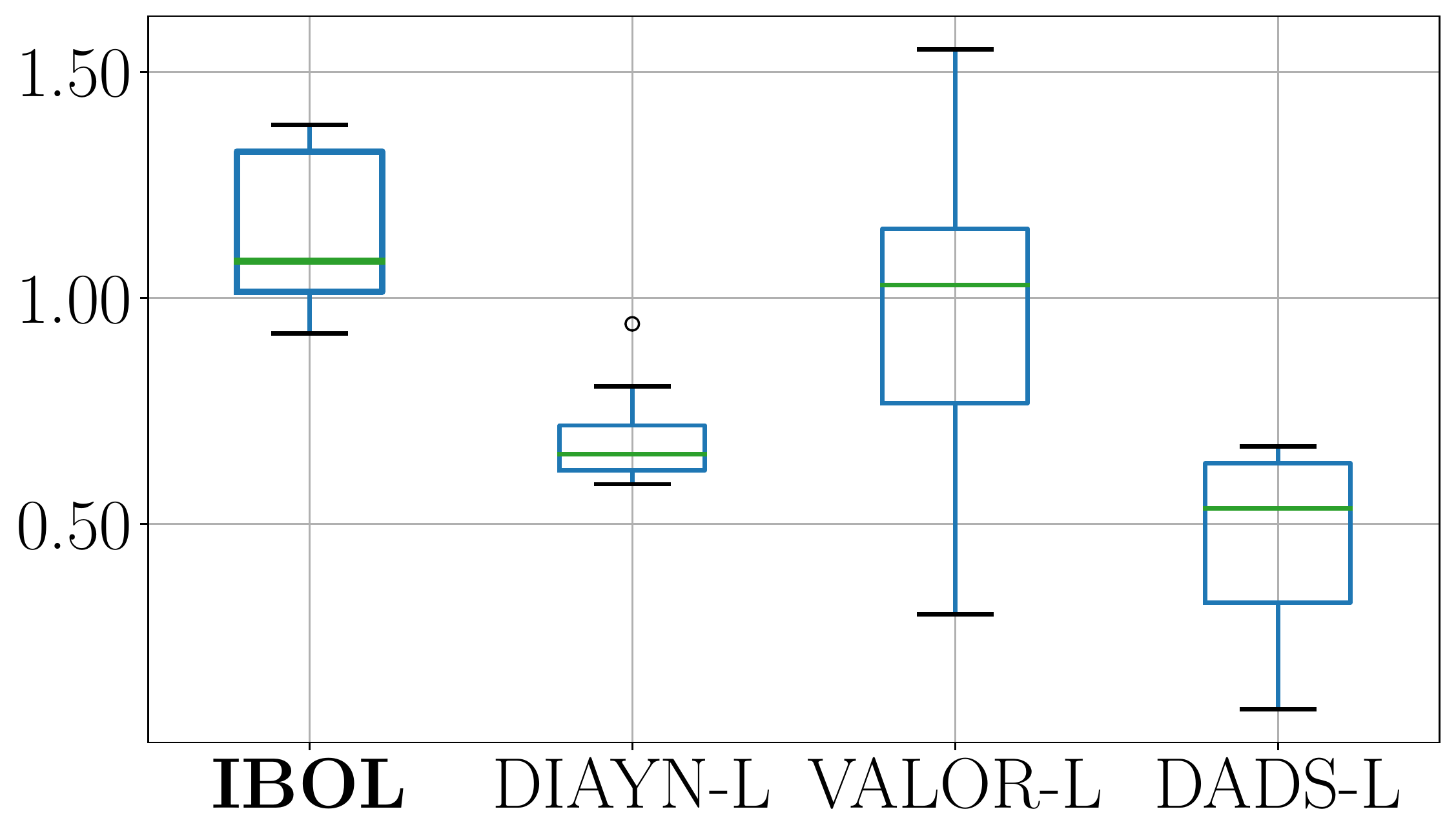}
  \end{subfigure}
  \begin{subfigure}[t]{0.1913187986\linewidth}
    \includegraphics[width=1.0\columnwidth]{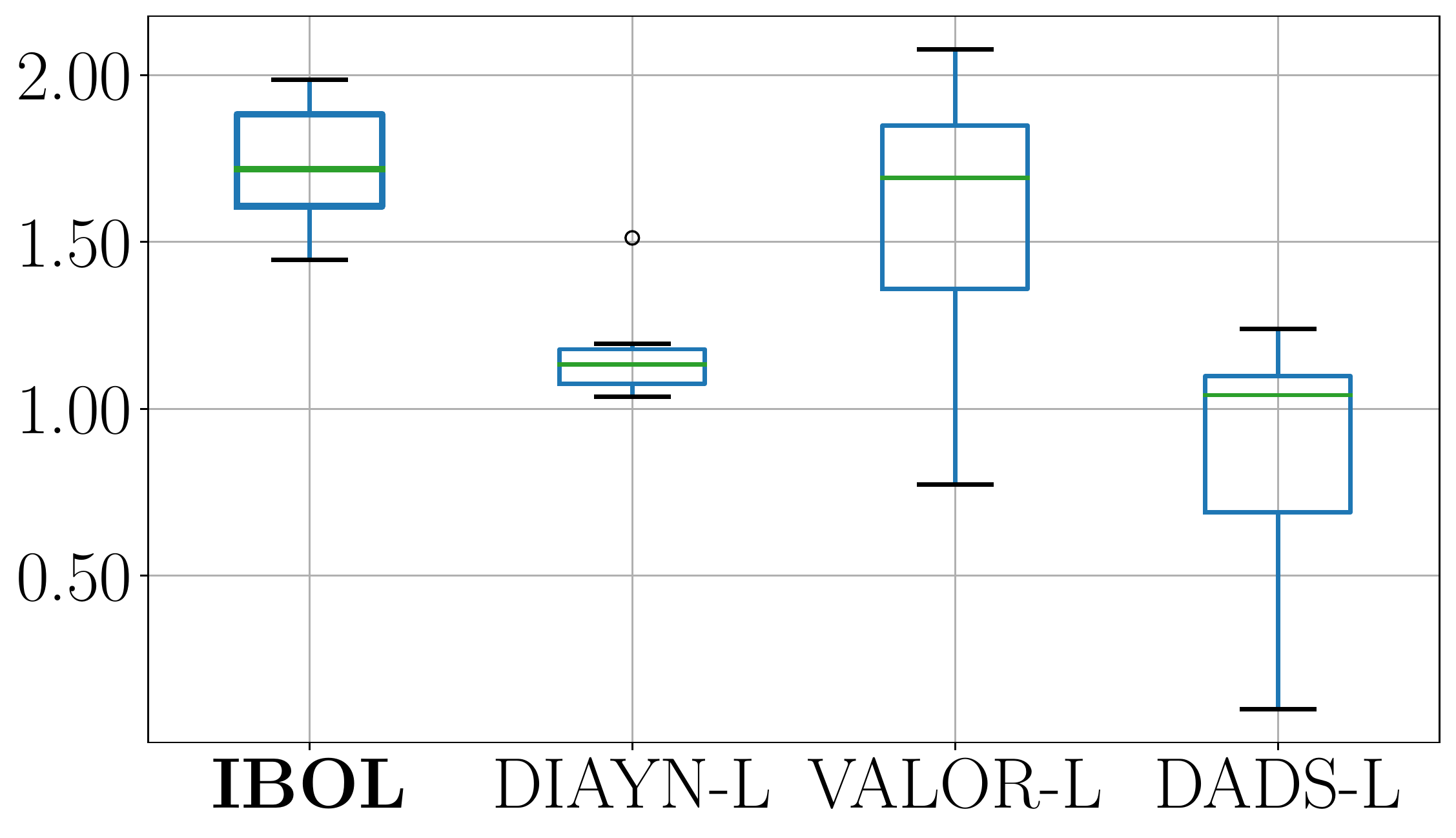}
  \end{subfigure}
  \begin{subfigure}[t]{0.1913187986\linewidth}
    \includegraphics[width=1.0\columnwidth]{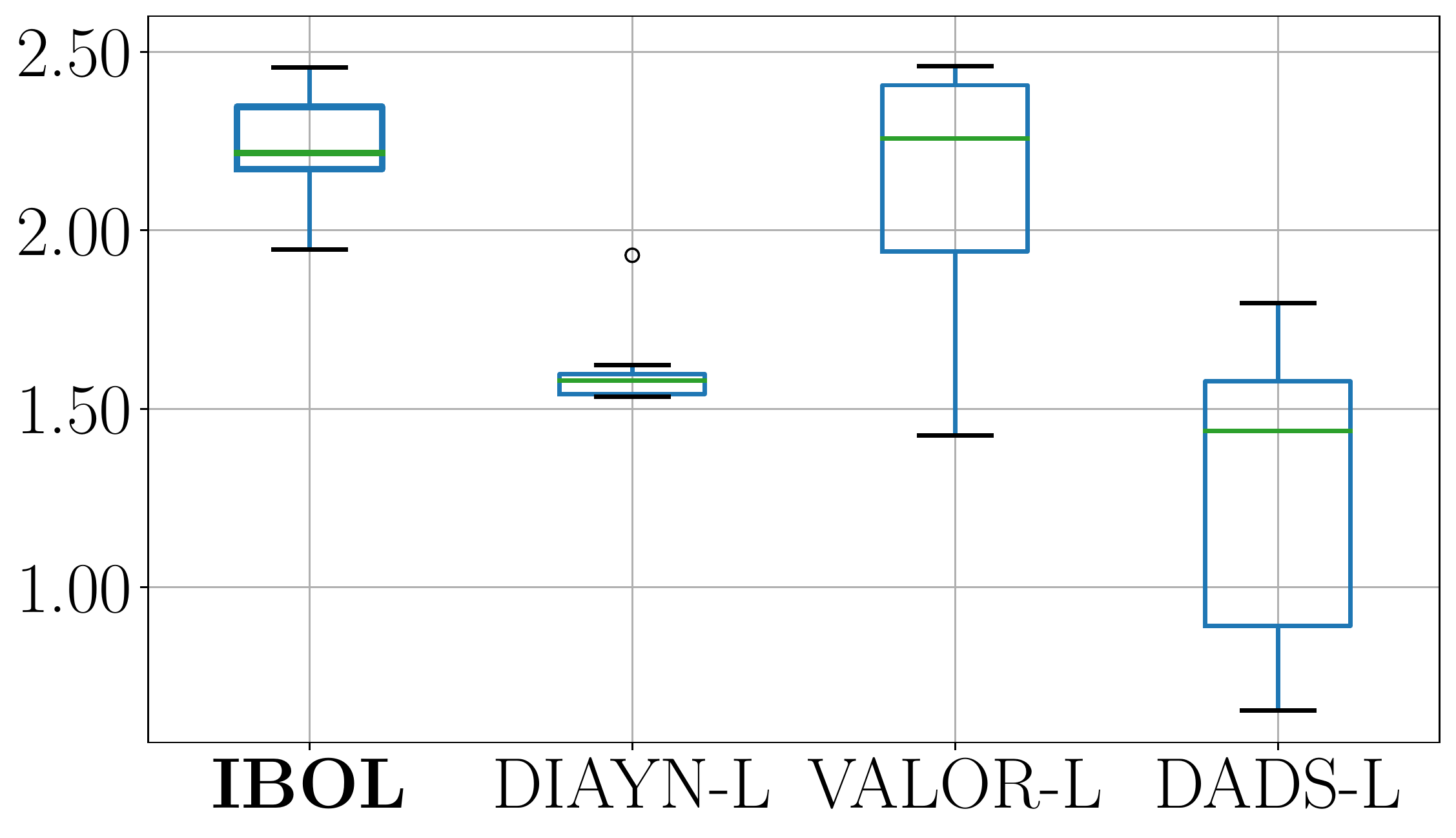}
  \end{subfigure}

  \hfill
  \begin{subfigure}[t]{0.2008490516\linewidth}
    \includegraphics[width=1.0\columnwidth]{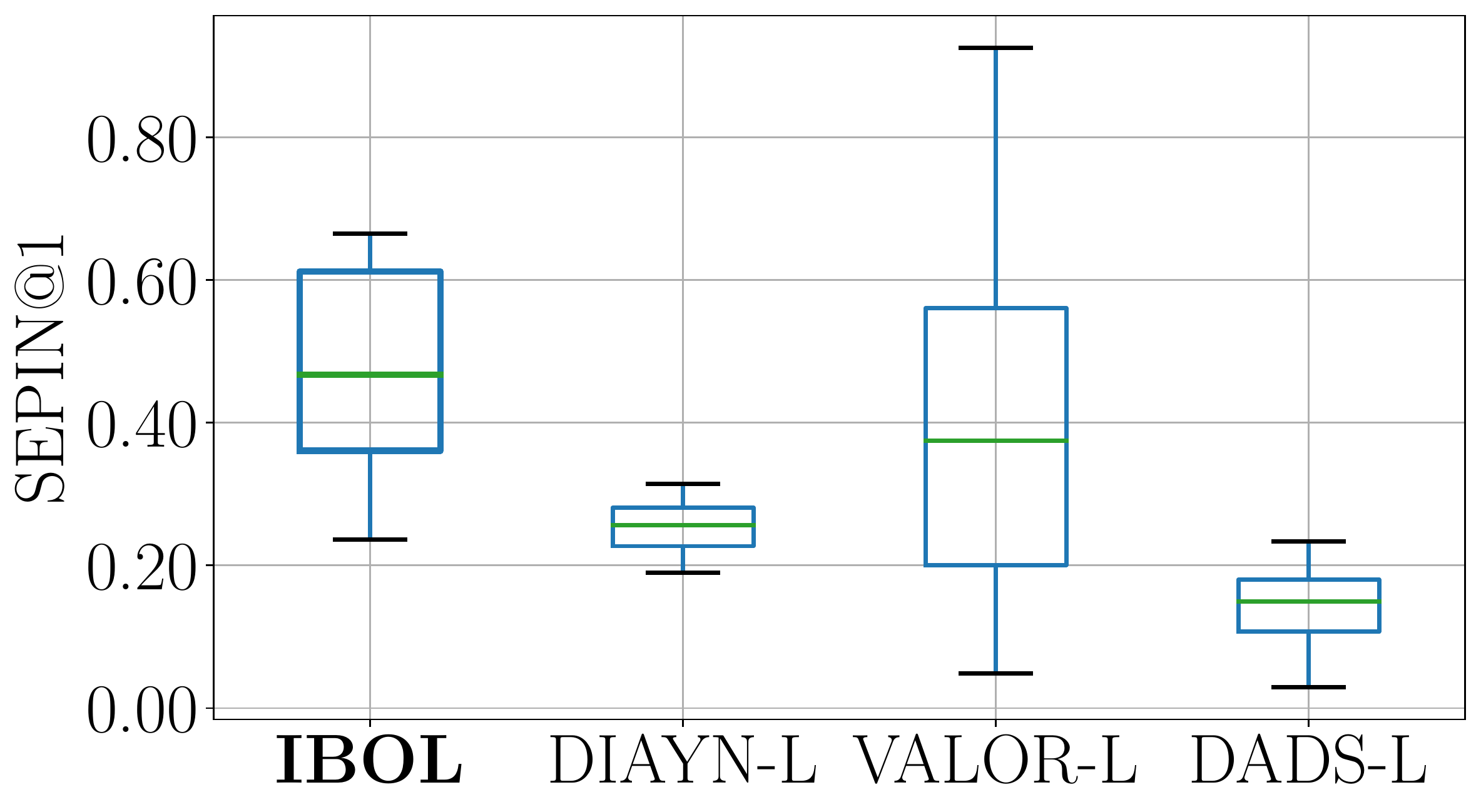}
    \caption{\# bins = 8}
  \end{subfigure}
  \begin{subfigure}[t]{0.1913187986\linewidth}
    \includegraphics[width=1.0\columnwidth]{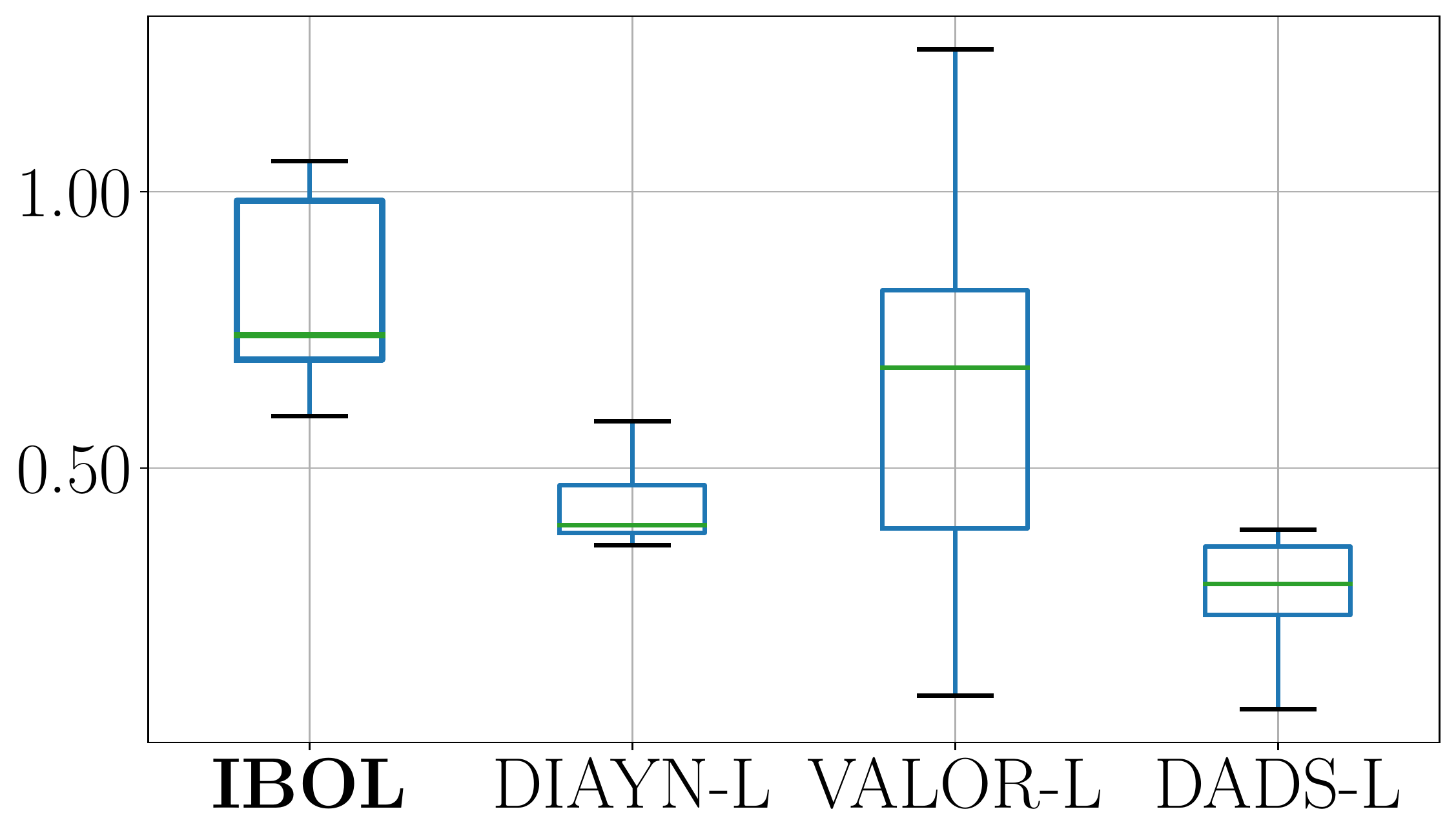}
    \caption{\# bins = 16}
  \end{subfigure}
  \begin{subfigure}[t]{0.1913187986\linewidth}
    \includegraphics[width=1.0\columnwidth]{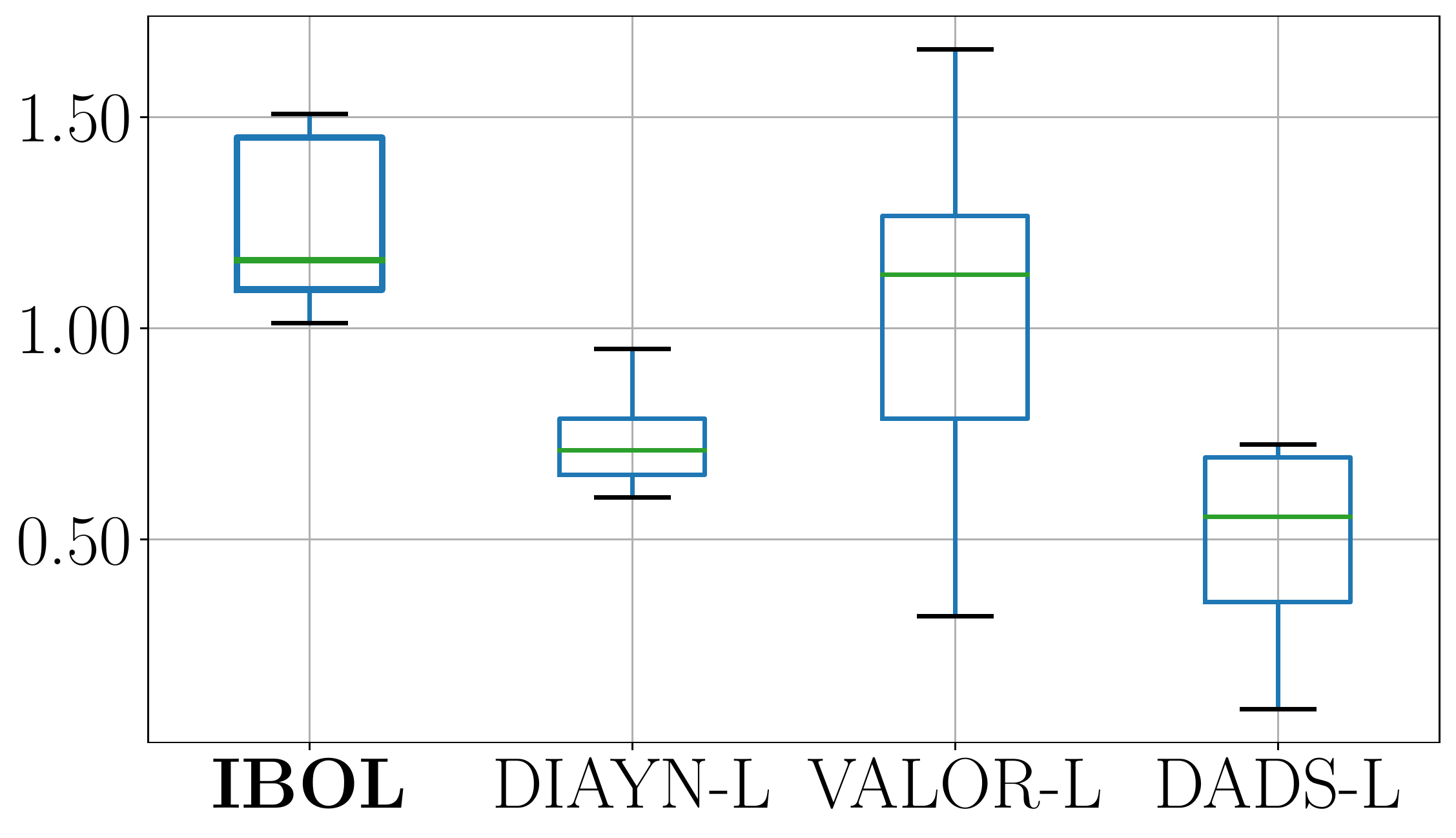}
    \caption{\# bins = 32}
  \end{subfigure}
  \begin{subfigure}[t]{0.1913187986\linewidth}
    \includegraphics[width=1.0\columnwidth]{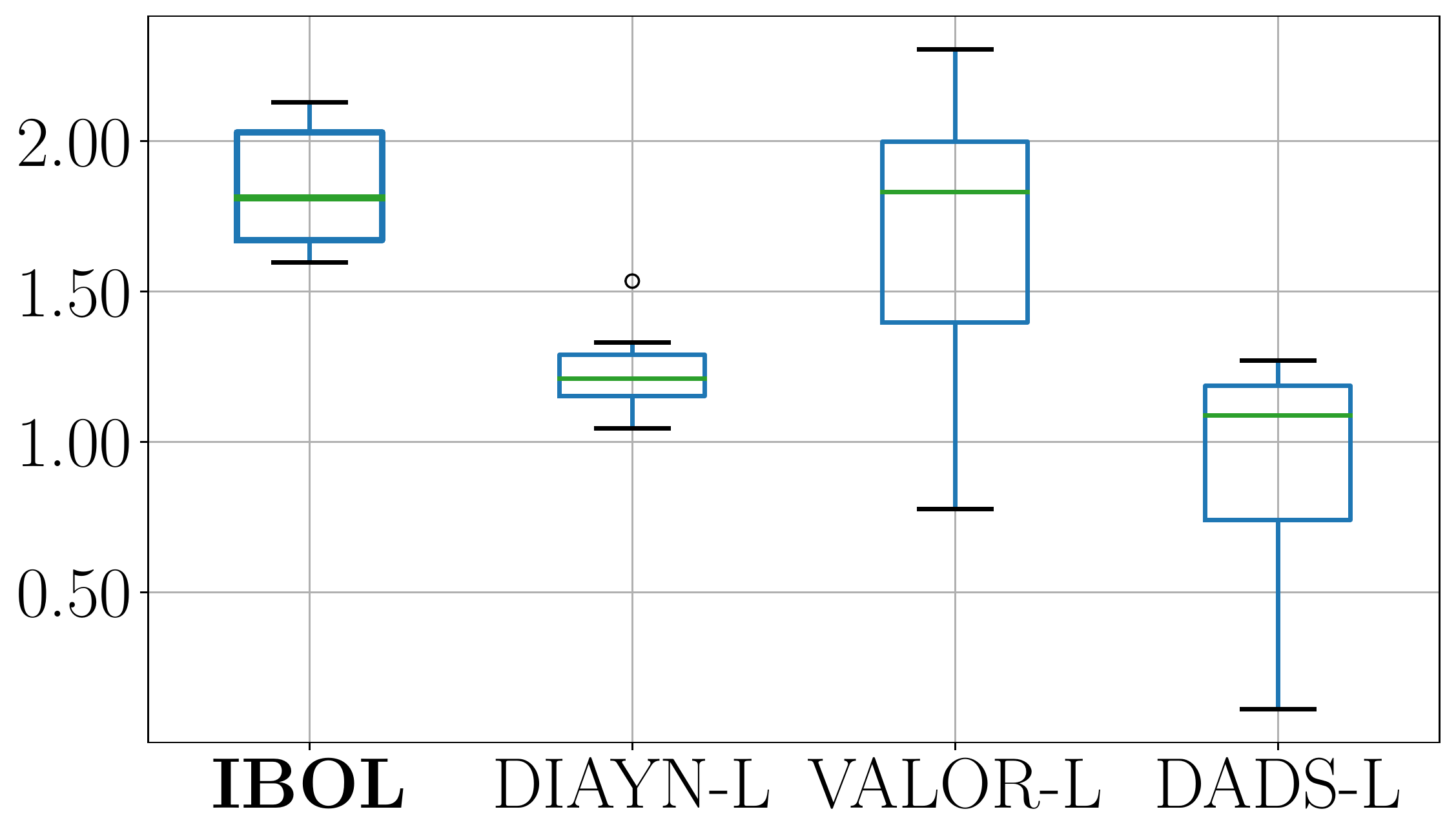}
    \caption{\# bins = 64}
  \end{subfigure}
  \begin{subfigure}[t]{0.1913187986\linewidth}
    \includegraphics[width=1.0\columnwidth]{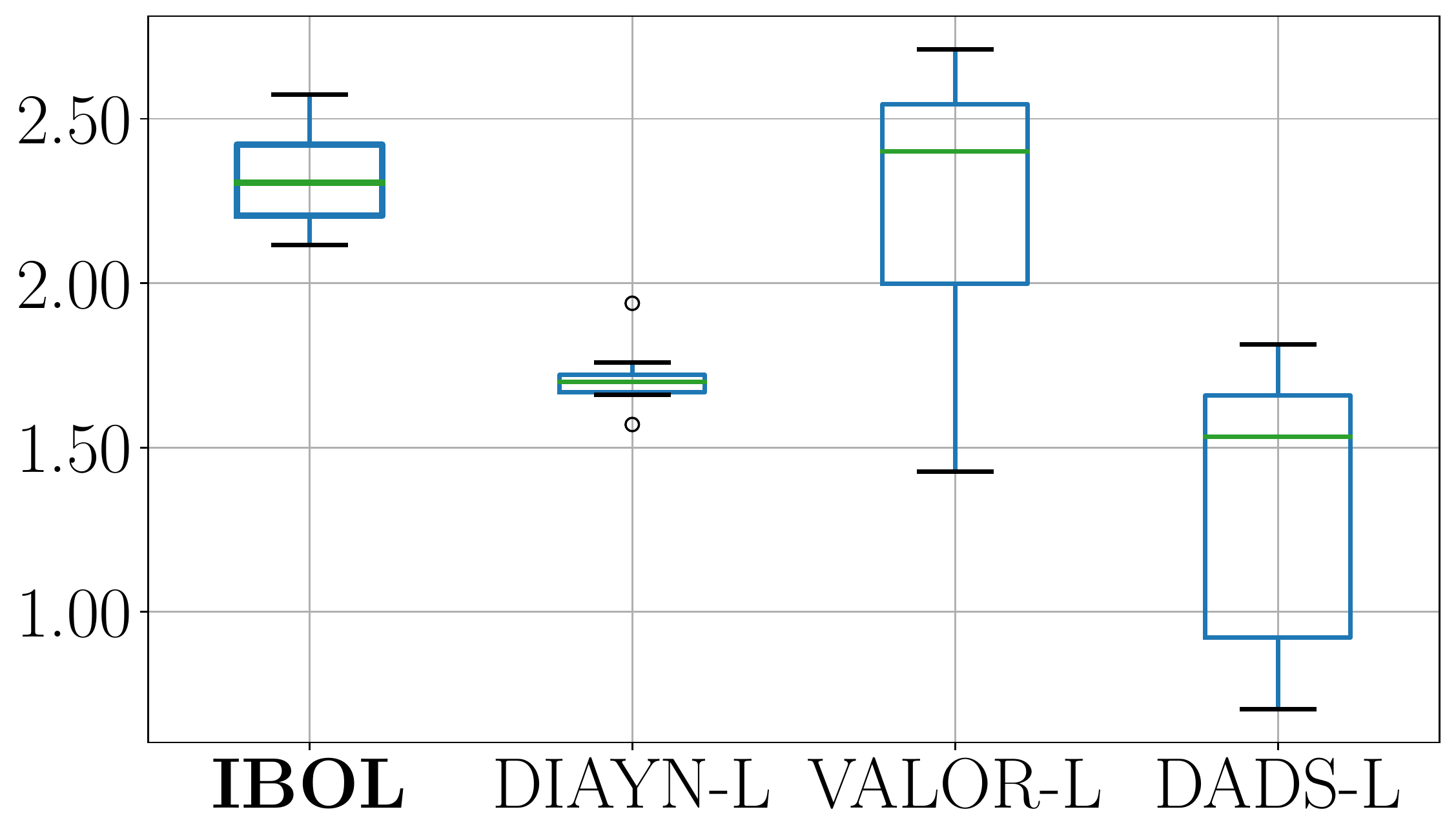}
    \caption{\# bins = 128}
  \end{subfigure}

  \caption{
    Comparison of IBOL (ours) with the baseline methods, DIAYN-L, VALOR-L and DADS-L,
    in the evaluation metrics of $I(Z; S_T^{\text{(loc)}})$, WSEPIN and SEPIN$@1$, on Hopper,
    with different bin counts for the range of each variable estimating mutual information.
    For each method, we use the eight trained skill policies.
  }
  \label{fig:eval_metrics_num_bins_hp}
\end{figure*}

\begin{figure*}[t!]
  \centering

  \hfill
  \begin{subfigure}[t]{0.2047248057\linewidth}
    \includegraphics[width=1.0\columnwidth]{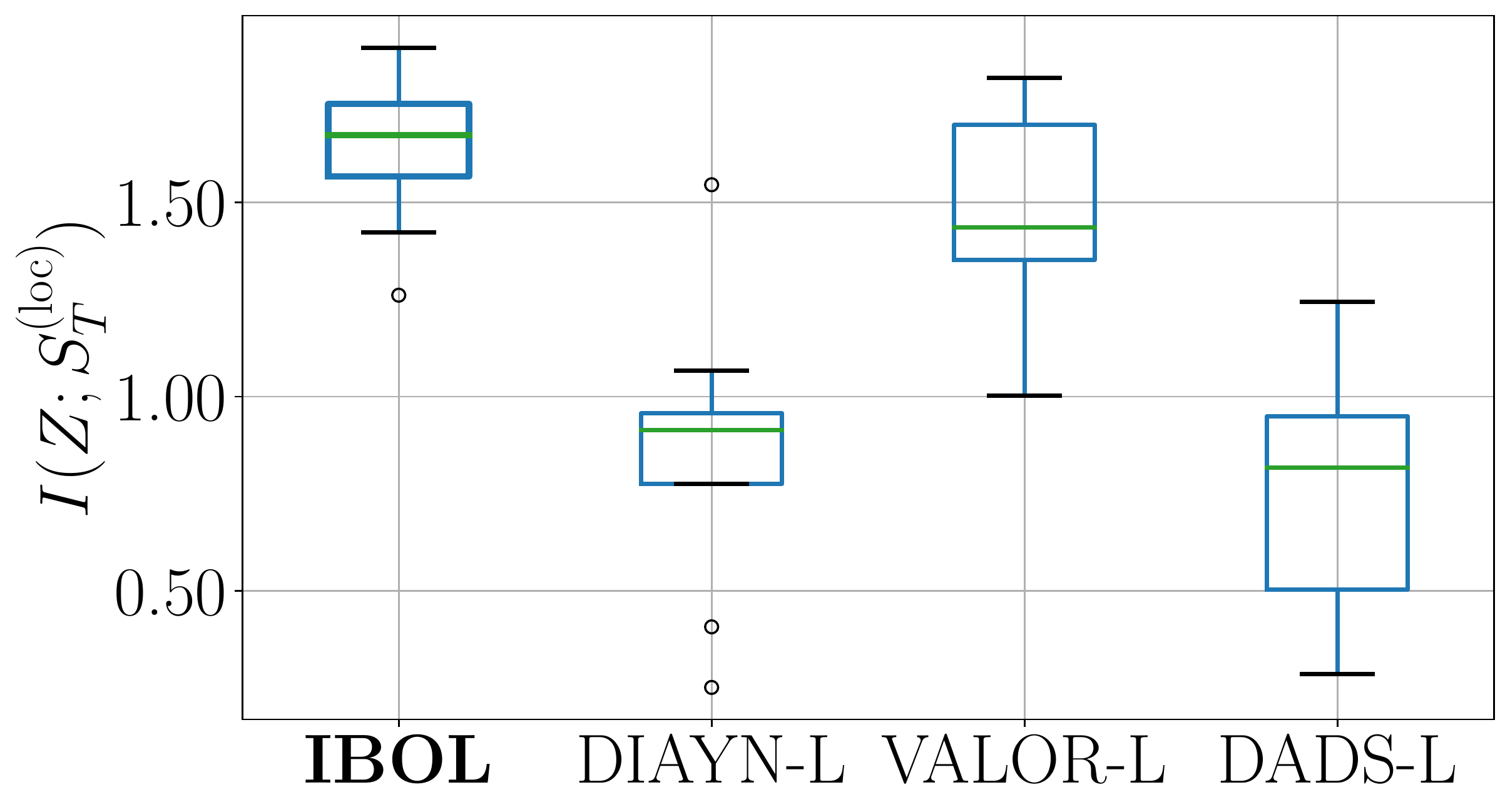}
  \end{subfigure}
  \begin{subfigure}[t]{0.1913187986\linewidth}
    \includegraphics[width=1.0\columnwidth]{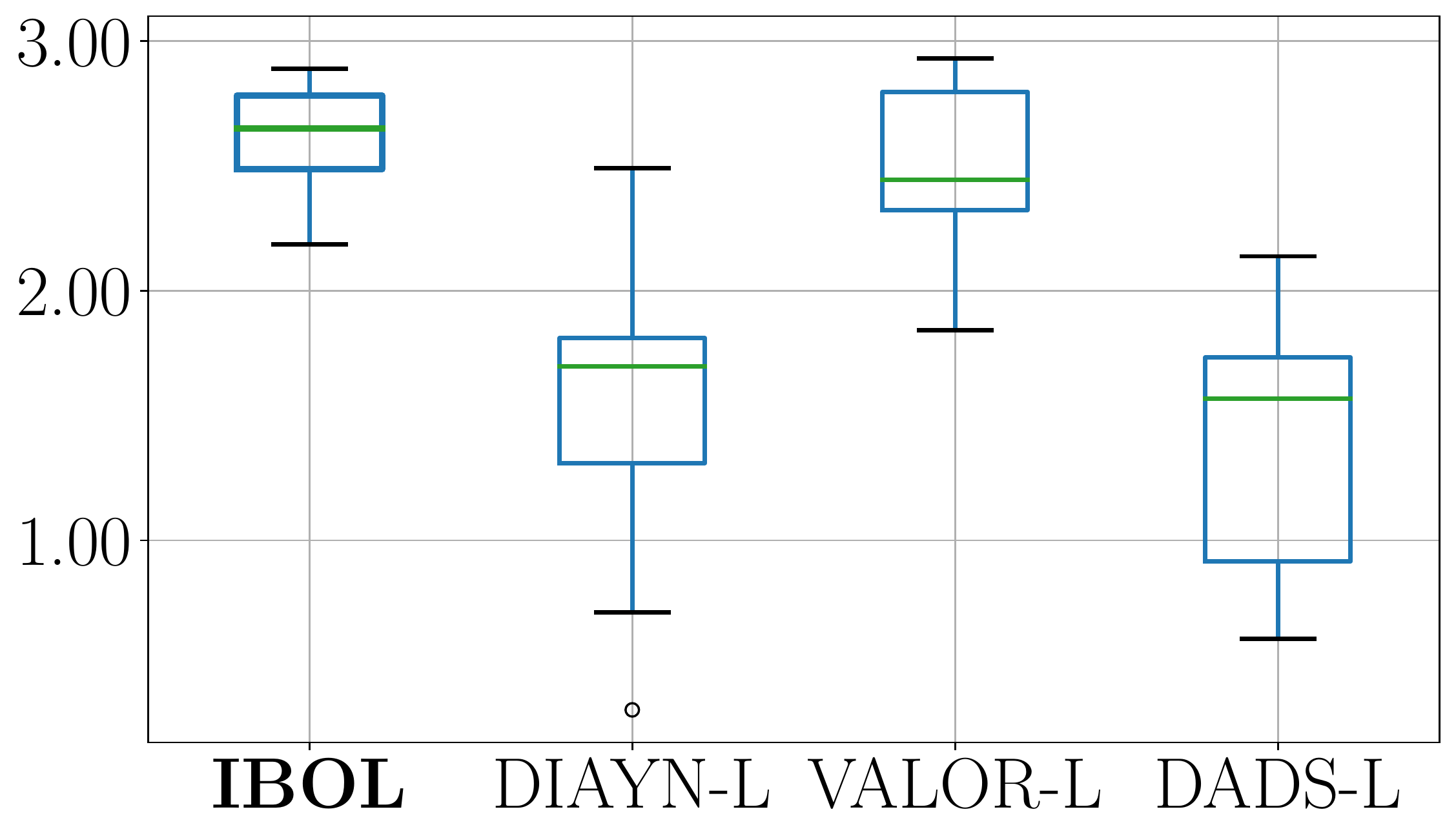}
  \end{subfigure}
  \begin{subfigure}[t]{0.1913187986\linewidth}
    \includegraphics[width=1.0\columnwidth]{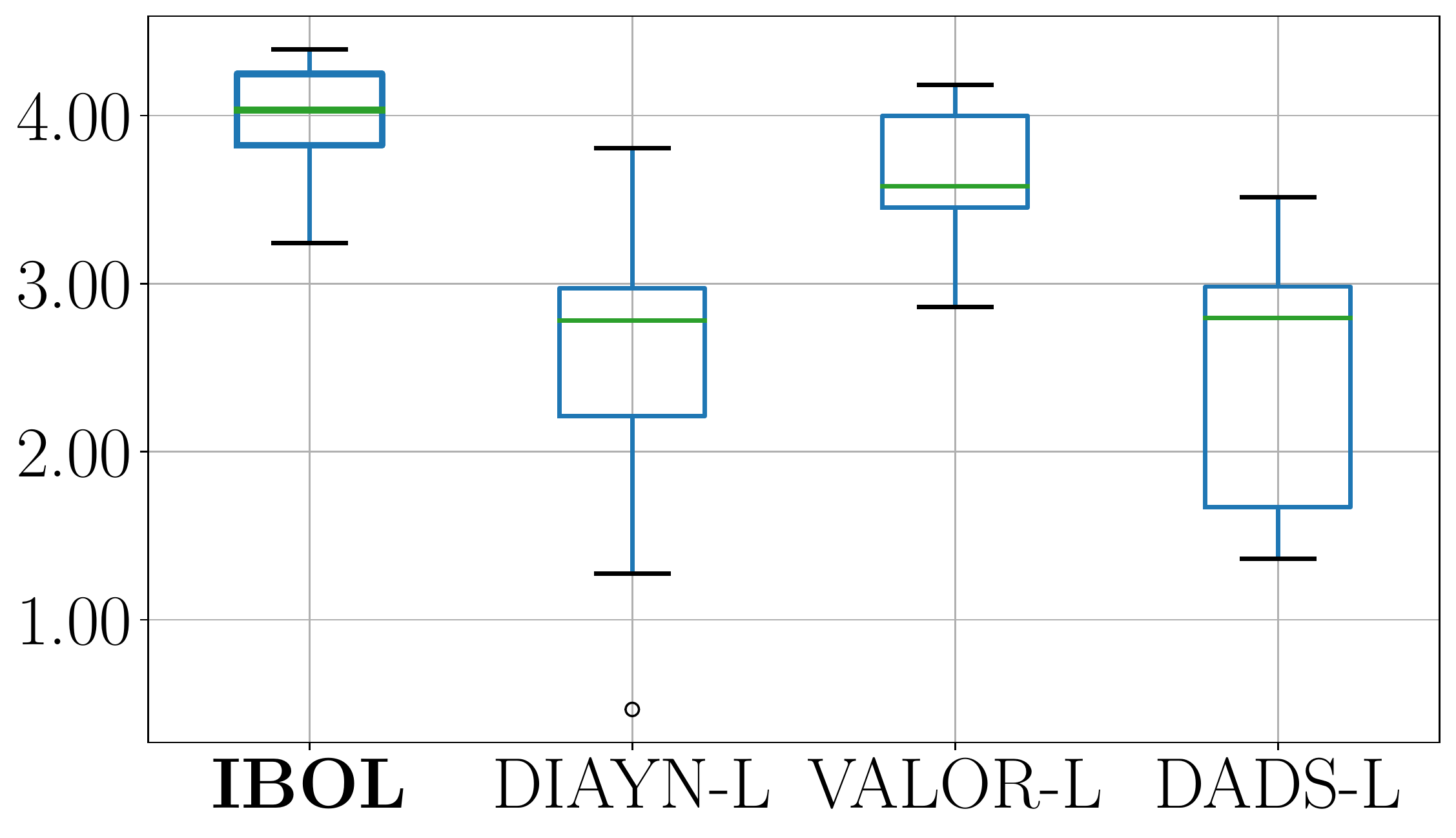}
  \end{subfigure}
  \begin{subfigure}[t]{0.1913187986\linewidth}
    \includegraphics[width=1.0\columnwidth]{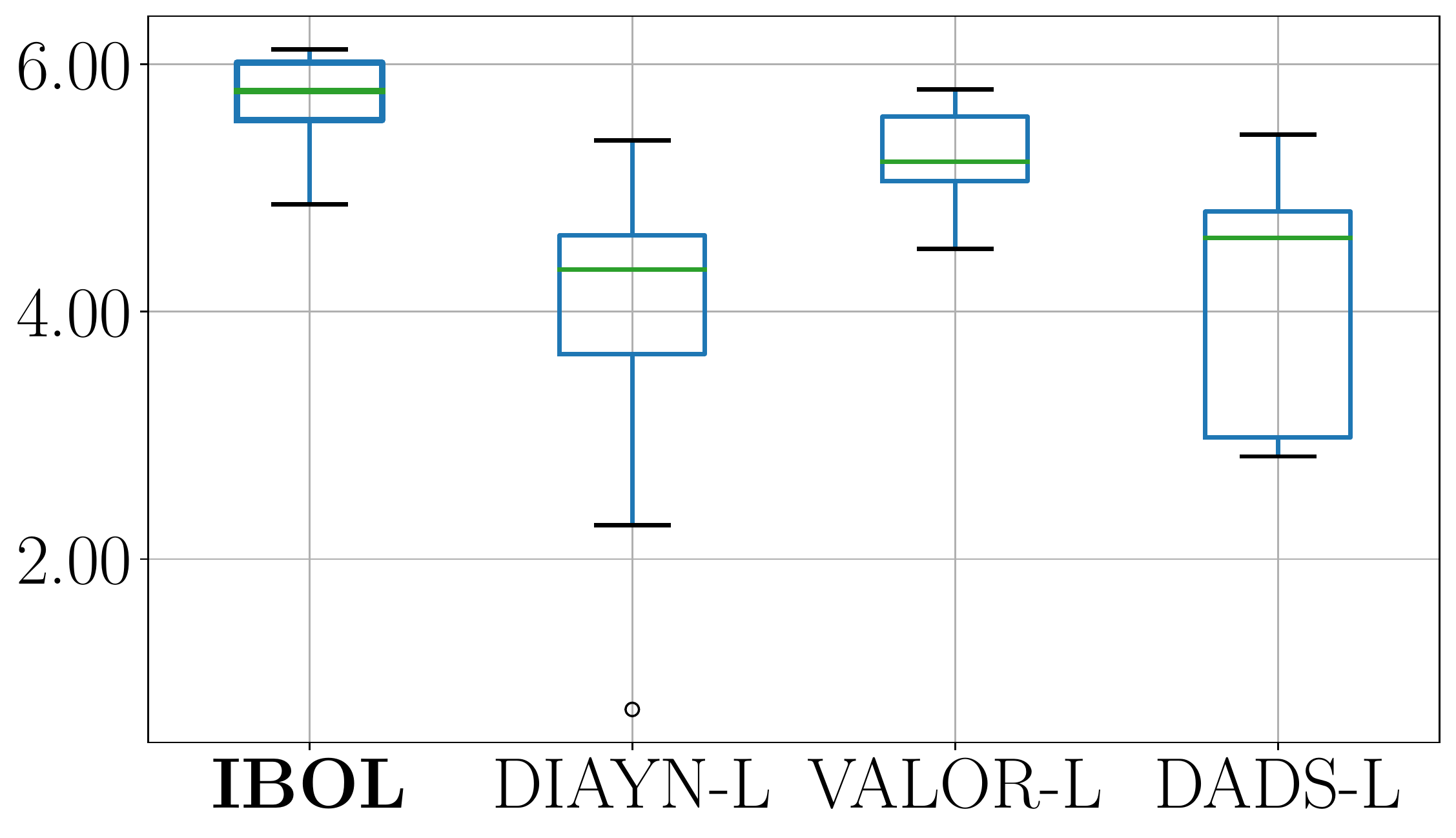}
  \end{subfigure}
  \begin{subfigure}[t]{0.1913187986\linewidth}
    \includegraphics[width=1.0\columnwidth]{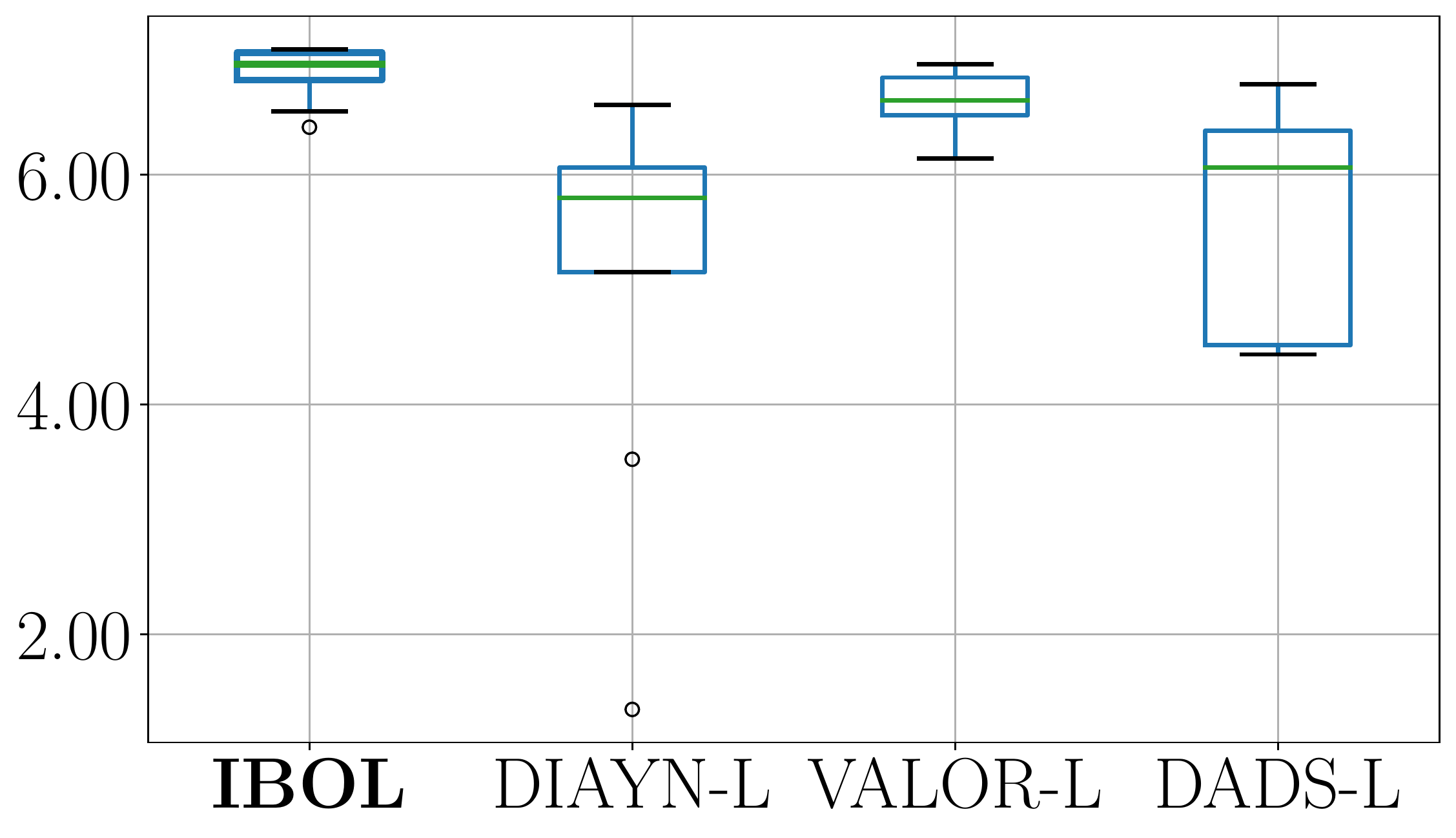}
  \end{subfigure}

  \hfill
  \begin{subfigure}[t]{0.2008490516\linewidth}
    \includegraphics[width=1.0\columnwidth]{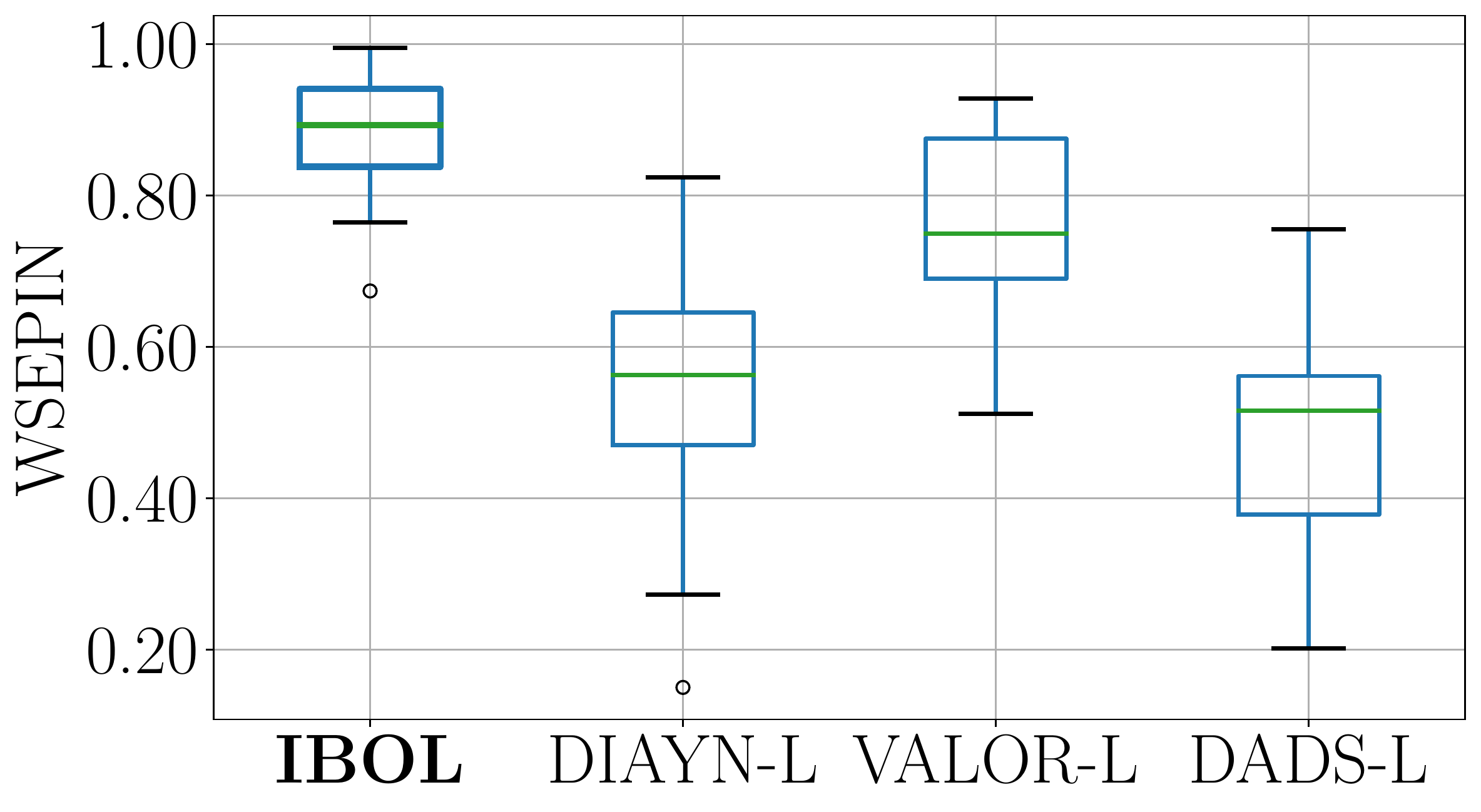}
  \end{subfigure}
  \begin{subfigure}[t]{0.1913187986\linewidth}
    \includegraphics[width=1.0\columnwidth]{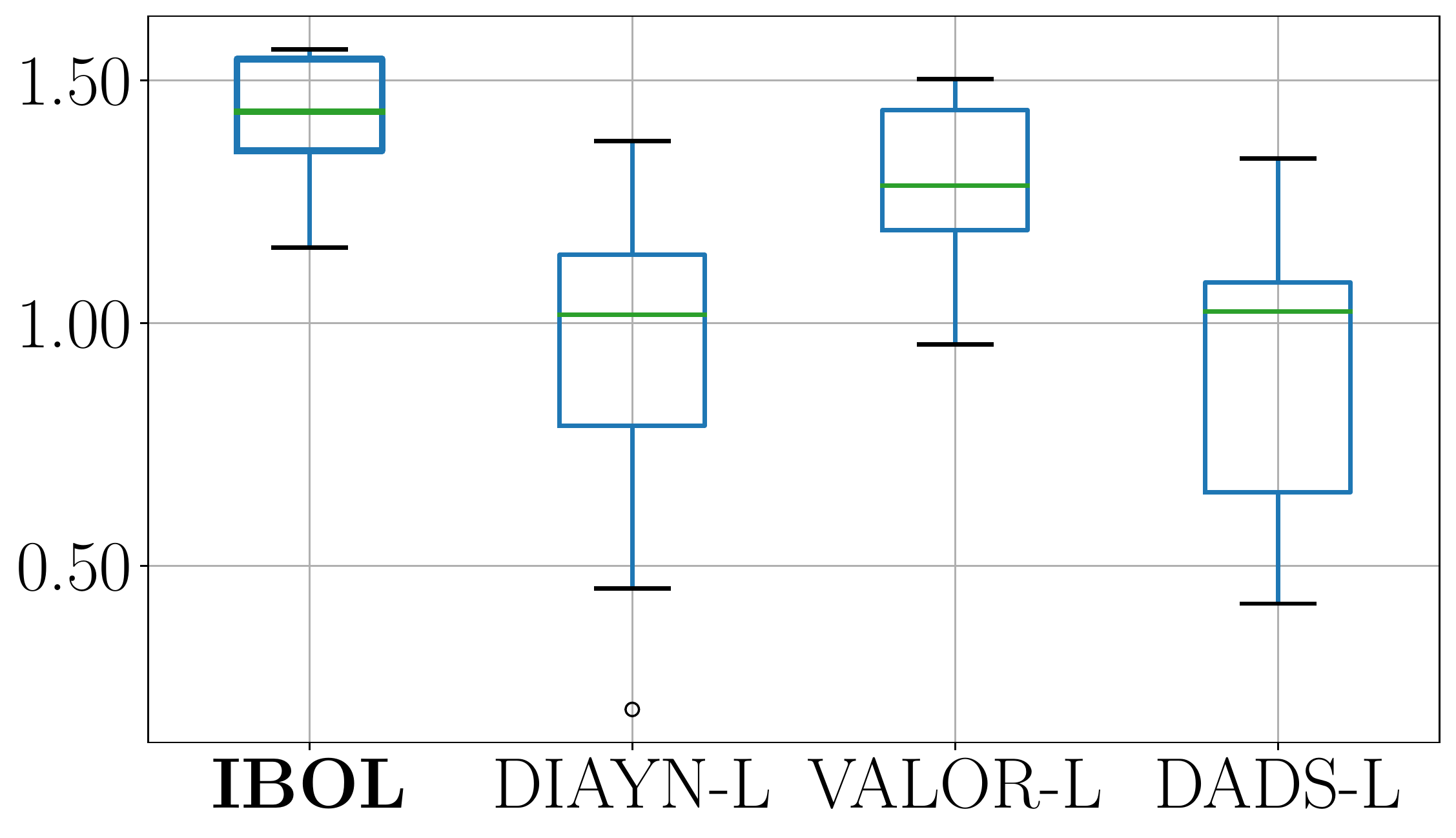}
  \end{subfigure}
  \begin{subfigure}[t]{0.1913187986\linewidth}
    \includegraphics[width=1.0\columnwidth]{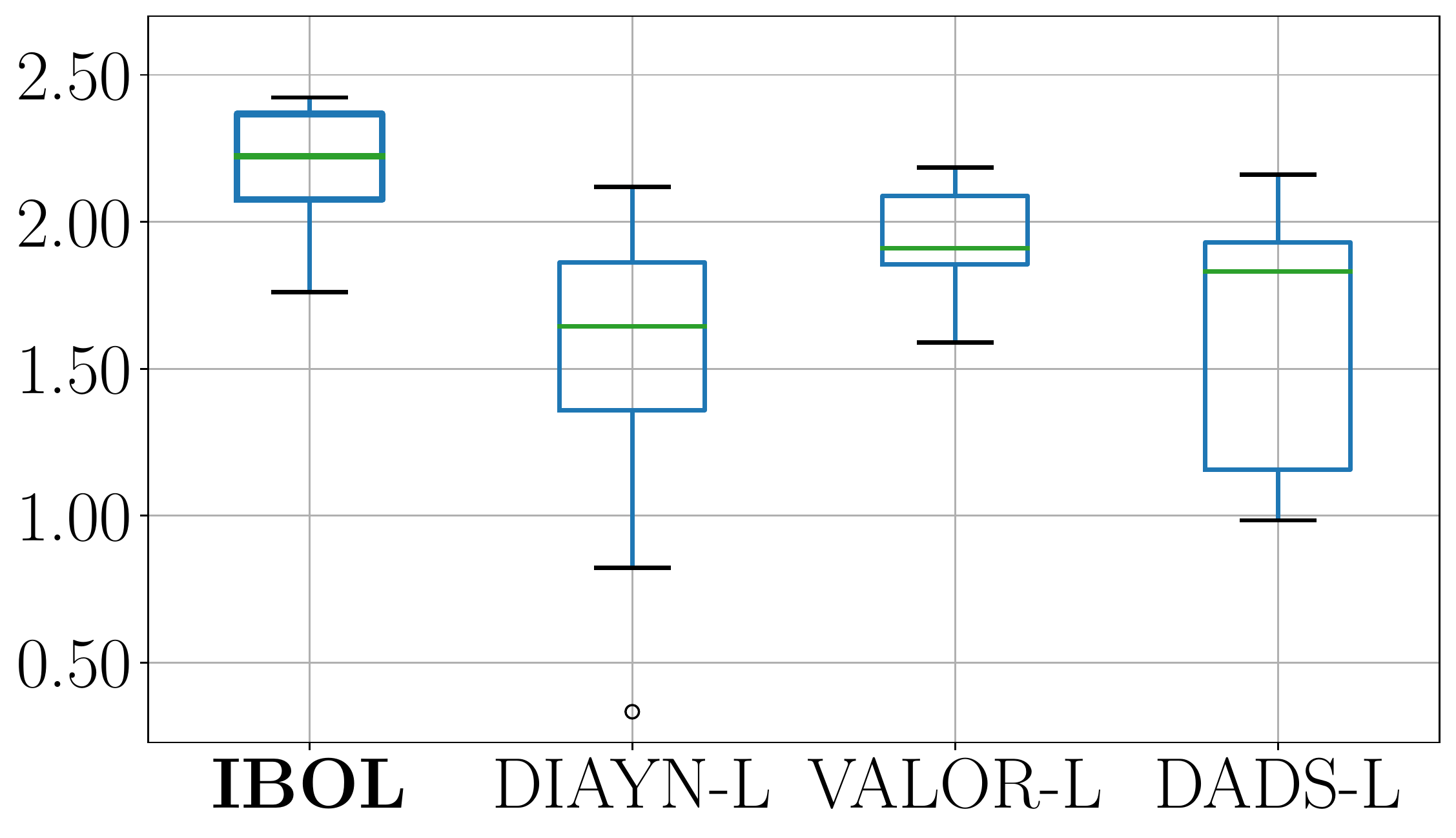}
  \end{subfigure}
  \begin{subfigure}[t]{0.1913187986\linewidth}
    \includegraphics[width=1.0\columnwidth]{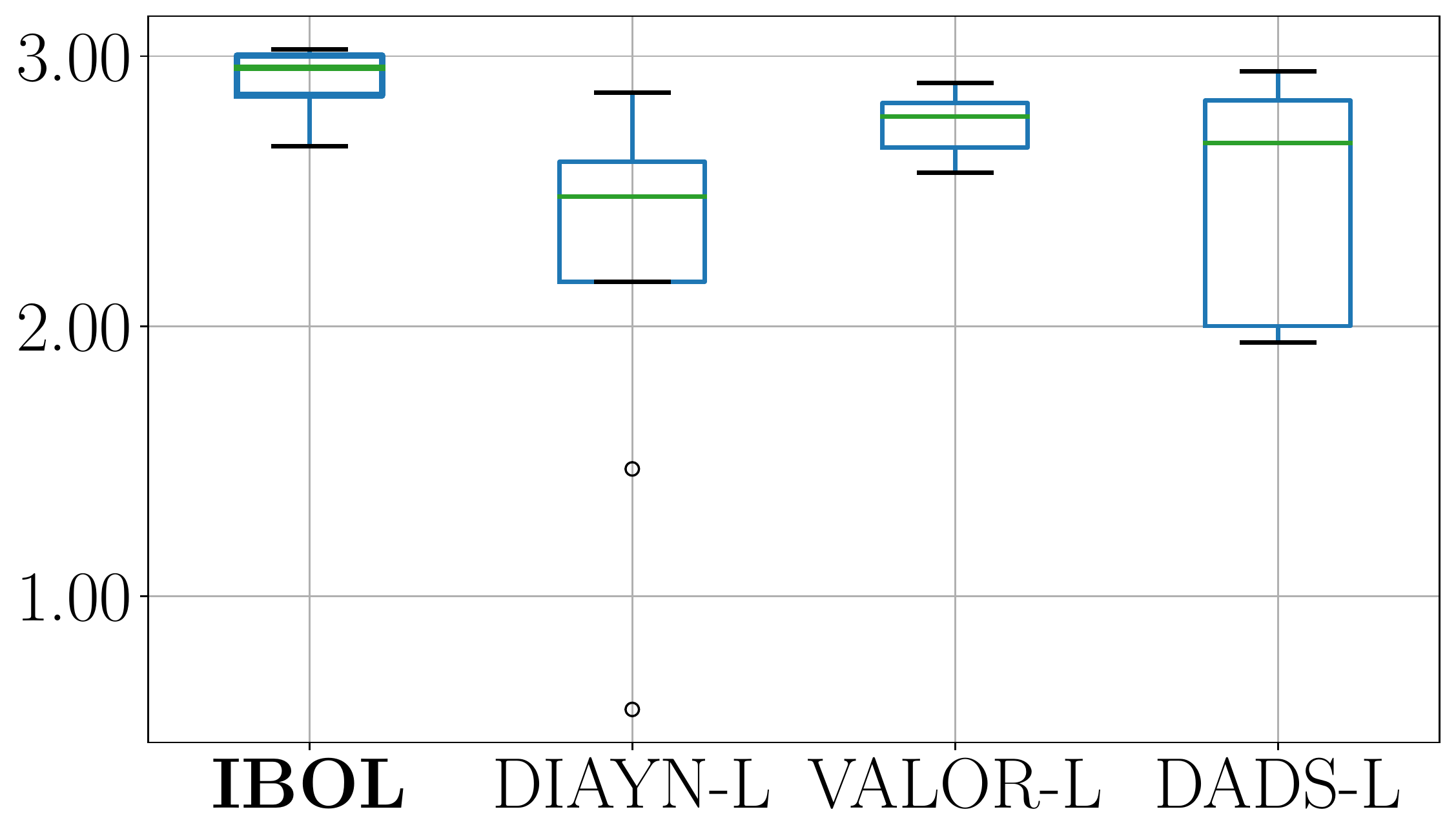}
  \end{subfigure}
  \begin{subfigure}[t]{0.1913187986\linewidth}
    \includegraphics[width=1.0\columnwidth]{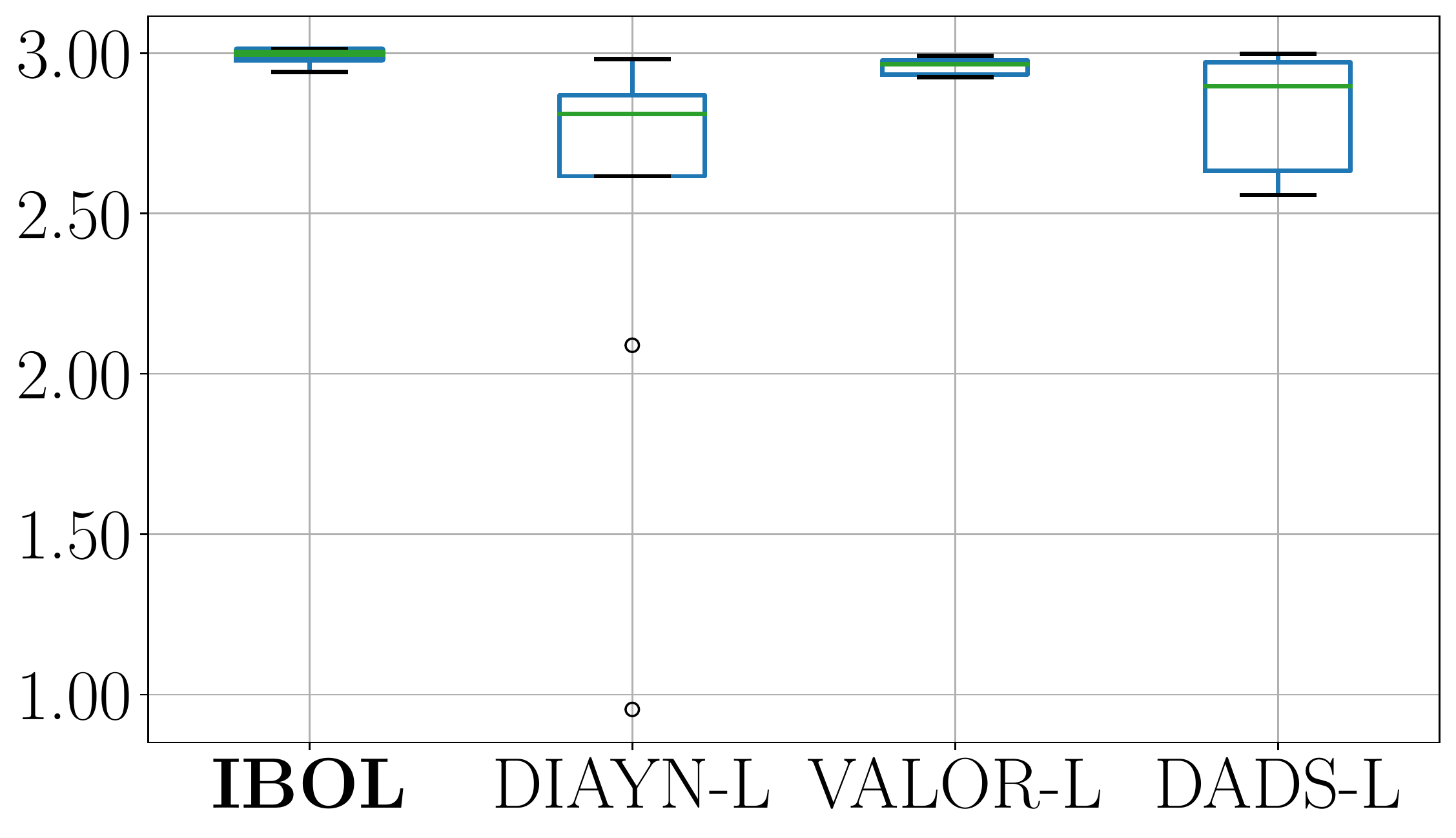}
  \end{subfigure}

  \hfill
  \begin{subfigure}[t]{0.2008490516\linewidth}
    \includegraphics[width=1.0\columnwidth]{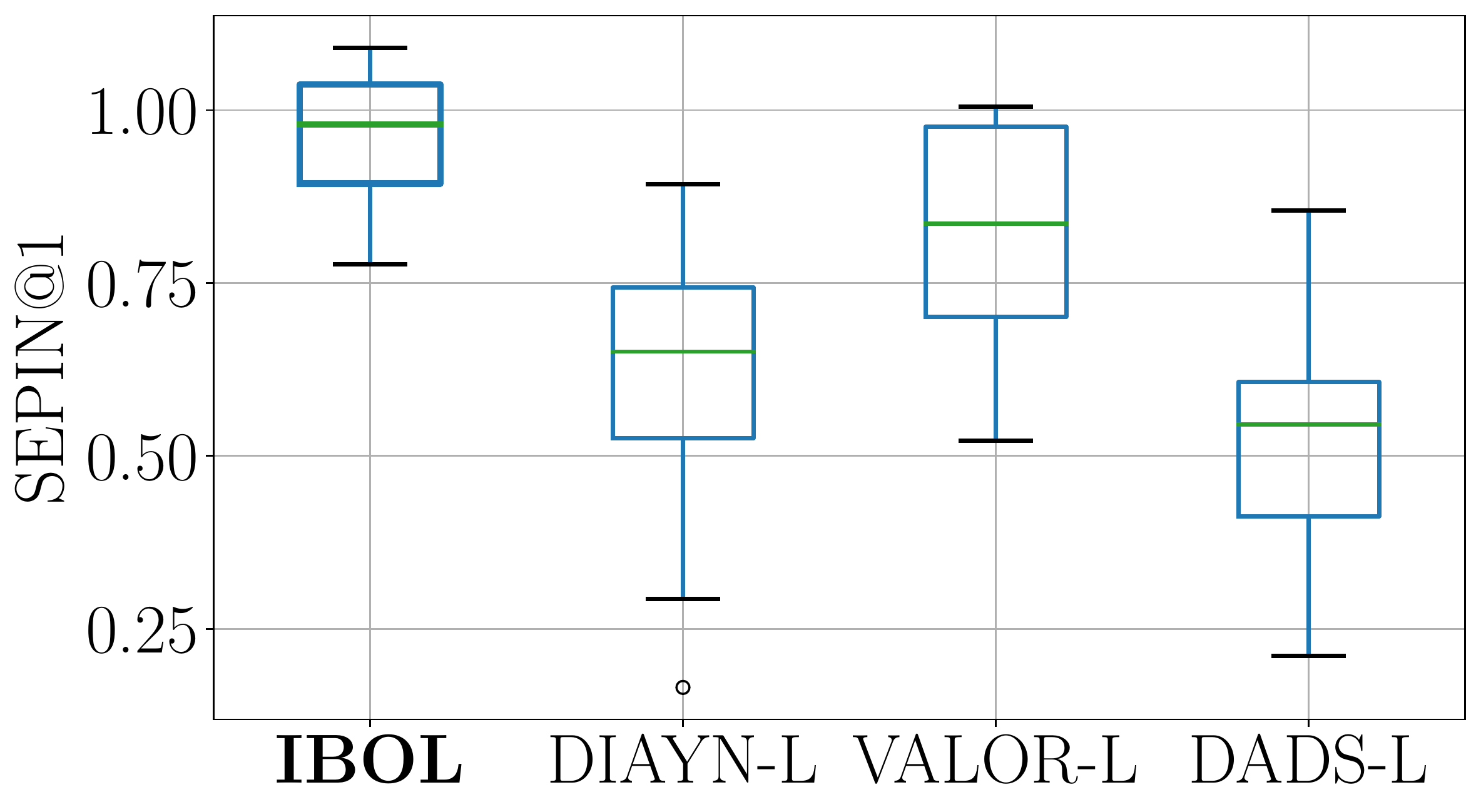}
    \caption{\# bins = 8}
  \end{subfigure}
  \begin{subfigure}[t]{0.1913187986\linewidth}
    \includegraphics[width=1.0\columnwidth]{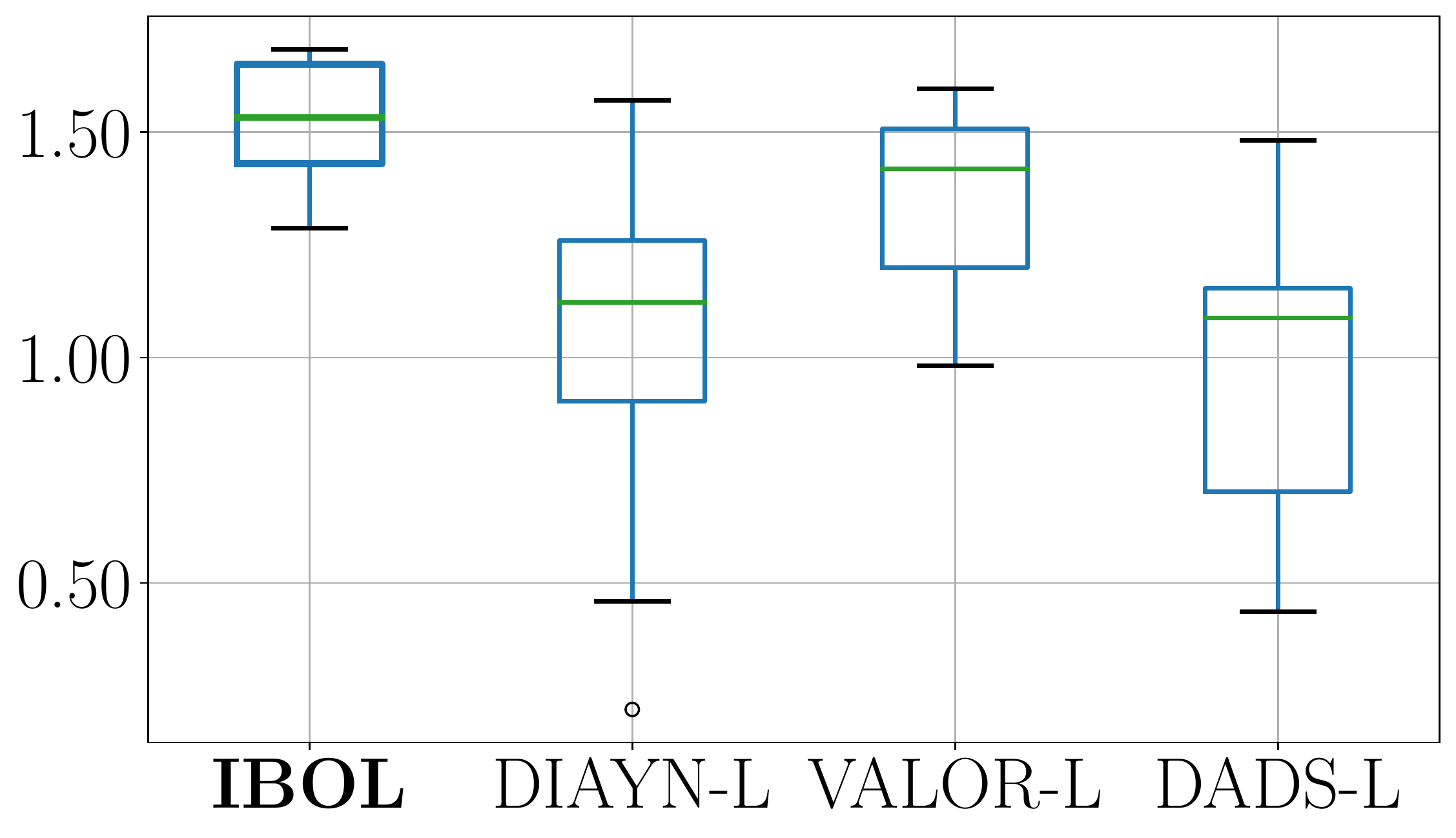}
    \caption{\# bins = 16}
  \end{subfigure}
  \begin{subfigure}[t]{0.1913187986\linewidth}
    \includegraphics[width=1.0\columnwidth]{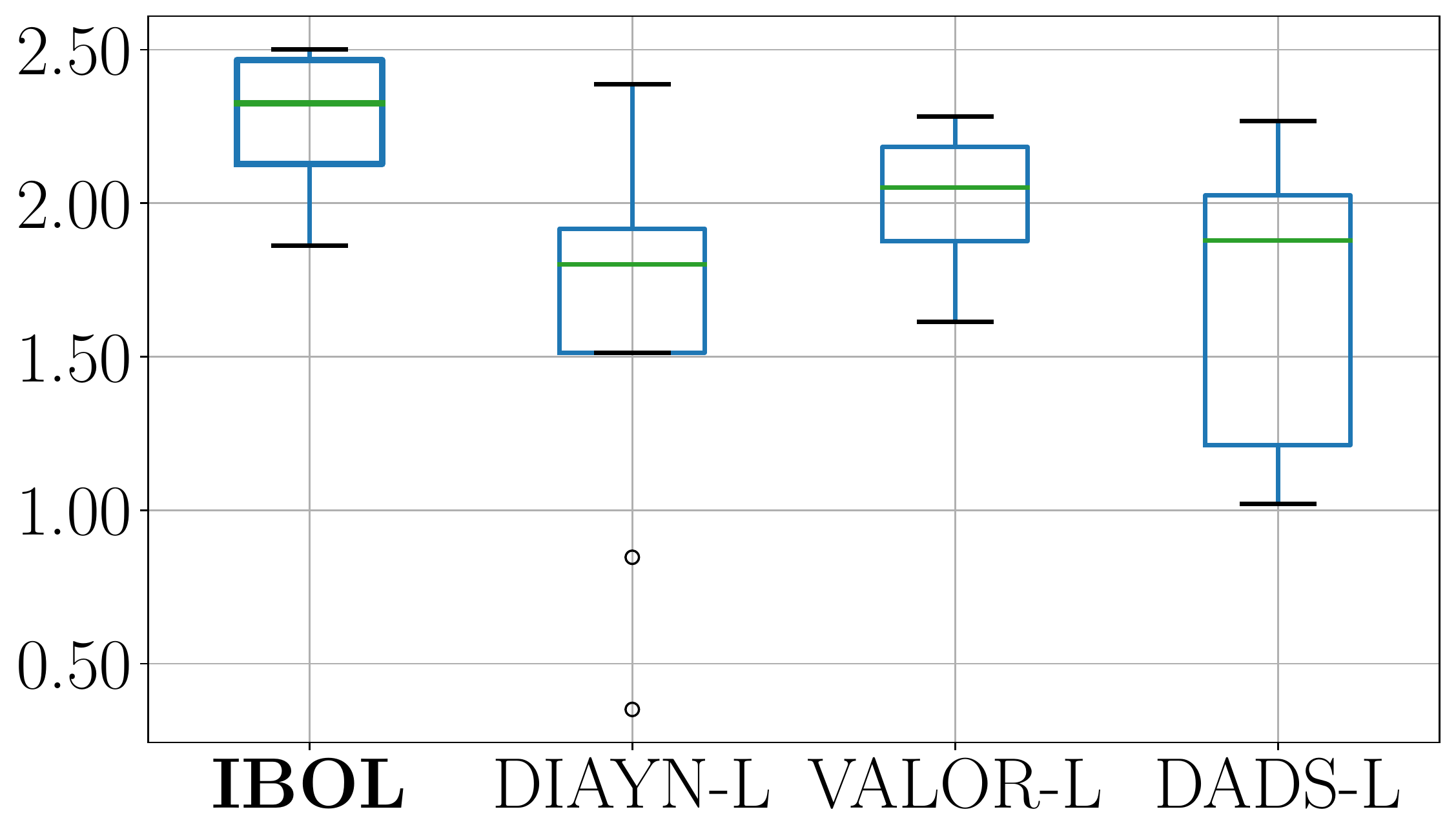}
    \caption{\# bins = 32}
  \end{subfigure}
  \begin{subfigure}[t]{0.1913187986\linewidth}
    \includegraphics[width=1.0\columnwidth]{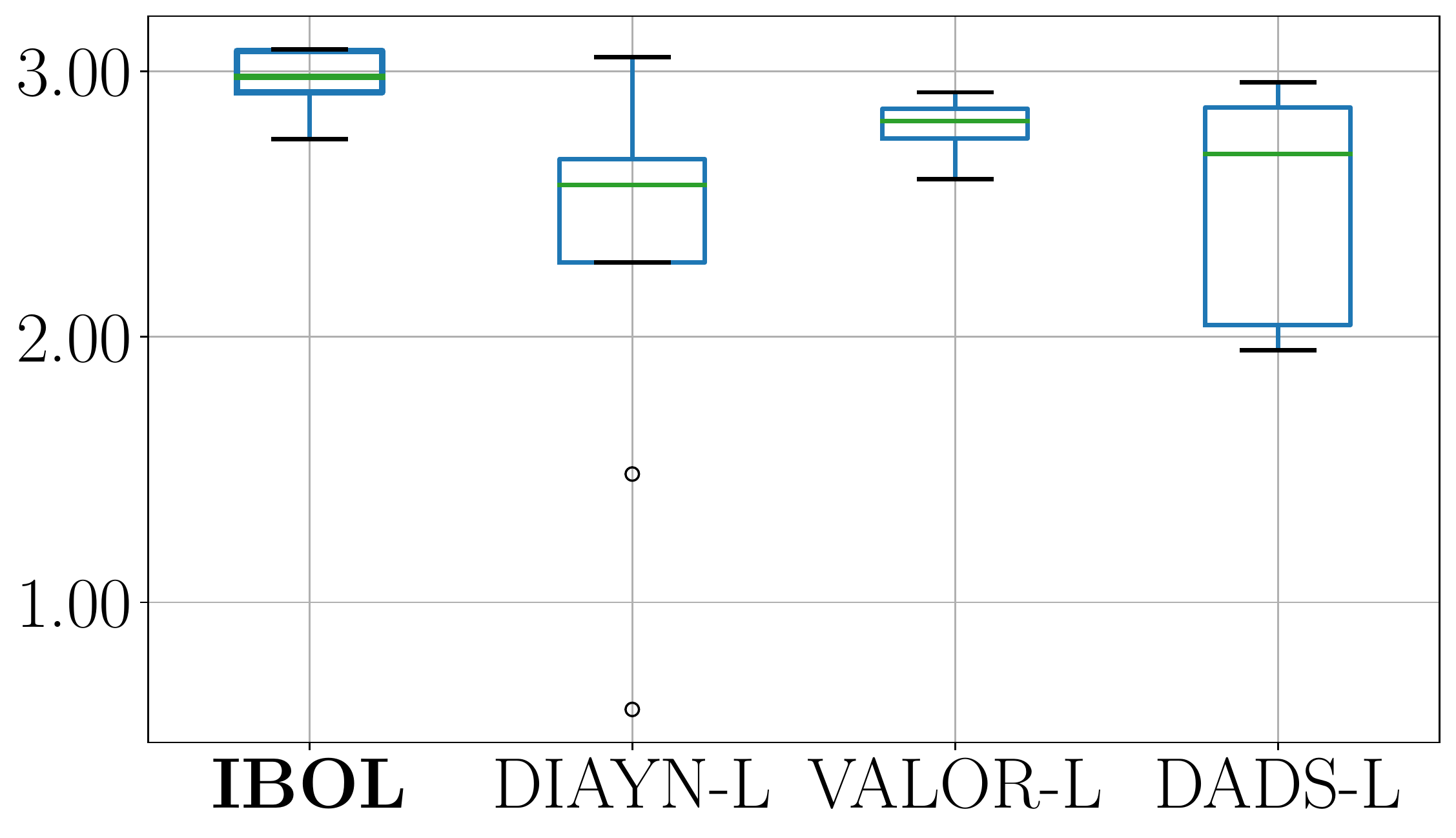}
    \caption{\# bins = 64}
  \end{subfigure}
  \begin{subfigure}[t]{0.1913187986\linewidth}
    \includegraphics[width=1.0\columnwidth]{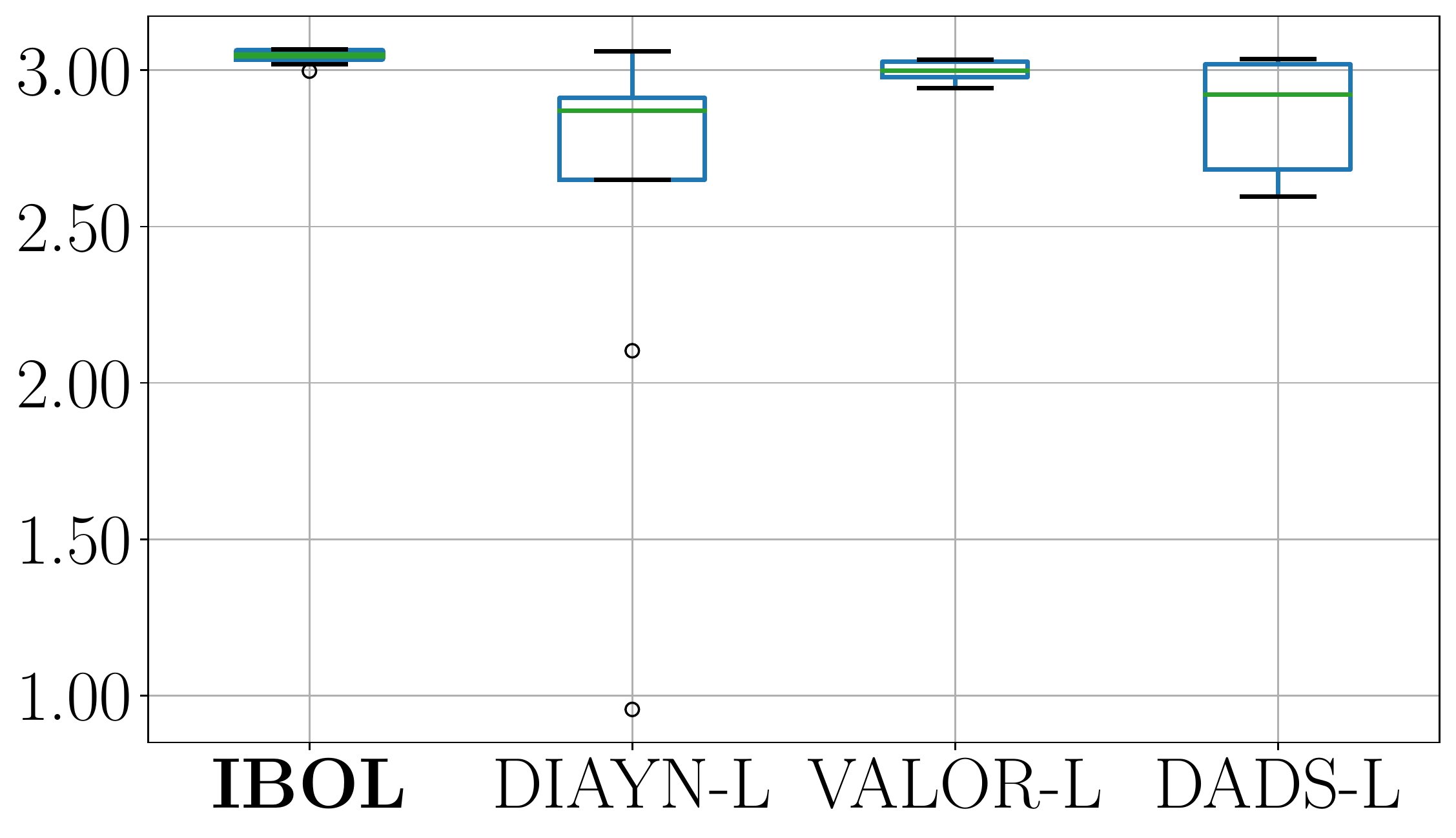}
    \caption{\# bins = 128}
  \end{subfigure}

  \caption{
    Comparison of IBOL (ours) with the baseline methods, DIAYN-L, VALOR-L and DADS-L,
    in the evaluation metrics of $I(Z; S_T^{\text{(loc)}})$, WSEPIN and SEPIN$@1$, on D'Kitty,
    with different bin counts for the range of each variable estimating mutual information.
    For each method, we use the eight trained skill policies.
  }
  \label{fig:eval_metrics_num_bins_dk}
\end{figure*}

We quantize variables for the estimation of mutual information for measuring the information-theoretic metrics in \Cref{sec:info_eval} from the main paper (see \Cref{sec:info_eval_details} for the details).
To show that IBOL outperforms the baseline skill discovery methods under different evaluation configurations,
we make a more comprehensive comparison between the methods using different numbers of bins for quantization. 

\Cref{fig:eval_metrics_num_bins_ant,,fig:eval_metrics_num_bins_hc,,fig:eval_metrics_num_bins_hp,,fig:eval_metrics_num_bins_dk} compare the skill discovery methods on Ant, HalfCheetah, Hopper and D'Kitty,
varying the number of bins for the mutual information estimation.
The results show that on all the four environments,
IBOL outperforms DIAYN-L, VALOR-L and DADS-L in the evaluation metrics of $I(Z; S_T^{\text{(loc)}})$, WSEPIN and SEPIN$@1$ regardless of binning,
which robustly supports IBOL's improved performance.

\section{Diversity of External Returns}

\begin{figure*}[t!]
  \begin{subfigure}[t]{0.08629021746\linewidth}
    \includegraphics[width=1.0\columnwidth]{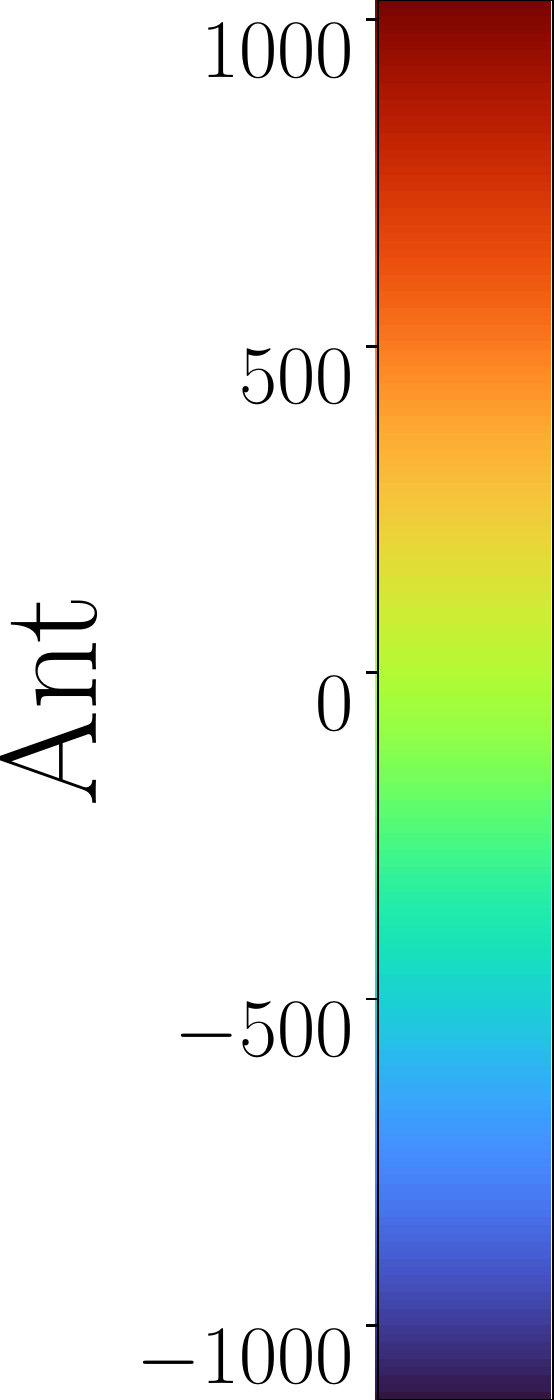}
  \end{subfigure}
  \hfill
  \begin{subfigure}[t]{0.2184274456\linewidth}
    \includegraphics[width=1.0\columnwidth]{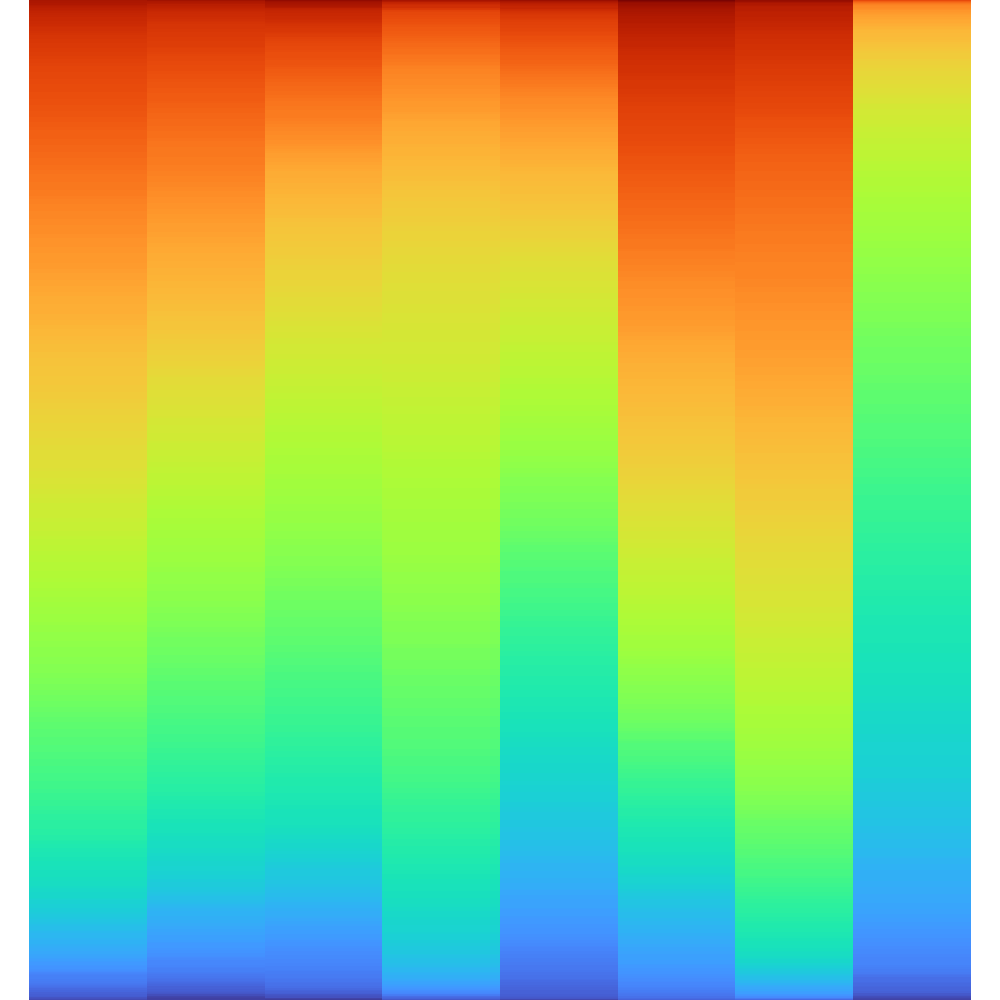}
  \end{subfigure}
  \begin{subfigure}[t]{0.2184274456\linewidth}
    \includegraphics[width=1.0\columnwidth]{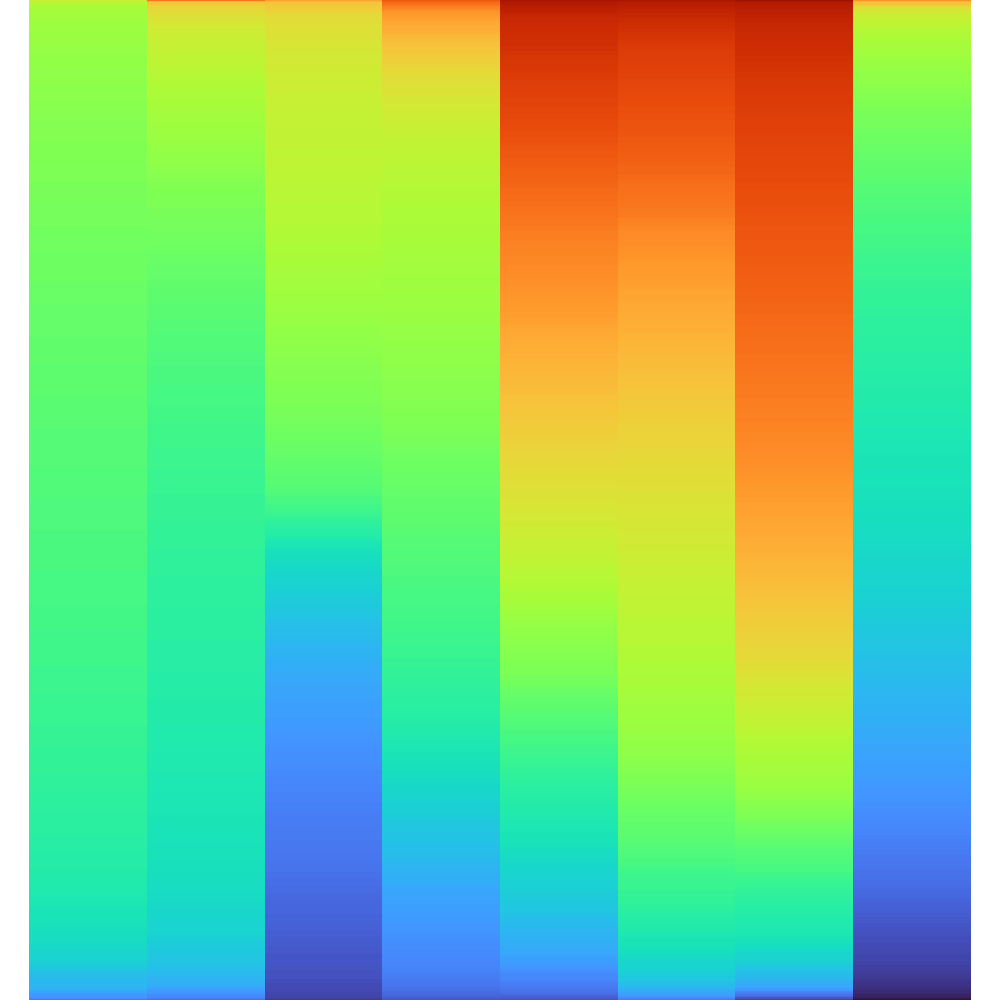}
  \end{subfigure}
  \begin{subfigure}[t]{0.2184274456\linewidth}
    \includegraphics[width=1.0\columnwidth]{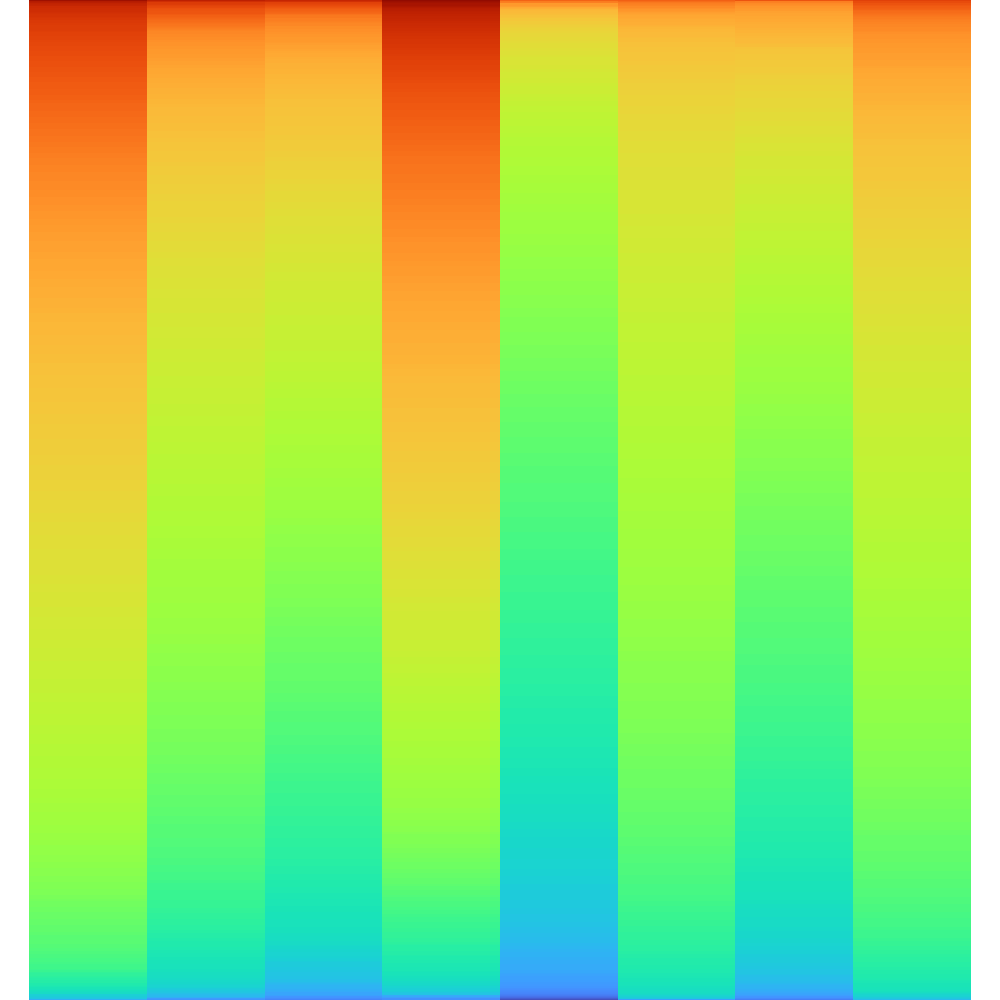}
  \end{subfigure}
  \begin{subfigure}[t]{0.2184274456\linewidth}
    \includegraphics[width=1.0\columnwidth]{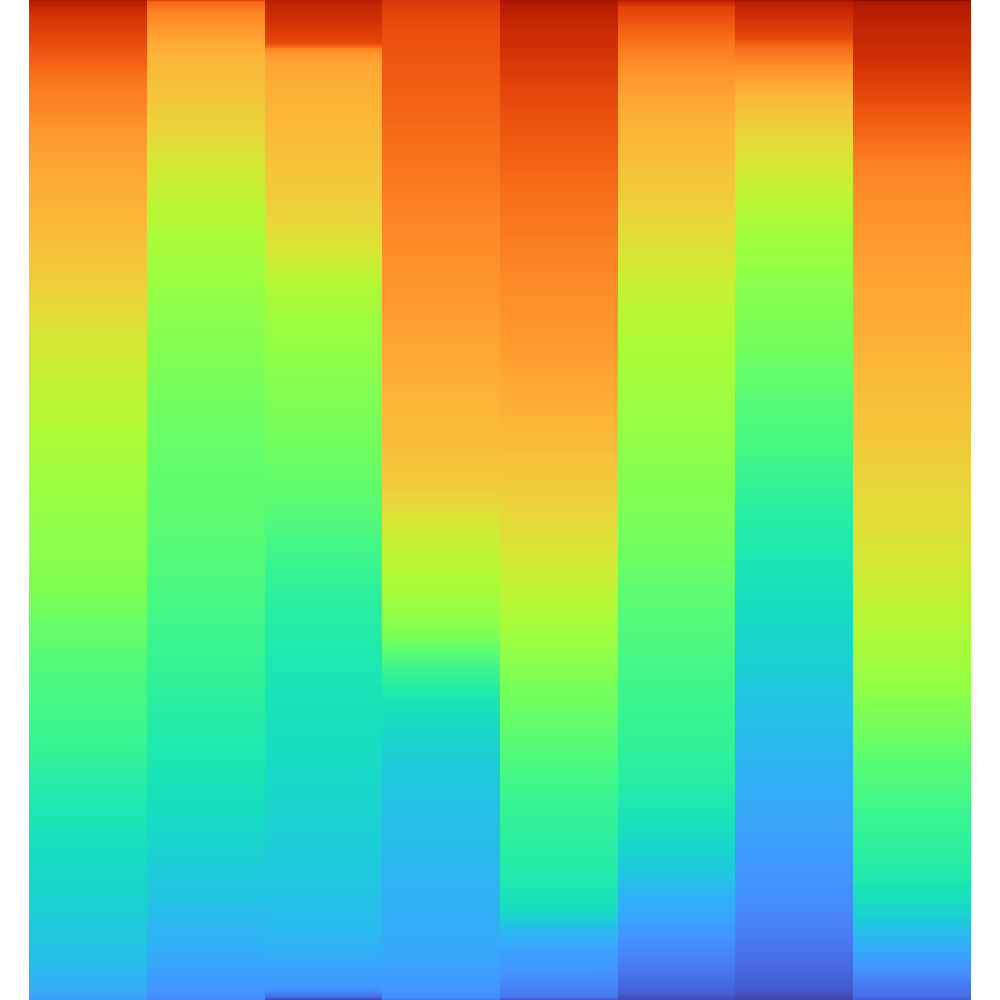}
  \end{subfigure}
  \\

  \begin{subfigure}[t]{0.08629021746\linewidth}
    \includegraphics[width=1.0\columnwidth]{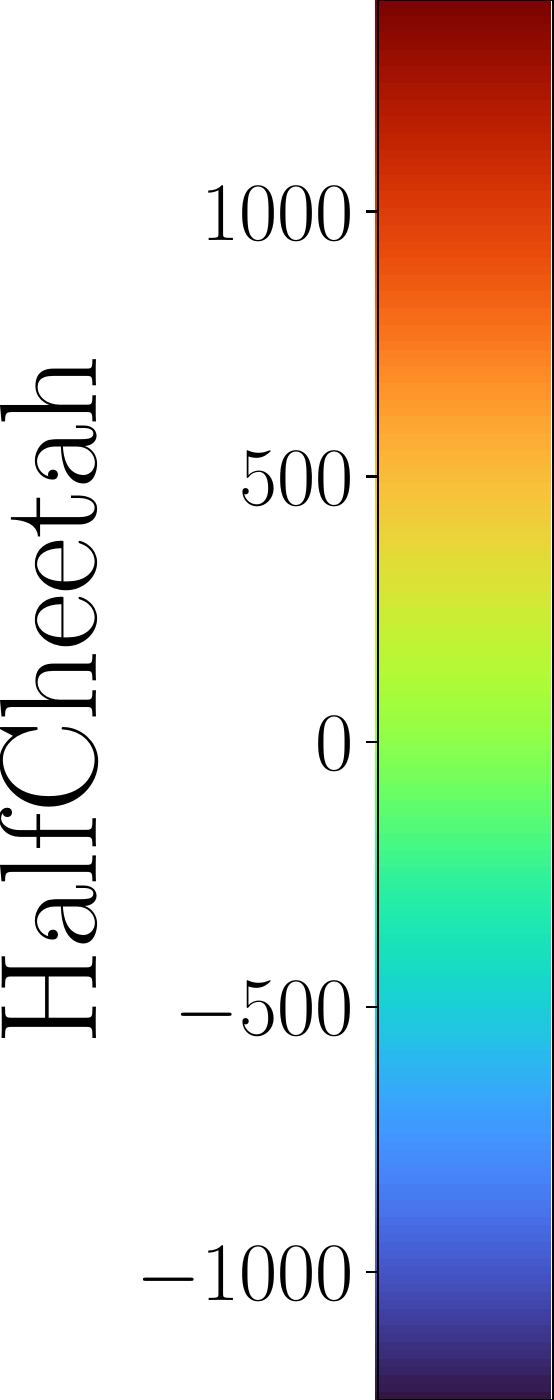}
  \end{subfigure}
  \hfill
  \begin{subfigure}[t]{0.2184274456\linewidth}
    \includegraphics[width=1.0\columnwidth]{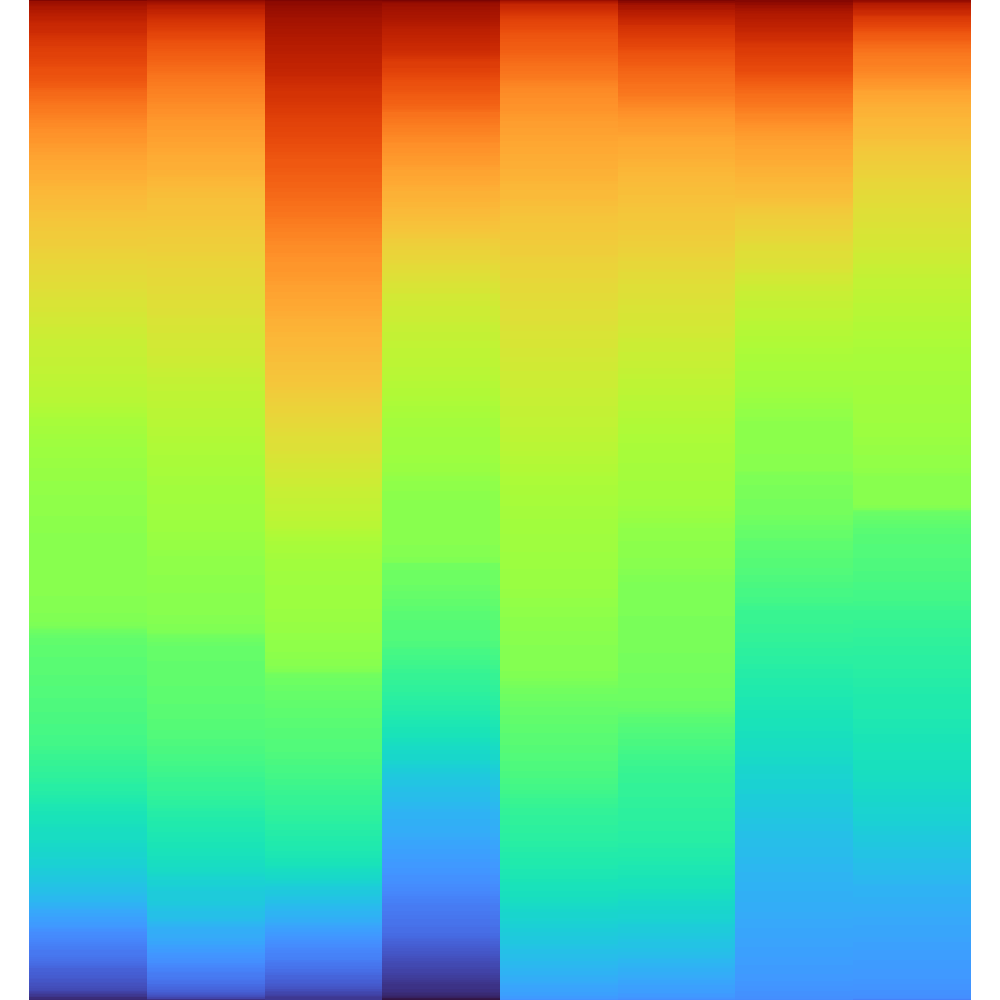}
  \end{subfigure}
  \begin{subfigure}[t]{0.2184274456\linewidth}
    \includegraphics[width=1.0\columnwidth]{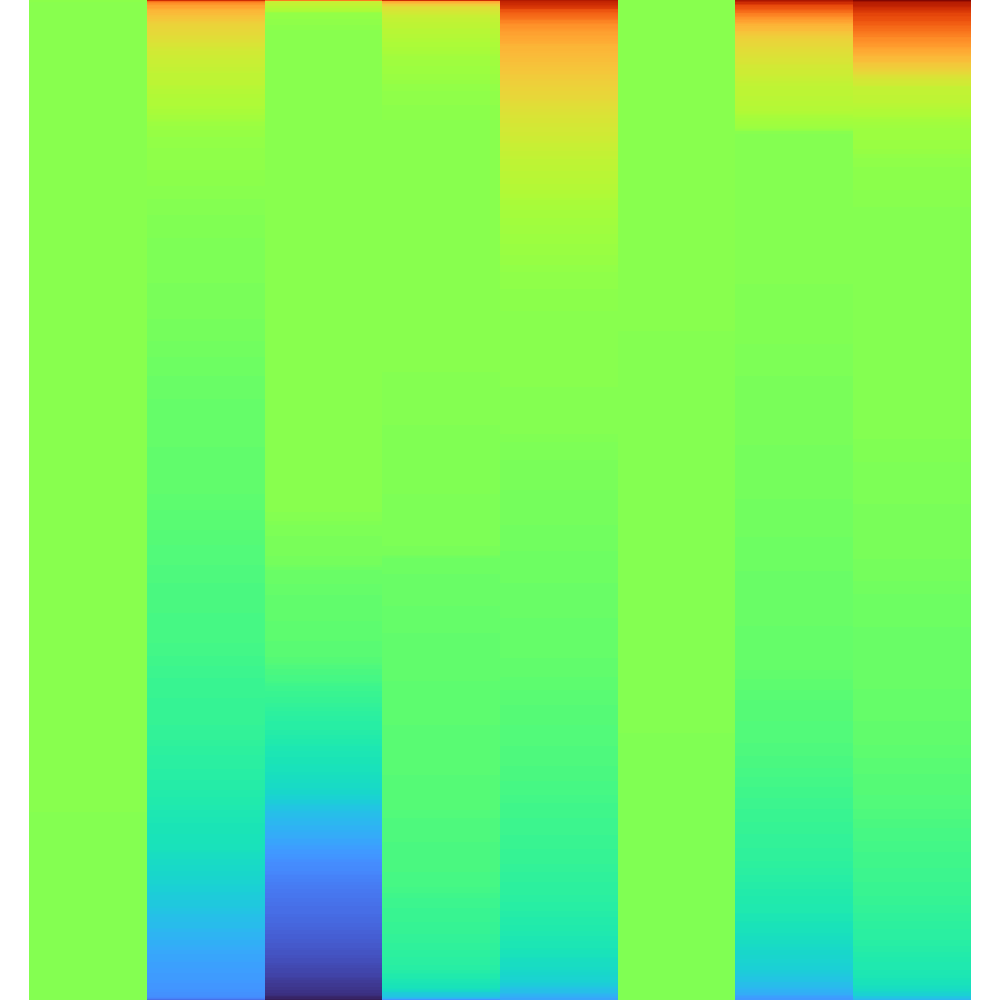}
  \end{subfigure}
  \begin{subfigure}[t]{0.2184274456\linewidth}
    \includegraphics[width=1.0\columnwidth]{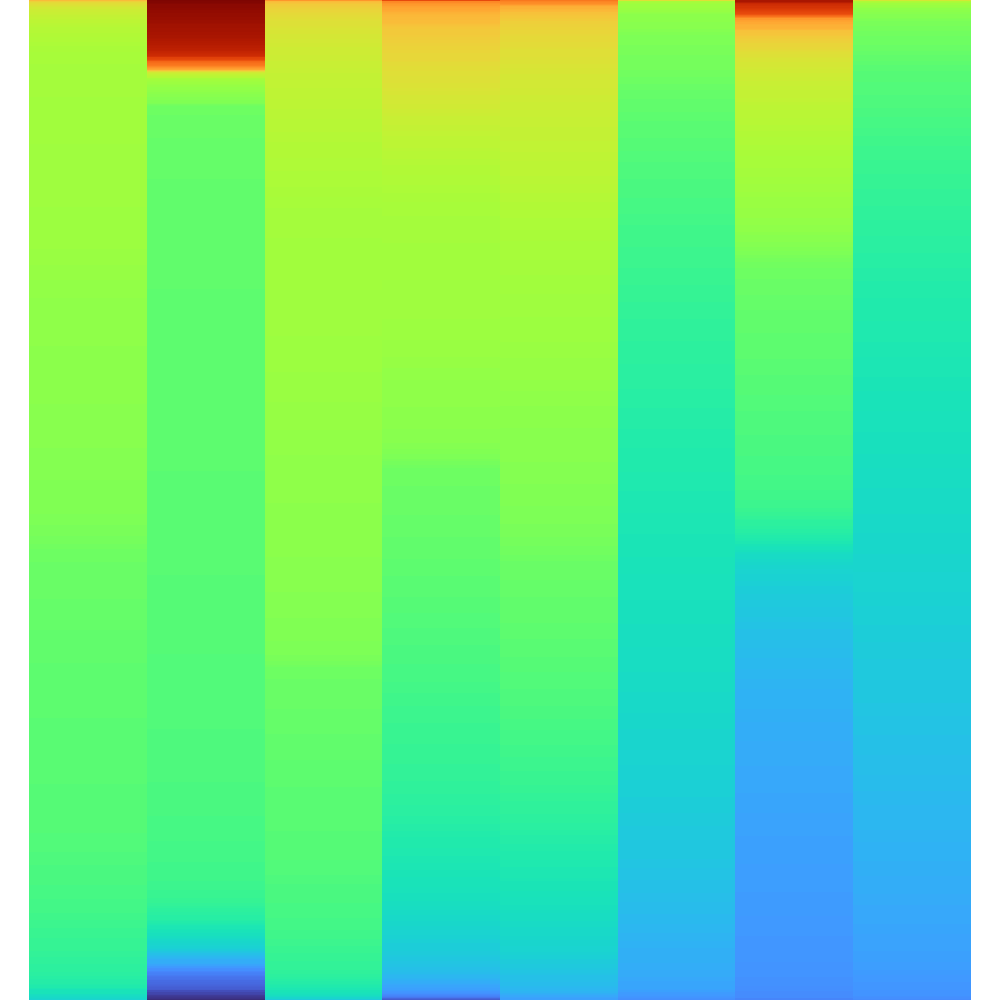}
  \end{subfigure}
  \begin{subfigure}[t]{0.2184274456\linewidth}
    \includegraphics[width=1.0\columnwidth]{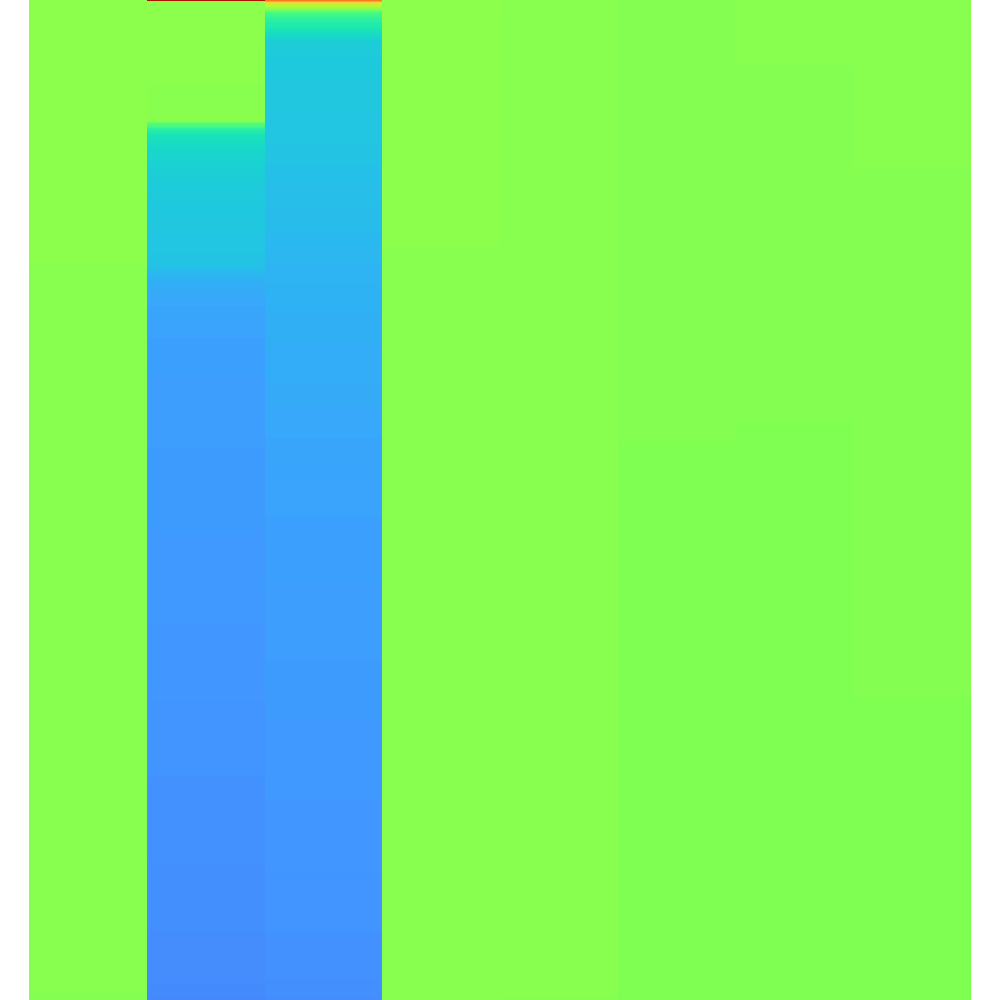}
  \end{subfigure}
  \\

  \begin{subfigure}[t]{0.08629021746\linewidth}
    \includegraphics[width=1.0\columnwidth]{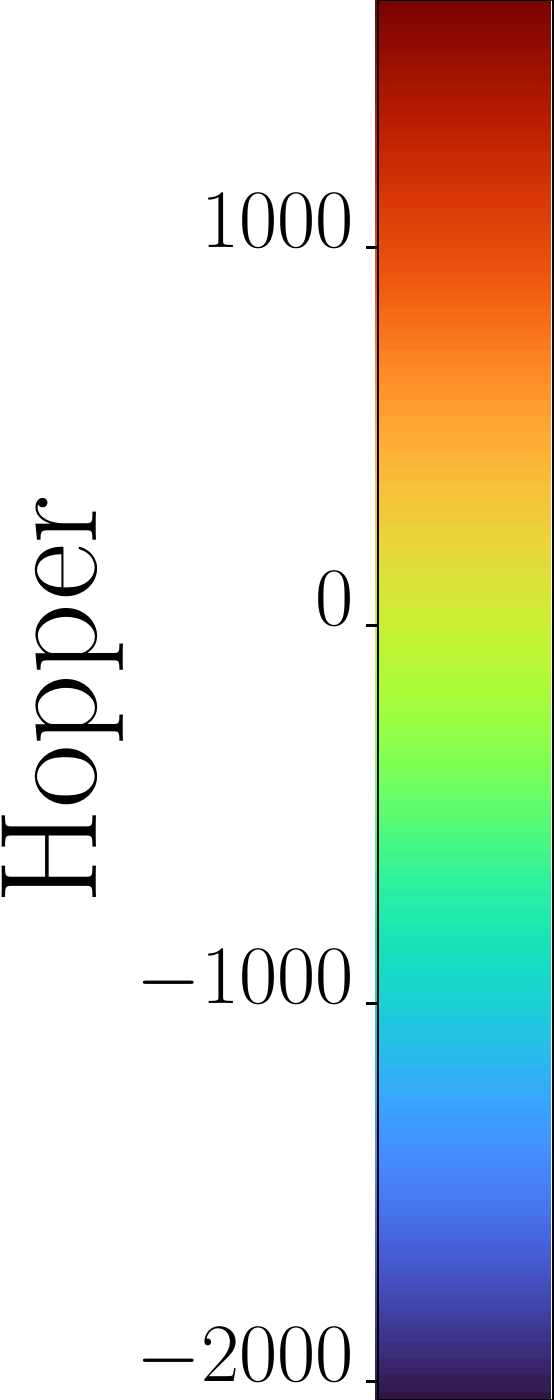}
  \end{subfigure}
  \hfill
  \begin{subfigure}[t]{0.2184274456\linewidth}
    \includegraphics[width=1.0\columnwidth]{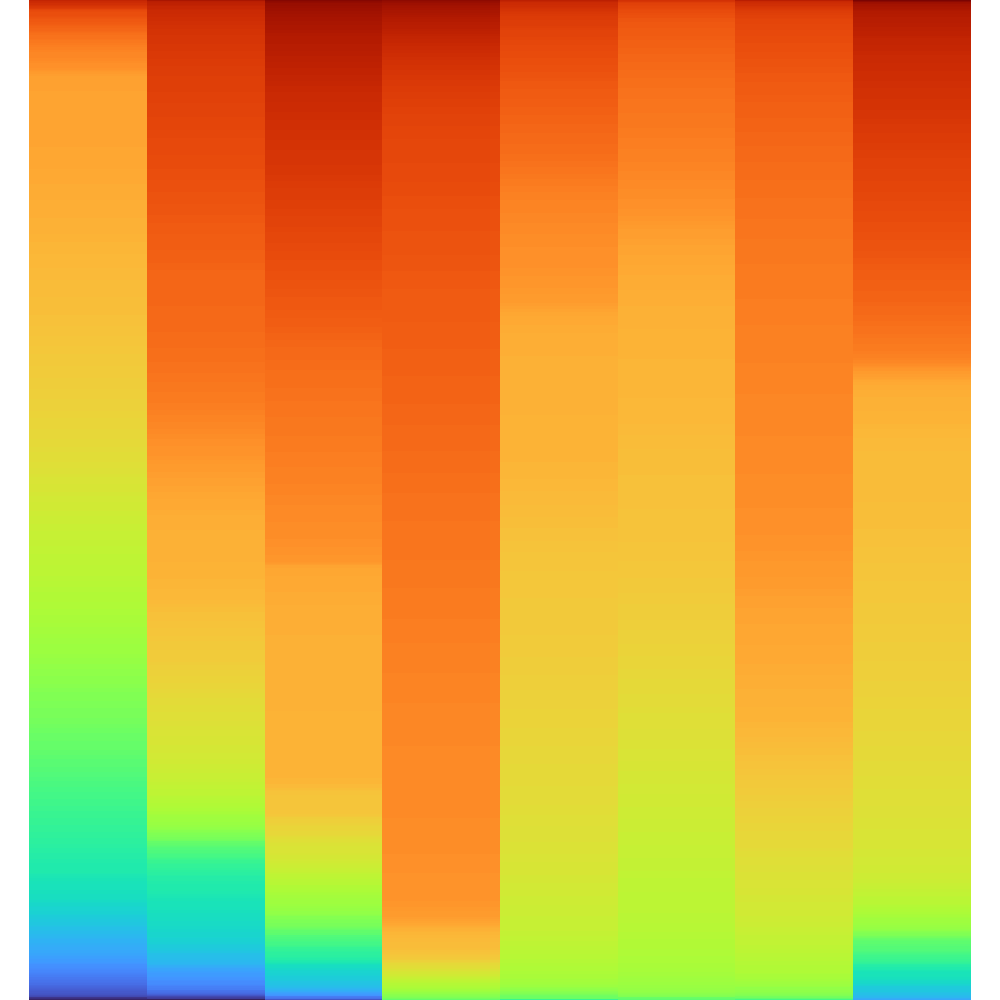}
    \caption{IBOL}
  \end{subfigure}
  \begin{subfigure}[t]{0.2184274456\linewidth}
    \includegraphics[width=1.0\columnwidth]{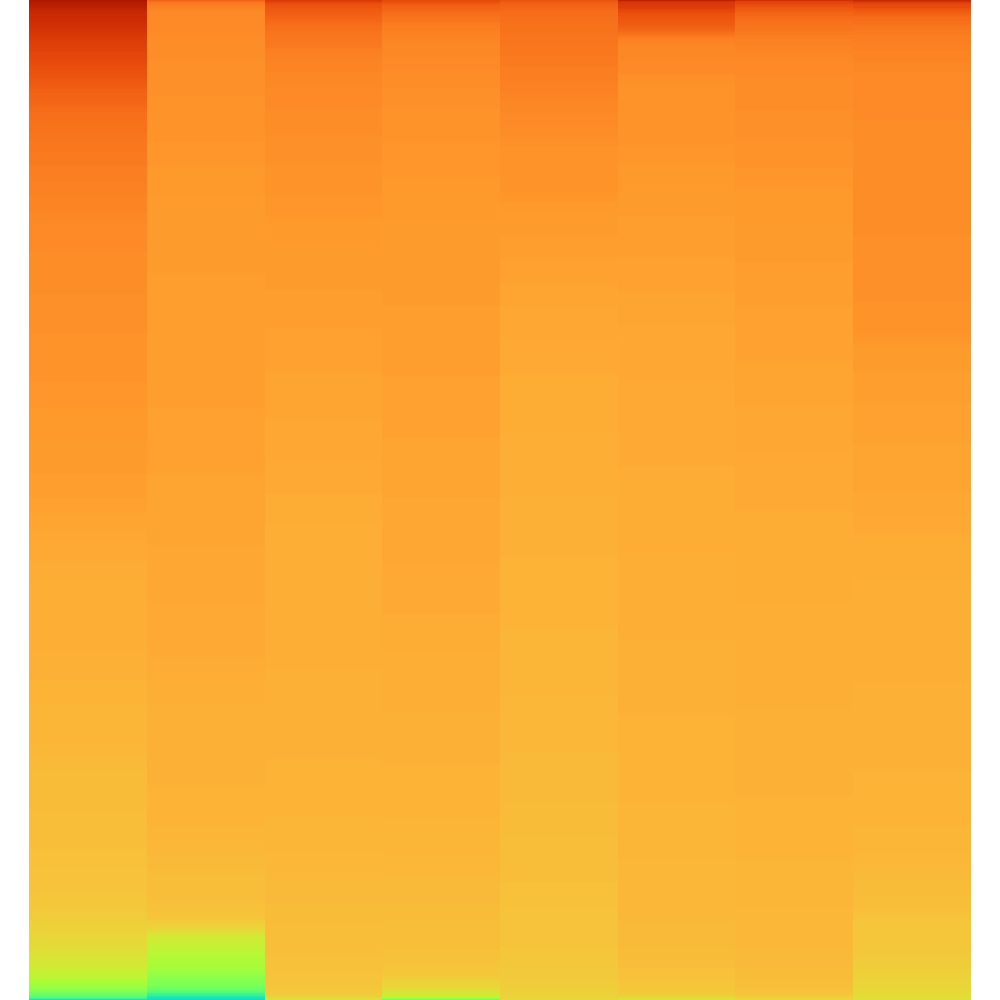}
    \caption{DIAYN-L}
  \end{subfigure}
  \begin{subfigure}[t]{0.2184274456\linewidth}
    \includegraphics[width=1.0\columnwidth]{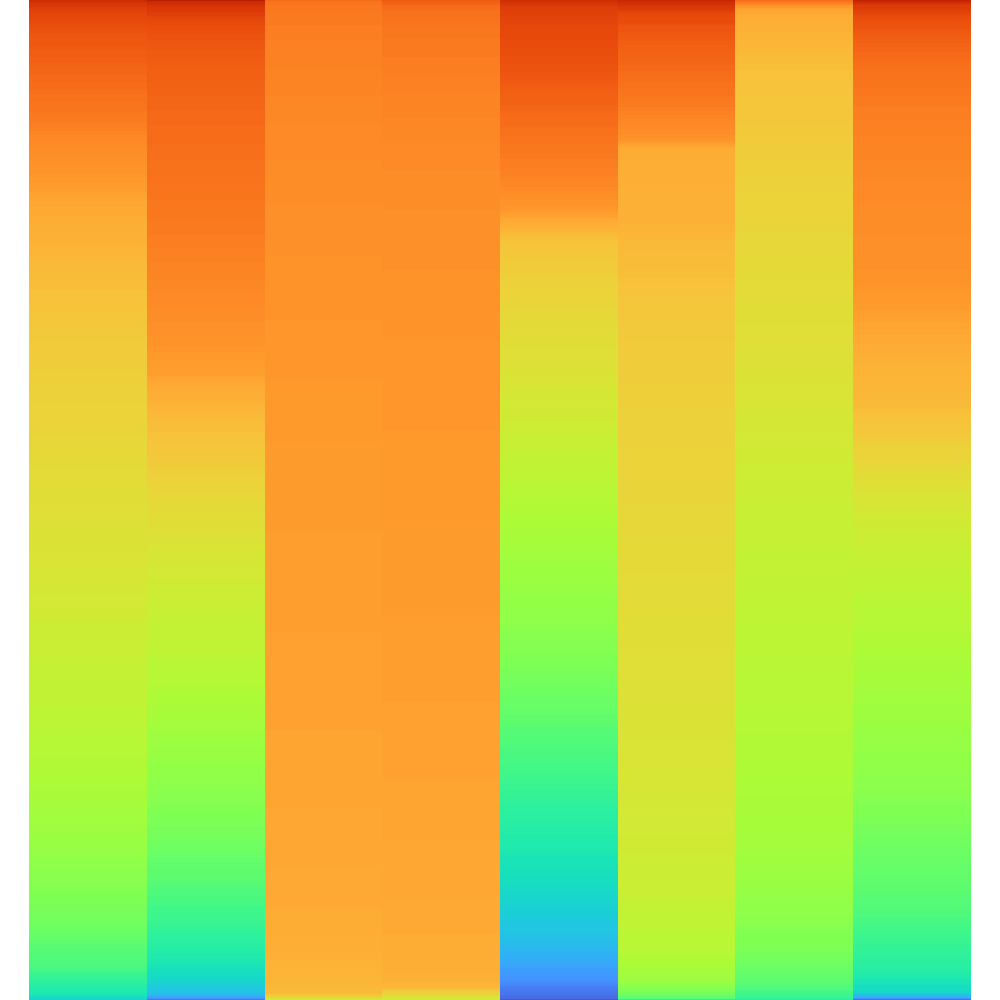}
    \caption{VALOR-L}
  \end{subfigure}
  \begin{subfigure}[t]{0.2184274456\linewidth}
    \includegraphics[width=1.0\columnwidth]{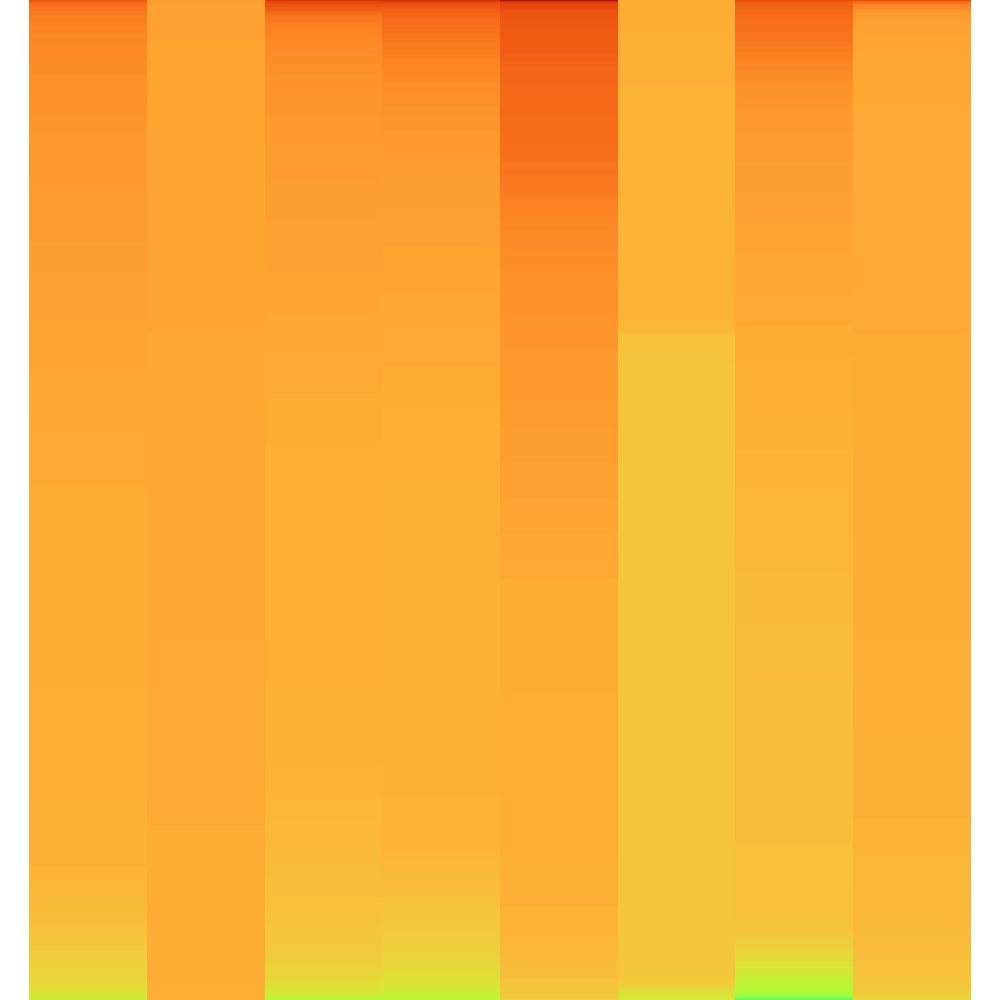}
    \caption{DADS-L}
  \end{subfigure}

  \caption{
    Comparison of IBOL (ours) with the baseline methods, DIAYN-L, VALOR-L and DADS-L,
    in the diversity of external returns for the skills discovered without any rewards.
    For every method, each of the eight vertical bars visualizes the external returns for $2000$ skills sampled randomly with one trained skill policy of the skill discovery method,
    as a stacked histogram with corresponding colors from the color bar on the left.
  }
  \label{fig:external_return_diversity}
\end{figure*}

We qualitatively demonstrate the diversity of external returns the methods receive for their skills.
As the skill discovery methods learn their skill policies without any external rewards,
examining their skills in regard to external returns can be used to 
evaluate the diversity of the skills as well as their usefulness on the original tasks.

We compare IBOL with DIAYN-L, VALOR-L and DADS-L in Ant, HalfCheetah and Hopper.
For every pair of a skill discovery method and an environment,
each of the eight skill policies learned by the method with $d = 2$ in the environment is used to sample trajectories 
given $2000$ random skill latents from their prior distribution, $p(z)$ \ie the standard normal distribution.
\Cref{fig:external_return_diversity} visualizes the results, 
where each vertically stacked histogram denotes the external returns for the $2000$ skills 
with corresponding colors from the color bar for the environment.
In the visualizations, the skills learned by IBOL exhibit not only wider but also more diverse ranges of external returns 
compared to the baseline methods, 
which suggests that IBOL can acquire a more varied and useful set of skills in the environments.

\section{Comparison of Reward Function Choices for the Linearizer}

\textbf{Prior work on hierarchical reinforcement learning.}
We first review previous works that train low-level policies similarly to ours.
SNN4HRL \cite{snn4hrl_florensa2016} trains a high-level policy on top of a context-conditioned low-level policy,
which is pre-trained with a task-related auxiliary reward function that facilitates the desired behaviors as well as exploration.
For example, as a reward for its low-level policy in locomotion tasks,
it uses the speed of the agent combined with an information-theoretic regularizer that encourages diversity.
FuN \cite{fun_vezhnevets2017} jointly trains both a high-level policy and a low-level goal-conditioned policy rewarded by the cosine similarity between goals and directions in its latent space.
HIRO \cite{hiro_nachum2018} takes a similar approach to FuN, but its high-level policy generates goals in the raw state space, without having a separate latent goal space.
Its low-level policy is guided by the Euclidean distance instead of the cosine similarity.
\citet{nearoptimal_nachum2019} train a goal-conditioned low-level policy with the Huber loss, which is a variant of the Euclidean distance, in a learned representation space within the framework of sub-optimality.

In contrast to these approaches,
we train the linearizer, which can be viewed as a low-level policy,
with the reward in the inner-product form.
Also, we reward the linearizer with the state difference between \textit{macro} time steps:
$(s_{(i + 1) \cdot \ell} - s_{i \cdot \ell})$, where $\ell$ is the interval of the macro step
(we use $\ell = 10$ in our experiments).

\begin{figure}[t!]
  \centering
  \includegraphics[width=\linewidth]{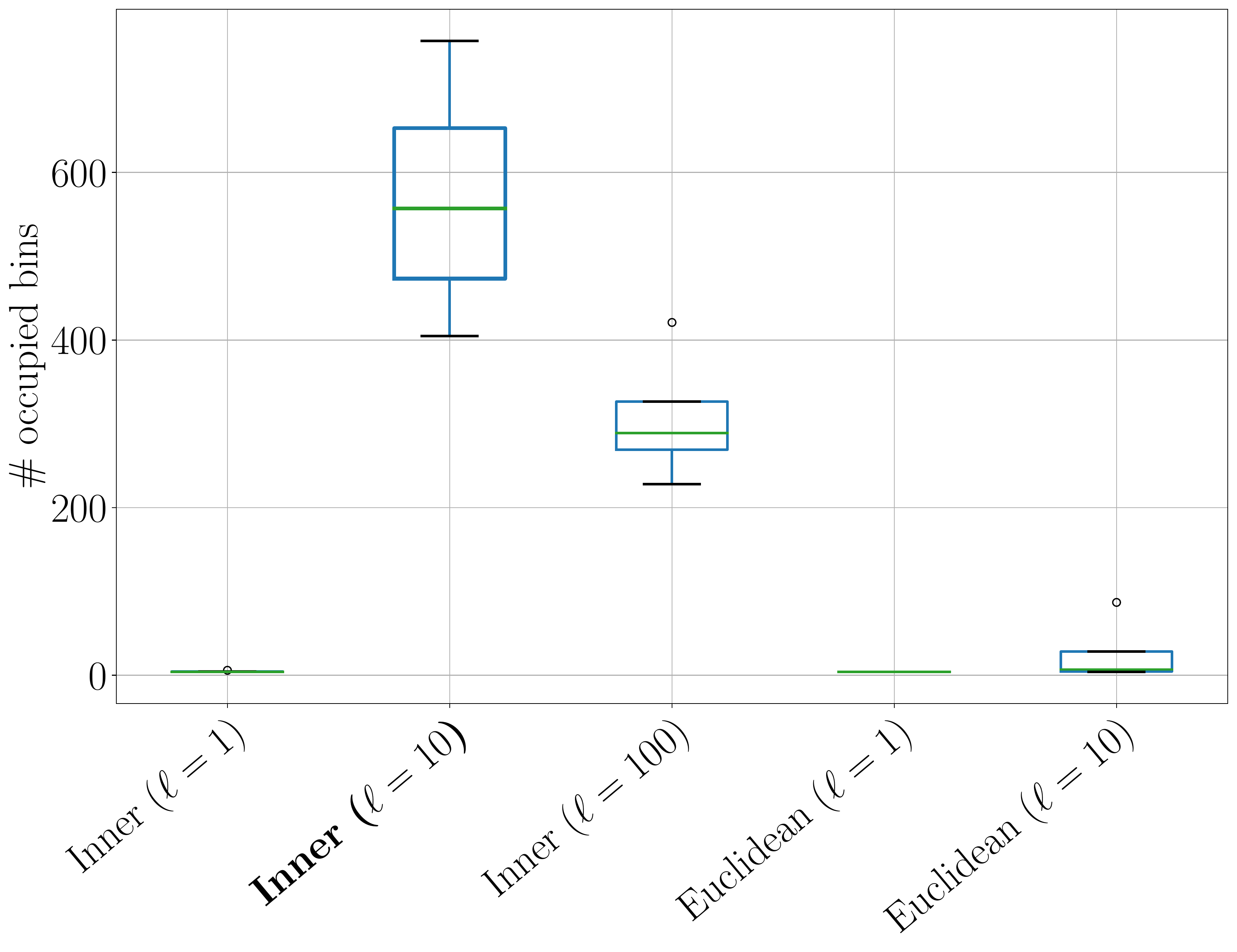}
  \caption{
      Comparison of various reward function choices for the linearizer.
      The box plot shows the state coverage of each reward function, measured by the number of bins
      occupied by the $2000$ trajectories in the state space.
      We use four random seeds for each method.
  }
  \label{fig:reward}
\end{figure}

\textbf{Comparison of different reward function choices.}
We now compare our reward function for the linearizer with other choices.
We experiment on Ant and evaluate them by their state coverage in the $x$-$y$ plane.
We sample $2000$ trajectories from each of the linearizers,
where we only change the values of $x$ and $y$ dimensions in the goal space and set the other dimensions' value to $0$.
We measure the state coverage by the number of bins occupied by the trajectories
out of $1024$ equally divided bins in the $x$-$y$ plane.
For the comparison, we test different values of $\ell = 1, 10, 100$ with our inner-product reward function,
as well as one in the form of the Euclidean distance as in HIRO \cite{hiro_nachum2018} with $\ell = 1, 10$.
Since the Euclidean distance reward function requires the specification of the valid goal ranges, we employ the goal range values used by HIRO.
As a consequence, we follow the practice of HIRO to exclude the state dimensions for velocities in specifying the goal space for the Euclidean distance reward function.
On the contrary, we use the full state dimensions to design the goal space for the inner-product reward function.

\Cref{fig:reward} compares the performances of the reward function choices.
It suggests that using an appropriate size of the macro step (\ie $\ell = 10$) improves the state coverage,
especially exhibiting drastic performance improvement over the case of $\ell = 1$.
We also observe that our inner-product reward function shows a better state coverage compared to the Euclidean reward function.

\section{Experimental Details}
\label{sec:exp_details}

\subsection{Implementation}

We employ garage \cite{garage_2019} and PyTorch \cite{pytorch_paszke2019} to implement
IBOL, DIAYN \cite{diayn_eysenbach2019}, VALOR \cite{valor_achiam2018} and DADS \cite{dads_sharma2020}.
We use the official implementations
for EDL\footnote{\url{https://github.com/victorcampos7/edl}} \cite{edl_campos2020}
and SeCTAR\footnote{\url{https://github.com/wyndwarrior/Sectar}} \cite{sectar_coreyes2018}
with additional tuning of hyperparameters to ensure fair comparisons.

\subsection{Environments}

We experiment with robot simulation environments in MuJoCo \cite{mujoco_todorov2012}:
Ant, HalfCheetah, Hopper and Humanoid from OpenAI Gym \cite{openaigym_brockman2016} adopting the configurations by \citet{dads_sharma2020} and 
D'Kitty with random dynamics from ROBEL \cite{robel_ahn2020} with the setups provided by \citet{sharma2020_emergent}.
We use a maximum episode horizon of $200$ environment steps for Ant, HalfCheetah and D'Kitty, $500$ for Hopper and $1000$ for Humanoid.
Note that D'Kitty and Humanoid have variable episode horizons, and we use an alive bonus of $3e-2$ at each step in the training of the linearizers for Humanoid to stabilize the training.

For the linearizer, we omit the locomotion coordinates of the torso ($x$ and $y$ for Ant, Humanoid and D'Kitty, and $x$ for the others) from the input of the policy.
Note that the linearizer could be agnostic to the agent's global location since its rewards are computed only with the change of the state.
On the other hand, we retain them for skill discovery policies and meta-controller policies since, without those coordinates,  the expressiveness of learnable skills may be restricted. %

\begin{figure}[t!]
  \centering
  \includegraphics[width=\linewidth]{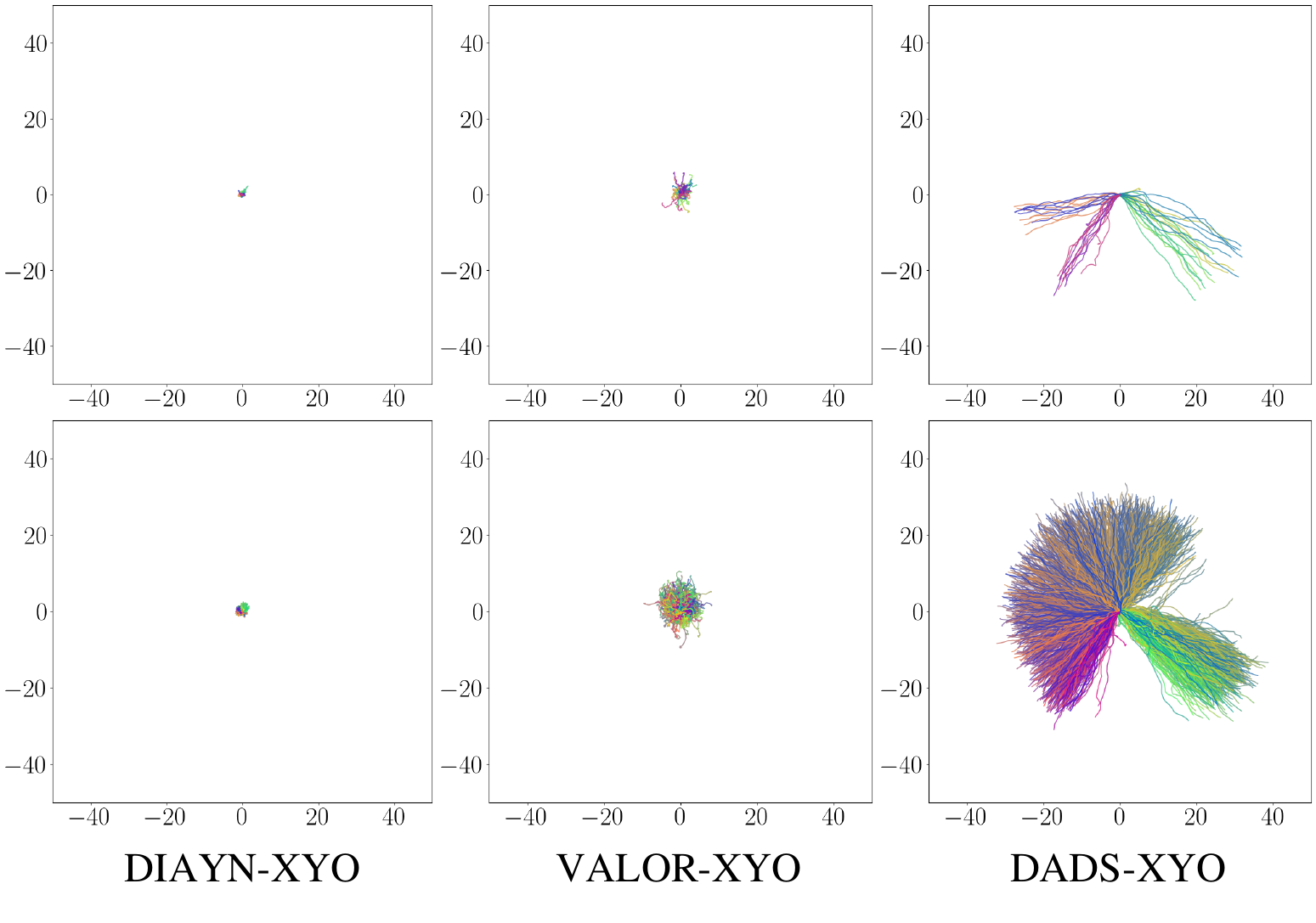}
  \caption{
    Visualization of the $x$-$y$ traces of the skills for Ant discovered by each baseline method trained with the omission of the $x$-$y$ coordinates from the inputs and the $x$-$y$ prior \cite{dads_sharma2020}.
    The same skill latents are used with Figure \ref{fig:ant_xy} of the main paper.
  }
  \label{fig:ant_xyo}
\end{figure}

However, as DADS originally omits the $x$-$y$ coordinates from the inputs in Ant \cite{dads_sharma2020}, we also test baseline methods of $d = 2$ with both the omission and the $x$-$y$ prior \cite{dads_sharma2020}, denoted with the suffix `-XYO', in \Cref{fig:ant_xyo}.

\subsection{Models}

In the experiments, we use an MLP with two hidden layers of $512$ dimensions for each non-recurrent learnable component 
except for the linearizer, which uses two hidden layers of $1024$ dimensions.
We use the $\tanh$ and ReLU nonlinearities for the policies and the others, respectively.
We model the outputs of the linearizer and the meta-controller for downstream tasks with the factorized Gaussian distribution followed by a $\tanh$ transformation to fit into the action space of environments.
We use the Beta distribution policies for the skill discovery methods.
To feed policies with the skill latent variable, we concatenate the skill latent $z$ for each episode with its state $s_t$ at every time step $t$.

For the trajectory encoder of IBOL and VALOR, we use a bidirectional LSTM with a $512$-dimensional hidden layer followed by two $512$-dimensional FC layers.
When training VALOR without the linearizer, we use a subset of the full state sequence of each trajectory with evenly spaced states to match the effective horizon with VALOR-L, following \citet{valor_achiam2018}.
We employ the original implementation choice of DADS to predict $\Delta s = s' - s$ (instead of $s'$) from $s$ and $z$ with its skill dynamics model \cite{dads_sharma2020}.
Both $s$ and $\Delta s$ are batch-normalized, with a fixed covariance matrix of $I$ and a Gaussian mixture model with four heads, again following \citet{dads_sharma2020}.

\subsection{Training}

We use the Adam optimizer \cite{adam_kingma2015} with a learning rate of $1e-4$ for skill discovery methods and $3e-4$ for the others.
We normalize each dimension of states, which is important since it helps skill discovery methods equally focus on every dimension of the state space
rather than solely on large-scale dimensions.
Note that while we observe that the skill discovery methods primarily focus on the locomotion dimensions in the absence of the $x$-$y$ prior \cite{dads_sharma2020} as in Figure \ref{fig:ant_xy} from the main paper,
this is not due to the scale of those dimensions, as all the state dimensions are normalized.
We hypothesize it is because the locomotion dimensions are those which can have high informativeness with the skill latent variable.
When training meta-controllers or skill policies with the linearizer,
we use the exponential moving average.
For the rest, we use the mean and standard deviation pre-computed from $10000$ trajectories with an episode length of $50$.
Meta-controllers for downstream tasks and skill policies use the mode of each output distribution from their lower-level policies.

At every epoch of the training of the linearizer or meta-controller for downstream tasks, we collect ten trajectories for Ant, HalfCheetah, Hopper and D'Kitty, and five trajectories for Humanoid.
For the skill discovery methods with the linearizer, at each epoch $64$ trajectories are sampled for Ant, HalfCheetah and D'Kitty and 
$32$ for Hopper and Humanoid.
When training the methods without the linearizer (\eg VALOR-XY), we collect ten trajectories for Ant, since their effective horizon is longer than that with the linearizer.

\textbf{The linearizer}.
We train the linearizer using SAC \cite{sac_haarnoja2018} with the automatic entropy adjustment \cite{essac_haarnoja2018}
for $8e4$ epochs for D'Kitty, $3e5$ epochs for Humanoid and $1e5$ epochs for the others.
We apply $4$ gradient steps and consider training with and without a replay buffer, 
where rewards are normalized with their exponential moving average without a buffer and $2048$-sized mini-batches are used with a buffer of $1e6$.
We set the initial entropy to $0.1$, the target entropy to $-dim(\mathcal{A}) / 2$, the target smoothing coefficient to $0.005$ and the discount rate to $0.99$.
We choose a prior goal distribution for each environment from $\{ \text{Beta}(1, 1), \text{Beta}(2, 2) \}$.
We determine the hyperparameters based on the state coverage of the trained linearizer.

\begin{figure*}[t!]
  \begin{subfigure}[t]{0.95\linewidth}
    \makebox[20pt]{\raisebox{50pt}{\rotatebox[origin=c]{90}{IBOL ($\lambda = 1.5$)}}}
    \centering
    \includegraphics[width=0.22\columnwidth]{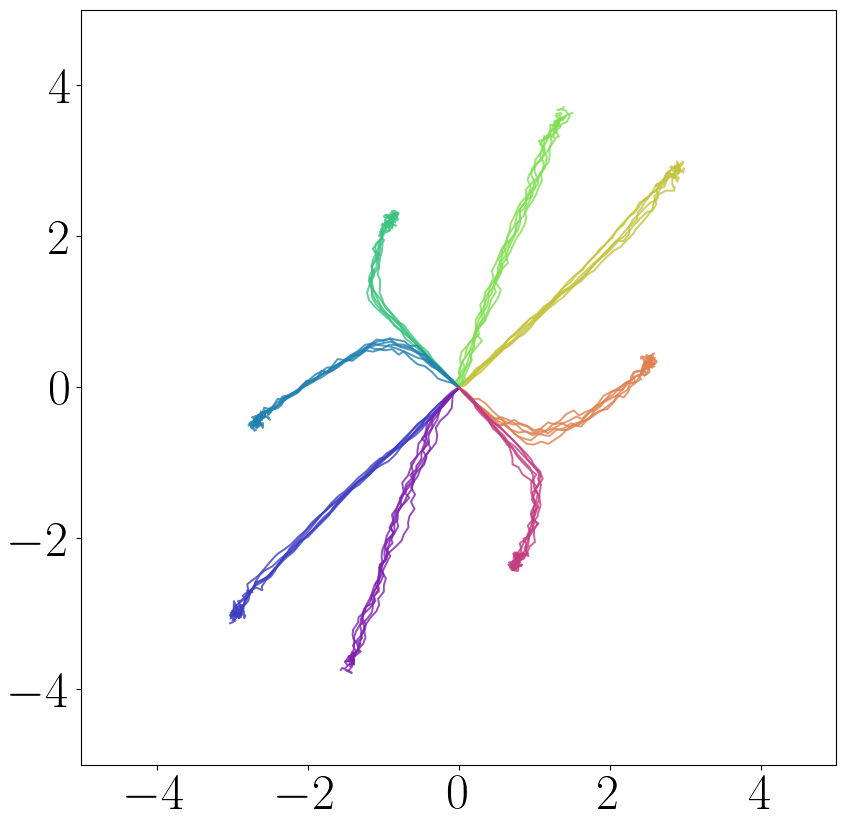}
    \includegraphics[width=0.22\columnwidth]{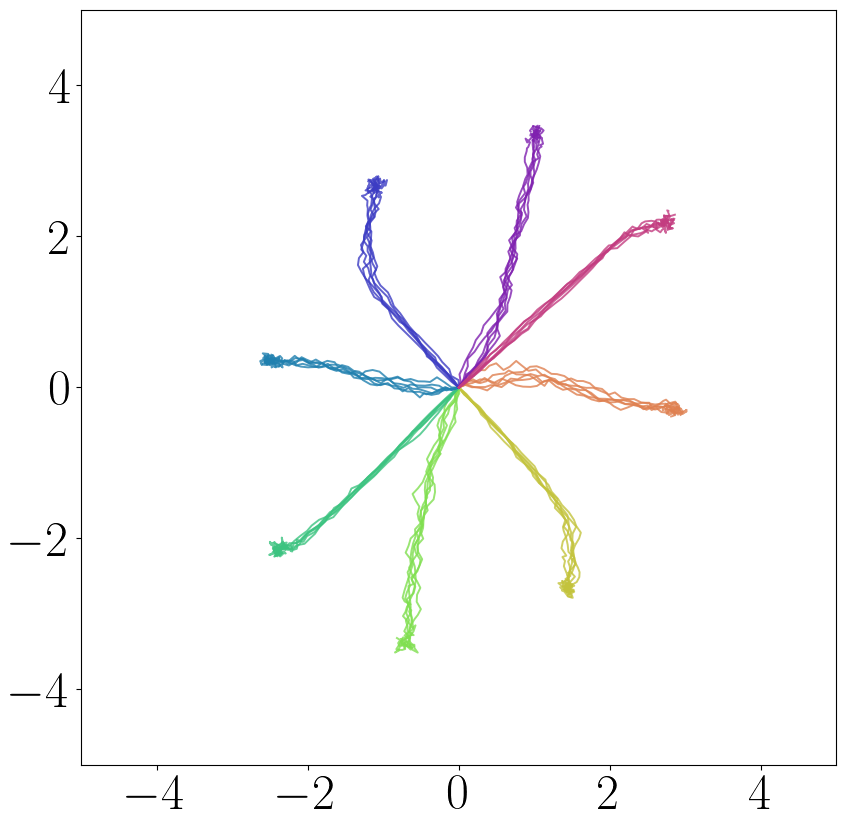}
    \includegraphics[width=0.22\columnwidth]{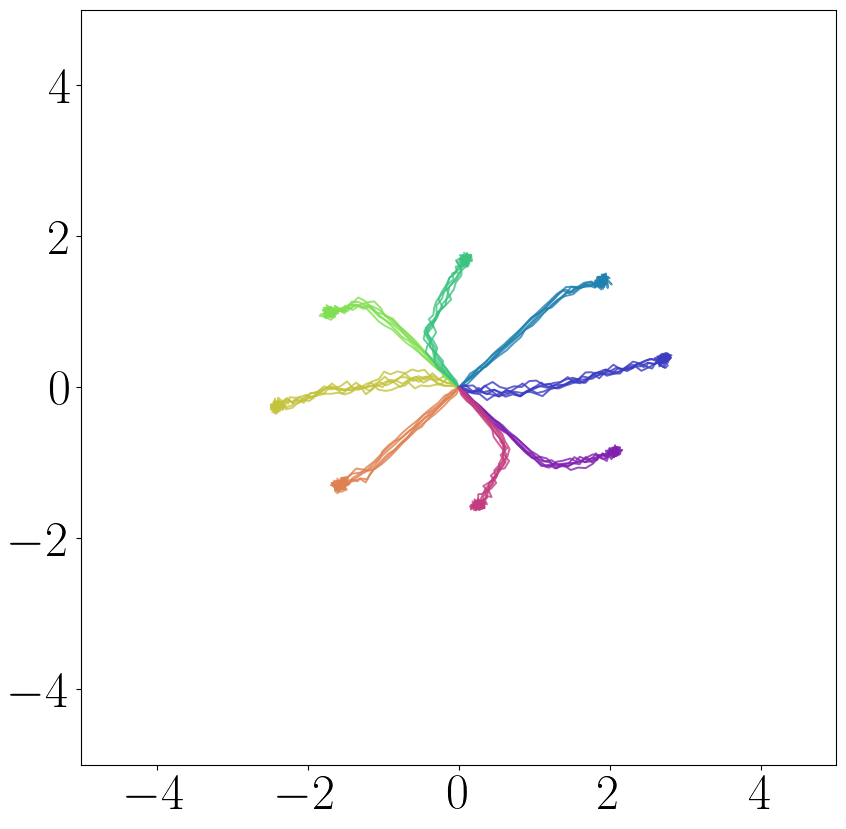}
  \end{subfigure}

  \begin{subfigure}[t]{0.95\linewidth}
    \makebox[20pt]{\raisebox{50pt}{\rotatebox[origin=c]{90}{IBOL ($\lambda = 0.45$)}}}
    \centering
    \includegraphics[width=0.22\columnwidth]{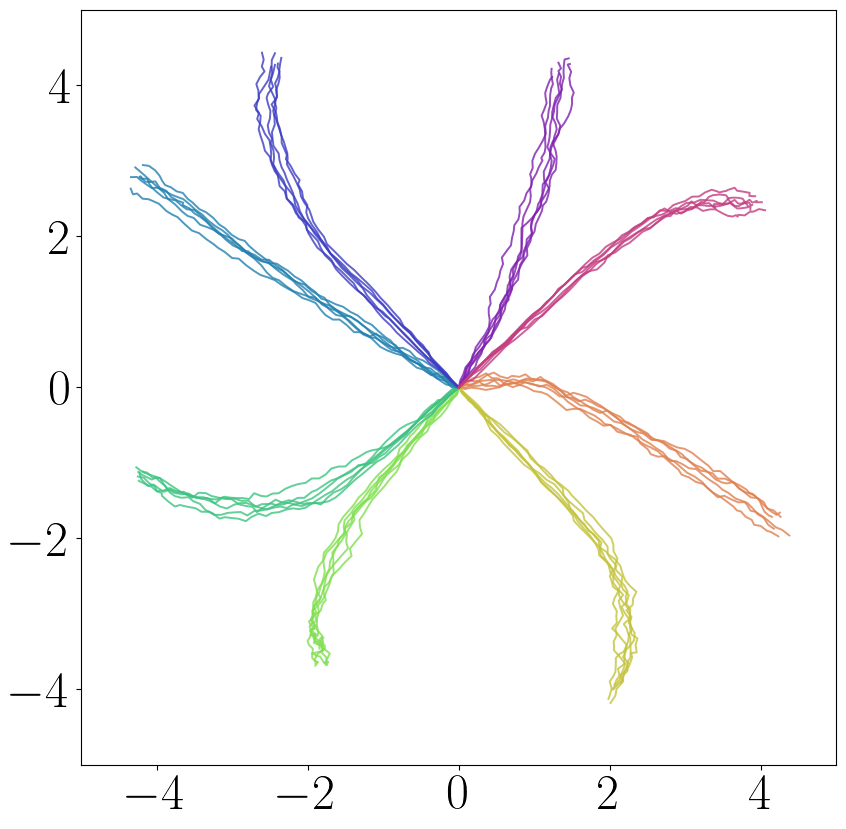}
    \includegraphics[width=0.22\columnwidth]{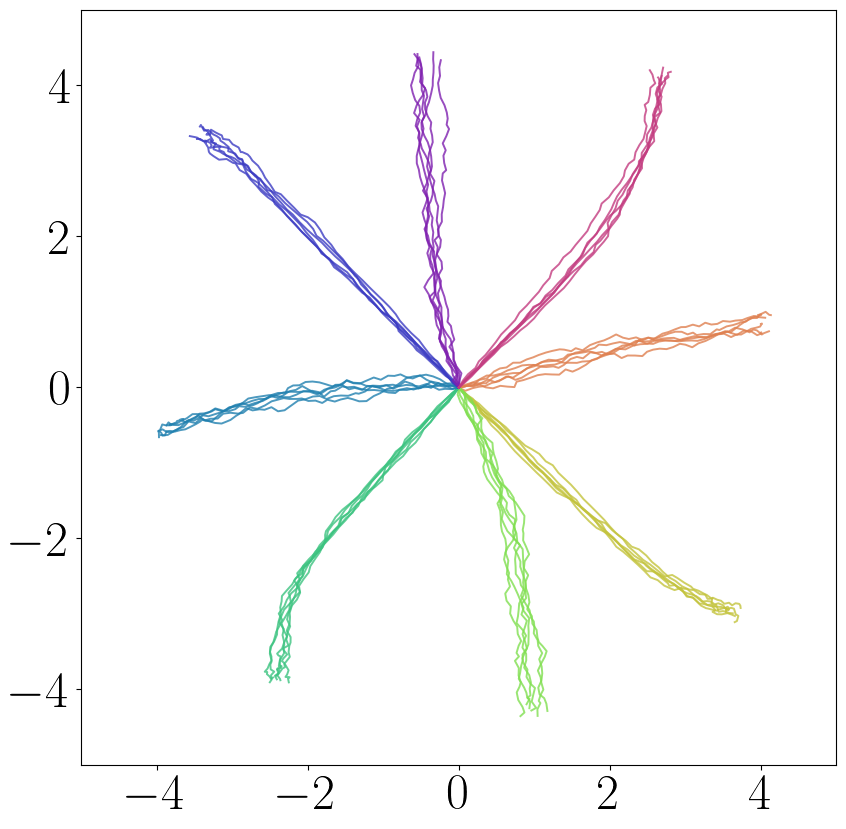}
    \includegraphics[width=0.22\columnwidth]{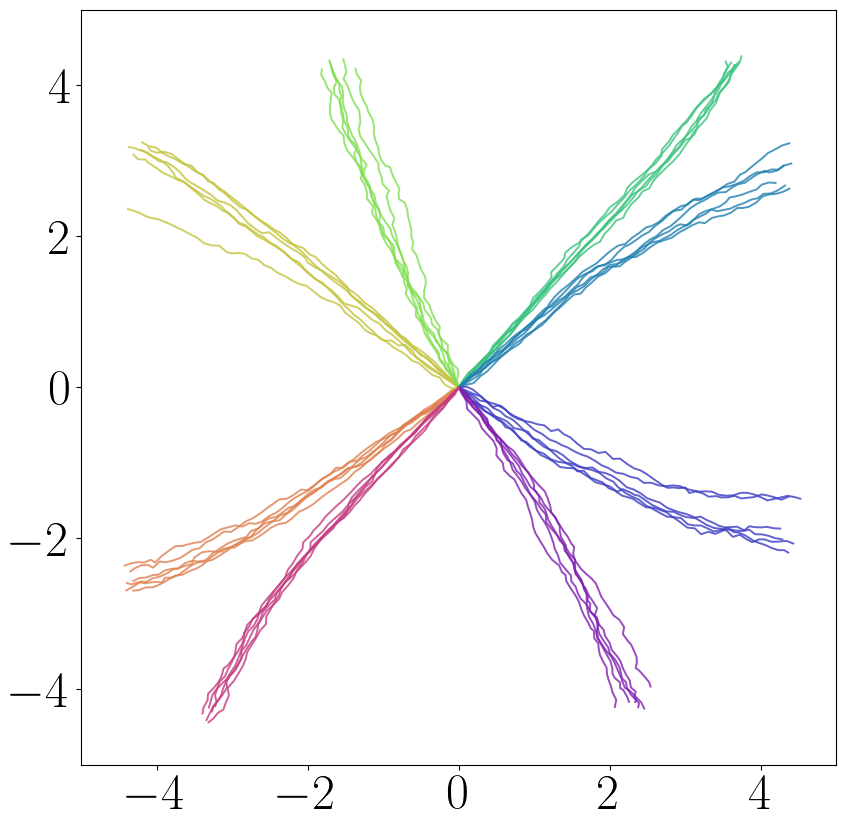}
  \end{subfigure}

  \begin{subfigure}[t]{0.95\linewidth}
    \makebox[20pt]{\raisebox{50pt}{\rotatebox[origin=c]{90}{IBOL ($\lambda = 0.15$)}}}
    \centering
    \includegraphics[width=0.22\columnwidth]{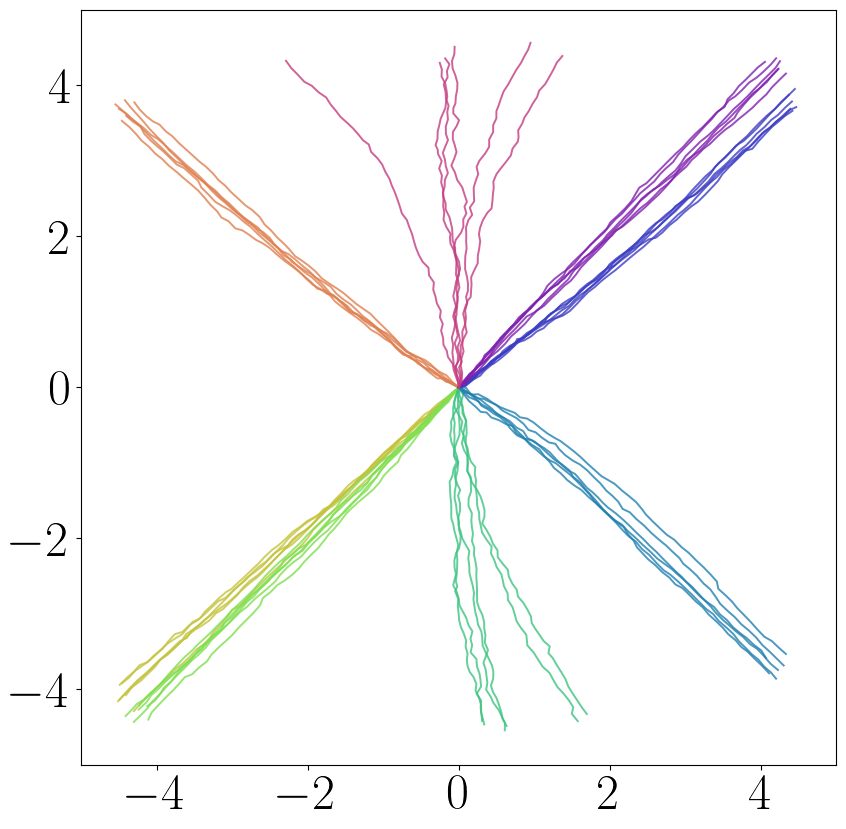}
    \includegraphics[width=0.22\columnwidth]{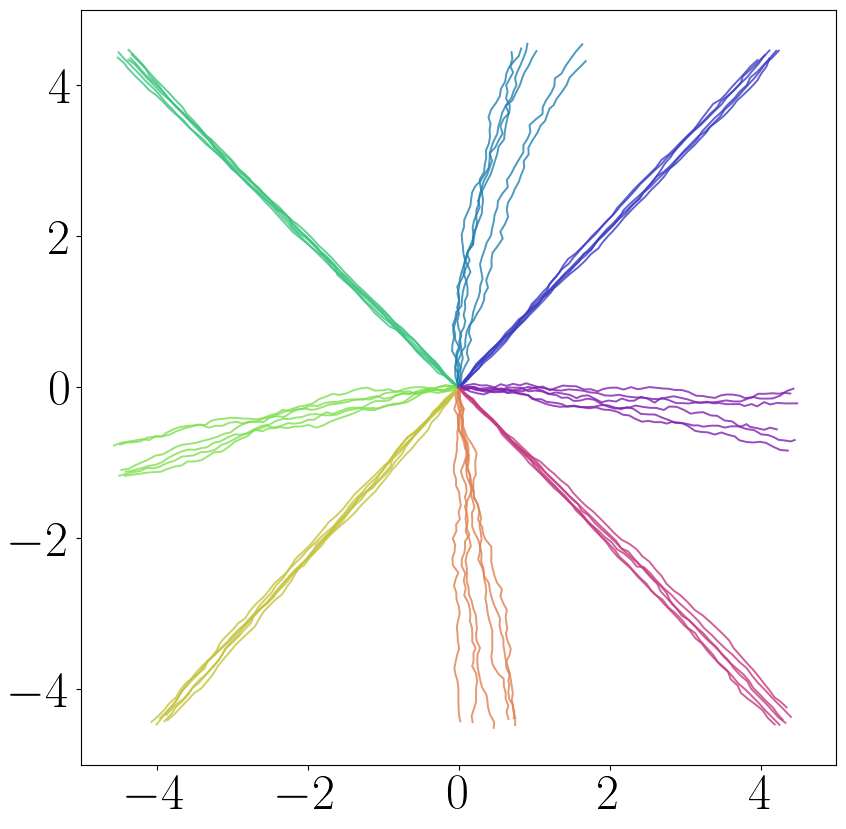}
    \includegraphics[width=0.22\columnwidth]{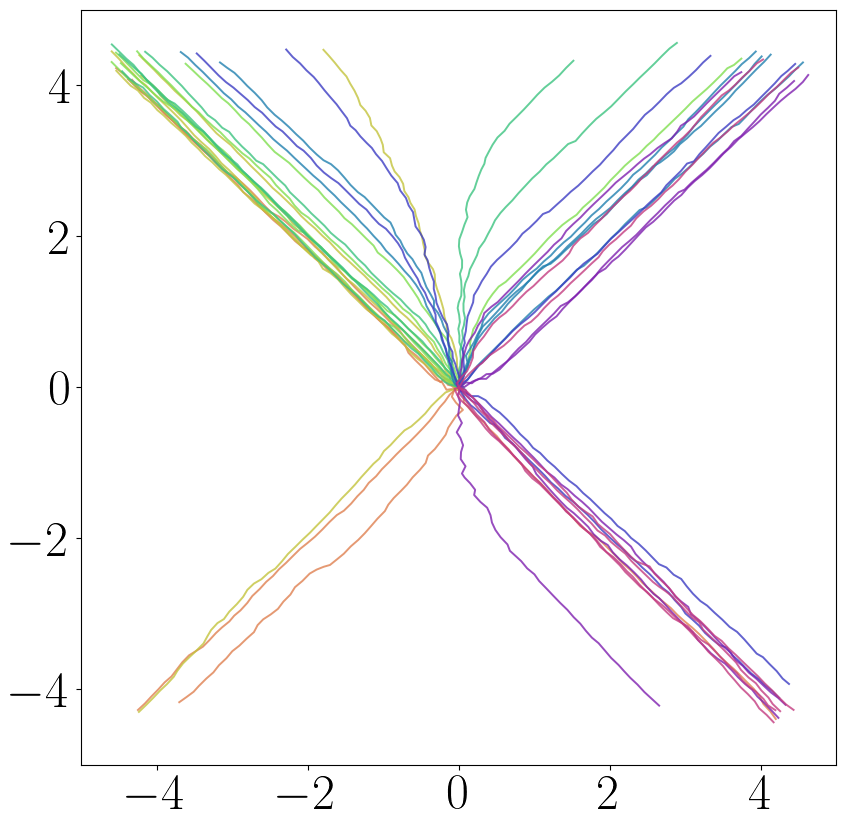}
  \end{subfigure}

  \begin{subfigure}[t]{0.95\linewidth}
    \makebox[20pt]{\raisebox{50pt}{\rotatebox[origin=c]{90}{IBOL W/o $u$}}}
    \centering
    \includegraphics[width=0.22\columnwidth]{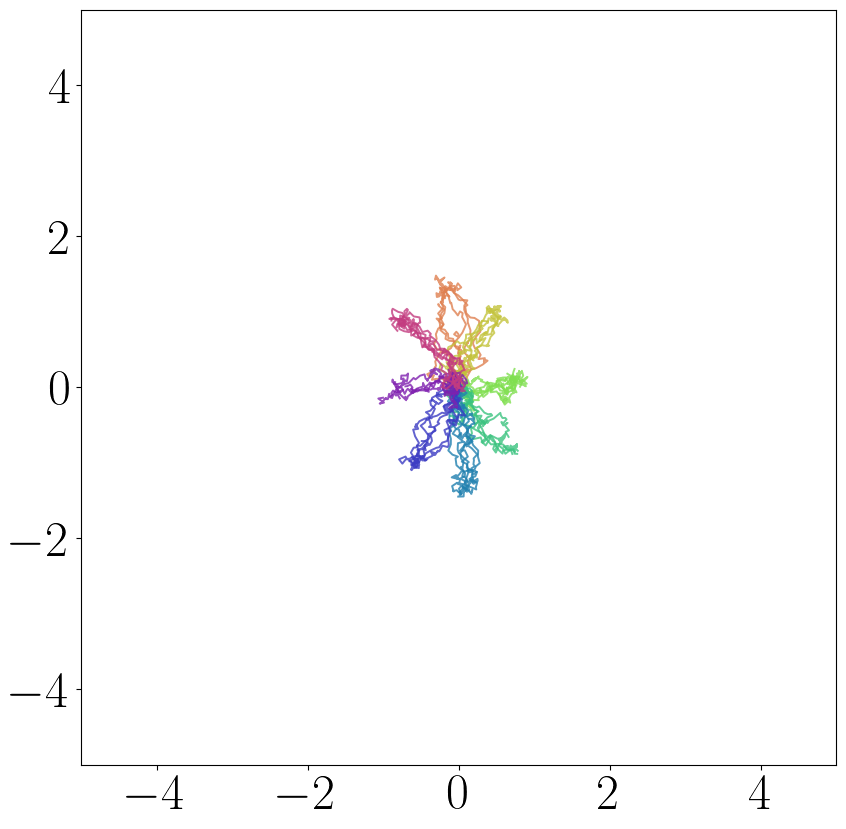}
    \includegraphics[width=0.22\columnwidth]{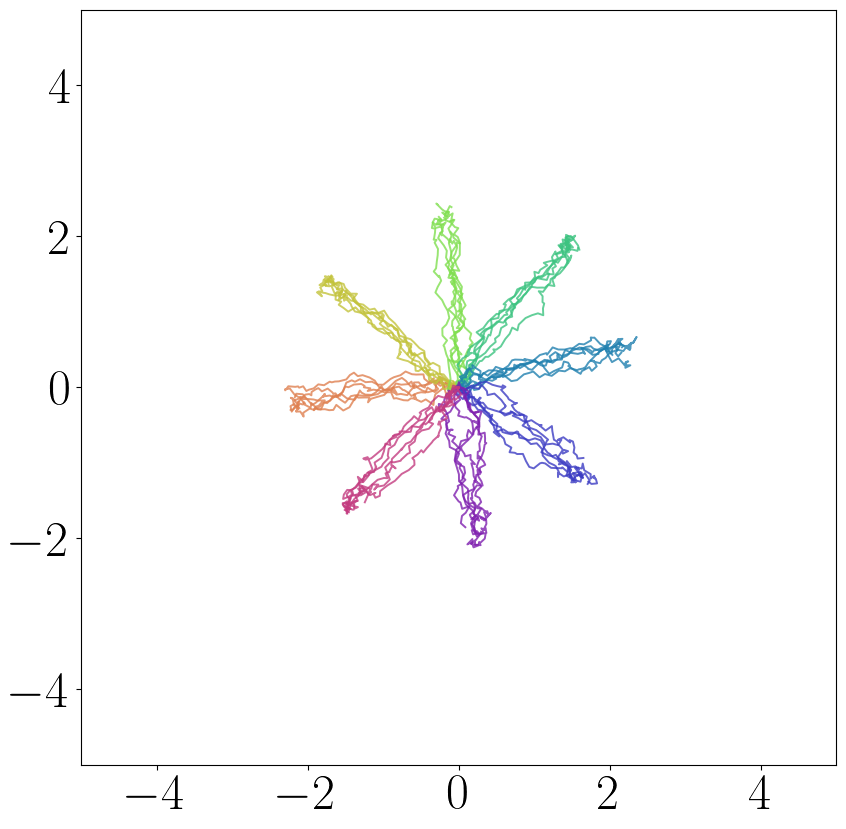}
    \includegraphics[width=0.22\columnwidth]{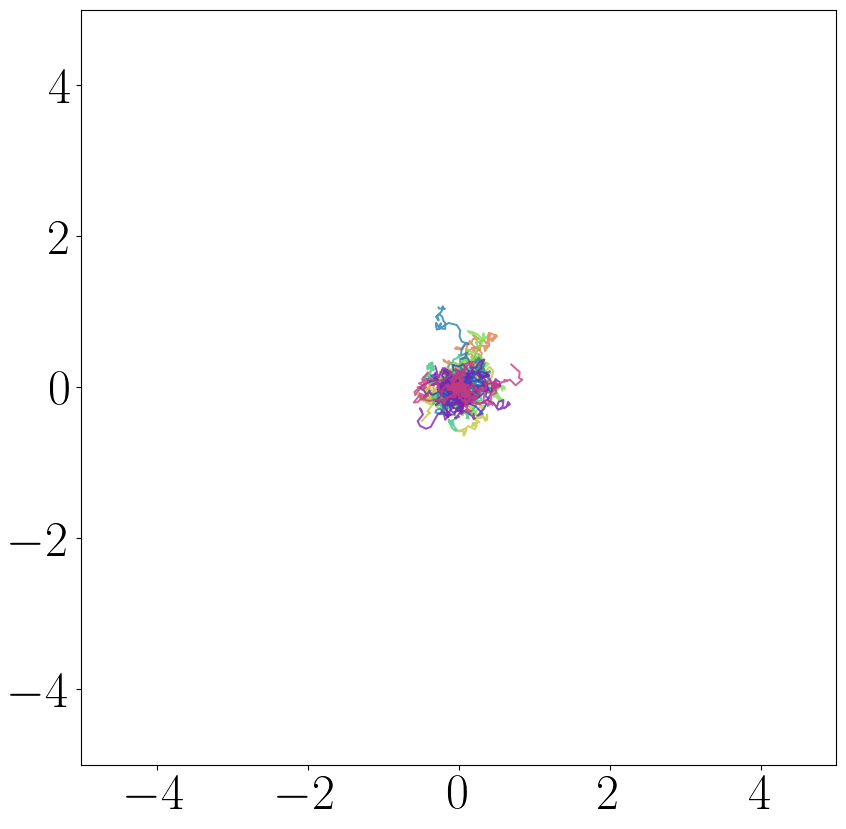}
  \end{subfigure}

  \captionsetup[subfigure]{labelformat=empty}
  \begin{subfigure}[t]{0.95\linewidth}
    \begin{subfigure}[t]{0.11\columnwidth}
        \caption{}
    \end{subfigure}
    \begin{subfigure}[t]{0.08\columnwidth}
        \caption{}
    \end{subfigure}
    \begin{subfigure}[t]{0.22\columnwidth}
        \caption{$\beta = 0$}
    \end{subfigure}
    \begin{subfigure}[t]{0.22\columnwidth}
        \caption{$\beta = 2.25e-3$}
    \end{subfigure}
    \begin{subfigure}[t]{0.22\columnwidth}
        \caption{$\beta = 2.25e-1$}
    \end{subfigure}
    \begin{subfigure}[t]{0.11\columnwidth}
        \caption{}
    \end{subfigure}
  \end{subfigure}

  \caption{
      Visualization of the $x$-$y$ traces of the skills discovered by IBOL in PointEnv with various hyperparameter settings.
      The fourth row corresponds to IBOL without $u$ and the auxiliary term, modelling $\pisample$ as a LSTM policy.
      The same skill latents are used with the top row of Figure \ref{fig:ant_xy} of the main paper.
  }
  \label{fig:point_ibol}
\end{figure*}

\begin{figure*}[t!]
  \begin{subfigure}[t]{0.95\linewidth}
    \makebox[20pt]{\raisebox{50pt}{\rotatebox[origin=c]{90}{VALOR}}}
    \centering
    \captionsetup[subfigure]{labelformat=empty}
    \hspace{0.22\columnwidth}
    \includegraphics[width=0.22\columnwidth]{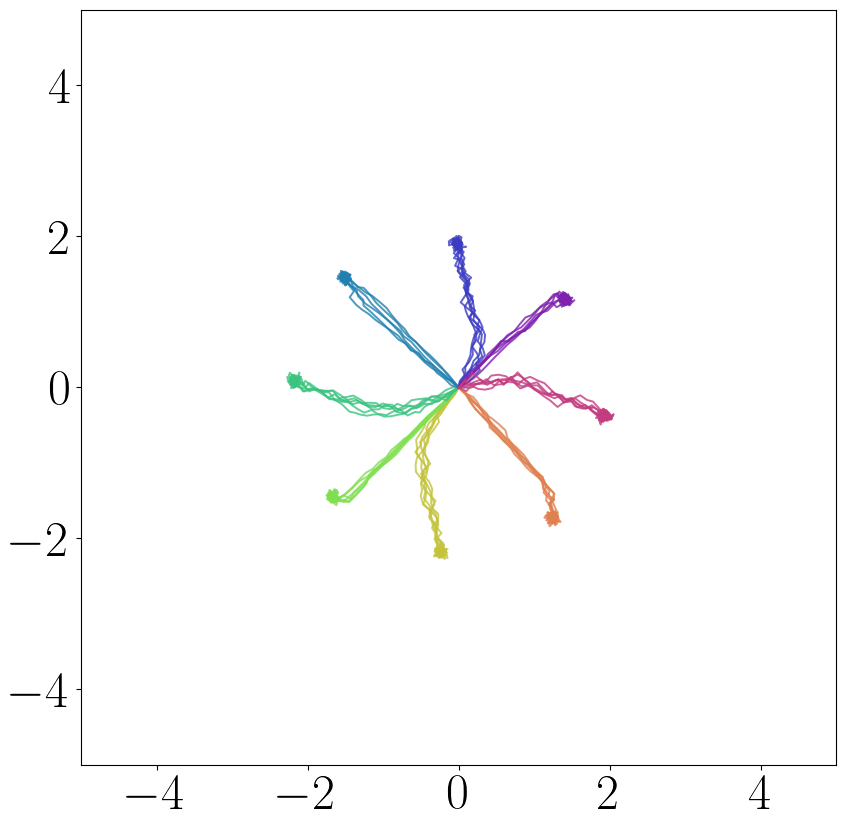}
    \includegraphics[width=0.22\columnwidth]{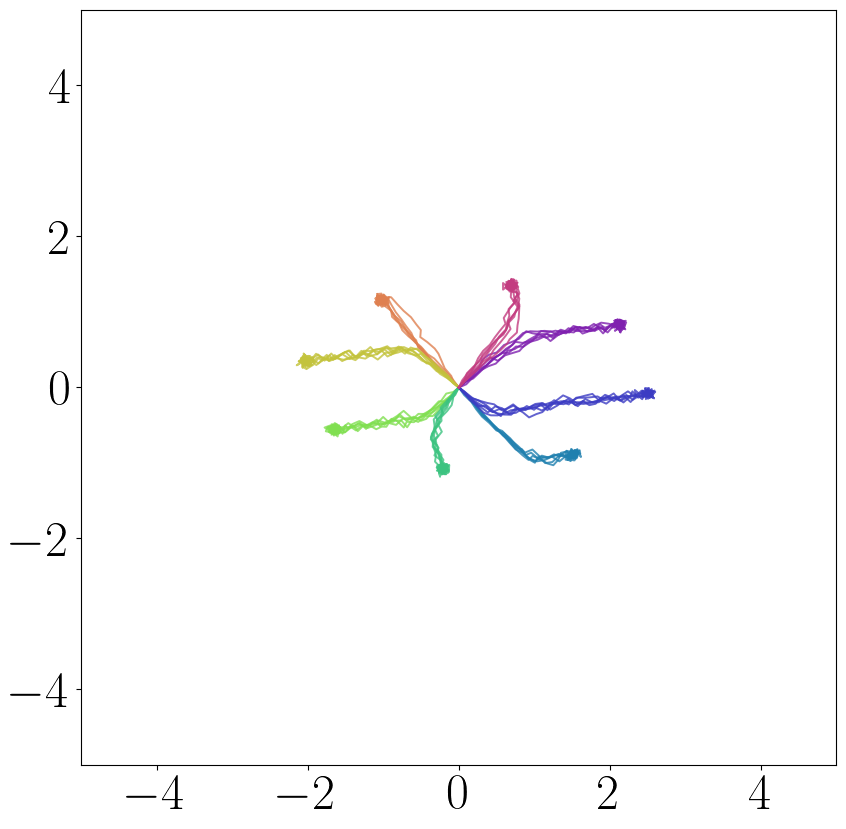}
    \includegraphics[width=0.22\columnwidth]{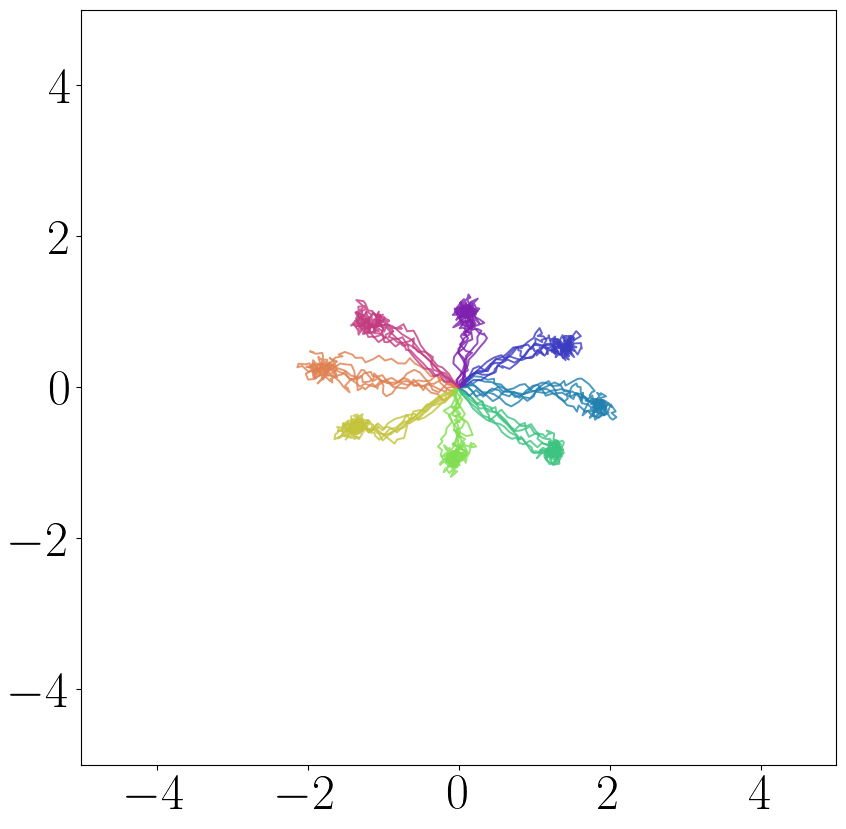}
  \end{subfigure}

  \begin{subfigure}[t]{0.95\linewidth}
    \makebox[20pt]{\raisebox{50pt}{\rotatebox[origin=c]{90}{DIAYN}}}
    \centering
    \includegraphics[width=0.22\columnwidth]{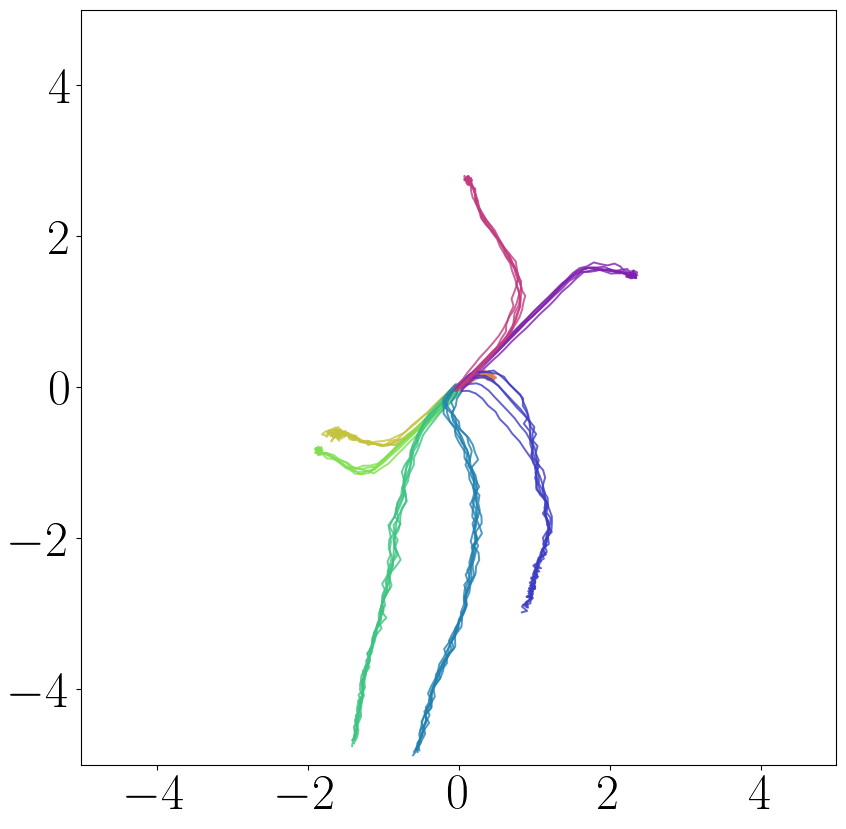}
    \includegraphics[width=0.22\columnwidth]{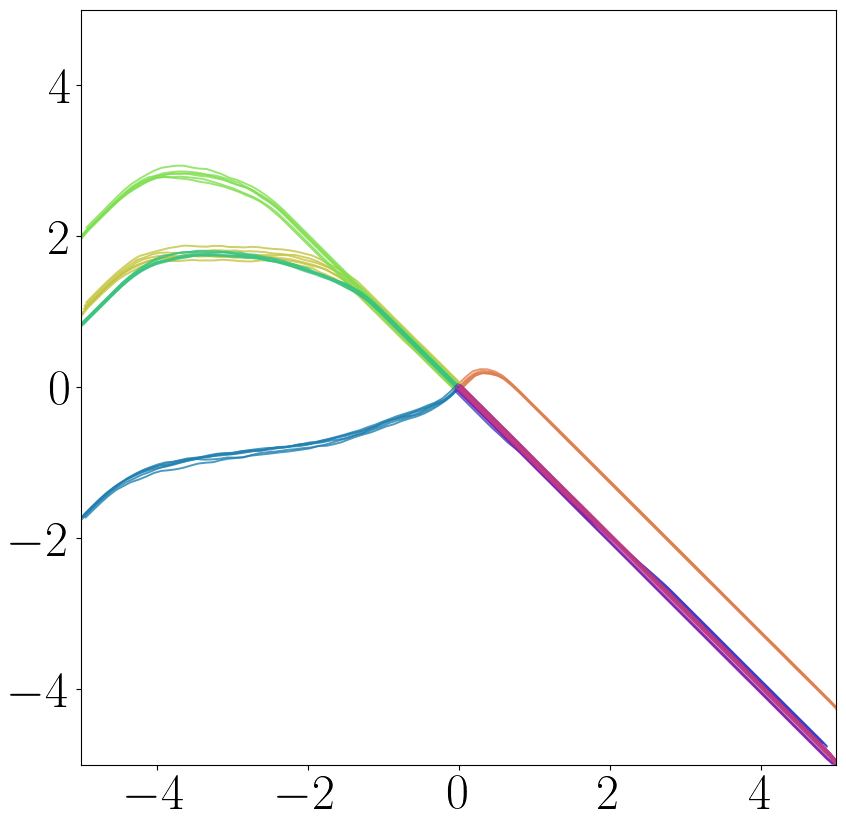}
    \includegraphics[width=0.22\columnwidth]{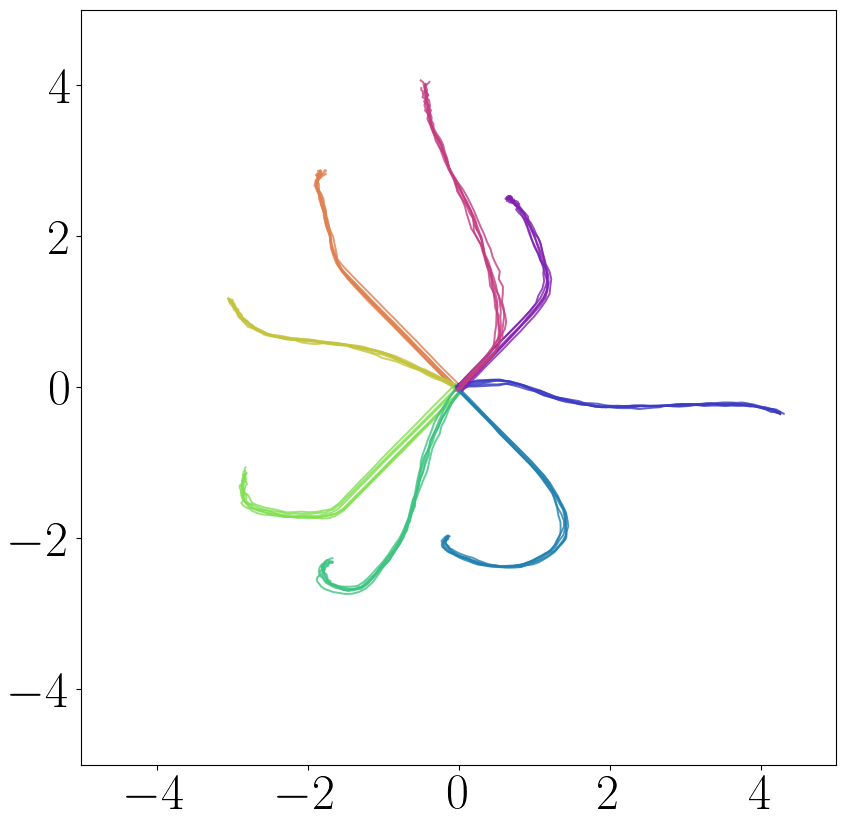}
    \includegraphics[width=0.22\columnwidth]{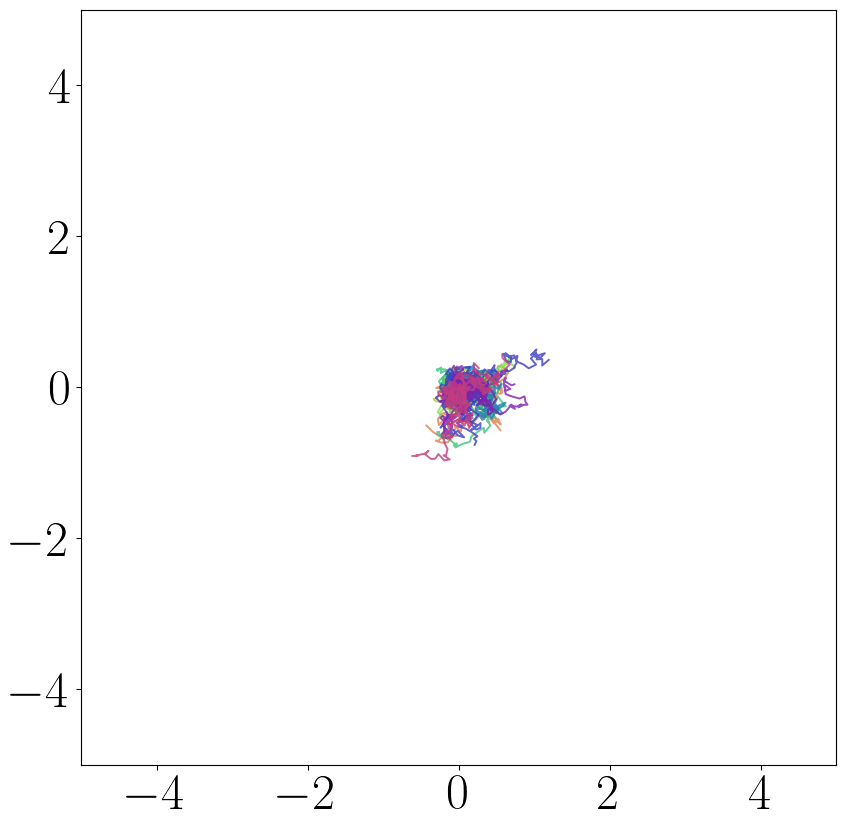}
  \end{subfigure}

  \begin{subfigure}[t]{0.95\linewidth}
    \makebox[20pt]{\raisebox{50pt}{\rotatebox[origin=c]{90}{DADS}}}
    \centering
    \includegraphics[width=0.22\columnwidth]{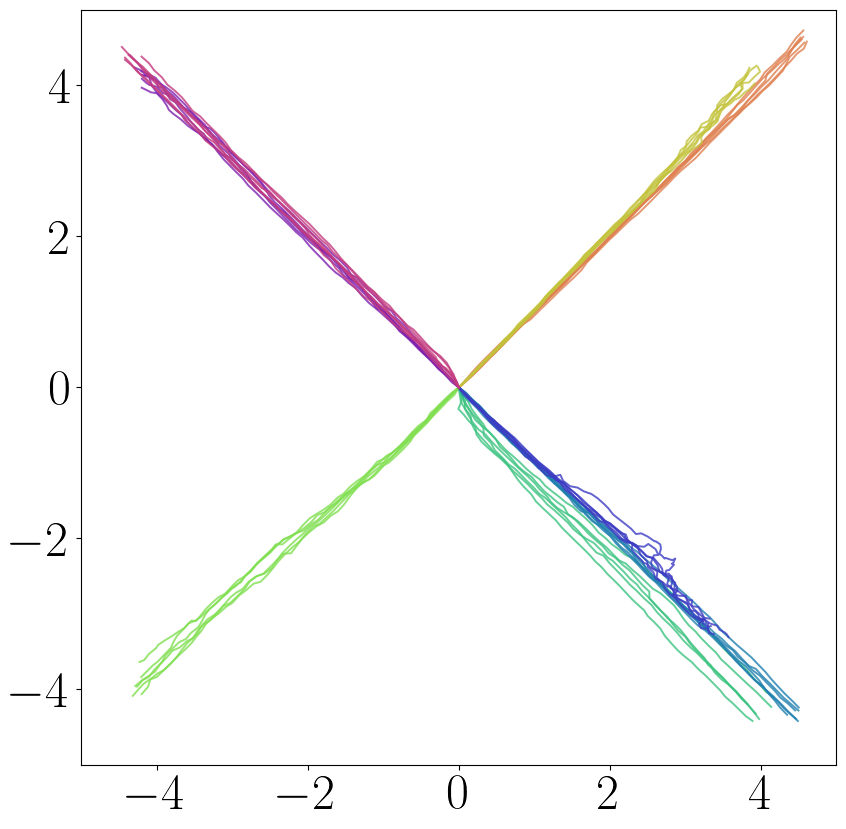}
    \includegraphics[width=0.22\columnwidth]{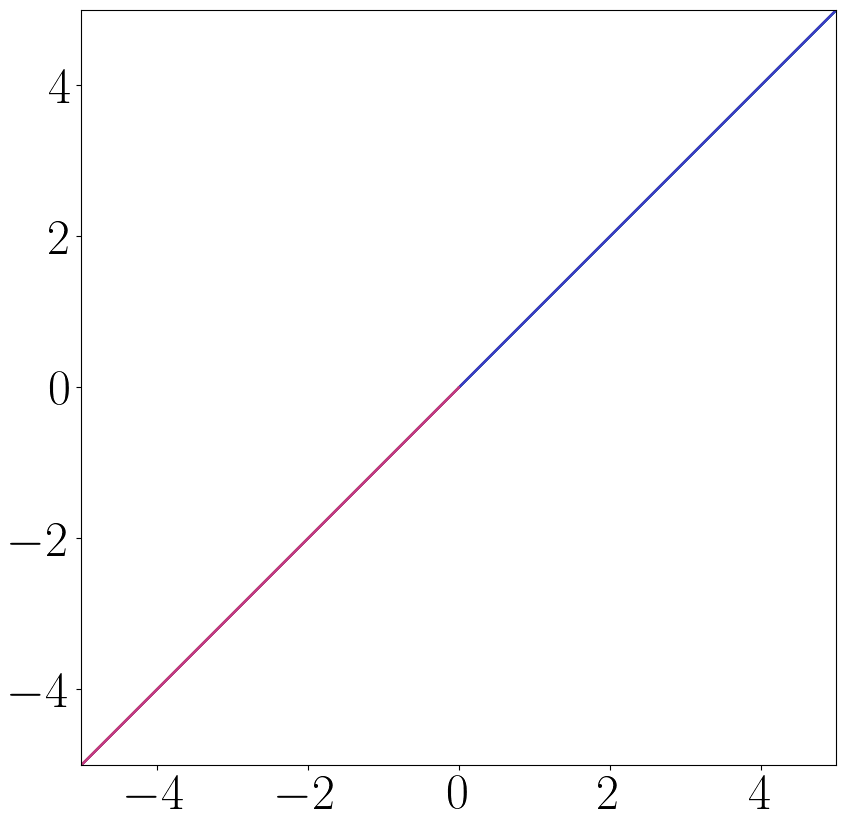}
    \includegraphics[width=0.22\columnwidth]{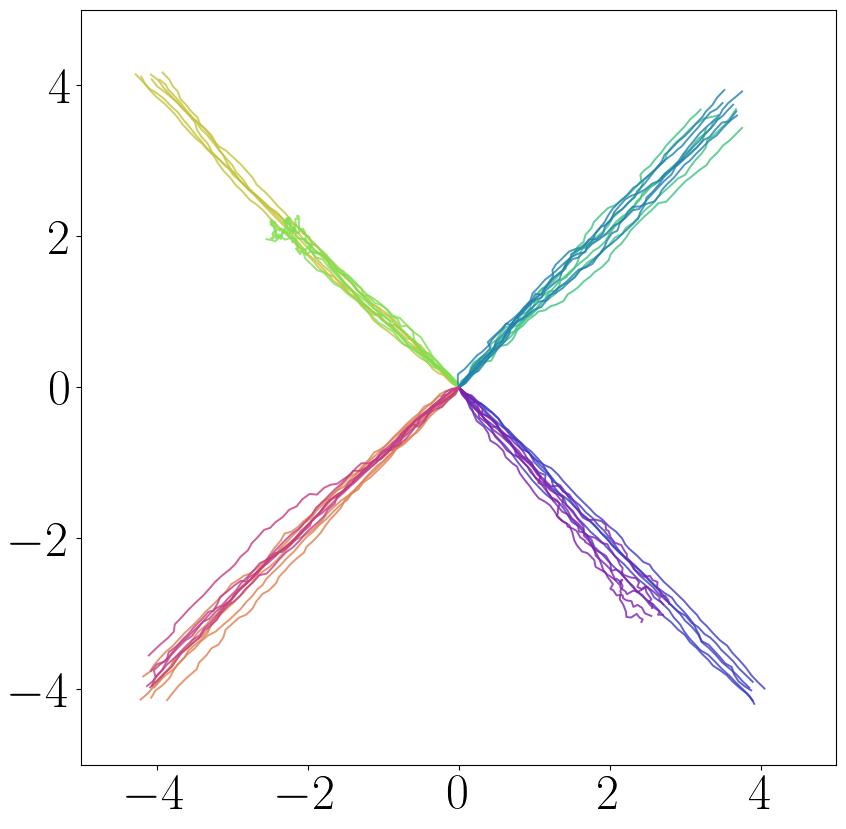}
    \includegraphics[width=0.22\columnwidth]{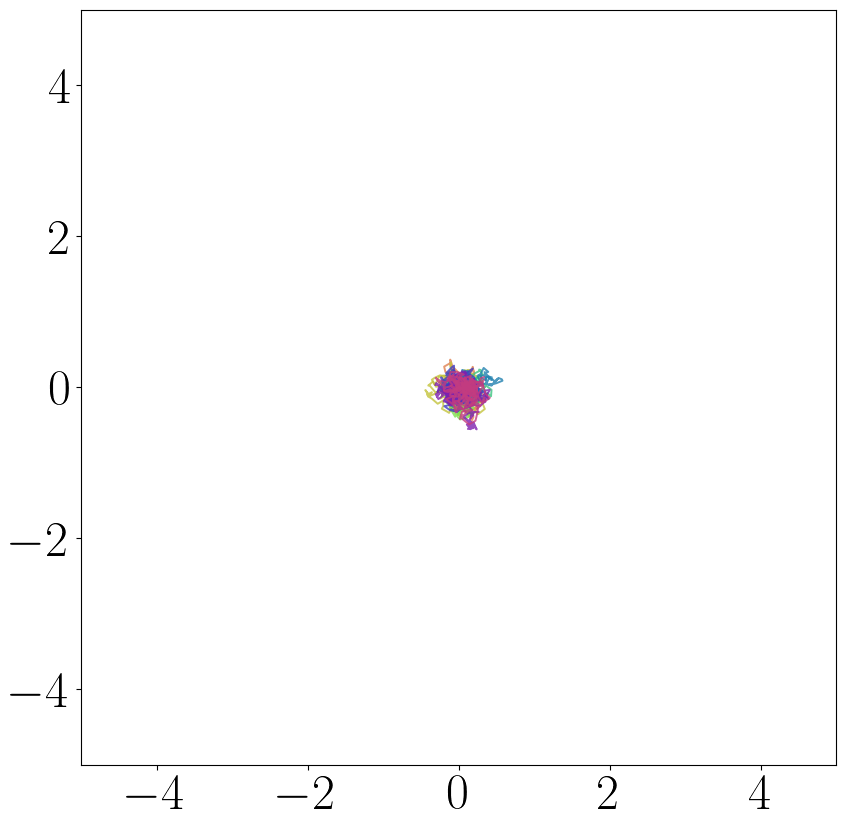}
  \end{subfigure}

  \captionsetup[subfigure]{labelformat=empty}
  \begin{subfigure}[t]{0.95\linewidth}
    \begin{subfigure}[t]{0.08\columnwidth}
        \caption{}
    \end{subfigure}
    \begin{subfigure}[t]{0.22\columnwidth}
        \caption{Auto-adjusted $\alpha$}
    \end{subfigure}
    \begin{subfigure}[t]{0.22\columnwidth}
        \caption{$\alpha = 1e-3$}
    \end{subfigure}
    \begin{subfigure}[t]{0.22\columnwidth}
        \caption{$\alpha = 1e-1$}
    \end{subfigure}
    \begin{subfigure}[t]{0.22\columnwidth}
        \caption{$\alpha = 1e+1$}
    \end{subfigure}
  \end{subfigure}

  \caption{
      Visualization of the $x$-$y$ traces of the skills discovered by VALOR, DIAYN and DADS in PointEnv with various hyperparameter settings.
      The same skill latents are used with the top row of Figure \ref{fig:ant_xy} of the main paper.
  }
  \label{fig:point_others}
\end{figure*}

\textbf{Skill discovery methods}.
We train IBOL and the `-L' variants of other skill discovery methods for $1e4$ epochs with $\ell = 10$,
while the methods without the linearizer are trained with the number of transitions that matches the total number of transitions for the training of both the linearizer and each skill discovery method on top of it.
We employ SAC for DADS and DIAYN, and the vanilla policy gradient (VPG) for IBOL and VALOR.
We set the entropy regularization coefficient to 1e-3 for VALOR and VALOR-L (searched over $\{1e-1, 1e-2, 1e-3, 0\}$),
and use the automatic entropy adjustment for DADS with an initial entropy coefficient of $1e-1$, DADS-L with $1e-3$, DIAYN with $1e-1$ and DIAYN-L with $1e-2$ (searched over $\{1e-1, 1e-2, 1e-3\}$ with and without the automatic regularization).
For those using VPG, we apply four gradient steps with the entire batch at each epoch.
For SAC, we apply $64$ gradient steps (or $32$ steps for the skill dynamics model in DADS) with $256$-sized mini-batches, since increased gradient steps expedite the training by exploiting the off-policy property of SAC. 
We use $L = 100$ for DADS and IBOL, and set $\lambda = 2$ (searched over $\{0.1, 1, 2\}$) and $\beta = 1e-2$ (searched over $\{1e-1, 1e-2, 1e-3\}$) for IBOL.

\textbf{Meta-controllers for downstream policies}.
SAC is used for training the meta-controllers. %
We fix the entropy coefficient to $0.01$, and apply four gradient steps with the full-sized batches.
The meta-controllers select skill latents in a range of $[-2, 2]$, where they are fed into the learned skill policies.

\subsection{Downstream Tasks}
In \textit{AntGoal}, we sample a goal $w \in [-50, 50]^2$ at the beginning of each roll-out.
In \textit{AntMultiGoals}, a goal $w$ is sampled from $[s^{(x)}-15, s^{(x)}+15] \times [s^{(y)}-15, s^{(y)}+15]$ at every $\eta = 50$ steps, where $[s^{(x)}, s^{(y)}]$ denotes the agent's position when the goal is about to be sampled.
In \textit{CheetahGoal}, we sample a goal $w \in [-60, 60]$ when each episode starts.

\subsection{Information-Theoretic Evaluations}
\label{sec:info_eval_details}
For each environment, we employ two pre-trained linearizers, %
and train every method four times for each linearizer, resulting in eight runs in total.
To measure the quantities, we sample $2000$ trajectories per run and use quantization,
where for each variable we divide the range of the values from all the runs into $32$ bins.

\subsection{Additional Settings}
For the rendered scenes of skills, we additionally consider excluding velocity dimensions defining the goal space for the linearizer as in HIRO \citep{hiro_nachum2018}, to get more visually diverse skills.
For learning the non-locomotion skills in \Cref{sec:additional_obs} from the main paper, 
we exclude the $x$ and $y$ dimensions from the input of each component.
Also, for the experiments with the goal space distortion in \Cref{sec:additional_obs}, 
we preserve only the $x$ and $y$ dimensions in the inputs.

\section{Ablation Study}

In this section, we demonstrate the effect of each hyperparameter of IBOL by showing qualitative results on a synthetic environment named \textit{PointEnv}, which is suitable for clear illustrations.
In PointEnv, a state $s \in \mathbb{R}^2$ is defined as the $x$-$y$ coordinates of the agent (point), and an action $a \in [-0.1, 0.1]^2$ indicates a vector by which the agent moves. 
The initial state is sampled from $[-0.05, 0.05]^2$ uniformly at random.
As PointEnv is already linearized, we do not use the linearizer for IBOL as well as other baseline methods.
We also reduce the common dimensionality of the neural networks to $32$ in lieu of $512$.
We train IBOL, DIAYN, VALOR and DADS for $5e3$ epochs with an episode length of $50$ and a learning rate of $3e-4$,
having two-dimensional skill latents with various hyperparameter settings on this environment.
For IBOL, we test $\beta \in \{2.25e-1, 2.25e-3, 0\}$ and $\lambda \in \{1.5, 0.45, 0.15\}$,
and we also consider the setting without the auxiliary parameter $u$ for the sampling policy $\pisample$, in which we model the sampling policy as an LSTM policy (instead of a non-recurrent policy) to compensate for the reduced expressiveness that comes from the dropping of $u$.
We examine the entropy regularization coefficient $\alpha \in \{1e+1, 1e-1, 1e-3\}$ for VALOR, DADS and DIAYN,
and we test the automatic entropy regularization for SAC \cite{essac_haarnoja2018} as well for the latter two.

\Cref{fig:point_ibol,fig:point_others} illustrate the $x$-$y$ traces of the skills discovered by each method with various settings.
First, we observe that an appropriate value of $\beta$ (especially $\beta = 2.25e-3$ in \Cref{fig:point_ibol}) for IBOL helps discover more disentangled and evenly distributed skills.
Also, since the auxiliary term
$\mathbb{E}_{u \sim p(u), \tau \sim p_{\samplesub}(\tau | u)} [ \lambda \cdot \trajencoder(u | s_{0:T}) ]$
encourages IBOL to discover skills that can be easily reconstructed from the trajectories, increasing $\lambda$ results in having relatively condensed trajectories.
The fourth row of \Cref{fig:point_ibol} shows that IBOL can still discover visually disentangled (yet slightly noisy) skills in the absence of $u$ and the auxiliary term.
\Cref{fig:point_others} presents that for the baseline methods, overly increasing $\alpha$ could result in collapsing while having a moderate value of $\alpha$ improves the quality of discovered skills.

\end{document}